%% file: paper.tex
\documentclass{article}
\pdfoutput=1 
\usepackage[nonatbib,preprint]{neurips_2024}
\usepackage[utf8]{inputenc} %
\usepackage[T1]{fontenc}    %
\usepackage{hyperref}       %
\usepackage{xcolor}
\hypersetup{colorlinks,linkcolor={red!50!black},citecolor={blue!50!black},urlcolor={blue!80!black}}
\usepackage{url}            %
\usepackage{booktabs}       %
\usepackage{amsfonts}       %
\usepackage{nicefrac}       %
\usepackage{microtype}      %
\usepackage{xcolor}         %
\usepackage{adjustbox}
\usepackage[numbers]{natbib}

\usepackage{algorithm}
\usepackage[noend]{algorithmic}

\usepackage{amsmath}
\usepackage{amssymb}
\usepackage{mathtools}
\usepackage{amsthm}
\usepackage{colonequals}
\usepackage{dsfont}

\usepackage{wrapfig}

\theoremstyle{plain}
\newtheorem{theorem}{Theorem}[section]

\newtheorem{lemma}[theorem]{Lemma}
\newtheorem{corollary}[theorem]{Corollary}
\theoremstyle{definition}

\theoremstyle{remark}

\usepackage{pgfplots}
\DeclareUnicodeCharacter{2212}{−}
\usepgfplotslibrary{groupplots,dateplot}
\usetikzlibrary{patterns,shapes.arrows}
\pgfplotsset{compat=newest}

\usepackage{physics}
\usepackage{amsmath}
\usepackage{tikz}
\usepackage{mathdots}
\usepackage{yhmath}
\usepackage{cancel}
\usepackage{color}
\usepackage{siunitx}
\usepackage{array}
\usepackage{multirow}
\usepackage{amssymb}
\usepackage{gensymb}
\usepackage{tabularx}
\usepackage{extarrows}
\usepackage{booktabs}
\usetikzlibrary{fadings}
\usetikzlibrary{patterns}
\usetikzlibrary{shadows.blur}
\usetikzlibrary{shapes}

\usetikzlibrary{arrows}
\usepackage{url}
\usepackage{booktabs}
\usepackage{subfigure}

\DeclareMathOperator*{\argmin}{arg\,min}

\let\L\undefined %
\newcommand{\L}{\mathcal{L}}
\newcommand{\barL}{\bar{\mathcal{L}}}
\newcommand{\Lw}{\L_w}
\newcommand{\barLw}{\barL_w}
\newcommand{\X}{\mathcal{X}}
\newcommand{\Y}{\mathcal{Y}}
\newcommand{\aug}{\mathrm{(aug)}}
\newcommand{\Xaug}{\mathcal{X}^{\aug}}
\newcommand{\Var}{\operatorname{Var}}
\newcommand{\Vmax}{V_{\max}}
\newcommand{\regLw}{L_w}

\usepackage{xspace}
\newcommand{\newowa}{ACOWA\xspace}

\title{Optimizing the Optimal Weighted Average: Efficient Distributed Sparse Classification}

\author{%
  Fred Lu$^*$ \\
  Booz Allen Hamilton \\
  \texttt{Lu\_Fred@bah.com} \\
  \And
  Ryan R. Curtin$^*$ \\
  Booz Allen Hamilton \\
  \texttt{ryan@ratml.org} \\
  \And
  Edward Raff \\
  Booz Allen Hamilton \\
  \And
  Francis Ferraro \\
  University of Maryland, Baltimore County \\
  \And
  James Holt \\
  Laboratory for Physical Sciences \\
}

\begin{document}

\maketitle

\def\thefootnote{*}\footnotetext{Equal contribution}

\begin{abstract}
While distributed training is often viewed as a solution to
optimizing linear models on increasingly large datasets,
inter-machine communication costs of popular distributed approaches
can dominate as data dimensionality increases.
Recent work on non-interactive algorithms shows that
approximate solutions for linear models can be obtained efficiently
with only a single round of communication among machines.
However, this approximation often degenerates as the number of machines increases.
In this paper,
building on the recent optimal weighted average method,
we introduce a new technique,
{\it \newowa},
that allows an extra round of communication %
to achieve noticeably better approximation quality
with minor runtime increases.
Results show that for sparse distributed logistic regression,
\newowa obtains solutions that are more faithful to the empirical risk minimizer and attain substantially higher accuracy than other distributed algorithms.
\end{abstract}

\input{sections/introduction.tex}
\input{sections/owa_problems.tex}
\input{sections/centroid_augmentation.tex}
\input{sections/fw_second_round.tex}
\input{sections/acowa.tex}
\input{sections/isoefficiency.tex}
\input{sections/experiments.tex}
\input{sections/conclusion.tex}

\bibliographystyle{unsrtnat}
\bibliography{refs}

\newpage
\appendix

\input{sections/appendix.tex}

\end{document}

%% file: sections/introduction.tex
\vspace*{-0.5em}
\section{Introduction}
\label{sec:introduction}
\vspace*{-0.3em}

Statistical and machine learning research trends have had one important underlying trend for the past few decades:
practitioners want to train models on larger and larger datasets~\cite{jordan2015machine, halevy2009unreasonable}.
Massive-scale datasets present significant computational issues,
regardless of the complexity of the models being used.
Even training linear models is a challenge
when the datasets get large and high-dimensional.
As an example, consider a simple logistic regression model,
which may be penalized with either the L1-regularizer for sparsity, 
the L2-regularizer to prevent overfitting,
or both (the `elastic net' \cite{zou2005regularization}).
Given a dataset $\X$ with $n$ points in $d$ dimensions,
and labels $\Y$ with value $-1$ or $1$,
we want to find
\begin{equation}
\hat{w} \colonequals \argmin L_w(\X) = \argmin_w \L_w(\X) + \lambda_1 \| w \|_1 + \lambda_2 \| w \|_2^2
\end{equation}
\noindent where $\L_w(\X)$ is the logistic regression objective:
\begin{equation}
\L_w(\X) \colonequals \sum_{(x_i, y_i) \in (\X, \Y)} \log(1 + e^{-y_i w^\top x_i}).
\end{equation}
\addtolength{\textfloatsep}{-1.0em}
\begin{figure}[!h]
    \centering
    \adjustbox{max width=\textwidth}{%
    \input{big_picture}
    }
    \vspace*{-1.0em}
    \caption{Vertical dashed lines show synchronization points between threads and boxes indicate different compute nodes. Approaches for many-core and distributed training of models with an $L_1$ penalty are either one-shot (left), or iterative (right), neither of which produce satisfying solutions of high accuracy in a limited time frame. Our \newowa strikes a careful balance of sharing information, such that a more accurate solution can be obtained with just two rounds of communication.}
    \label{fig:big_picture}
\end{figure}
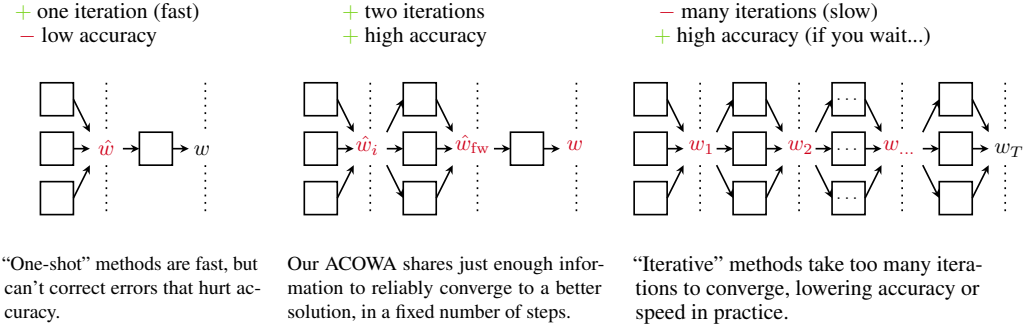
\addtolength{\textfloatsep}{1.0em}
Solving this problem on moderately-sized datasets is fast and easy~\cite{samadian2020unconditional},
but on datasets with millions or more of samples or features (or both!),
this is computationally challenging.
Having observed this,
we wish to accelerate the training of linear models on large-scale data.
Here we will consider the logistic regression objective,
but our approach can be easily adapted for more general models.

{\bf Early singlethreaded attempts}.  We are far from the first to consider
accelerating the training of logistic regression models.
For smaller datasets, the problem has been intensively studied~\cite{hastie2015statistical}.
There is extra difficulty when considering the L1 penalty or elastic net
(e.g. when $\lambda_1 > 0$),
as this causes $L_w(\X)$ to be non-differentiable
and thus simple gradient descent techniques cannot be directly applied.
Instead, algorithms such as FISTA and FASTA~\cite{goldstein2014field}, based on proximal gradient descent, are often used.
Proximal Newton techniques, using coordinate descent to solve the quadratic approximation, are the most popular approach,
with the {\it GLMNET}~\cite{friedman2010regularization} and {\it newGLMNET}~\cite{yuan2011improved} algorithms
offering fast convergence.
{\it newGLMNET} is specifically tuned for expensive objective functions such as logistic regression,
and through its implementation in LIBLINEAR~\cite{fan2008liblinear} has become likely the most widely-used solver in practice.

{\bf Multithreaded single-system approaches.}
As the number of cores on processors has increased,
interest in single-system parallelism has also.
LIBLINEAR-MP~\cite{zhuang2018naive} is a modified multi-core newGLMNET implementation.
Hogwild~\cite{recht2011hogwild} and the more recent SAUS~\cite{raff2018linear}
use lock-free parallelism to prevent conflicts during gradient updates.

{\bf Iterative distributed approaches}.  However, even with a multithreaded approach,
very large datasets may be larger than the memory of a single system,
and thus a distributed approach is required.
The use of distributed algorithms to
train logistic regression models has been studied extensively~\cite{gopal2013distributed, lin2014large, zhuang2015distributed}.
In typical approaches, the dataset $\X$ is partitioned across $p$ nodes,
and then the model is learned iteratively.
The simplest approach is to partition by data points;
this partitioning strategy has been paired with
distributed Newton methods~\cite{shamir2014communication,zhuang2015distributed}
and also ADMM~\cite{boyd2011distributed}. Block coordinate descent methods which split over features have also been shown to work \cite{trofimov2015distributed,richtarik2016distributed}.
A large disadvantage of these techniques is that they involve significant communication overhead:
at every iteration, the gradients from each machine must be communicated back to the main node.
These communication costs become very painful
as the dimensionality $d$ of the problem increases.

A number of approaches have been developed to reduce communication costs.
The CoCoA~\cite{jaggi2014communication,smith2018cocoa} and ProxCoCoA+~\cite{smith2015l1} frameworks are two notable examples.
ProxCoCoA+ uses the dual of the logistic regression objective function,
partitioning the data by dimension instead of points.
In each iteration,
each worker solves a local quadratic approximation of the objective,
communicating its solution back to the main node for aggregation.
The DANE and CSL frameworks~\cite{shamir2014communication,jordan2015machine,wang2017efficient}
first find single-partition solutions independently,
which are averaged as an initial estimator.
Global gradients are then collected and
combined with local higher-order derivatives on each partition to solve a surrogate likelihood function
that has bounded loss with respect to the true likelihood.

{\bf Non-interactive algorithms}.
Still, even with efforts to reduce the amount of communication,
the iterative nature of all the previous algorithms
presents a problematic overhead when the number of iterations is large.
Hence, there has recently been increased interest in {\it non-interactive} or {\it one-shot} algorithms~\cite{mcdonald2009efficient,zhang2012communication,zhang2013divide},
which use only a single round of communication.
These methods tend to be extremely fast,
but produce models with a larger amount of approximation.
The recent {\it optimal weighted average} (OWA) approach
is a compelling example of a non-interactive algorithm~\cite{izbicki2020distributed}.
In the OWA approach, data is partitioned along samples;
each worker trains a model independently on its data partition
and returns its trained model weights to the main node;
then, the final model is a learned linear combination of each partition's model.

{\bf Our contribution: \newowa}.
We have observed the compelling speedups of non-interactive algorithms,
but found ourselves disappointed by the approximation quality---especially as the number of partitions $p$ grows large,
and as the dataset becomes sparse.
Aiming to trade a small amount of speed for a much better approximation,
we relax the one-shot requirement,
and use two rounds of communication.
Starting from OWA~\cite{izbicki2020distributed},
we introduce \newowa with a number of novel improvements:

\begin{itemize}
    \item We augment each partition's data with summary information from other partitions, reducing the variance of each partition's model.
    This step has compelling theoretical support,
    and we lay out reasonable conditions under which it is guaranteed to improve the model.

    \item We allow a second round of weighted distributed learning, similar to the iterated LASSO~\cite{huang2008iterated}.  This improves approximation quality by ensuring that \newowa selects only features that have support across many partitions' models.

    \item We show that \newowa has isoefficiency (a measure of communication efficiency) comparable to the original OWA, and thus retains its favorable scaling properties.

    \item Our experimental results demonstrate the significant quality increases that \newowa yields,
    on a variety of datasets,
    with only a modest additional runtime cost.
\end{itemize}

%% file: big_picture.tex
\tikzset{every picture/.style={line width=0.75pt}} %

\begin{tikzpicture}[x=0.75pt,y=0.75pt,yscale=-1,xscale=1]
\draw   (40,70) -- (60,70) -- (60,90) -- (40,90) -- cycle ;
\draw   (40,100) -- (60,100) -- (60,120) -- (40,120) -- cycle ;
\draw   (40,130) -- (60,130) -- (60,150) -- (40,150) -- cycle ;

\draw    (60,140) -- (68.66,122.68) ;
\draw [shift={(70,120)}, rotate = 116.57] [fill={rgb, 255:red, 0; green, 0; blue, 0 }  ][line width=0.08]  [draw opacity=0] (5.36,-2.57) -- (0,0) -- (5.36,2.57) -- (3.56,0) -- cycle    ;
\draw    (60,110) -- (67,110) ;
\draw [shift={(70,110)}, rotate = 180] [fill={rgb, 255:red, 0; green, 0; blue, 0 }  ][line width=0.08]  [draw opacity=0] (5.36,-2.57) -- (0,0) -- (5.36,2.57) -- (3.56,0) -- cycle    ;
\draw    (60,80) -- (68.66,97.32) ;
\draw [shift={(70,100)}, rotate = 243.43] [fill={rgb, 255:red, 0; green, 0; blue, 0 }  ][line width=0.08]  [draw opacity=0] (5.36,-2.57) -- (0,0) -- (5.36,2.57) -- (3.56,0) -- cycle    ;
\draw  [dash pattern={on 0.84pt off 2.51pt}]  (80,70) -- (80,100) ;
\draw  [dash pattern={on 0.84pt off 2.51pt}]  (80,120) -- (80,150) ;
\draw   (100,100) -- (120,100) -- (120,120) -- (100,120) -- cycle ;
\draw    (120,110) -- (127,110) ;
\draw [shift={(130,110)}, rotate = 180] [fill={rgb, 255:red, 0; green, 0; blue, 0 }  ][line width=0.08]  [draw opacity=0] (5.36,-2.57) -- (0,0) -- (5.36,2.57) -- (3.56,0) -- cycle    ;
\draw    (90,110) -- (97,110) ;
\draw [shift={(100,110)}, rotate = 180] [fill={rgb, 255:red, 0; green, 0; blue, 0 }  ][line width=0.08]  [draw opacity=0] (5.36,-2.57) -- (0,0) -- (5.36,2.57) -- (3.56,0) -- cycle    ;
\draw   (520,70) -- (540,70) -- (540,90) -- (520,90) -- cycle ;
\draw   (520,100) -- (540,100) -- (540,120) -- (520,120) -- cycle ;
\draw   (520,130) -- (540,130) -- (540,150) -- (520,150) -- cycle ;
\draw   (460,70) -- (480,70) -- (480,90) -- (460,90) -- cycle ;
\draw   (460,100) -- (480,100) -- (480,120) -- (460,120) -- cycle ;
\draw   (460,130) -- (480,130) -- (480,150) -- (460,150) -- cycle ;
\draw   (585,70) -- (605,70) -- (605,90) -- (585,90) -- cycle ;
\draw   (585,100) -- (605,100) -- (605,120) -- (585,120) -- cycle ;
\draw   (585,130) -- (605,130) -- (605,150) -- (585,150) -- cycle ;
\draw   (400,70) -- (420,70) -- (420,90) -- (400,90) -- cycle ;
\draw   (400,100) -- (420,100) -- (420,120) -- (400,120) -- cycle ;
\draw   (400,130) -- (420,130) -- (420,150) -- (400,150) -- cycle ;
\draw    (420,140) -- (428.66,122.68) ;
\draw [shift={(430,120)}, rotate = 116.57] [fill={rgb, 255:red, 0; green, 0; blue, 0 }  ][line width=0.08]  [draw opacity=0] (5.36,-2.57) -- (0,0) -- (5.36,2.57) -- (3.56,0) -- cycle    ;
\draw    (420,110) -- (427,110) ;
\draw [shift={(430,110)}, rotate = 180] [fill={rgb, 255:red, 0; green, 0; blue, 0 }  ][line width=0.08]  [draw opacity=0] (5.36,-2.57) -- (0,0) -- (5.36,2.57) -- (3.56,0) -- cycle    ;
\draw    (420,80) -- (428.66,97.32) ;
\draw [shift={(430,100)}, rotate = 243.43] [fill={rgb, 255:red, 0; green, 0; blue, 0 }  ][line width=0.08]  [draw opacity=0] (5.36,-2.57) -- (0,0) -- (5.36,2.57) -- (3.56,0) -- cycle    ;
\draw    (450,100) -- (458.66,82.68) ;
\draw [shift={(460,80)}, rotate = 116.57] [fill={rgb, 255:red, 0; green, 0; blue, 0 }  ][line width=0.08]  [draw opacity=0] (5.36,-2.57) -- (0,0) -- (5.36,2.57) -- (3.56,0) -- cycle    ;
\draw    (450,110) -- (457,110) ;
\draw [shift={(460,110)}, rotate = 180] [fill={rgb, 255:red, 0; green, 0; blue, 0 }  ][line width=0.08]  [draw opacity=0] (5.36,-2.57) -- (0,0) -- (5.36,2.57) -- (3.56,0) -- cycle    ;
\draw    (450,120) -- (458.66,137.32) ;
\draw [shift={(460,140)}, rotate = 243.43] [fill={rgb, 255:red, 0; green, 0; blue, 0 }  ][line width=0.08]  [draw opacity=0] (5.36,-2.57) -- (0,0) -- (5.36,2.57) -- (3.56,0) -- cycle    ;
\draw  [dash pattern={on 0.84pt off 2.51pt}]  (500,70) -- (500,100) ;
\draw    (480,140) -- (488.66,122.68) ;
\draw [shift={(490,120)}, rotate = 116.57] [fill={rgb, 255:red, 0; green, 0; blue, 0 }  ][line width=0.08]  [draw opacity=0] (5.36,-2.57) -- (0,0) -- (5.36,2.57) -- (3.56,0) -- cycle    ;
\draw    (480,110) -- (487,110) ;
\draw [shift={(490,110)}, rotate = 180] [fill={rgb, 255:red, 0; green, 0; blue, 0 }  ][line width=0.08]  [draw opacity=0] (5.36,-2.57) -- (0,0) -- (5.36,2.57) -- (3.56,0) -- cycle    ;
\draw    (480,80) -- (488.66,97.32) ;
\draw [shift={(490,100)}, rotate = 243.43] [fill={rgb, 255:red, 0; green, 0; blue, 0 }  ][line width=0.08]  [draw opacity=0] (5.36,-2.57) -- (0,0) -- (5.36,2.57) -- (3.56,0) -- cycle    ;
\draw    (510,100) -- (518.66,82.68) ;
\draw [shift={(520,80)}, rotate = 116.57] [fill={rgb, 255:red, 0; green, 0; blue, 0 }  ][line width=0.08]  [draw opacity=0] (5.36,-2.57) -- (0,0) -- (5.36,2.57) -- (3.56,0) -- cycle    ;
\draw    (510,110) -- (517,110) ;
\draw [shift={(520,110)}, rotate = 180] [fill={rgb, 255:red, 0; green, 0; blue, 0 }  ][line width=0.08]  [draw opacity=0] (5.36,-2.57) -- (0,0) -- (5.36,2.57) -- (3.56,0) -- cycle    ;
\draw    (510,120) -- (518.66,137.32) ;
\draw [shift={(520,140)}, rotate = 243.43] [fill={rgb, 255:red, 0; green, 0; blue, 0 }  ][line width=0.08]  [draw opacity=0] (5.36,-2.57) -- (0,0) -- (5.36,2.57) -- (3.56,0) -- cycle    ;
\draw  [dash pattern={on 0.84pt off 2.51pt}]  (562.5,70) -- (562.5,100) ;
\draw    (540,140) -- (548.66,122.68) ;
\draw [shift={(550,120)}, rotate = 116.57] [fill={rgb, 255:red, 0; green, 0; blue, 0 }  ][line width=0.08]  [draw opacity=0] (5.36,-2.57) -- (0,0) -- (5.36,2.57) -- (3.56,0) -- cycle    ;
\draw    (540,110) -- (547,110) ;
\draw [shift={(550,110)}, rotate = 180] [fill={rgb, 255:red, 0; green, 0; blue, 0 }  ][line width=0.08]  [draw opacity=0] (5.36,-2.57) -- (0,0) -- (5.36,2.57) -- (3.56,0) -- cycle    ;
\draw    (540,80) -- (548.66,97.32) ;
\draw [shift={(550,100)}, rotate = 243.43] [fill={rgb, 255:red, 0; green, 0; blue, 0 }  ][line width=0.08]  [draw opacity=0] (5.36,-2.57) -- (0,0) -- (5.36,2.57) -- (3.56,0) -- cycle    ;
\draw    (575,100) -- (583.66,82.68) ;
\draw [shift={(585,80)}, rotate = 116.57] [fill={rgb, 255:red, 0; green, 0; blue, 0 }  ][line width=0.08]  [draw opacity=0] (5.36,-2.57) -- (0,0) -- (5.36,2.57) -- (3.56,0) -- cycle    ;
\draw    (575,110) -- (582,110) ;
\draw [shift={(585,110)}, rotate = 180] [fill={rgb, 255:red, 0; green, 0; blue, 0 }  ][line width=0.08]  [draw opacity=0] (5.36,-2.57) -- (0,0) -- (5.36,2.57) -- (3.56,0) -- cycle    ;
\draw    (575,120) -- (583.66,137.32) ;
\draw [shift={(585,140)}, rotate = 243.43] [fill={rgb, 255:red, 0; green, 0; blue, 0 }  ][line width=0.08]  [draw opacity=0] (5.36,-2.57) -- (0,0) -- (5.36,2.57) -- (3.56,0) -- cycle    ;
\draw  [dash pattern={on 0.84pt off 2.51pt}]  (562.5,120) -- (562.5,150) ;
\draw  [dash pattern={on 0.84pt off 2.51pt}]  (500,120) -- (500,150) ;
\draw  [dash pattern={on 0.84pt off 2.51pt}]  (440,70) -- (440,100) ;
\draw  [dash pattern={on 0.84pt off 2.51pt}]  (440,120) -- (440,150) ;
\draw   (325,100) -- (345,100) -- (345,120) -- (325,120) -- cycle ;
\draw   (260,70) -- (280,70) -- (280,90) -- (260,90) -- cycle ;
\draw   (260,100) -- (280,100) -- (280,120) -- (260,120) -- cycle ;
\draw   (260,130) -- (280,130) -- (280,150) -- (260,150) -- cycle ;
\draw   (200,70) -- (220,70) -- (220,90) -- (200,90) -- cycle ;
\draw   (200,100) -- (220,100) -- (220,120) -- (200,120) -- cycle ;
\draw   (200,130) -- (220,130) -- (220,150) -- (200,150) -- cycle ;
\draw    (220,140) -- (228.66,122.68) ;
\draw [shift={(230,120)}, rotate = 116.57] [fill={rgb, 255:red, 0; green, 0; blue, 0 }  ][line width=0.08]  [draw opacity=0] (5.36,-2.57) -- (0,0) -- (5.36,2.57) -- (3.56,0) -- cycle    ;
\draw    (220,110) -- (227,110) ;
\draw [shift={(230,110)}, rotate = 180] [fill={rgb, 255:red, 0; green, 0; blue, 0 }  ][line width=0.08]  [draw opacity=0] (5.36,-2.57) -- (0,0) -- (5.36,2.57) -- (3.56,0) -- cycle    ;
\draw    (220,80) -- (228.66,97.32) ;
\draw [shift={(230,100)}, rotate = 243.43] [fill={rgb, 255:red, 0; green, 0; blue, 0 }  ][line width=0.08]  [draw opacity=0] (5.36,-2.57) -- (0,0) -- (5.36,2.57) -- (3.56,0) -- cycle    ;
\draw    (250,100) -- (258.66,82.68) ;
\draw [shift={(260,80)}, rotate = 116.57] [fill={rgb, 255:red, 0; green, 0; blue, 0 }  ][line width=0.08]  [draw opacity=0] (5.36,-2.57) -- (0,0) -- (5.36,2.57) -- (3.56,0) -- cycle    ;
\draw    (250,110) -- (257,110) ;
\draw [shift={(260,110)}, rotate = 180] [fill={rgb, 255:red, 0; green, 0; blue, 0 }  ][line width=0.08]  [draw opacity=0] (5.36,-2.57) -- (0,0) -- (5.36,2.57) -- (3.56,0) -- cycle    ;
\draw    (250,120) -- (258.66,137.32) ;
\draw [shift={(260,140)}, rotate = 243.43] [fill={rgb, 255:red, 0; green, 0; blue, 0 }  ][line width=0.08]  [draw opacity=0] (5.36,-2.57) -- (0,0) -- (5.36,2.57) -- (3.56,0) -- cycle    ;
\draw  [dash pattern={on 0.84pt off 2.51pt}]  (305,70) -- (305,100) ;
\draw    (280,140) -- (288.66,122.68) ;
\draw [shift={(290,120)}, rotate = 116.57] [fill={rgb, 255:red, 0; green, 0; blue, 0 }  ][line width=0.08]  [draw opacity=0] (5.36,-2.57) -- (0,0) -- (5.36,2.57) -- (3.56,0) -- cycle    ;
\draw    (280,110) -- (287,110) ;
\draw [shift={(290,110)}, rotate = 180] [fill={rgb, 255:red, 0; green, 0; blue, 0 }  ][line width=0.08]  [draw opacity=0] (5.36,-2.57) -- (0,0) -- (5.36,2.57) -- (3.56,0) -- cycle    ;
\draw    (280,80) -- (288.66,97.32) ;
\draw [shift={(290,100)}, rotate = 243.43] [fill={rgb, 255:red, 0; green, 0; blue, 0 }  ][line width=0.08]  [draw opacity=0] (5.36,-2.57) -- (0,0) -- (5.36,2.57) -- (3.56,0) -- cycle    ;
\draw    (315,110) -- (322,110) ;
\draw [shift={(325,110)}, rotate = 180] [fill={rgb, 255:red, 0; green, 0; blue, 0 }  ][line width=0.08]  [draw opacity=0] (5.36,-2.57) -- (0,0) -- (5.36,2.57) -- (3.56,0) -- cycle    ;
\draw  [dash pattern={on 0.84pt off 2.51pt}]  (365,70) -- (365,100) ;
\draw    (345,110) -- (352,110) ;
\draw [shift={(355,110)}, rotate = 180] [fill={rgb, 255:red, 0; green, 0; blue, 0 }  ][line width=0.08]  [draw opacity=0] (5.36,-2.57) -- (0,0) -- (5.36,2.57) -- (3.56,0) -- cycle    ;
\draw  [dash pattern={on 0.84pt off 2.51pt}]  (365,120) -- (365,150) ;
\draw  [dash pattern={on 0.84pt off 2.51pt}]  (305,120) -- (305,150) ;
\draw  [dash pattern={on 0.84pt off 2.51pt}]  (240,70) -- (240,100) ;
\draw  [dash pattern={on 0.84pt off 2.51pt}]  (240,120) -- (240,150) ;
\draw  [dash pattern={on 0.84pt off 2.51pt}]  (140,70) -- (140,100) ;
\draw  [dash pattern={on 0.84pt off 2.51pt}]  (140,120) -- (140,150) ;
\draw  [dash pattern={on 0.84pt off 2.51pt}]  (625,70) -- (625,100) ;
\draw  [dash pattern={on 0.84pt off 2.51pt}]  (625,120) -- (625,150) ;
\draw    (605,140) -- (613.66,122.68) ;
\draw [shift={(615,120)}, rotate = 116.57] [fill={rgb, 255:red, 0; green, 0; blue, 0 }  ][line width=0.08]  [draw opacity=0] (5.36,-2.57) -- (0,0) -- (5.36,2.57) -- (3.56,0) -- cycle    ;
\draw    (605,110) -- (612,110) ;
\draw [shift={(615,110)}, rotate = 180] [fill={rgb, 255:red, 0; green, 0; blue, 0 }  ][line width=0.08]  [draw opacity=0] (5.36,-2.57) -- (0,0) -- (5.36,2.57) -- (3.56,0) -- cycle    ;
\draw    (605,80) -- (613.66,97.32) ;
\draw [shift={(615,100)}, rotate = 243.43] [fill={rgb, 255:red, 0; green, 0; blue, 0 }  ][line width=0.08]  [draw opacity=0] (5.36,-2.57) -- (0,0) -- (5.36,2.57) -- (3.56,0) -- cycle    ;

\draw (131.5,106.4) node [anchor=north west][inner sep=0.75pt]    {$w$};
\draw (627.5,111) node    {$w_{T}$};
\draw (530,80) node  [font=\scriptsize]  {$\dotsc $};
\draw (530,110) node  [font=\scriptsize]  {$\dotsc $};
\draw (530,140) node  [font=\scriptsize]  {$\dotsc $};
\draw (80,109.19) node [color={rgb, 255:red, 208; green, 2; blue, 27 }  ,opacity=1 ] {$\hat{w}$};
\draw (440.86,109.19) node  [color={rgb, 255:red, 208; green, 2; blue, 27 }  ,opacity=1 ]  {$w_{1}$};
\draw (500.86,109.19) node  [color={rgb, 255:red, 208; green, 2; blue, 27 }  ,opacity=1 ]  {$w_{2}$};
\draw (560.84,109.19) node  [color={rgb, 255:red, 208; green, 2; blue, 27 }  ,opacity=1 ]  {$w_{\dotsc}$};
\draw (238.22,108.0) node  [color={rgb, 255:red, 208; green, 2; blue, 27 }  ,opacity=1 ]  {$\hat{w}_i$};
\draw (302.13,107.2) node  [color={rgb, 255:red, 208; green, 2; blue, 27 }  ,opacity=1 ]  {$\hat{w}_{\mathrm{fw}}$};
\draw (364.13,108.0) node  [color={rgb, 255:red, 208; green, 2; blue, 27 }  ,opacity=1 ]  {$w$};
\draw (95,195.5) node   [align=left] {\begin{minipage}[lt]{115.26pt}\setlength\topsep{0pt}
{\small ``One-shot'' methods are fast, but can't correct errors that hurt accuracy.}
\end{minipage}};
\draw (505,195.5) node   [align=left] {\begin{minipage}[lt]{156.15pt}\setlength\topsep{0pt}
``Iterative'' methods take too many iterations to converge, lowering accuracy or speed in practice.
\end{minipage}};
\draw (285,195.5) node   [align=left] {\begin{minipage}[lt]{142.67pt}\setlength\topsep{0pt}
{\small Our ACOWA shares just enough information to reliably converge to a better solution, in a fixed number of steps.}
\end{minipage}};
\draw (95,35) node   [align=left] {\begin{minipage}[lt]{102pt}\setlength\topsep{0pt}
$\displaystyle \textcolor[rgb]{0.49,0.83,0.13}{\mathbf{+}}$ one iteration (fast)\\\textcolor[rgb]{0.82,0.01,0.11}{$\mathbf{\displaystyle-}$} low accuracy
\end{minipage}};
\draw (510,35) node   [align=left] {\begin{minipage}[lt]{142.8pt}\setlength\topsep{0pt}
\textcolor[rgb]{0.82,0.01,0.11}{$\mathbf{\displaystyle-}$} many iterations (slow)\\$\displaystyle \textcolor[rgb]{0.49,0.83,0.13}{\mathbf{+}}$ high accuracy (if you wait...)
\end{minipage}};
\draw (280,35) node   [align=left] {\begin{minipage}[lt]{81.6pt}\setlength\topsep{0pt}
$\displaystyle \textcolor[rgb]{0.49,0.83,0.13}{\mathbf{+}}$ two iterations\\$\displaystyle \textcolor[rgb]{0.49,0.83,0.13}{\mathbf{+}}$ high accuracy
\end{minipage}};

\end{tikzpicture}

%% file: sections/owa_problems.tex
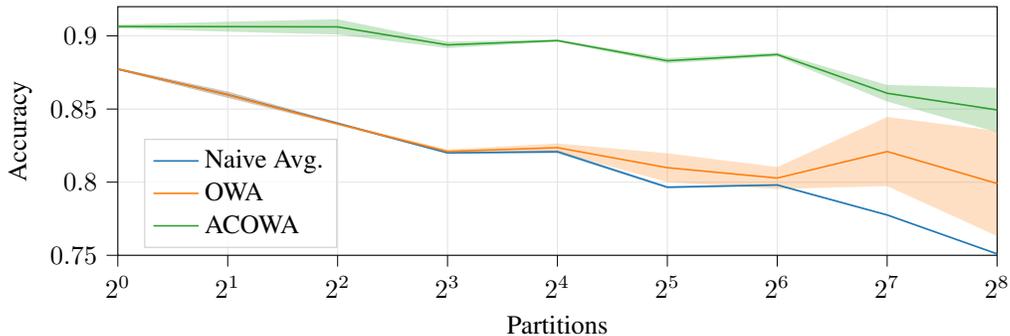
\begin{figure}[b!]
    \adjustbox{max width=\textwidth}{%
    \input{figs/p_sweep_ember100k_v2.tex}
    }
    \caption{Accuracy on
    held-out test set for different numbers of partitions $p$, when sparsity is fixed.  The quality of the naive averaging and OWA models degrades significantly as $p$ increases. Our method \newowa improves accuracy across all levels of $p$.}
    \vspace*{-1.0em}
\label{fig:partition-sweep}
\end{figure}

\vspace*{-0.5em}
\section{Problems with OWA}
\label{sec:owa_problems}
\vspace*{-0.3em}

Before describing OWA in detail,
it is helpful to consider the simplest one-shot approach:
{\it naive averaging}.
In both approaches, the dataset $(\X, \Y)$
is split into $p$ equal-sized partitions $(\X_i, \Y_i)$.
For naive averaging,
each worker $i$ learns a model $\hat{w}_i$ independently
on its partition,
and then all models are collected on the main worker and simply averaged.
Specifically,
$\hat{w}_{\mathrm{na}} \colonequals (1 / p) \sum_{i \in [p]} \hat{w}_i$.
Naive averaging is trivial to implement,
only involves one round of communication,
and gives reasonable approximate solutions to the true $\hat{w}$.
While naive averaging has been shown to be asymptotically optimal for nearly unbiased linear models~\cite{zhang2012communication},
in high-dimensional models higher-order loss terms cause increasing approximation error~\cite{rosenblatt2016optimality}.
Specifically, naive aggregates show greater error as the number of partitions $p$ grows, likely due to increasing bias of the subsampled estimators.

OWA~\citep{izbicki2020distributed} works similarly to naive averaging,
but has an improved merge step that results in an optimal $\mathcal{O}(\sqrt{d / n})$ approximation rate.
To reduce the bias of $\hat{w}_{\mathrm{na}}$,
OWA instead learns a weighted linear combination of each $\hat{w}_i$:
after each $\hat{w}_i$ is computed and returned to the main node,
a small subsample $\X_C \subseteq \X$ and $\Y_C \subseteq \Y$ is computed.
This sample can be as small as $pn / d$ points,
which in most settings is substantially smaller than $n$.
Then, the optimal weighted average is defined as
$\hat{w}_{\mathrm{owa}} \colonequals \hat{W} \hat{v}$,
where $\hat{W} \colonequals [ \hat{w}_1, \ldots, \hat{w}_p ]$.
For logistic regression, $\hat{v}$ is the linear combination of models found by solving the merge optimization
\begin{equation}
\hat{v} \colonequals \argmin_{v \in \mathbb{R}^p} \sum_{(x_i, y_i) \in (\X_c, \Y_c)} \log(1 + e^{-y_i (\hat{W} v)^\top x_i}) + \lambda_{\mathrm{cv}} \| v \|_2^2.
\label{eqn:cv_obj}
\end{equation}
Here, the original penalty is replaced with an L2 penalty for $v$.
This term can be taken as a surrogate for the `true' penalty term
$\lambda_1 \| \hat{W} v \|_1 + \lambda_2 \| \hat{W} v \|_2^2$,
and it is suggested that $\lambda_{\mathrm{cv}}$ be set by cross-validation.
Because $\X_C$ and $\Y_C$ are small subsets,
the cost of cross-validation is generally small as compared to
the cost of training each $\hat{w}_i$.
Adapting the OWA strategy to other statistical problems
simply involves reworking the original objective function to learn $v$ instead of $\hat{W} v$.

While the OWA estimator $\hat{w}_{\mathrm{owa}}$ tends to improve over the naive average $\hat{w}_{\mathrm{na}}$,
in practice the accuracy of the resulting model also degrades significantly
when the number of partitions $p$ becomes large.
This may be due to the need for a subsample for the second optimization---which will maintain bias.
In addition, the variance of the OWA estimator increases.
See Figure~\ref{fig:partition-sweep}
for a representative example.

%% file: figs/p_sweep_ember100k_v2.tex
\begin{tikzpicture}

\definecolor{darkorange25512714}{RGB}{255,127,14}
\definecolor{forestgreen4416044}{RGB}{44,160,44}
\definecolor{gainsboro229}{RGB}{229,229,229}
\definecolor{lightgray204}{RGB}{204,204,204}
\definecolor{steelblue31119180}{RGB}{31,119,180}

\begin{axis}[
width=0.95\textwidth,
height=0.35\textwidth,
legend cell align={left},
legend style={
  fill opacity=0.8,
  draw opacity=1,
  text opacity=1,
  at={(0.03,0.03)},
  anchor=south west,
  draw=lightgray204
},
log basis x={2},
tick align=outside,
tick pos=left,
x grid style={gainsboro229},
xlabel={Partitions},
xmajorgrids,
xmin=1, xmax=256,
xminorgrids,
xmode=log,
xtick style={color=black},
y grid style={gainsboro229},
ylabel={Accuracy},
ymajorgrids,
ymin=0.75,
ymax=0.92,
yminorgrids,
ytick style={color=black}
]
\path [fill=steelblue31119180, fill opacity=0.25]
(axis cs:1,0.878096869003469)
--(axis cs:1,0.876586730996531)
--(axis cs:2,0.857647836997387)
--(axis cs:4,0.839289392469112)
--(axis cs:8,0.819515089563976)
--(axis cs:16,0.8196116636728)
--(axis cs:32,0.795648726879535)
--(axis cs:64,0.797498154464848)
--(axis cs:128,0.777034805598043)
--(axis cs:256,0.750434534939892)
--(axis cs:256,0.751462265060108)
--(axis cs:256,0.751462265060108)
--(axis cs:128,0.778081594401957)
--(axis cs:64,0.798645445535152)
--(axis cs:32,0.797434073120465)
--(axis cs:16,0.8219683363272)
--(axis cs:8,0.820478110436024)
--(axis cs:4,0.841100607530887)
--(axis cs:2,0.862022163002613)
--(axis cs:1,0.878096869003469)
--cycle;

\path [fill=darkorange25512714, fill opacity=0.25]
(axis cs:1,0.878096869003469)
--(axis cs:1,0.876586730996531)
--(axis cs:2,0.857615538110808)
--(axis cs:4,0.838889256879917)
--(axis cs:8,0.819638176263559)
--(axis cs:16,0.820671769889183)
--(axis cs:32,0.800041047072381)
--(axis cs:64,0.795325191767899)
--(axis cs:128,0.797321354157186)
--(axis cs:256,0.763142805424661)
--(axis cs:256,0.835007194575339)
--(axis cs:256,0.835007194575339)
--(axis cs:128,0.844565445842814)
--(axis cs:64,0.810308008232102)
--(axis cs:32,0.819728552927619)
--(axis cs:16,0.826475030110817)
--(axis cs:8,0.822408623736441)
--(axis cs:4,0.840767543120083)
--(axis cs:2,0.861837661889192)
--(axis cs:1,0.878096869003469)
--cycle;

\path [fill=forestgreen4416044, fill opacity=0.25]
(axis cs:1,0.907652667850135)
--(axis cs:1,0.905160932149865)
--(axis cs:2,0.902909291604613)
--(axis cs:4,0.900950714209055)
--(axis cs:8,0.891669394179781)
--(axis cs:16,0.895952933197424)
--(axis cs:32,0.881211914193587)
--(axis cs:64,0.885942829325219)
--(axis cs:128,0.855104302506908)
--(axis cs:256,0.834183271345894)
--(axis cs:256,0.864566728654106)
--(axis cs:256,0.864566728654106)
--(axis cs:128,0.866468897493092)
--(axis cs:64,0.888597170674782)
--(axis cs:32,0.884791685806413)
--(axis cs:16,0.897637066802576)
--(axis cs:8,0.896043805820219)
--(axis cs:4,0.911402485790945)
--(axis cs:2,0.909757108395387)
--(axis cs:1,0.907652667850135)
--cycle;

\addplot [semithick,steelblue31119180, opacity=1.0]
table {%
1 0.8773418
2 0.859835
4 0.840195
8 0.8199966
16 0.82079
32 0.7965414
64 0.7980718
128 0.7775582
256 0.7509484
};
\addlegendentry{Naive Avg.}
\addplot [semithick,darkorange25512714, opacity=1.0]
table {%
1 0.8773418
2 0.8597266
4 0.8398284
8 0.8210234
16 0.8235734
32 0.8098848
64 0.8028166
128 0.8209434
256 0.799075
};
\addlegendentry{OWA}
\addplot [semithick,forestgreen4416044, opacity=1.0]
table {%
1 0.9064068
2 0.9063332
4 0.9061766
8 0.8938566
16 0.896795
32 0.8830018
64 0.88727
128 0.8607866
256 0.849375
};
\addlegendentry{ACOWA}
\end{axis}

\end{tikzpicture}

%% file: sections/centroid_augmentation.tex
\section{First Improvement: Centroid Augmentation}
\label{sec:centroid_augmentation}

The increasing variance of OWA as $p$ increases 
is a result of degradation in each $\hat{w}_i$:
the smaller $\X_i$ is,
the more likely $\hat{w}_i$ is to be further from the true $\hat{w}$.
As a first line of reasoning,
view $\X_i$ as a uniformly randomly sampled coreset of $\X$:
then, as $| \X_i |$ shrinks,
the bound on the relative error increases quadratically~\cite{samadian2020unconditional}.
As a second line of reasoning,
consider that a coreset of a high-dimensional sparse dataset
may contain features that do not have any nonzero values at all.
These `dead' features are then effectively ignored by any model learned on that coreset.
From both of these viewpoints,
we see $\X_i$ may not contain enough information
to reliably produce an accurate approximation to the true $\hat{w}$.

Consider a scheme where we augment $\X_i$ with centroids of other partitions:
\begin{equation}
\Xaug_i \colonequals \X_i \cup \left(\bigcup_{j \in [p] \setminus \{i\}} \mu^+_j \cup \mu^-_j\right)
\label{eqn:data_aug}
\end{equation}
\noindent with positive and negative centroids $\mu^+$ and $\mu^-$ defined as
\begin{equation}
\mu^+_j \colonequals \frac{1}{| \X_j^+ |} \sum_{x_k \in \X_j^+} x_k,\;\;
\mu^-_j \colonequals \frac{1}{| \X_j^- |} \sum_{x_k \in \X_j^-} x_k, \label{eqn:mu}
\end{equation}
\noindent where $\X_j^+$ is the subset of $\X_j$ with positive labels in $\Y_j$,
and correspondingly for $\X_j^-$.
The weight of any point in $\X_i$ is taken as $1$,
and the weight of any $\mu_j^+$ or $\mu_j^-$ is $| \X_j^+ |$ or $| \X_j^- |$, respectively.
(This scheme is trivially adaptable to the multiclass case.)
In order to theoretically justify this idea,
we show that augmenting $\X_i$ with $2p$ centroids
is guaranteed to produce a better approximation to $\hat{w}$
than increasing the size of $\X_i$ by sampling $2p$ additional points from $\X$.

\begin{lemma}
Suppose a partition $\X_i$ of size $k$ is an $\epsilon$-coreset with probability at least $1 - \delta$.
Define $\X^{(+2)}_i \colonequals \X_i \cup \{ x_{r1}, x_{r2} \}$
where $x_{r1}$ and $x_{r2}$ are uniformly randomly sampled points from the dataset.
$\X^{(+2)}_i$ is an $\epsilon^{(+2)}$-coreset also with probability at least $1 - \delta$, where
$
\epsilon^{(+2)} \colonequals \sqrt{k / k + 2} \epsilon$.
\label{lem:add_points_bound}
\end{lemma}
\begin{proof}
Due to space constraints,
all proofs in this paper are deferred to Appendix~\ref{app:proofs}.
\end{proof}

For $\Xaug$ to be worthwhile,
the loss bound must be less than $\epsilon^{+(2p)}$.
Let us therefore turn our attention to the loss on a centroid.
Here we focus on the case of positively-labeled points,
but the results are identical for negatively-labeled points.
First we bound the difference between the logistic loss on a coreset,
and the logistic loss on its centroid.

\begin{lemma}
For any coreset of positively-labeled points $\X_i$ with centroid $\mu_i$,
define the (unregularized) centroid loss for some weights $w$ as
$\barLw(\X_i) \colonequals | \X_i | \log(1 + e^{-w^\top \mu_i})$.
Then,
taking $Z_i$ as the vector of activations
$Z_i \colonequals \{ -w^\top x : x \in \X_i \}$,
\begin{equation}
0 \le \Lw(\X_i) - \barLw(\X_i) \le \frac{| \X_i |}{8} \Var(Z_i).
\end{equation}
\label{lem:centroid_loss_bound}
\end{lemma}

This individual bound can then be used to derive the final result.

\begin{theorem}
Given a centroid-augmented partition $\Xaug_i$
as defined in Eq.~\ref{eqn:data_aug},
where each partition $\X_i$ is an $\epsilon$-coreset with probability at least $1 - \delta$,
\begin{equation}
\frac{| \Lw(\X) - \Lw(\Xaug_i) |}{\Lw(\X)} \le \epsilon^{\aug}, \;\;\;\;
\epsilon^{\aug} \colonequals \frac{(p - 1)}{p} \frac{n \Vmax}{4 p \regLw(\X)}
\end{equation}
also with probability $1 - \delta$,
with $\Vmax$ defined as the maximum activation variance among centroid sets:
$%
\Vmax \colonequals \max_j ( \max( \Var(Z_j^+), \Var(Z_j^-) ) ).
$%
\label{thm:aug_loss_bound}
\end{theorem}

The important part of the result is the scaling behavior:
if the partition size $n / p$ remains constant,
$\epsilon^{\aug}$ increases only by a factor related to $1 / p$.
That is: as the dataset scales,
so long as $p$ is increased,
the bound remains (effectively) the same.
Next, consider the term $\regLw(\X)$ (the full data loss)
in the denominator.
As $\lambda$ increases, the optimal $\regLw(\X)$ with respect to $w$ will also increase,
tightening the bound.
Further, if $\lambda$ scales as $\mathcal{O}(\sqrt{n})$
as suggested by standard machine learning texts and other works~\cite{shalev2014understanding, negahban2012unified},
and the regularization term dominates the loss
(reasonable if $w$ is a good model),
then the bound takes a favorable $\mathcal{O}(\sqrt{n}/p)$ dependence.

Lastly, consider the maximum variance of activations $\Vmax$.
Regularized logistic regression models penalize large activations,
balancing $\log(1 + e^z)$ and the penalty by keeping $z$ (and thus $w$) small.
Thus, with higher $\lambda$s,
and as $w$ converges to the optimal solution,
we can expect $\Vmax$ to decrease.

Overall the intuition is reasonable:
as regularization increases,
and as the model improves,
centroids of other partitions provide better samples than other random samples.
We now formalize the conditions where
centroid augmentation is superior to
a coreset of size $n/p + 2p$.

\begin{corollary}
Define $\X_M$ as a uniform random sample from $\X$
of size $M \colonequals |\X_i|\!+\!2p$.
This is a $\epsilon^{(2p+)}$-coreset with probability $1 - \delta$
(Lemma~\ref{lem:add_points_bound}).
Then, given $\Xaug_i$ as in Eq.~\ref{eqn:data_aug},
$\epsilon^{\mathrm{aug}} \le \epsilon^{(2p+)}$
under the two conditions that $(n / p) \ge 2p$
(that is, if the number of points in each partition is at least twice the number of partitions),
and
\begin{equation}
\Vmax \le \frac{4p \regLw(\mathcal{X})}{n} \frac{1}{\sqrt{2}} \epsilon.
\end{equation}
\end{corollary}

%% file: sections/fw_second_round.tex
\section{Second Improvement: Feature Weighting}
\label{sec:fw_second_round}

When $\X$ is high-dimensional and sparse,
not only do we have the problem of `dead features'
as previously discussed,
but we also may have the situation where
individual features are significantly over- or under-represented
in any $\X_i$.
This can also cause greater variance in the performance of OWA,
as seen in Figure~\ref{fig:partition-sweep}.
This phenomenon is compounded by the fact that
L1-regularized logistic regression (and also elastic net) is not guaranteed to be
a consistent estimator or possess the oracle property~\cite{zou2006adaptive}.
Even in low dimensions, L1-regularized LASSO-type procedures are known to be inconsistent
in variable selection~\cite{fu2000asymptotics,leng2006note}.
Thus, if an $\X_i$ significantly over- or under-represents a feature in its sample of $\X$,
the effects on variable selection can be even worse.

\citet{zou2006adaptive} proposed a solution
for the consistency of the simple linear regression Lasso estimator with the adaptive Lasso,
which makes variable selection consistent by applying weights to each feature.
This was then extended to the case of
L1-regularized logistic regression with the `iterated Lasso'~\cite{huang2008iterated},
where a first model is trained on the data,
and then its weights are used to weight each feature for a second round of learning.
The iterated Lasso can be shown to be a selection-consistent estimator
and possesses the oracle property
under a few general assumptions on the data.

The strategy of the iterated Lasso
is straightforward to adapt to the distributed case:
we relax the one-round communication constraint
and allow an additional round of feature-weighted learning,
using weights from the first round of learning.
Given first-round models $\hat{w}_i$,
we can compute the percentage of models $\hat{w}_i$ that used a particular feature $j$:
$%
P_j \colonequals (1 / p) \sum\nolimits_{i \in [p]} \mathds{1}(\hat{w}_{ij} \ne 0).
$%
Then, we can define the weight for feature $j$ as
$\alpha_j \colonequals 1 + \beta P_j$
where $\beta$ is a tunable parameter that controls the severity of the feature weighting.
We then solve an adaptive feature-weighted second round optimization
in each partition:
\begin{equation}
\hat{w}^{\mathrm{fw}} \colonequals \argmin_{w \in \mathbb{R}^d} \sum_{i = 1}^{n} \ell(y_i, x_i^\top w) + \lambda_1 \sum_{j \in [d]} \alpha_j^{-1} | w_j | + \lambda_2 \sum_{j \in [d]} (\alpha_j^{-1} w_j)^2 .
\label{eqn:fw_obj}
\end{equation}

Note that this problem is equivalent to scaling each dimension $j$ of $\mathcal{X}$
by $\alpha_j$ with a rescaled penalty parameter
$\lambda_{fw} \colonequals \lambda d / (\sum_{i \in [d]} \alpha_i^{-1})$.
The use of a second round of individual model learning is an additional improvement over standard OWA, and
in our experiments, it improves stability
by acting as a soft feature-selection step across partitions.
This matches our expectation,
as the iterated Lasso can be understood as doing the same thing.

%% file: sections/acowa.tex
\newlength{\textfloatsepsave}
\setlength{\textfloatsepsave}{\textfloatsep}
\setlength{\textfloatsep}{0.0em}
\begin{figure}[t!]
\begin{algorithm}[H]%
\caption{\newowa.}
\label{alg:newowa}
\begin{algorithmic}[1]
\STATE {\bf Input}: $\mathcal{X} \in \mathbb{R}^{n \times d}$, $\mathcal{Y} \in \{ 0, 1 \}^n$, regularization penalty $\lambda$, $p$ processors, $\beta$ \\
\STATE {\bf Output}: learned model $\hat{w}_{\mathrm{acowa}}$ \\
\medskip
\STATE split $(\mathcal{X}, \mathcal{Y})$ into $p$ partitions and distribute to $p$ workers \\
\FOR{$i \in [p]$ {\bf in parallel}} \label{alg:cent_start}
    \STATE compute $\mu^+_i$ and $\mu^-_i$ using Eq.~\ref{eqn:mu} \\
\ENDFOR
\STATE distribute all $\mu^+_i$ and $\mu^-_i$ to all $p$ workers \label{alg:cent_end} \\
\FOR{$i \in [p]$ {\bf in parallel}}
    \STATE learn $\hat{w}_i$ using local optimizer on $\mathcal{X}_i^{\mathrm{aug}}$ (Eq.~\ref{eqn:data_aug})
\ENDFOR
\STATE compute $\alpha_j$ for all $j \in [d]$
\FOR{$i \in [p]$ {\bf in parallel}}
    \STATE learn $\hat{w}_i^{\mathrm{fw}}$ (Eq.~\ref{eqn:fw_obj}) using local optimizer on $\mathcal{X}_i^{\mathrm{aug}}$
\ENDFOR
\STATE form $(\mathcal{X}_C, \mathcal{Y}_C)$ as a sample of size $\min(n / p, pn / d)$ from $(\mathcal{X}, \mathcal{Y})$
\STATE compute $\hat{w}_{\mathrm{acowa}}$ as the solution to Eq.~\ref{eqn:cv_obj} with $\hat{W} = \{ \hat{w}_1^{\mathrm{fw}}, \ldots, \hat{w}_p^{\mathrm{fw}} \}$
\end{algorithmic}

\end{algorithm}
\vspace*{-2.0em}
\end{figure}
\setlength{\textfloatsep}{\textfloatsepsave}

\section{ACOWA}
\label{sec:acowa}

With these two major improvements over OWA,
we can now introduce \newowa,
shown in Algorithm~\ref{alg:newowa}.
\newowa uses the centroid augmentation of Sec.~\ref{sec:centroid_augmentation}
for the first distributed round of learning,
and then computes feature weights as described in Sec.~\ref{sec:fw_second_round}
for a second distributed round of learning.
Then, the standard OWA merge step (Eqn.~\ref{eqn:cv_obj}) is applied.
We highlight two additional improvements:

\textbf{Larger merge set.}
OWA's merge step
uses a dataset $\mathcal{X}_C$, of size $pn / d$.
For high-dimensional problems, this set can be extremely small ($<100$ samples),
which causes $\hat{w}_{\mathrm{owa}}$
to be highly variable.
Thus, it makes sense to increase the subsample size.
Given that data is already distributed across partitions,
we can simply use the main node's partition
(with size $n / p$)
as the second round.
This significantly reduces the variance of $\hat{w}_{\mathrm{owa}}$,
and has negligible runtime cost
(see Appendix~\ref{app:runtime-breakdown}).

\textbf{Better optimizer.}
Our implementation
(described further in Section~\ref{sec:experiments})
uses {\it newGLMNET} as provided by LIBLINEAR~\cite{fan2008liblinear}.
We found that relaxing the optimizer
by reducing the number of inner coordinate descent iterations to 50
and the number of outer Newton iterations to 20
gives good speedup without loss in accuracy.
Since the first round of learning can intuitively be understood as
a `soft' feature selection step,
it is not necessary to run the first round of optimization to full convergence.

\textbf{Extensions.}
In our exposition,
for the sake of simplicity,
we have specifically considered the regularized logistic regression problem,
and our theoretical results have been restricted to that problem.
However, ACOWA is a general algorithm,
and as such it is straightforward to substitute any other type of linear model
(such as, e.g., the linear SVM).
Theoretical results for centroid augmentation still apply,
if Lemma~\ref{lem:centroid_loss_bound} can be adapted to the objective function of interest.
Theoretical scaling results (in the next section) apply so long as the individual solvers
for each partition scale similarly to {\it newGLMNET}
(and if not, the results can be adapted).

%% file: sections/isoefficiency.tex
\vspace*{-0.5em}
\section{Scalability Analysis: Isoefficiency}
\label{sec:isoefficiency}
\vspace*{-0.3em}

In order to understand the scalability of \newowa in terms of data size $z$ and the number of partitions $p$,
we analyze a parallel performance metric known as the \textit{isoefficiency function}~\cite{grama1993isoefficiency}.
To the best of our knowledge, this is the first isoefficiency analysis of distributed logistic regression.

Generally, when $p$ is increased linearly,
the speedup $S$ of a system increases slower than linearly,
decreasing its overall efficiency $E$.
On the other hand, increasing the problem size $z$ super-linearly will increase the speedup and efficiency
because of the time-savings of parallelism relative to overhead.
We are interested in deriving the growth rate of $z$ compared to $p$ which keeps a fixed asymptotic efficiency;
the smaller this isoefficiency function, the more scalable the algorithm.

For a given data size $z$ (number of non-zero entries in $\mathcal{X}$), let $T_1(z)$ be the time complexity of solving non-parallel logistic regression and $T_p(z)$ be the time complexity of solving it with $p$ processors. To simplify the notation we suppress the dependence on $z$ in the remaining text. We have the relation
$p T_p = T_1 + T_0$,
where $T_0$ represents the overhead of the distributed algorithm~\cite{grama1993isoefficiency}.
Then, the speedup and efficiency are defined as
$S  = T_1 / T_p$ and $E = S/p$
respectively.
We need to ensure that the data size $z$ is defined such that it scales linearly with $T_1$.
When using an iterative solver such as {\it newGLMNET},
Lemma \ref{lem:lasso_problem_size} confirms this for $z$ being number of non-zeros.

Because $z$ can scale in various ways with the underlying dataset dimensions,
the relative growth rates of $n$ and $d$ can affect an algorithm's scaling.
We keep in mind two regimes:
(1) bounded $d$ where $z \propto n$, and
(2) $z \propto nd$.
(1) implies we get more samples,
while in (2) we get more samples and features,
with the same underlying sparsity ratio.
After distributed training, we define the support set $S$ to be the union of all non-zero features across the partitions,
and let $s=|S|$.
Our first result concerns naive averaging.

\begin{theorem}
    Given distributed sparse logistic regression with sparsity level $s$, naive averaging maintains isoefficiency if $z = \Theta(s p \log p) $. If we suppose that (1) $d$ is constant, or (2) $d(z) \to D$ for some bounded $D$, then $z = \Theta(p \log p)$. On the other hand, if $n$ and $d$ grow at the same rate such that $z \propto nd$, then the isoefficiency function is $z = \Theta(p^2 \log^2 p)$.
    \label{thm:navg_isoefficiency}
\end{theorem}

Moving on to consider OWA,
we find that the behavior not only depends on growth rates of the dataset,
as in naive averaging,
but also on certain parameters involved in the algorithm.
In particular, the second round on the subsampled set of size $n_c$ can pose a challenge to scalability
if $n_c$ grows on par with $z$,
due to the overhead of solving the objective while the other processors are idle.
In practice, $n_c$ can be kept small as $z$ grows without loss in accuracy.

\begin{theorem}
    Consider the OWA algorithm with subsampled second round dataset $\mathcal{X}_C$ with $n_c$ rows.
    If the growth rate of $n_c$ is such that $n_c \propto z^{\alpha}$ for some $0 < \alpha < 1$,
    then OWA maintains isoefficiency if
$%
z = \Theta(\max\{p^2, p^{2 / (1 - \alpha)}\})
$. %
When $n \propto z$, we have $n_c \propto n^{\alpha}$.
When $n \propto \sqrt{z}$, we have $n_c \propto n^{2\alpha}$.

\label{thm:owa_isoefficiency}
\end{theorem}

Finally, we observe that \newowa has the same scalability as OWA.

\begin{theorem}
    Consider the \newowa algorithm with subsampled second round dataset $\mathcal{X}_C$ with $n_c$ rows.
    Suppose the growth rate of $n_c$ is such that $n_c \propto z^{\alpha}$ for some $0 < \alpha < 1$.
    Then \newowa maintains isoefficiency if
$%
z = \Theta(\max\{p^2, p^{2 / (1 - \alpha)}\}).
$%
\label{thm:acowa_isoefficiency}
\end{theorem}

%% file: sections/experiments.tex
\section{Experiments}
\label{sec:experiments}

\begin{figure*}[t!]
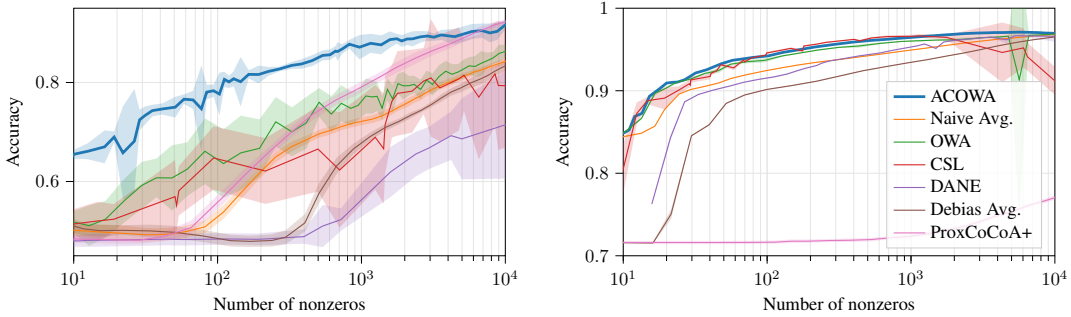

    \subfigure[{\it newsgroups}, single-node multicore, 256 partitions.]{
        \input{figs/newsgroups-no-liblinear-test-acc.tex}
    }
    \subfigure[{\it amazon7}, single-node multicore, 256 partitions.]{
        \input{figs/amazon7-no-liblinear-test-acc.tex}
    }
    \vspace*{-1.0em}
    \caption{Number of nonzeros vs. test set accuracy in the single-node parallel setting.
    \newowa has consistently better performance than other distributed methods, especially for sparser solutions on \textit{newsgroups}. It generally also performs the best on \textit{amazon7} across a range of sparsities, compared to the second best method (CSL).
    }
    \label{fig:nnz-sweep-single-node}
\end{figure*}

We conducted experiments to thoroughly compare the performance and runtime of \newowa
with competitive baselines:
standard OWA~\cite{izbicki2020distributed},
naive averaging,
ProxCoCoA+~\cite{smith2015l1},
CSL~\cite{jordan2018communication},
DANE~\cite{shamir2014communication},
and debiased averaging~\cite{zhang2012communication}.
In Appendix~\ref{app:liblinear-compare},
we also compare against {\em newGLMNET}
as provided by LIBLINEAR~\cite{fan2008liblinear}
on sufficiently small datasets;
although {\em newGLMNET} is a serial algorithm
and thus cannot scale to larger data,
we include it to show how close \newowa can get to the full-data solution.

We implemented \newowa, OWA, naive averaging, and debiased averaging in C++
with OpenMP and MPI
using the Armadillo linear algebra library~\cite{sanderson2016armadillo},
the ensmallen optimization library~\cite{curtin2021ensmallen},
and adapted parts of the mlpack machine learning library~\cite{curtin2023mlpack}.
Our OWA implementation is tuned to allow 
a higher optimizer tolerance and a larger merge set size (like \newowa).
CSL and DANE were implemented similarly,
using $\hat{w}_{\mathrm{owa}}$ as the initial solution
and OWL-QN~\cite{andrew2007scalable} from libLBFGS\footnote{See \url{https://github.com/chokkan/liblbfgs}.}
as the per-iteration solver.
To keep communication costs similar to OWA and \newowa,
we only ran one iteration of CSL and DANE.
(In our experiments, further iterations did not improve the model significantly,
and caused CSL and DANE to take much longer.)
For ProxCoCoA+ we used the Scala implementation from the authors,
which required adaptation to logistic regression.

\addtolength{\textfloatsep}{-1.0em}
\begin{figure}[t]
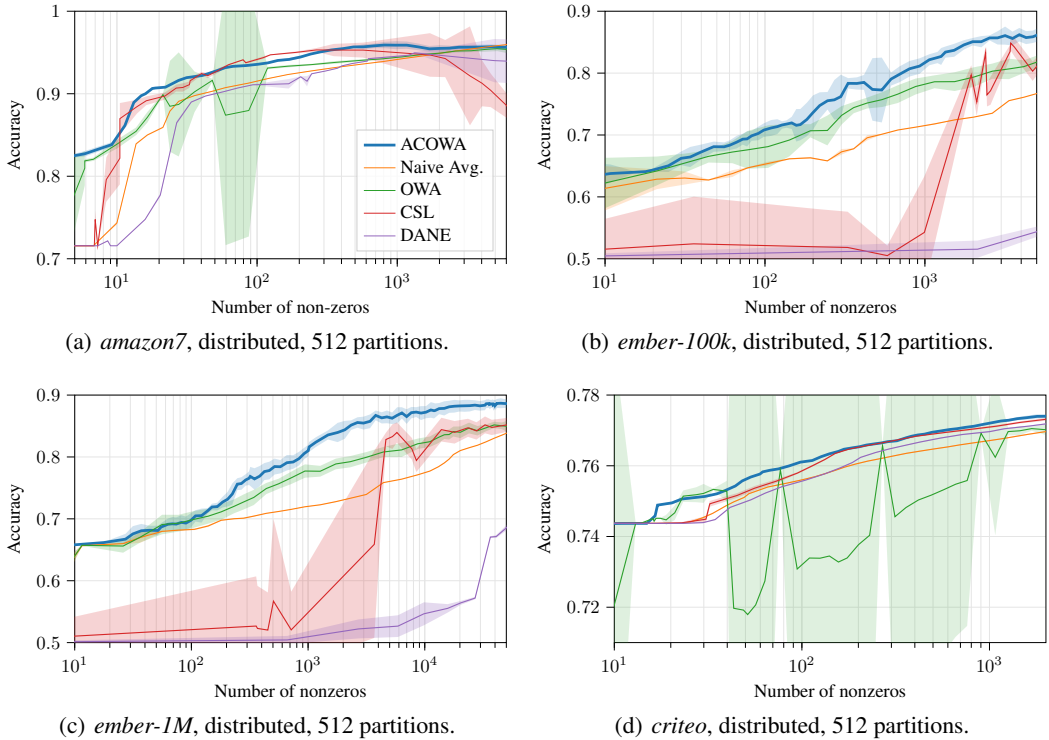

    \subfigure[{\it amazon7}, distributed, 512 partitions.]{
        \input{figs/mpi_amazon7_nnz_vs_test_acc.tex}
    }
    \subfigure[{\it ember-100k}, distributed, 512 partitions.]{
        \input{figs/ember-n8-d100k-512-test-acc.tex}
    }

    \subfigure[{\it ember-1M}, distributed, 512 partitions.]{
        \input{figs/ember-n8-d1M-512-test-acc.tex}
    }
    \subfigure[{\it criteo}, distributed, 512 partitions.]{
        \input{figs/criteo-512-test-acc.tex}
    }
    \vspace*{-1.0em}
    \caption{Number of nonzeros vs. test set accuracy in the multi-node distributed setting.  \newowa outperforms, again especially for sparser solutions.
    OWA on the criteo dataset exhibited significant variance.
    We were unable to run ProxCoCoA+ in this setting due to memory usage issues and extremely long runtimes.}
    \label{fig:nnz-sweep-distributed}
\end{figure}
\addtolength{\textfloatsep}{1.0em}

For datasets,
we used several large-scale real-world datasets
with sizes from approximately 10MB to 250GB.
All except the EMBER dataset~\cite{anderson2018ember}
are available on the LIBSVM website~\cite{chang2011libsvm}
or UCI dataset repository~\cite{ucimlrepository}.
For the EMBER dataset, we compute length-8 $n$-grams~\cite{raff2018investigation}
and keep the most common $100$k and $1$M
to produce the ember-100k and ember-1M datasets.

We are interested in two settings:
{\it (1)} single-node multicore, and
{\it (2)} fully distributed.
The first setting is relevant in modern environments,
as modern systems can have very many cores available.
In our case, we used a powerful server with 256 cores and 4TB of RAM
for our single-node experiments.
Simple parallelized solvers such as the OpenMP version of LIBLINEAR struggle in this setting,
as they were designed for only a few threads
and cannot distribute large enough work chunks
to very large numbers of cores.
Applying distributed algorithms in this context
is an effective strategy;
the algorithms operate the same as in the fully distributed setting,
but communication costs are lower as no network latency is incurred.
For our fully distributed setting, we use a cluster with 16 nodes, using up to 32 cores and 1TB of RAM on each node.

\subsection{Approximation Error}

In the first set of experiments,
we sweep over a logarithmic grid of $\lambda_1$
and compute the model's accuracy on a held-out test set,
with $\lambda_2$ set to 0.
We perform 10 trials with random partitions
and different random seeds.
Because practitioners often try to tune sparse logistic regression
to optimize accuracy at a certain level of sparsity,
we plot the number of nonzeros in the solution versus accuracy.
This gives us a good picture of how each algorithm behaves at different sparsity levels.

Figure~\ref{fig:nnz-sweep-single-node} shows the results of the $\lambda$ sweep
on smaller datasets
in the single-node multi-core environment.
\newowa (blue) consistently attains higher test set accuracy,
especially as the solution becomes more sparse
(our setting of interest).
This is also true in the fully distributed setting (Figure~\ref{fig:nnz-sweep-distributed}).
We found that CSL and DANE struggled on some datasets to produce sparse solutions;
this can be seen, for instance, on the EMBER datasets,
where the models produced by CSL and DANE only become competitive when they are dense
(not our setting of interest).

\newowa, due to the centroid augmentation and feature reweighting,
is able to identify relevant features for the full model.
This is especially true for sparser models,
where other methods struggle due to the variance of models produced by each partition.
In Appendix~\ref{app:ablation},
we perform an ablation study and show that {\em both} centroid augmentation and feature reweighting
are necessary for the improved approximation that \newowa gives.
Approximation results are similar for $\lambda_2 \ne 0$ (the elastic net); see Appendix~\ref{app:elasticnet}.
\newowa is also robust to the choice of $\beta$; see Appendix~\ref{app:beta_comparison}.

\subsection{Runtime}

In the second set of experiments,
we characterize the runtime of \newowa.
We expect \newowa to be slower than other one-shot algorithms we compare against,
as we chose to increase the amount of communication modestly
in order to achieve significant improvements in model performance.
We tune $\lambda_1$ to produce approximately $1000$ nonzeros in the final model,
and take $\lambda_2 = 0$.
We record the time taken to learn the model
(excluding data loading and unrelated preprocessing).

\addtolength{\textfloatsep}{-3.5em}
\begin{table}[!t]
    \centering
    \subfigure[single-node parallel]{
    \begin{tabular}{@{}lcccrrrrr@{}}
    \toprule
    \multicolumn{1}{c}{\textbf{dataset}} & \multicolumn{1}{c}{$n$} & \multicolumn{1}{c}{$d$} & \multicolumn{1}{c}{\textbf{nnz}} & \multicolumn{1}{c}{ProxCoCoA+} & \multicolumn{1}{c}{Naive Avg.} & \multicolumn{1}{c}{OWA} & \multicolumn{1}{c}{\newowa} \\ \midrule
    newsgroups & 11k & 54k & 1.5M & 40.129s & 0.226s & 0.242s & 2.154s \\
    amazon7 & 1.3M & 262k & 133M & 531.778s & 2.356s & 14.933s & 31.675s \\
    criteo & 45M & 1M & 1.78B & \textit{Fail} & 17.085s & 218.092s & 264.200s \\
    ember-100k & 600k & 100k & 8.48B & \textit{Fail} & 7.811s & 13.245s & 80.814s \\ \midrule
    \bottomrule
    \end{tabular}
    }
    \subfigure[fully distributed]{
    \begin{tabular}{@{}lcccrrrrr@{}}
    \toprule
    \multicolumn{1}{c}{\textbf{dataset}} & \multicolumn{1}{c}{$n$} & \multicolumn{1}{c}{$d$} & \multicolumn{1}{c}{\textbf{nnz}} & \multicolumn{1}{c}{CSL} & \multicolumn{1}{c}{DANE} & \multicolumn{1}{c}{Naive Avg.} & \multicolumn{1}{c}{OWA} & \multicolumn{1}{c}{\newowa} \\ \midrule
    ember-100k & 600k & 100k & 8.48B & 28.831s & 40.318s & 0.863s & 1.129s & 19.147s \\
    ember-1M & 600k & 1M & 38.0B & 81.634s & 68.975s & 6.176s & 5.836s & 145.051s \\
    criteo & 45M & 1M & 1.78B & 36.069s & 65.618s & 2.349s & 25.885s & 150.383s \\
    \bottomrule
    \end{tabular}%
    }
    \caption{Runtime results for different techniques.  {\it Fail} indicates the method took over two hours or had an out-of-memory issue.  Although \newowa takes longer to converge than naive averaging and OWA, it provides significantly better performance (see Fig.~\ref{fig:nnz-sweep-single-node}). This also generally holds when comparing with CSL and DANE.}
    \label{tab:runtimes}
\end{table}
\addtolength{\textfloatsep}{3.5em}

Results are shown for each dataset in Table~\ref{tab:runtimes}.
These results match expectations:
\newowa is slower than the other one-shot algorithms,
because it involves an additional round of communication,
plus the initial communication of the centroids.
ProxCoCoA+ is unable to complete within two hours for many datasets;
we believe this to be a result of high communication overhead.
In the distributed setting,
the increased complexity of solving the surrogate loss function for CSL
(and similar for DANE)
causes slowdowns.
As mentioned earlier,
in our experiments we only use one iteration of CSL and DANE.
Were we to run those algorithms to convergence,
the runtime (and the communication costs)
would be substantially higher.

In Appendix~\ref{app:runtime-breakdown},
we conduct a detailed breakdown of the runtime of each step of \newowa.

%% file: figs/mpi_amazon7_nnz_vs_test_acc.tex
\begin{tikzpicture}[scale=0.7]

\definecolor{crimson2143940}{RGB}{214,39,40}
\definecolor{darkorange25512714}{RGB}{255,127,14}
\definecolor{forestgreen4416044}{RGB}{44,160,44}
\definecolor{gainsboro229}{RGB}{229,229,229}
\definecolor{mediumpurple148103189}{RGB}{148,103,189}
\definecolor{steelblue31119180}{RGB}{31,119,180}
\definecolor{lightgray204}{RGB}{204,204,204}

\begin{axis}[
width=0.7\textwidth,
height=0.45\textwidth,
legend cell align={left},
legend style={
  fill opacity=0.8,
  draw opacity=1,
  text opacity=1,
  at={(0.97,0.03)},
  anchor=south east,
  draw=lightgray204
},
log basis x={10},
tick align=outside,
tick pos=left,
x grid style={gainsboro229},
xlabel={Number of non-zeros},
xmajorgrids,
xmin=5, xmax=6000,
xminorgrids,
xmode=log,
xtick style={color=black},
y grid style={gainsboro229},
ylabel={Accuracy},
ymajorgrids,
ymin=0.7, ymax=1,
yminorgrids,
ytick style={color=black}
]
\path [fill=steelblue31119180, fill opacity=0.2]
(axis cs:1.9,0.71603545118715)
--(axis cs:1.9,0.71543854881285)
--(axis cs:1.9,0.71543854881285)
--(axis cs:2.4,0.70423942730076)
--(axis cs:2.9,0.814317805512623)
--(axis cs:2.9,0.814353995976352)
--(axis cs:2.9,0.81434810864245)
--(axis cs:2.9,0.814352843350764)
--(axis cs:4.4,0.821862115797791)
--(axis cs:5.9,0.823890329404942)
--(axis cs:7,0.828340673407221)
--(axis cs:9.2,0.835436115514646)
--(axis cs:11.4,0.859173274508626)
--(axis cs:13.2,0.887985875659434)
--(axis cs:14.9,0.896757624197768)
--(axis cs:16,0.899017096455612)
--(axis cs:17.9,0.904530443778656)
--(axis cs:18.8,0.905827199673861)
--(axis cs:22.2,0.907293629544425)
--(axis cs:27.6,0.914654123445727)
--(axis cs:31.4,0.918407004694119)
--(axis cs:40.6,0.920736065889301)
--(axis cs:51.6,0.926573631708644)
--(axis cs:59.2,0.930069032336894)
--(axis cs:67.4,0.931851000601521)
--(axis cs:83,0.932605129870567)
--(axis cs:101.1,0.934563328488445)
--(axis cs:119.4,0.936288190863966)
--(axis cs:140.8,0.93913663234478)
--(axis cs:169.4,0.939913104777352)
--(axis cs:204.9,0.942899910880902)
--(axis cs:241.5,0.947320595588943)
--(axis cs:281.6,0.949742239064542)
--(axis cs:337.8,0.952355236592607)
--(axis cs:412.6,0.953765664456537)
--(axis cs:486.7,0.955814936578705)
--(axis cs:574.4,0.953782758192987)
--(axis cs:685.2,0.955275889046553)
--(axis cs:810.7,0.955191662429342)
--(axis cs:977.7,0.953973266966211)
--(axis cs:1168.8,0.954460319170161)
--(axis cs:1406.9,0.952759199699167)
--(axis cs:1695.6,0.950938141982972)
--(axis cs:2029.9,0.95103769961658)
--(axis cs:2365.9,0.952099253953886)
--(axis cs:2438.3,0.953119841081568)
--(axis cs:2757.4,0.95383226936398)
--(axis cs:3222.6,0.953501992469658)
--(axis cs:3920.6,0.953619233404792)
--(axis cs:4713.4,0.954085471900776)
--(axis cs:5474.6,0.953359244805406)
--(axis cs:6181.8,0.953356558912566)
--(axis cs:6558.6,0.952820217618567)
--(axis cs:7186.2,0.952860450703186)
--(axis cs:8315.9,0.952590818769614)
--(axis cs:9677.3,0.951411351226885)
--(axis cs:11757.3,0.951871532009206)
--(axis cs:14938.2,0.950773385254837)
--(axis cs:19512,0.949095528263853)
--(axis cs:22777.1,0.949081740358328)
--(axis cs:23940.4,0.947686741226419)
--(axis cs:24885.1,0.947790592053574)
--(axis cs:28206.4,0.947169151571034)
--(axis cs:32613.5,0.946348159160413)
--(axis cs:36300.3,0.945522668675592)
--(axis cs:39919.1,0.944642198917901)
--(axis cs:43898.1,0.943697675627965)
--(axis cs:48399,0.942862095169416)
--(axis cs:53809.6,0.941531948931789)
--(axis cs:62655.5,0.939631466835372)
--(axis cs:80750.3,0.935086433321698)
--(axis cs:98870.3,0.929900066727174)
--(axis cs:112273.5,0.926621973374787)
--(axis cs:124459.4,0.925588040728029)
--(axis cs:136975.9,0.921415143842998)
--(axis cs:149834.2,0.919403451822939)
--(axis cs:162584.5,0.917395858956586)
--(axis cs:174627,0.915076249307324)
--(axis cs:185226.2,0.913117829402229)
--(axis cs:194272.9,0.911060298308366)
--(axis cs:194272.9,0.917593101691634)
--(axis cs:194272.9,0.917593101691634)
--(axis cs:185226.2,0.919490370597771)
--(axis cs:174627,0.921161750692675)
--(axis cs:162584.5,0.923355741043414)
--(axis cs:149834.2,0.925343948177061)
--(axis cs:136975.9,0.927305056157002)
--(axis cs:124459.4,0.931313759271971)
--(axis cs:112273.5,0.931973026625213)
--(axis cs:98870.3,0.936051733272826)
--(axis cs:80750.3,0.942153366678301)
--(axis cs:62655.5,0.944999133164628)
--(axis cs:53809.6,0.946980051068211)
--(axis cs:48399,0.948422504830584)
--(axis cs:43898.1,0.949607524372035)
--(axis cs:39919.1,0.950524801082099)
--(axis cs:36300.3,0.951150331324409)
--(axis cs:32613.5,0.951580840839587)
--(axis cs:28206.4,0.952098848428966)
--(axis cs:24885.1,0.952665607946426)
--(axis cs:23940.4,0.952608058773581)
--(axis cs:22777.1,0.953522859641672)
--(axis cs:19512,0.954706671736147)
--(axis cs:14938.2,0.955822814745163)
--(axis cs:11757.3,0.957356067990794)
--(axis cs:9677.3,0.957850648773115)
--(axis cs:8315.9,0.958408181230386)
--(axis cs:7186.2,0.958937949296814)
--(axis cs:6558.6,0.959177382381433)
--(axis cs:6181.8,0.959265841087434)
--(axis cs:5474.6,0.959459555194594)
--(axis cs:4713.4,0.959876328099224)
--(axis cs:3920.6,0.959605566595208)
--(axis cs:3222.6,0.959286007530342)
--(axis cs:2757.4,0.95886773063602)
--(axis cs:2438.3,0.957861558918432)
--(axis cs:2365.9,0.958292946046114)
--(axis cs:2029.9,0.95889390038342)
--(axis cs:1695.6,0.957902458017028)
--(axis cs:1406.9,0.960870000300833)
--(axis cs:1168.8,0.963310280829839)
--(axis cs:977.7,0.963732533033789)
--(axis cs:810.7,0.962896737570658)
--(axis cs:685.2,0.960879310953447)
--(axis cs:574.4,0.959957441807014)
--(axis cs:486.7,0.957603463421295)
--(axis cs:412.6,0.955726335543464)
--(axis cs:337.8,0.953771363407393)
--(axis cs:281.6,0.951458760935458)
--(axis cs:241.5,0.948965204411057)
--(axis cs:204.9,0.946268289119098)
--(axis cs:169.4,0.943305495222648)
--(axis cs:140.8,0.94157116765522)
--(axis cs:119.4,0.938489009136034)
--(axis cs:101.1,0.936822071511555)
--(axis cs:83,0.935637070129433)
--(axis cs:67.4,0.934592199398479)
--(axis cs:59.2,0.933139967663106)
--(axis cs:51.6,0.929127968291356)
--(axis cs:40.6,0.923400334110699)
--(axis cs:31.4,0.920560995305881)
--(axis cs:27.6,0.917751276554273)
--(axis cs:22.2,0.908604770455575)
--(axis cs:18.8,0.908011000326139)
--(axis cs:17.9,0.908232356221344)
--(axis cs:16,0.901028103544388)
--(axis cs:14.9,0.898391175802232)
--(axis cs:13.2,0.890896724340566)
--(axis cs:11.4,0.863615325491374)
--(axis cs:9.2,0.841307684485354)
--(axis cs:7,0.834732326592779)
--(axis cs:5.9,0.830593670595058)
--(axis cs:4.4,0.824774884202209)
--(axis cs:2.9,0.816066956649236)
--(axis cs:2.9,0.81606649135755)
--(axis cs:2.9,0.816073004023648)
--(axis cs:2.9,0.816025194487377)
--(axis cs:2.4,0.80643757269924)
--(axis cs:1.9,0.71603545118715)
--(axis cs:1.9,0.71603545118715)
--cycle;

\path [fill=darkorange25512714, fill opacity=0.2]
(axis cs:1,0.71603545118715)
--(axis cs:1,0.71543854881285)
--(axis cs:1,0.71543854881285)
--(axis cs:1,0.71543854881285)
--(axis cs:1,0.71543854881285)
--(axis cs:1,0.71543854881285)
--(axis cs:1,0.71543854881285)
--(axis cs:1.4,0.71543854881285)
--(axis cs:1.9,0.715439767265034)
--(axis cs:1.9,0.715439767265034)
--(axis cs:1.9,0.715439767265034)
--(axis cs:1.9,0.715439767265034)
--(axis cs:2.9,0.715439767265034)
--(axis cs:5.9,0.715441620368576)
--(axis cs:5.9,0.715441620368576)
--(axis cs:5.9,0.715441620368576)
--(axis cs:5.9,0.715441620368576)
--(axis cs:5.9,0.715441620368576)
--(axis cs:6.8,0.715441620368576)
--(axis cs:10,0.741592782132641)
--(axis cs:13.7,0.838225929735506)
--(axis cs:16.5,0.849146648406168)
--(axis cs:21.3,0.85838154761574)
--(axis cs:23.9,0.878073547440342)
--(axis cs:27.6,0.890049236013311)
--(axis cs:35.8,0.897180686588475)
--(axis cs:47.9,0.902395432516599)
--(axis cs:59.6,0.906944276990219)
--(axis cs:87.3,0.912432835090159)
--(axis cs:118.7,0.91781486242074)
--(axis cs:166.3,0.922930995552983)
--(axis cs:244.8,0.927225931519828)
--(axis cs:359,0.931246026001756)
--(axis cs:531.9,0.934969432821178)
--(axis cs:784.9,0.938469608188759)
--(axis cs:1090.4,0.941977744214656)
--(axis cs:1506.9,0.945271612019271)
--(axis cs:2012.3,0.948579737346392)
--(axis cs:2612.3,0.951421590454878)
--(axis cs:3360.1,0.953730688612897)
--(axis cs:4222.5,0.955976966289796)
--(axis cs:5167.4,0.957807027965889)
--(axis cs:6177.2,0.95926632940869)
--(axis cs:7363.6,0.960619883462697)
--(axis cs:8602,0.961650165332503)
--(axis cs:9858,0.962712713234921)
--(axis cs:11612.5,0.963507651309002)
--(axis cs:13347.3,0.964456185891243)
--(axis cs:15032.3,0.96531264328365)
--(axis cs:16870.9,0.965856036201892)
--(axis cs:19041.6,0.966458356299566)
--(axis cs:21530.7,0.96670713716579)
--(axis cs:24358.7,0.966686891916825)
--(axis cs:27416.9,0.966403855188214)
--(axis cs:30324.3,0.96618727988758)
--(axis cs:32761.5,0.965918586399181)
--(axis cs:33827.6,0.966126638849798)
--(axis cs:34281,0.966378516757114)
--(axis cs:37416.6,0.966054207302583)
--(axis cs:42481.5,0.965285226050331)
--(axis cs:49031,0.964381517484229)
--(axis cs:57561.9,0.963155151390005)
--(axis cs:71534.8,0.960527734947479)
--(axis cs:83569,0.957514680443512)
--(axis cs:91241.4,0.955853400185758)
--(axis cs:94944.8,0.955701994556109)
--(axis cs:95011.4,0.956270407355422)
--(axis cs:98689.3,0.955696630911241)
--(axis cs:107836.6,0.954260393021221)
--(axis cs:118930.6,0.952417611374759)
--(axis cs:127220.6,0.951045694148635)
--(axis cs:135441.2,0.949714220200625)
--(axis cs:144103.3,0.948295368175634)
--(axis cs:153328.7,0.946582096869873)
--(axis cs:165106.4,0.9444111615369)
--(axis cs:187037.9,0.94010591331795)
--(axis cs:203752.8,0.935659747162624)
--(axis cs:211870.3,0.933446925175322)
--(axis cs:216838.4,0.932216828092637)
--(axis cs:220710.9,0.93125137757599)
--(axis cs:220710.9,0.93193722242401)
--(axis cs:220710.9,0.93193722242401)
--(axis cs:216838.4,0.932889771907363)
--(axis cs:211870.3,0.934081074824678)
--(axis cs:203752.8,0.936321452837376)
--(axis cs:187037.9,0.94070108668205)
--(axis cs:165106.4,0.944941838463099)
--(axis cs:153328.7,0.947135503130127)
--(axis cs:144103.3,0.948785031824366)
--(axis cs:135441.2,0.950217979799375)
--(axis cs:127220.6,0.951481305851365)
--(axis cs:118930.6,0.952860588625241)
--(axis cs:107836.6,0.954744406978779)
--(axis cs:98689.3,0.956244369088759)
--(axis cs:95011.4,0.956832192644577)
--(axis cs:94944.8,0.956438605443891)
--(axis cs:91241.4,0.956565199814242)
--(axis cs:83569,0.958200519556488)
--(axis cs:71534.8,0.961395065052521)
--(axis cs:57561.9,0.963778848609995)
--(axis cs:49031,0.965091282515771)
--(axis cs:42481.5,0.966156973949669)
--(axis cs:37416.6,0.966873392697417)
--(axis cs:34281,0.967266083242886)
--(axis cs:33827.6,0.967054361150202)
--(axis cs:32761.5,0.966821813600819)
--(axis cs:30324.3,0.96701592011242)
--(axis cs:27416.9,0.967360344811786)
--(axis cs:24358.7,0.967555508083175)
--(axis cs:21530.7,0.96770446283421)
--(axis cs:19041.6,0.967404843700434)
--(axis cs:16870.9,0.966818963798108)
--(axis cs:15032.3,0.96627675671635)
--(axis cs:13347.3,0.965390214108757)
--(axis cs:11612.5,0.964455148690998)
--(axis cs:9858,0.963557686765079)
--(axis cs:8602,0.962611234667497)
--(axis cs:7363.6,0.961498116537303)
--(axis cs:6177.2,0.96021787059131)
--(axis cs:5167.4,0.958732572034111)
--(axis cs:4222.5,0.956964633710203)
--(axis cs:3360.1,0.954736711387103)
--(axis cs:2612.3,0.952273809545123)
--(axis cs:2012.3,0.949496462653608)
--(axis cs:1506.9,0.946201587980729)
--(axis cs:1090.4,0.943023255785345)
--(axis cs:784.9,0.939410591811241)
--(axis cs:531.9,0.935869767178822)
--(axis cs:359,0.932187773998244)
--(axis cs:244.8,0.928159668480172)
--(axis cs:166.3,0.923700804447017)
--(axis cs:118.7,0.91871533757926)
--(axis cs:87.3,0.913440964909841)
--(axis cs:59.6,0.907917123009781)
--(axis cs:47.9,0.903573567483401)
--(axis cs:35.8,0.898266913411525)
--(axis cs:27.6,0.891535563986689)
--(axis cs:23.9,0.879848852559658)
--(axis cs:21.3,0.86009545238426)
--(axis cs:16.5,0.850878151593832)
--(axis cs:13.7,0.840203670264494)
--(axis cs:10,0.745173817867359)
--(axis cs:6.8,0.716035979631424)
--(axis cs:5.9,0.716035979631424)
--(axis cs:5.9,0.716035979631424)
--(axis cs:5.9,0.716035979631424)
--(axis cs:5.9,0.716035979631424)
--(axis cs:5.9,0.716035979631424)
--(axis cs:2.9,0.716035832734966)
--(axis cs:1.9,0.716035832734966)
--(axis cs:1.9,0.716035832734966)
--(axis cs:1.9,0.716035832734966)
--(axis cs:1.9,0.716035832734966)
--(axis cs:1.4,0.71603545118715)
--(axis cs:1,0.71603545118715)
--(axis cs:1,0.71603545118715)
--(axis cs:1,0.71603545118715)
--(axis cs:1,0.71603545118715)
--(axis cs:1,0.71603545118715)
--(axis cs:1,0.71603545118715)
--cycle;

\path [fill=forestgreen4416044, fill opacity=0.2]
(axis cs:1,0.716035832734966)
--(axis cs:1,0.715439767265034)
--(axis cs:1,0.715439767265034)
--(axis cs:1,0.715439767265034)
--(axis cs:1,0.715439767265034)
--(axis cs:1,0.715439767265034)
--(axis cs:1,0.715439767265034)
--(axis cs:1.2,0.715439767265034)
--(axis cs:1.9,0.715439767265034)
--(axis cs:1.9,0.715439767265034)
--(axis cs:1.9,0.5494284912082)
--(axis cs:1.9,0.5494284912082)
--(axis cs:2.9,0.558951761659767)
--(axis cs:5.9,0.788910162972054)
--(axis cs:5.9,0.809896531260603)
--(axis cs:5.9,0.811936106672626)
--(axis cs:5.9,0.813478417774283)
--(axis cs:5.9,0.81676242706554)
--(axis cs:6.8,0.818618734629549)
--(axis cs:10,0.833468907371768)
--(axis cs:13.7,0.849777125673134)
--(axis cs:16.5,0.864979449402414)
--(axis cs:21.3,0.897932154154537)
--(axis cs:23.9,0.829751468675853)
--(axis cs:27.6,0.823942579415336)
--(axis cs:35.8,0.877689699891877)
--(axis cs:47.9,0.914155781702759)
--(axis cs:59.6,0.716454525855463)
--(axis cs:87.3,0.726959741661271)
--(axis cs:118.7,0.929631532394818)
--(axis cs:166.3,0.931891977536882)
--(axis cs:244.8,0.934339929986731)
--(axis cs:359,0.936224089423915)
--(axis cs:531.9,0.938949633085932)
--(axis cs:784.9,0.941478902437811)
--(axis cs:1090.4,0.943495357689519)
--(axis cs:1506.9,0.946462253872638)
--(axis cs:2012.3,0.947537456194095)
--(axis cs:2612.3,0.949950619163139)
--(axis cs:3360.1,0.950422961468309)
--(axis cs:4222.5,0.950852330782912)
--(axis cs:5167.4,0.951629065284348)
--(axis cs:6177.2,0.948961558396502)
--(axis cs:7363.6,0.949728767441141)
--(axis cs:8602,0.946182366298552)
--(axis cs:9858,0.947362343986301)
--(axis cs:11612.5,0.946393867925541)
--(axis cs:13347.3,0.947211900321256)
--(axis cs:15032.3,0.947140181738919)
--(axis cs:16870.9,0.948742177318243)
--(axis cs:19041.6,0.949990032205902)
--(axis cs:21530.7,0.949886811122671)
--(axis cs:24358.7,0.950609721159103)
--(axis cs:27416.9,0.950880304649834)
--(axis cs:30324.3,0.951199236084678)
--(axis cs:32761.5,0.950947546784614)
--(axis cs:33827.6,0.951799910264384)
--(axis cs:34281,0.950736931273274)
--(axis cs:37416.6,0.95056826124693)
--(axis cs:42481.5,0.94963187447844)
--(axis cs:49031,0.949321218638316)
--(axis cs:57561.9,0.947848931397898)
--(axis cs:71534.8,0.945359803333154)
--(axis cs:83569,0.941922196145679)
--(axis cs:91241.4,0.939161141693629)
--(axis cs:94944.8,0.939599349060773)
--(axis cs:95011.4,0.938333876700243)
--(axis cs:98689.3,0.937083461180183)
--(axis cs:107836.6,0.935487465622217)
--(axis cs:118930.6,0.932909238573354)
--(axis cs:127220.6,0.931251102482322)
--(axis cs:135441.2,0.929219300601338)
--(axis cs:144103.3,0.927407310241693)
--(axis cs:153328.7,0.925123553308334)
--(axis cs:165106.4,0.923347966316282)
--(axis cs:187037.9,0.918584895466187)
--(axis cs:203752.8,0.91269002921004)
--(axis cs:211870.3,0.909095104506383)
--(axis cs:216838.4,0.907047764419619)
--(axis cs:220710.9,0.905963641652609)
--(axis cs:220710.9,0.918881558347391)
--(axis cs:220710.9,0.918881558347391)
--(axis cs:216838.4,0.920146435580381)
--(axis cs:211870.3,0.921814095493617)
--(axis cs:203752.8,0.92482677078996)
--(axis cs:187037.9,0.933111304533813)
--(axis cs:165106.4,0.933923633683718)
--(axis cs:153328.7,0.935347646691666)
--(axis cs:144103.3,0.937470089758307)
--(axis cs:135441.2,0.938385899398662)
--(axis cs:127220.6,0.940286097517677)
--(axis cs:118930.6,0.941630361426646)
--(axis cs:107836.6,0.943172534377783)
--(axis cs:98689.3,0.944665938819817)
--(axis cs:95011.4,0.945908323299757)
--(axis cs:94944.8,0.945878450939227)
--(axis cs:91241.4,0.946890658306371)
--(axis cs:83569,0.950125403854321)
--(axis cs:71534.8,0.952717196666846)
--(axis cs:57561.9,0.954926868602103)
--(axis cs:49031,0.955825781361684)
--(axis cs:42481.5,0.95638392552156)
--(axis cs:37416.6,0.95648593875307)
--(axis cs:34281,0.956477868726726)
--(axis cs:33827.6,0.956325689735616)
--(axis cs:32761.5,0.957066653215386)
--(axis cs:30324.3,0.956971763915322)
--(axis cs:27416.9,0.956481695350167)
--(axis cs:24358.7,0.956694078840897)
--(axis cs:21530.7,0.956817988877329)
--(axis cs:19041.6,0.955672567794098)
--(axis cs:16870.9,0.954536822681757)
--(axis cs:15032.3,0.954323418261081)
--(axis cs:13347.3,0.954713699678744)
--(axis cs:11612.5,0.954654332074459)
--(axis cs:9858,0.955870656013699)
--(axis cs:8602,0.957456033701448)
--(axis cs:7363.6,0.957454432558859)
--(axis cs:6177.2,0.958153241603498)
--(axis cs:5167.4,0.957689734715653)
--(axis cs:4222.5,0.956481469217088)
--(axis cs:3360.1,0.953774238531691)
--(axis cs:2612.3,0.953531780836861)
--(axis cs:2012.3,0.950822343805905)
--(axis cs:1506.9,0.948662946127362)
--(axis cs:1090.4,0.946040842310481)
--(axis cs:784.9,0.943126097562189)
--(axis cs:531.9,0.941617366914068)
--(axis cs:359,0.939644710576085)
--(axis cs:244.8,0.936281870013268)
--(axis cs:166.3,0.934897222463118)
--(axis cs:118.7,0.932578267605182)
--(axis cs:87.3,1.03245765833873)
--(axis cs:59.6,1.03155447414454)
--(axis cs:47.9,0.918161018297241)
--(axis cs:35.8,0.926923900108123)
--(axis cs:27.6,0.950160620584664)
--(axis cs:23.9,0.939832531324147)
--(axis cs:21.3,0.899865845845463)
--(axis cs:16.5,0.877882750597586)
--(axis cs:13.7,0.859066274326865)
--(axis cs:10,0.844171292628232)
--(axis cs:6.8,0.822565265370451)
--(axis cs:5.9,0.82056157293446)
--(axis cs:5.9,0.818283782225717)
--(axis cs:5.9,0.818618293327374)
--(axis cs:5.9,0.816642868739397)
--(axis cs:5.9,0.830616237027946)
--(axis cs:2.9,0.797165038340233)
--(axis cs:1.9,0.8021123087918)
--(axis cs:1.9,0.8021123087918)
--(axis cs:1.9,0.716035832734966)
--(axis cs:1.9,0.716035832734966)
--(axis cs:1.2,0.716035832734966)
--(axis cs:1,0.716035832734966)
--(axis cs:1,0.716035832734966)
--(axis cs:1,0.716035832734966)
--(axis cs:1,0.716035832734966)
--(axis cs:1,0.716035832734966)
--(axis cs:1,0.716035832734966)
--cycle;

\path [fill=crimson2143940, fill opacity=0.2]
(axis cs:1,0.716035832734966)
--(axis cs:1,0.715439767265034)
--(axis cs:1,0.715439767265034)
--(axis cs:1,0.715439767265034)
--(axis cs:1,0.715439767265034)
--(axis cs:1,0.715439767265034)
--(axis cs:1,0.715439767265034)
--(axis cs:1,0.715439767265034)
--(axis cs:1,0.715439767265034)
--(axis cs:1,0.715439767265034)
--(axis cs:1,0.715439767265034)
--(axis cs:1,0.715439767265034)
--(axis cs:2.7,0.712546114950572)
--(axis cs:4.8,0.715441620368576)
--(axis cs:4.9,0.715439767265034)
--(axis cs:5.3,0.715439767265034)
--(axis cs:6.4,0.715439767265034)
--(axis cs:6.9,0.715439767265034)
--(axis cs:7,0.73682238296495)
--(axis cs:7.3,0.715439767265034)
--(axis cs:8.4,0.769139038262412)
--(axis cs:10.5,0.790150925180048)
--(axis cs:10.5,0.850884363061027)
--(axis cs:13.9,0.875931938499349)
--(axis cs:16.2,0.886174820055348)
--(axis cs:22,0.892833299823863)
--(axis cs:26.5,0.9026986457722)
--(axis cs:29.5,0.90246535666211)
--(axis cs:32.9,0.907128779044266)
--(axis cs:33.3,0.911791974688164)
--(axis cs:40.2,0.922517039500629)
--(axis cs:44.5,0.918817890081579)
--(axis cs:53.5,0.926007090640371)
--(axis cs:57.1,0.929837358062289)
--(axis cs:62.5,0.934253533677291)
--(axis cs:79.5,0.940452935059425)
--(axis cs:83.1,0.936334884852018)
--(axis cs:111.2,0.942396106175022)
--(axis cs:124.4,0.946554787283136)
--(axis cs:189.8,0.949024769626248)
--(axis cs:264.5,0.950817413437417)
--(axis cs:351.9,0.945520216481281)
--(axis cs:569.8,0.943815630534435)
--(axis cs:1711.4,0.920575127781031)
--(axis cs:2198.8,0.920131406139017)
--(axis cs:3301.9,0.86166631915938)
--(axis cs:4020.2,0.884774947585819)
--(axis cs:4642,0.883753372946883)
--(axis cs:5955,0.871397935057073)
--(axis cs:6826,0.867351077036784)
--(axis cs:7549.6,0.858574553392354)
--(axis cs:7775.8,0.853655945283859)
--(axis cs:7821.8,0.860156055210895)
--(axis cs:8116,0.86709905399176)
--(axis cs:8241.7,0.875228871171959)
--(axis cs:9334.4,0.877092762809609)
--(axis cs:10154.4,0.882065026942343)
--(axis cs:12229.4,0.884131402058042)
--(axis cs:13488.1,0.895436611419244)
--(axis cs:15524.3,0.900373458195162)
--(axis cs:18523.9,0.910826804300405)
--(axis cs:22297,0.916199450486103)
--(axis cs:26586.3,0.924087521502625)
--(axis cs:32288,0.916018436513868)
--(axis cs:38817,0.91711017873279)
--(axis cs:45093.9,0.921098132548289)
--(axis cs:50417.8,0.933336000786075)
--(axis cs:62466.5,0.937036489085956)
--(axis cs:68751,0.924402479698778)
--(axis cs:75240.2,0.919273370430166)
--(axis cs:92823.6,0.915032504154615)
--(axis cs:101364.1,0.917240632903647)
--(axis cs:107341.8,0.915431852930318)
--(axis cs:112758.4,0.913244616351657)
--(axis cs:117934.4,0.910630643345692)
--(axis cs:123601.2,0.922288664217667)
--(axis cs:130435.6,0.905206516872243)
--(axis cs:135357.9,0.899644153773729)
--(axis cs:139103.1,0.898401117997245)
--(axis cs:142563.6,0.881107891309968)
--(axis cs:142563.6,0.929210908690032)
--(axis cs:142563.6,0.929210908690032)
--(axis cs:139103.1,0.933123082002756)
--(axis cs:135357.9,0.922565846226271)
--(axis cs:130435.6,0.920341483127757)
--(axis cs:123601.2,0.952388935782333)
--(axis cs:117934.4,0.942609556654308)
--(axis cs:112758.4,0.929109183648344)
--(axis cs:107341.8,0.938281947069682)
--(axis cs:101364.1,0.949631167096353)
--(axis cs:92823.6,0.952945095845386)
--(axis cs:75240.2,0.943815829569833)
--(axis cs:68751,0.951285720301222)
--(axis cs:62466.5,0.954261910914044)
--(axis cs:50417.8,0.946451799213925)
--(axis cs:45093.9,0.947225467451711)
--(axis cs:38817,0.94431222126721)
--(axis cs:32288,0.948223363486132)
--(axis cs:26586.3,0.942484478497375)
--(axis cs:22297,0.935336349513897)
--(axis cs:18523.9,0.932839195699595)
--(axis cs:15524.3,0.932675741804838)
--(axis cs:13488.1,0.924922788580755)
--(axis cs:12229.4,0.925114797941958)
--(axis cs:10154.4,0.914842773057657)
--(axis cs:9334.4,0.916361037190391)
--(axis cs:8241.7,0.902114528828041)
--(axis cs:8116,0.90960334600824)
--(axis cs:7821.8,0.894767344789105)
--(axis cs:7775.8,0.897771054716141)
--(axis cs:7549.6,0.881371646607645)
--(axis cs:6826,0.883705322963216)
--(axis cs:5955,0.901610664942927)
--(axis cs:4642,0.924457027053117)
--(axis cs:4020.2,0.940754252414181)
--(axis cs:3301.9,0.98159908084062)
--(axis cs:2198.8,0.965157793860983)
--(axis cs:1711.4,0.973409272218969)
--(axis cs:569.8,0.961947369465565)
--(axis cs:351.9,0.96038258351872)
--(axis cs:264.5,0.955345386562583)
--(axis cs:189.8,0.951227030373752)
--(axis cs:124.4,0.947750412716864)
--(axis cs:111.2,0.945298293824978)
--(axis cs:83.1,0.938829915147982)
--(axis cs:79.5,0.941827264940575)
--(axis cs:62.5,0.936033466322709)
--(axis cs:57.1,0.933167841937711)
--(axis cs:53.5,0.931253509359629)
--(axis cs:44.5,0.924991709918421)
--(axis cs:40.2,0.926533560499371)
--(axis cs:33.3,0.919630625311836)
--(axis cs:32.9,0.914866220955734)
--(axis cs:29.5,0.91268404333789)
--(axis cs:26.5,0.9110475542278)
--(axis cs:22,0.901576500176137)
--(axis cs:16.2,0.896801779944652)
--(axis cs:13.9,0.892694261500651)
--(axis cs:10.5,0.888730636938973)
--(axis cs:10.5,0.853430674819952)
--(axis cs:8.4,0.822698961737588)
--(axis cs:7.3,0.716035832734966)
--(axis cs:7,0.75855021703505)
--(axis cs:6.9,0.716035832734966)
--(axis cs:6.4,0.716035832734966)
--(axis cs:5.3,0.716035832734966)
--(axis cs:4.9,0.716035832734966)
--(axis cs:4.8,0.716035979631424)
--(axis cs:2.7,0.721809885049428)
--(axis cs:1,0.716035832734966)
--(axis cs:1,0.716035832734966)
--(axis cs:1,0.716035832734966)
--(axis cs:1,0.716035832734966)
--(axis cs:1,0.716035832734966)
--(axis cs:1,0.716035832734966)
--(axis cs:1,0.716035832734966)
--(axis cs:1,0.716035832734966)
--(axis cs:1,0.716035832734966)
--(axis cs:1,0.716035832734966)
--(axis cs:1,0.716035832734966)
--cycle;

\path [fill=mediumpurple148103189, fill opacity=0.2]
(axis cs:1,0.716035832734966)
--(axis cs:1,0.715439767265034)
--(axis cs:1,0.715439767265034)
--(axis cs:1,0.715439767265034)
--(axis cs:1,0.715439767265034)
--(axis cs:1,0.715439767265034)
--(axis cs:1,0.715439767265034)
--(axis cs:1,0.715439767265034)
--(axis cs:1,0.715439767265034)
--(axis cs:1,0.715439767265034)
--(axis cs:1,0.715439767265034)
--(axis cs:5,0.715439767265034)
--(axis cs:6,0.715441620368576)
--(axis cs:7.3,0.715439767265034)
--(axis cs:8.6,0.720975359936244)
--(axis cs:9,0.715439767265034)
--(axis cs:9,0.715439767265034)
--(axis cs:9.5,0.715439767265034)
--(axis cs:10,0.715439767265034)
--(axis cs:15.9,0.746898680842377)
--(axis cs:20.4,0.775169042450904)
--(axis cs:26.7,0.862636190920037)
--(axis cs:33.9,0.889192103659156)
--(axis cs:42.9,0.896593570181533)
--(axis cs:58.2,0.902079567164925)
--(axis cs:72.8,0.90640154688148)
--(axis cs:93.4,0.910370572051175)
--(axis cs:168.1,0.906293946081953)
--(axis cs:204.9,0.920069788000152)
--(axis cs:227.2,0.908847265288691)
--(axis cs:245.3,0.92378866817482)
--(axis cs:309.6,0.927272318194904)
--(axis cs:375.1,0.930313486044812)
--(axis cs:402.5,0.933223540099255)
--(axis cs:475.9,0.936003763287002)
--(axis cs:611.7,0.938840111007757)
--(axis cs:623.1,0.941740681628312)
--(axis cs:1198.8,0.945708281925142)
--(axis cs:1330.7,0.948953526559142)
--(axis cs:4925.8,0.913932517266558)
--(axis cs:11996.7,0.909765487977383)
--(axis cs:21147.9,0.880030521113367)
--(axis cs:30049.9,0.900138053940246)
--(axis cs:72961.7,0.896433063301406)
--(axis cs:79788.1,0.897690337253881)
--(axis cs:90713.3,0.90674540280498)
--(axis cs:123186.6,0.905668166637025)
--(axis cs:131628.2,0.906738762170557)
--(axis cs:144764.5,0.902937745943888)
--(axis cs:155816.2,0.898885399193661)
--(axis cs:156954.2,0.909992650789164)
--(axis cs:157491.3,0.90323304077775)
--(axis cs:157563.8,0.908842465331643)
--(axis cs:157863.1,0.91866674881865)
--(axis cs:159121,0.898105378752367)
--(axis cs:159675.2,0.918774427030845)
--(axis cs:160091.1,0.897594269010964)
--(axis cs:160986.8,0.893870954843942)
--(axis cs:161982.9,0.925090587634954)
--(axis cs:162043.7,0.892911499959637)
--(axis cs:162449.1,0.890921861849266)
--(axis cs:165655.1,0.926462935088498)
--(axis cs:169992.1,0.935025446304169)
--(axis cs:175124.6,0.941023628005294)
--(axis cs:180415.7,0.943931784172744)
--(axis cs:183103.3,0.946488027093341)
--(axis cs:185704.8,0.950234435778554)
--(axis cs:190039.8,0.949399291683129)
--(axis cs:194284.8,0.948819108889743)
--(axis cs:199015.5,0.946546390572159)
--(axis cs:202685.1,0.94595806396212)
--(axis cs:205895.9,0.945106920609384)
--(axis cs:208775.1,0.944487640321327)
--(axis cs:211621.4,0.942968766201586)
--(axis cs:213721.7,0.942904431402299)
--(axis cs:215948.1,0.943117838295233)
--(axis cs:218747.7,0.944752827498319)
--(axis cs:220524.9,0.9438975893983)
--(axis cs:221788.2,0.944108885671239)
--(axis cs:222863.4,0.944557823048208)
--(axis cs:222863.4,0.953763576951791)
--(axis cs:222863.4,0.953763576951791)
--(axis cs:221788.2,0.953938314328761)
--(axis cs:220524.9,0.9538374106017)
--(axis cs:218747.7,0.954514972501681)
--(axis cs:215948.1,0.954700761704767)
--(axis cs:213721.7,0.953790368597701)
--(axis cs:211621.4,0.954319233798415)
--(axis cs:208775.1,0.954456159678673)
--(axis cs:205895.9,0.957117879390616)
--(axis cs:202685.1,0.95603533603788)
--(axis cs:199015.5,0.957794409427841)
--(axis cs:194284.8,0.960616091110257)
--(axis cs:190039.8,0.961721108316871)
--(axis cs:185704.8,0.961496764221445)
--(axis cs:183103.3,0.960620172906659)
--(axis cs:180415.7,0.960517815827255)
--(axis cs:175124.6,0.956928171994705)
--(axis cs:169992.1,0.953174353695831)
--(axis cs:165655.1,0.949908664911502)
--(axis cs:162449.1,0.913848938150734)
--(axis cs:162043.7,0.915869900040363)
--(axis cs:161982.9,0.946909812365046)
--(axis cs:160986.8,0.913167045156058)
--(axis cs:160091.1,0.920207530989036)
--(axis cs:159675.2,0.945644772969155)
--(axis cs:159121,0.925308221247633)
--(axis cs:157863.1,0.94035545118135)
--(axis cs:157563.8,0.930802134668357)
--(axis cs:157491.3,0.93088835922225)
--(axis cs:156954.2,0.938813949210836)
--(axis cs:155816.2,0.91647380080634)
--(axis cs:144764.5,0.919200654056112)
--(axis cs:131628.2,0.923705037829443)
--(axis cs:123186.6,0.935131633362975)
--(axis cs:90713.3,0.96023399719502)
--(axis cs:79788.1,0.964484862746119)
--(axis cs:72961.7,0.963041736698594)
--(axis cs:30049.9,0.963281946059754)
--(axis cs:21147.9,0.968701078886633)
--(axis cs:11996.7,0.965021912022617)
--(axis cs:4925.8,0.966349882733442)
--(axis cs:1330.7,0.950160073440858)
--(axis cs:1198.8,0.947447118074857)
--(axis cs:623.1,0.943261518371688)
--(axis cs:611.7,0.940322488992243)
--(axis cs:475.9,0.937173836712998)
--(axis cs:402.5,0.934309859900745)
--(axis cs:375.1,0.931567513955188)
--(axis cs:309.6,0.927992281805096)
--(axis cs:245.3,0.92465433182518)
--(axis cs:227.2,0.921455134711309)
--(axis cs:204.9,0.921019811999849)
--(axis cs:168.1,0.917847253918047)
--(axis cs:93.4,0.911379427948825)
--(axis cs:72.8,0.90754445311852)
--(axis cs:58.2,0.903170032835076)
--(axis cs:42.9,0.897881429818467)
--(axis cs:33.9,0.890613496340844)
--(axis cs:26.7,0.866999409079963)
--(axis cs:20.4,0.779328157549095)
--(axis cs:15.9,0.748432119157623)
--(axis cs:10,0.716035832734966)
--(axis cs:9.5,0.716035832734966)
--(axis cs:9,0.716035832734966)
--(axis cs:9,0.716035832734966)
--(axis cs:8.6,0.722400240063756)
--(axis cs:7.3,0.716035832734966)
--(axis cs:6,0.716035979631424)
--(axis cs:5,0.716035832734966)
--(axis cs:1,0.716035832734966)
--(axis cs:1,0.716035832734966)
--(axis cs:1,0.716035832734966)
--(axis cs:1,0.716035832734966)
--(axis cs:1,0.716035832734966)
--(axis cs:1,0.716035832734966)
--(axis cs:1,0.716035832734966)
--(axis cs:1,0.716035832734966)
--(axis cs:1,0.716035832734966)
--(axis cs:1,0.716035832734966)
--cycle;

\addplot [steelblue31119180, line width=1.5, opacity=1.0, mark=*, mark size=0, mark options={solid}]
table {%
1.9 0.715737
1.9 0.715737
2.4 0.7553385
2.9 0.8151715
2.9 0.8152135
2.9 0.8152073
2.9 0.8152099
4.4 0.8233185
5.9 0.827242
7 0.8315365
9.2 0.8383719
11.4 0.8613943
13.2 0.8894413
14.9 0.8975744
16 0.9000226
17.9 0.9063814
18.8 0.9069191
22.2 0.9079492
27.6 0.9162027
31.4 0.919484
40.6 0.9220682
51.6 0.9278508
59.2 0.9316045
67.4 0.9332216
83 0.9341211
101.1 0.9356927
119.4 0.9373886
140.8 0.9403539
169.4 0.9416093
204.9 0.9445841
241.5 0.9481429
281.6 0.9506005
337.8 0.9530633
412.6 0.954746
486.7 0.9567092
574.4 0.9568701
685.2 0.9580776
810.7 0.9590442
977.7 0.9588529
1168.8 0.9588853
1406.9 0.9568146
1695.6 0.9544203
2029.9 0.9549658
2365.9 0.9551961
2438.3 0.9554907
2757.4 0.95635
3222.6 0.956394
3920.6 0.9566124
4713.4 0.9569809
5474.6 0.9564094
6181.8 0.9563112
6558.6 0.9559988
7186.2 0.9558992
8315.9 0.9554995
9677.3 0.954631
11757.3 0.9546138
14938.2 0.9532981
19512 0.9519011
22777.1 0.9513023
23940.4 0.9501474
24885.1 0.9502281
28206.4 0.949634
32613.5 0.9489645
36300.3 0.9483365
39919.1 0.9475835
43898.1 0.9466526
48399 0.9456423
53809.6 0.944256
62655.5 0.9423153
80750.3 0.9386199
98870.3 0.9329759
112273.5 0.9292975
124459.4 0.9284509
136975.9 0.9243601
149834.2 0.9223737
162584.5 0.9203758
174627 0.918119
185226.2 0.9163041
194272.9 0.9143267
};
\addlegendentry{ACOWA}
\addplot [darkorange25512714, opacity=1.0, mark=*, mark size=0, mark options={solid}]
table {%
1 0.715737
1 0.715737
1 0.715737
1 0.715737
1 0.715737
1 0.715737
1.4 0.715737
1.9 0.7157378
1.9 0.7157378
1.9 0.7157378
1.9 0.7157378
2.9 0.7157378
5.9 0.7157388
5.9 0.7157388
5.9 0.7157388
5.9 0.7157388
5.9 0.7157388
6.8 0.7157388
10 0.7433833
13.7 0.8392148
16.5 0.8500124
21.3 0.8592385
23.9 0.8789612
27.6 0.8907924
35.8 0.8977238
47.9 0.9029845
59.6 0.9074307
87.3 0.9129369
118.7 0.9182651
166.3 0.9233159
244.8 0.9276928
359 0.9317169
531.9 0.9354196
784.9 0.9389401
1090.4 0.9425005
1506.9 0.9457366
2012.3 0.9490381
2612.3 0.9518477
3360.1 0.9542337
4222.5 0.9564708
5167.4 0.9582698
6177.2 0.9597421
7363.6 0.961059
8602 0.9621307
9858 0.9631352
11612.5 0.9639814
13347.3 0.9649232
15032.3 0.9657947
16870.9 0.9663375
19041.6 0.9669316
21530.7 0.9672058
24358.7 0.9671212
27416.9 0.9668821
30324.3 0.9666016
32761.5 0.9663702
33827.6 0.9665905
34281 0.9668223
37416.6 0.9664638
42481.5 0.9657211
49031 0.9647364
57561.9 0.963467
71534.8 0.9609614
83569 0.9578576
91241.4 0.9562093
94944.8 0.9560703
95011.4 0.9565513
98689.3 0.9559705
107836.6 0.9545024
118930.6 0.9526391
127220.6 0.9512635
135441.2 0.9499661
144103.3 0.9485402
153328.7 0.9468588
165106.4 0.9446765
187037.9 0.9404035
203752.8 0.9359906
211870.3 0.933764
216838.4 0.9325533
220710.9 0.9315943
};
\addlegendentry{Naive Avg.}
\addplot [forestgreen4416044, opacity=1.0, mark=*, mark size=0, mark options={solid}]
table {%
1 0.7157378
1 0.7157378
1 0.7157378
1 0.7157378
1 0.7157378
1 0.7157378
1.2 0.7157378
1.9 0.7157378
1.9 0.7157378
1.9 0.6757704
1.9 0.6757704
2.9 0.6780584
5.9 0.8097632
5.9 0.8132697
5.9 0.8152772
5.9 0.8158811
5.9 0.818662
6.8 0.820592
10 0.8388201
13.7 0.8544217
16.5 0.8714311
21.3 0.898899
23.9 0.884792
27.6 0.8870516
35.8 0.9023068
47.9 0.9161584
59.6 0.8740045
87.3 0.8797087
118.7 0.9311049
166.3 0.9333946
244.8 0.9353109
359 0.9379344
531.9 0.9402835
784.9 0.9423025
1090.4 0.9447681
1506.9 0.9475626
2012.3 0.9491799
2612.3 0.9517412
3360.1 0.9520986
4222.5 0.9536669
5167.4 0.9546594
6177.2 0.9535574
7363.6 0.9535916
8602 0.9518192
9858 0.9516165
11612.5 0.9505241
13347.3 0.9509628
15032.3 0.9507318
16870.9 0.9516395
19041.6 0.9528313
21530.7 0.9533524
24358.7 0.9536519
27416.9 0.953681
30324.3 0.9540855
32761.5 0.9540071
33827.6 0.9540628
34281 0.9536074
37416.6 0.9535271
42481.5 0.9530079
49031 0.9525735
57561.9 0.9513879
71534.8 0.9490385
83569 0.9460238
91241.4 0.9430259
94944.8 0.9427389
95011.4 0.9421211
98689.3 0.9408747
107836.6 0.93933
118930.6 0.9372698
127220.6 0.9357686
135441.2 0.9338026
144103.3 0.9324387
153328.7 0.9302356
165106.4 0.9286358
187037.9 0.9258481
203752.8 0.9187584
211870.3 0.9154546
216838.4 0.9135971
220710.9 0.9124226
};
\addlegendentry{OWA}
\addplot [crimson2143940, opacity=1.0, mark=*, mark size=0, mark options={solid}]
table {%
1 0.7157378
1 0.7157378
1 0.7157378
1 0.7157378
1 0.7157378
1 0.7157378
1 0.7157378
1 0.7157378
1 0.7157378
1 0.7157378
1 0.7157378
2.7 0.717178
4.8 0.7157388
4.9 0.7157378
5.3 0.7157378
6.4 0.7157378
6.9 0.7157378
7 0.7476863
7.3 0.7157378
8.4 0.795919
10.5 0.8217908
10.5 0.8698075
13.9 0.8843131
16.2 0.8914883
22 0.8972049
26.5 0.9068731
29.5 0.9075747
32.9 0.9109975
33.3 0.9157113
40.2 0.9245253
44.5 0.9219048
53.5 0.9286303
57.1 0.9315026
62.5 0.9351435
79.5 0.9411401
83.1 0.9375824
111.2 0.9438472
124.4 0.9471526
189.8 0.9501259
264.5 0.9530814
351.9 0.9529514
569.8 0.9528815
1711.4 0.9469922
2198.8 0.9426446
3301.9 0.9216327
4020.2 0.9127646
4642 0.9041052
5955 0.8865043
6826 0.8755282
7549.6 0.8699731
7775.8 0.8757135
7821.8 0.8774617
8116 0.8883512
8241.7 0.8886717
9334.4 0.8967269
10154.4 0.8984539
12229.4 0.9046231
13488.1 0.9101797
15524.3 0.9165246
18523.9 0.921833
22297 0.9257679
26586.3 0.933286
32288 0.9321209
38817 0.9307112
45093.9 0.9341618
50417.8 0.9398939
62466.5 0.9456492
68751 0.9378441
75240.2 0.9315446
92823.6 0.9339888
101364.1 0.9334359
107341.8 0.9268569
112758.4 0.9211769
117934.4 0.9266201
123601.2 0.9373388
130435.6 0.912774
135357.9 0.911105
139103.1 0.9157621
142563.6 0.9051594
};
\addlegendentry{CSL}
\addplot [mediumpurple148103189, opacity=1.0, mark=*, mark size=0, mark options={solid}]
table {%
1 0.7157378
1 0.7157378
1 0.7157378
1 0.7157378
1 0.7157378
1 0.7157378
1 0.7157378
1 0.7157378
1 0.7157378
1 0.7157378
5 0.7157378
6 0.7157388
7.3 0.7157378
8.6 0.7216878
9 0.7157378
9 0.7157378
9.5 0.7157378
10 0.7157378
15.9 0.7476654
20.4 0.7772486
26.7 0.8648178
33.9 0.8899028
42.9 0.8972375
58.2 0.9026248
72.8 0.906973
93.4 0.910875
168.1 0.9120706
204.9 0.9205448
227.2 0.9151512
245.3 0.9242215
309.6 0.9276323
375.1 0.9309405
402.5 0.9337667
475.9 0.9365888
611.7 0.9395813
623.1 0.9425011
1198.8 0.9465777
1330.7 0.9495568
4925.8 0.9401412
11996.7 0.9373937
21147.9 0.9243658
30049.9 0.93171
72961.7 0.9297374
79788.1 0.9310876
90713.3 0.9334897
123186.6 0.9203999
131628.2 0.9152219
144764.5 0.9110692
155816.2 0.9076796
156954.2 0.9244033
157491.3 0.9170607
157563.8 0.9198223
157863.1 0.9295111
159121 0.9117068
159675.2 0.9322096
160091.1 0.9089009
160986.8 0.903519
161982.9 0.9360002
162043.7 0.9043907
162449.1 0.9023854
165655.1 0.9381858
169992.1 0.9440999
175124.6 0.9489759
180415.7 0.9522248
183103.3 0.9535541
185704.8 0.9558656
190039.8 0.9555602
194284.8 0.9547176
199015.5 0.9521704
202685.1 0.9509967
205895.9 0.9511124
208775.1 0.9494719
211621.4 0.948644
213721.7 0.9483474
215948.1 0.9489093
218747.7 0.9496339
220524.9 0.9488675
221788.2 0.9490236
222863.4 0.9491607
};
\addlegendentry{DANE}
\end{axis}

\end{tikzpicture}

%% file: figs/criteo-512-test-acc.tex
\begin{tikzpicture}[scale=0.7]

\definecolor{crimson2143940}{RGB}{214,39,40}
\definecolor{darkgray176}{RGB}{229,229,229}
\definecolor{darkorange25512714}{RGB}{255,127,14}
\definecolor{forestgreen4416044}{RGB}{44,160,44}
\definecolor{lightgray204}{RGB}{204,204,204}
\definecolor{mediumpurple148103189}{RGB}{148,103,189}
\definecolor{steelblue31119180}{RGB}{31,119,180}

\begin{axis}[
width=0.7\textwidth,
height=0.45\textwidth,
legend cell align={left},
legend style={
  fill opacity=0.8,
  draw opacity=1,
  text opacity=1,
  at={(0.97,0.03)},
  anchor=south east,
  draw=lightgray204
},
log basis x={10},
log basis y={10},
tick align=outside,
tick pos=left,
x grid style={darkgray176},
xlabel={Number of nonzeros},
xmajorgrids,
xmin=10, xmax=2000,
xminorgrids,
xmode=log,
xtick style={color=black},
y grid style={darkgray176},
ylabel={Accuracy},
ymajorgrids,
ymin=0.71, ymax=0.78,
yminorgrids,
ymode=linear,
ytick style={color=black}
]
\path [fill=steelblue31119180, fill opacity=0.2]
(axis cs:4,0.740402439329884)
--(axis cs:4,0.740108560670116)
--(axis cs:4,0.740108560670116)
--(axis cs:5,0.741837598547759)
--(axis cs:6,0.742620075610815)
--(axis cs:6.4,0.742620075610815)
--(axis cs:7,0.742620075610815)
--(axis cs:7,0.742620075610815)
--(axis cs:8.4,0.742967932571663)
--(axis cs:9.5,0.743478178202892)
--(axis cs:10.9,0.743558584843554)
--(axis cs:12.1,0.743568261416705)
--(axis cs:14.9,0.743568666050846)
--(axis cs:16.5,0.745194079332333)
--(axis cs:17,0.748712600214708)
--(axis cs:20.3,0.749104061046631)
--(axis cs:22.9,0.750204028854443)
--(axis cs:30.5,0.751139795644336)
--(axis cs:35.3,0.75183236347699)
--(axis cs:39,0.752793749030254)
--(axis cs:44.2,0.753988204295907)
--(axis cs:48.1,0.755505920611067)
--(axis cs:53.4,0.756596493506)
--(axis cs:59.8,0.757443581988929)
--(axis cs:61.8,0.758042448417682)
--(axis cs:67.4,0.758498164766307)
--(axis cs:75,0.758899784984675)
--(axis cs:85.4,0.759711558862324)
--(axis cs:98.1,0.760783200970762)
--(axis cs:113.6,0.761293602399651)
--(axis cs:126.5,0.762298690247477)
--(axis cs:140.2,0.763119617625418)
--(axis cs:156.4,0.763814047962895)
--(axis cs:167.7,0.7644344797945)
--(axis cs:182.3,0.764664626474023)
--(axis cs:202.9,0.765184903708199)
--(axis cs:220.7,0.765667999010815)
--(axis cs:240.1,0.765917404051868)
--(axis cs:265,0.766179310403427)
--(axis cs:286.6,0.766567949219575)
--(axis cs:320.9,0.766761457338813)
--(axis cs:358.9,0.767612373631217)
--(axis cs:383.3,0.76814353281206)
--(axis cs:414.8,0.768518728627888)
--(axis cs:451,0.768765455032704)
--(axis cs:494.2,0.769280913856791)
--(axis cs:548.3,0.76961830288851)
--(axis cs:610.3,0.770007529507635)
--(axis cs:670.8,0.7703111506717)
--(axis cs:741.8,0.770695424139792)
--(axis cs:827,0.771015690595983)
--(axis cs:919.1,0.771478819911463)
--(axis cs:1029.8,0.771964858262189)
--(axis cs:1148.7,0.772293972457701)
--(axis cs:1265.5,0.772528164513563)
--(axis cs:1405.3,0.77285217653434)
--(axis cs:1589.2,0.773059856795235)
--(axis cs:1793.9,0.773349175972828)
--(axis cs:2057.6,0.773529268882044)
--(axis cs:2382.7,0.77378503860063)
--(axis cs:2785.7,0.773669962791921)
--(axis cs:3256.3,0.773423416070942)
--(axis cs:3843.6,0.773141231927064)
--(axis cs:4528.1,0.773075499505904)
--(axis cs:5384,0.77244846823738)
--(axis cs:6435.3,0.771530450376308)
--(axis cs:7751.4,0.770824550378638)
--(axis cs:9351.5,0.768206659564837)
--(axis cs:11297.5,0.767305943806053)
--(axis cs:13678.7,0.764836998888205)
--(axis cs:16572.8,0.762825969661268)
--(axis cs:20066.3,0.759407008943507)
--(axis cs:24460.2,0.756796684417705)
--(axis cs:29962.1,0.753551822806034)
--(axis cs:36548.6,0.750286296447762)
--(axis cs:45839.9,0.747149683586106)
--(axis cs:58173.5,0.742312414348354)
--(axis cs:70872,0.739240176560699)
--(axis cs:82961,0.734768029168377)
--(axis cs:99604.2,0.730445029552712)
--(axis cs:150376.8,0.727551381654227)
--(axis cs:150376.8,0.729543018345773)
--(axis cs:150376.8,0.729543018345773)
--(axis cs:99604.2,0.732451170447288)
--(axis cs:82961,0.736889170831623)
--(axis cs:70872,0.741248223439301)
--(axis cs:58173.5,0.747018785651646)
--(axis cs:45839.9,0.748645316413894)
--(axis cs:36548.6,0.754162103552238)
--(axis cs:29962.1,0.757777777193966)
--(axis cs:24460.2,0.760926315582295)
--(axis cs:20066.3,0.766418791056493)
--(axis cs:16572.8,0.769361430338732)
--(axis cs:13678.7,0.770682401111795)
--(axis cs:11297.5,0.773063856193947)
--(axis cs:9351.5,0.771883340435163)
--(axis cs:7751.4,0.775637849621362)
--(axis cs:6435.3,0.775700349623692)
--(axis cs:5384,0.77657893176262)
--(axis cs:4528.1,0.776192300494096)
--(axis cs:3843.6,0.775911368072936)
--(axis cs:3256.3,0.774929983929059)
--(axis cs:2785.7,0.774907037208079)
--(axis cs:2382.7,0.77527536139937)
--(axis cs:2057.6,0.774491731117956)
--(axis cs:1793.9,0.774577624027172)
--(axis cs:1589.2,0.774173743204765)
--(axis cs:1405.3,0.773778423465659)
--(axis cs:1265.5,0.773224835486437)
--(axis cs:1148.7,0.773005027542299)
--(axis cs:1029.8,0.772369941737811)
--(axis cs:919.1,0.771873380088537)
--(axis cs:827,0.771347709404017)
--(axis cs:741.8,0.771008575860208)
--(axis cs:670.8,0.7706622493283)
--(axis cs:610.3,0.770467870492365)
--(axis cs:548.3,0.770064897111491)
--(axis cs:494.2,0.769655686143209)
--(axis cs:451,0.769139944967296)
--(axis cs:414.8,0.768898671372112)
--(axis cs:383.3,0.76853786718794)
--(axis cs:358.9,0.768199626368783)
--(axis cs:320.9,0.767361342661187)
--(axis cs:286.6,0.767063050780425)
--(axis cs:265,0.766727089596573)
--(axis cs:240.1,0.766376995948132)
--(axis cs:220.7,0.766054400989185)
--(axis cs:202.9,0.765599896291801)
--(axis cs:182.3,0.765325373525977)
--(axis cs:167.7,0.7649267202055)
--(axis cs:156.4,0.764299352037105)
--(axis cs:140.2,0.763636782374582)
--(axis cs:126.5,0.762818109752523)
--(axis cs:113.6,0.761684397600349)
--(axis cs:98.1,0.761258199029238)
--(axis cs:85.4,0.760200041137676)
--(axis cs:75,0.759334615015325)
--(axis cs:67.4,0.758946035233692)
--(axis cs:61.8,0.758512751582318)
--(axis cs:59.8,0.757888018011071)
--(axis cs:53.4,0.757036906494)
--(axis cs:48.1,0.755953879388933)
--(axis cs:44.2,0.754587995704093)
--(axis cs:39,0.753203650969746)
--(axis cs:35.3,0.75233363652301)
--(axis cs:30.5,0.751538004355664)
--(axis cs:22.9,0.751005571145557)
--(axis cs:20.3,0.749907738953369)
--(axis cs:17,0.749043799785292)
--(axis cs:16.5,0.746455920667667)
--(axis cs:14.9,0.743850133949154)
--(axis cs:12.1,0.743849538583295)
--(axis cs:10.9,0.743839815156446)
--(axis cs:9.5,0.743800821797108)
--(axis cs:8.4,0.743646267428337)
--(axis cs:7,0.742900324389185)
--(axis cs:7,0.742900324389185)
--(axis cs:6.4,0.742900324389185)
--(axis cs:6,0.742900324389185)
--(axis cs:5,0.742113801452241)
--(axis cs:4,0.740402439329884)
--(axis cs:4,0.740402439329884)
--cycle;

\path [fill=darkorange25512714, fill opacity=0.2]
(axis cs:2.1,0.697473058161386)
--(axis cs:2.1,0.652607341838613)
--(axis cs:4,0.742508835689981)
--(axis cs:4,0.742508835689981)
--(axis cs:4,0.742508835689981)
--(axis cs:4,0.742508835689981)
--(axis cs:4,0.742508835689981)
--(axis cs:6,0.743420534450217)
--(axis cs:6,0.743420534450217)
--(axis cs:6,0.743420534450217)
--(axis cs:6.1,0.743429987419205)
--(axis cs:7,0.743521098145535)
--(axis cs:7,0.743521098145535)
--(axis cs:7.2,0.743521098145535)
--(axis cs:9.1,0.743554930697607)
--(axis cs:13.1,0.74357001334583)
--(axis cs:14.7,0.74357001334583)
--(axis cs:16,0.74357001334583)
--(axis cs:16.1,0.74357001334583)
--(axis cs:16.8,0.74357001334583)
--(axis cs:17.4,0.74357001334583)
--(axis cs:18.3,0.74357001334583)
--(axis cs:19.1,0.74357001334583)
--(axis cs:22.8,0.74357001334583)
--(axis cs:25.3,0.743957765029595)
--(axis cs:29.4,0.745094298247693)
--(axis cs:33.4,0.746374856843836)
--(axis cs:39.4,0.748525067890448)
--(axis cs:43.3,0.749631035257122)
--(axis cs:49.2,0.750889289057856)
--(axis cs:52.1,0.751932219319989)
--(axis cs:58,0.752806059813234)
--(axis cs:65.8,0.7538327423615)
--(axis cs:83.6,0.754952206667501)
--(axis cs:100.5,0.755862399613444)
--(axis cs:117.8,0.756765962625615)
--(axis cs:133.5,0.757635130619868)
--(axis cs:150.8,0.758548719252333)
--(axis cs:169.1,0.759455932254631)
--(axis cs:193.2,0.760324696601383)
--(axis cs:224.7,0.761166185406964)
--(axis cs:266.7,0.761937444117604)
--(axis cs:313.5,0.762699148580143)
--(axis cs:361.9,0.763364275166219)
--(axis cs:429.3,0.763999843127229)
--(axis cs:506,0.764602607241596)
--(axis cs:596.5,0.765158150428632)
--(axis cs:690.1,0.765729396010347)
--(axis cs:816.4,0.766293916846091)
--(axis cs:983.4,0.76690294158735)
--(axis cs:1178,0.767524306023285)
--(axis cs:1377.7,0.768137705206235)
--(axis cs:1607.8,0.768727874871138)
--(axis cs:1908.1,0.769338514439845)
--(axis cs:2285.2,0.769924114354378)
--(axis cs:2774.9,0.770538371562275)
--(axis cs:3371.2,0.771151101558062)
--(axis cs:4096.4,0.771776895081464)
--(axis cs:5018.7,0.772410873874044)
--(axis cs:6168.6,0.773063443201881)
--(axis cs:7565.1,0.773717509111923)
--(axis cs:9308.3,0.774401649631738)
--(axis cs:11446,0.775098488170789)
--(axis cs:14073,0.775783269704232)
--(axis cs:17274.4,0.776446881147515)
--(axis cs:21232.1,0.77708937309455)
--(axis cs:26115.9,0.777709100431468)
--(axis cs:32277.9,0.778320773353698)
--(axis cs:40967.7,0.778898852814827)
--(axis cs:49878.1,0.779467517337093)
--(axis cs:66407.4,0.780011218867024)
--(axis cs:92515.9,0.780525349187382)
--(axis cs:116037,0.781009228085657)
--(axis cs:134654.2,0.781430895495123)
--(axis cs:153022.9,0.781830182445539)
--(axis cs:310360.5,0.782195644849376)
--(axis cs:444410.3,0.78255198755355)
--(axis cs:516245.8,0.782847954991761)
--(axis cs:555577.6,0.783100213497333)
--(axis cs:578720.3,0.783281838759934)
--(axis cs:593668.8,0.783402778860072)
--(axis cs:593668.8,0.783726621139928)
--(axis cs:593668.8,0.783726621139928)
--(axis cs:578720.3,0.783614761240066)
--(axis cs:555577.6,0.783435986502667)
--(axis cs:516245.8,0.783209045008239)
--(axis cs:444410.3,0.78291801244645)
--(axis cs:310360.5,0.782570955150624)
--(axis cs:153022.9,0.782192817554461)
--(axis cs:134654.2,0.781779904504877)
--(axis cs:116037,0.781350971914343)
--(axis cs:92515.9,0.780891450812618)
--(axis cs:66407.4,0.780360781132976)
--(axis cs:49878.1,0.779819282662907)
--(axis cs:40967.7,0.779258947185173)
--(axis cs:32277.9,0.778661226646302)
--(axis cs:26115.9,0.778041299568532)
--(axis cs:21232.1,0.77742142690545)
--(axis cs:17274.4,0.776786318852485)
--(axis cs:14073,0.776118330295768)
--(axis cs:11446,0.775415511829212)
--(axis cs:9308.3,0.774726750368262)
--(axis cs:7565.1,0.774038690888077)
--(axis cs:6168.6,0.773370356798119)
--(axis cs:5018.7,0.772716726125956)
--(axis cs:4096.4,0.772090904918536)
--(axis cs:3371.2,0.771471298441938)
--(axis cs:2774.9,0.770863828437725)
--(axis cs:2285.2,0.770262285645622)
--(axis cs:1908.1,0.769679685560155)
--(axis cs:1607.8,0.769067925128862)
--(axis cs:1377.7,0.768455694793765)
--(axis cs:1178,0.767842693976715)
--(axis cs:983.4,0.76721725841265)
--(axis cs:816.4,0.76661748315391)
--(axis cs:690.1,0.766040203989653)
--(axis cs:596.5,0.765475049571368)
--(axis cs:506,0.764914992758404)
--(axis cs:429.3,0.764335756872771)
--(axis cs:361.9,0.763698924833782)
--(axis cs:313.5,0.763035451419857)
--(axis cs:266.7,0.762271355882396)
--(axis cs:224.7,0.761487814593036)
--(axis cs:193.2,0.760651303398617)
--(axis cs:169.1,0.759789667745369)
--(axis cs:150.8,0.758890880747667)
--(axis cs:133.5,0.757969269380132)
--(axis cs:117.8,0.757104437374385)
--(axis cs:100.5,0.756210200386556)
--(axis cs:83.6,0.7552719933325)
--(axis cs:65.8,0.7541614576385)
--(axis cs:58,0.753119940186766)
--(axis cs:52.1,0.752250580680011)
--(axis cs:49.2,0.751192510942144)
--(axis cs:43.3,0.749936564742878)
--(axis cs:39.4,0.748830132109552)
--(axis cs:33.4,0.746673943156164)
--(axis cs:29.4,0.745377101752307)
--(axis cs:25.3,0.744239434970405)
--(axis cs:22.8,0.74385098665417)
--(axis cs:19.1,0.74385098665417)
--(axis cs:18.3,0.74385098665417)
--(axis cs:17.4,0.74385098665417)
--(axis cs:16.8,0.74385098665417)
--(axis cs:16.1,0.74385098665417)
--(axis cs:16,0.74385098665417)
--(axis cs:14.7,0.74385098665417)
--(axis cs:13.1,0.74385098665417)
--(axis cs:9.1,0.743835269302393)
--(axis cs:7.2,0.743803301854465)
--(axis cs:7,0.743803301854465)
--(axis cs:7,0.743803301854465)
--(axis cs:6.1,0.743713412580795)
--(axis cs:6,0.743703065549783)
--(axis cs:6,0.743703065549783)
--(axis cs:6,0.743703065549783)
--(axis cs:4,0.742795964310019)
--(axis cs:4,0.742795964310019)
--(axis cs:4,0.742795964310019)
--(axis cs:4,0.742795964310019)
--(axis cs:4,0.742795964310019)
--(axis cs:2.1,0.697473058161386)
--cycle;

\path [fill=forestgreen4416044, fill opacity=0.15]
(axis cs:2,0.668094129902267)
--(axis cs:2,0.667812070097733)
--(axis cs:3.8,0.740284077455645)
--(axis cs:4,0.742508835689981)
--(axis cs:4,0.742508835689981)
--(axis cs:4,0.742508835689981)
--(axis cs:4.1,0.742508835689981)
--(axis cs:6,0.743420403581865)
--(axis cs:6,0.743420534450217)
--(axis cs:6,0.743420534450217)
--(axis cs:6,0.743420534450217)
--(axis cs:6.2,0.738783577644401)
--(axis cs:7,0.743521098145535)
--(axis cs:7,0.727220155744813)
--(axis cs:7.2,0.743522580344731)
--(axis cs:9.1,0.613324970535362)
--(axis cs:13,0.743569661511796)
--(axis cs:13.8,0.74357001334583)
--(axis cs:16.1,0.744837840040233)
--(axis cs:16.3,0.743712829157572)
--(axis cs:16.7,0.745022531651387)
--(axis cs:17.4,0.745025108787382)
--(axis cs:17.6,0.745025454080999)
--(axis cs:18.9,0.743286934605236)
--(axis cs:21,0.744647533668151)
--(axis cs:23.3,0.749792950336922)
--(axis cs:27.3,0.749696606878731)
--(axis cs:29.9,0.749503883912976)
--(axis cs:33.7,0.753081360557195)
--(axis cs:39.8,0.749882938845555)
--(axis cs:43.3,0.609482163745598)
--(axis cs:48.8,0.606157295736455)
--(axis cs:51.4,0.596162437021755)
--(axis cs:56.7,0.604500619028408)
--(axis cs:63.5,0.629590199673286)
--(axis cs:76.8,0.756548277082645)
--(axis cs:94.4,0.640546430923442)
--(axis cs:109.7,0.649960149564104)
--(axis cs:124.9,0.647194565975152)
--(axis cs:140.2,0.647030400582404)
--(axis cs:156.7,0.63807671700059)
--(axis cs:174.3,0.640505240809305)
--(axis cs:200.4,0.651753306978806)
--(axis cs:229.7,0.661358056210726)
--(axis cs:269.1,0.765306953983133)
--(axis cs:313.5,0.682418334083423)
--(axis cs:359,0.690502795460954)
--(axis cs:418.1,0.69884084441519)
--(axis cs:488.8,0.702434754791024)
--(axis cs:569.8,0.707139615234625)
--(axis cs:659.9,0.711602790180265)
--(axis cs:760.4,0.714715630899329)
--(axis cs:902.2,0.768278152673169)
--(axis cs:1073.5,0.740314674548327)
--(axis cs:1257.5,0.768590684370952)
--(axis cs:1449.8,0.768639023914003)
--(axis cs:1680.2,0.769905147217177)
--(axis cs:1974.9,0.768757264019229)
--(axis cs:2344.1,0.76936888695199)
--(axis cs:2808.3,0.768968816441528)
--(axis cs:3371.2,0.769622063264243)
--(axis cs:4047.9,0.769540466205481)
--(axis cs:4881.3,0.769501197640865)
--(axis cs:5928.3,0.769249539211178)
--(axis cs:7167.1,0.769306996798331)
--(axis cs:8672.8,0.768712699045997)
--(axis cs:10549.5,0.76759673829775)
--(axis cs:12765.5,0.771087155800989)
--(axis cs:15471.9,0.771010343412605)
--(axis cs:18767.8,0.769444501856577)
--(axis cs:22776.4,0.769962135375219)
--(axis cs:27560.3,0.768342106696336)
--(axis cs:33851.2,0.76546862926318)
--(axis cs:42228.9,0.764184096385862)
--(axis cs:50555.5,0.762845778493518)
--(axis cs:66407.4,0.761169536158104)
--(axis cs:90759.8,0.757624963916751)
--(axis cs:113202.9,0.754696750519067)
--(axis cs:131246,0.75134223114369)
--(axis cs:146113.4,0.74971369888107)
--(axis cs:249508.7,0.747002063418574)
--(axis cs:249508.7,0.750850136581426)
--(axis cs:249508.7,0.750850136581426)
--(axis cs:146113.4,0.75443150111893)
--(axis cs:131246,0.756010168856309)
--(axis cs:113202.9,0.758618849480933)
--(axis cs:90759.8,0.762556836083249)
--(axis cs:66407.4,0.765571463841896)
--(axis cs:50555.5,0.767322021506482)
--(axis cs:42228.9,0.768658503614138)
--(axis cs:33851.2,0.76954137073682)
--(axis cs:27560.3,0.772343293303664)
--(axis cs:22776.4,0.772786864624781)
--(axis cs:18767.8,0.773364498143423)
--(axis cs:15471.9,0.773781256587395)
--(axis cs:12765.5,0.773503644199011)
--(axis cs:10549.5,0.77510886170225)
--(axis cs:8672.8,0.774721300954003)
--(axis cs:7167.1,0.774795803201669)
--(axis cs:5928.3,0.774156660788822)
--(axis cs:4881.3,0.773706602359135)
--(axis cs:4047.9,0.773204933794519)
--(axis cs:3371.2,0.772574536735757)
--(axis cs:2808.3,0.772274183558472)
--(axis cs:2344.1,0.77204951304801)
--(axis cs:1974.9,0.771669935980771)
--(axis cs:1680.2,0.771114852782823)
--(axis cs:1449.8,0.771148176085997)
--(axis cs:1257.5,0.770743315629048)
--(axis cs:1073.5,0.784494725451673)
--(axis cs:902.2,0.770030647326831)
--(axis cs:760.4,0.796934969100671)
--(axis cs:659.9,0.797957809819735)
--(axis cs:569.8,0.799306184765375)
--(axis cs:488.8,0.800938445208976)
--(axis cs:418.1,0.80181455558481)
--(axis cs:359,0.805886804539046)
--(axis cs:313.5,0.808773665916577)
--(axis cs:269.1,0.765961446016867)
--(axis cs:229.7,0.819326743789274)
--(axis cs:200.4,0.823627493021194)
--(axis cs:174.3,0.827343959190695)
--(axis cs:156.7,0.82748228299941)
--(axis cs:140.2,0.821929999417596)
--(axis cs:124.9,0.820519434024848)
--(axis cs:109.7,0.817859050435896)
--(axis cs:94.4,0.821068169076558)
--(axis cs:76.8,0.761032722917355)
--(axis cs:63.5,0.825221800326714)
--(axis cs:56.7,0.836789780971592)
--(axis cs:51.4,0.839682762978245)
--(axis cs:48.8,0.832230504263545)
--(axis cs:43.3,0.829913636254402)
--(axis cs:39.8,0.755745461154445)
--(axis cs:33.7,0.753556039442805)
--(axis cs:29.9,0.754856316087024)
--(axis cs:27.3,0.753927193121269)
--(axis cs:23.3,0.753025649663078)
--(axis cs:21,0.748749266331849)
--(axis cs:18.9,0.746116465394764)
--(axis cs:17.6,0.745276545919001)
--(axis cs:17.4,0.745276091212618)
--(axis cs:16.7,0.745276868348613)
--(axis cs:16.3,0.744993770842428)
--(axis cs:16.1,0.745313959959767)
--(axis cs:13.8,0.74385098665417)
--(axis cs:13,0.743850738488204)
--(axis cs:9.1,0.811428229464638)
--(axis cs:7.2,0.743808019655269)
--(axis cs:7,0.752211244255187)
--(axis cs:7,0.743803301854465)
--(axis cs:6.2,0.746085822355599)
--(axis cs:6,0.743703065549783)
--(axis cs:6,0.743703065549783)
--(axis cs:6,0.743703065549783)
--(axis cs:6,0.743702996418135)
--(axis cs:4.1,0.742795964310019)
--(axis cs:4,0.742795964310019)
--(axis cs:4,0.742795964310019)
--(axis cs:4,0.742795964310019)
--(axis cs:3.8,0.743531722544355)
--(axis cs:2,0.668094129902267)
--cycle;

\path [fill=crimson2143940, fill opacity=0.2]
(axis cs:1,0.25642998665417)
--(axis cs:1,0.25614901334583)
--(axis cs:1,0.25614901334583)
--(axis cs:1,0.25614901334583)
--(axis cs:1,0.25614901334583)
--(axis cs:1,0.25614901334583)
--(axis cs:1,0.25614901334583)
--(axis cs:1,0.25614901334583)
--(axis cs:6.4,0.743503618483168)
--(axis cs:8.9,0.743560674057678)
--(axis cs:11.4,0.743567522878385)
--(axis cs:12.5,0.743564946366365)
--(axis cs:14.7,0.74357001334583)
--(axis cs:16,0.74357001334583)
--(axis cs:16.8,0.74357001334583)
--(axis cs:22.9,0.743685759197662)
--(axis cs:30.8,0.744012896505353)
--(axis cs:32.3,0.748431256828598)
--(axis cs:45.2,0.750951076152884)
--(axis cs:55.9,0.753014656228393)
--(axis cs:73.2,0.755113973389789)
--(axis cs:94.7,0.757692719439545)
--(axis cs:113.4,0.759692998849516)
--(axis cs:132.6,0.761547513178336)
--(axis cs:150.6,0.763261325012036)
--(axis cs:179.7,0.764628180837099)
--(axis cs:232,0.765718913492601)
--(axis cs:310.3,0.766901285469608)
--(axis cs:406.9,0.768152471710241)
--(axis cs:578.9,0.769203854403581)
--(axis cs:784.8,0.769958117011529)
--(axis cs:1091,0.770903603853373)
--(axis cs:1447.3,0.771928485989492)
--(axis cs:1967.6,0.772705390979187)
--(axis cs:2703,0.77360149374829)
--(axis cs:3612.3,0.774497266496627)
--(axis cs:4811.5,0.775463999901589)
--(axis cs:6871.3,0.776394124666851)
--(axis cs:9889.8,0.776620538164184)
--(axis cs:15466.5,0.776772777029978)
--(axis cs:25401,0.775232481611154)
--(axis cs:33746.7,0.774994623944803)
--(axis cs:59603.8,0.765617854160232)
--(axis cs:77059.3,0.762870467682732)
--(axis cs:90662.2,0.755748329082939)
--(axis cs:101728.7,0.749951458271721)
--(axis cs:111916.7,0.748223657524629)
--(axis cs:124374.8,0.74603075536761)
--(axis cs:137641.9,0.743109530254953)
--(axis cs:152559.1,0.74241241638278)
--(axis cs:152559.1,0.75082638361722)
--(axis cs:152559.1,0.75082638361722)
--(axis cs:137641.9,0.758303669745047)
--(axis cs:124374.8,0.75662024463239)
--(axis cs:111916.7,0.75713014247537)
--(axis cs:101728.7,0.761347741728279)
--(axis cs:90662.2,0.764224670917061)
--(axis cs:77059.3,0.768634532317268)
--(axis cs:59603.8,0.768788145839768)
--(axis cs:33746.7,0.777184776055197)
--(axis cs:25401,0.776383318388846)
--(axis cs:15466.5,0.777338222970022)
--(axis cs:9889.8,0.777228661835816)
--(axis cs:6871.3,0.77680247533315)
--(axis cs:4811.5,0.776412400098411)
--(axis cs:3612.3,0.775307533503373)
--(axis cs:2703,0.77414730625171)
--(axis cs:1967.6,0.773494809020813)
--(axis cs:1447.3,0.772552714010508)
--(axis cs:1091,0.771497596146627)
--(axis cs:784.8,0.770571082988471)
--(axis cs:578.9,0.76962074559642)
--(axis cs:406.9,0.768576328289759)
--(axis cs:310.3,0.767353914530392)
--(axis cs:232,0.766057686507399)
--(axis cs:179.7,0.764888019162901)
--(axis cs:150.6,0.763728674987964)
--(axis cs:132.6,0.762002286821664)
--(axis cs:113.4,0.760234801150484)
--(axis cs:94.7,0.758182480560455)
--(axis cs:73.2,0.756209026610211)
--(axis cs:55.9,0.754275943771607)
--(axis cs:45.2,0.752278723847116)
--(axis cs:32.3,0.750064943171402)
--(axis cs:30.8,0.745765103494647)
--(axis cs:22.9,0.744049640802338)
--(axis cs:16.8,0.74385098665417)
--(axis cs:16,0.74385098665417)
--(axis cs:14.7,0.74385098665417)
--(axis cs:12.5,0.743849253633635)
--(axis cs:11.4,0.743847677121615)
--(axis cs:8.9,0.743835925942322)
--(axis cs:6.4,0.743829181516832)
--(axis cs:1,0.25642998665417)
--(axis cs:1,0.25642998665417)
--(axis cs:1,0.25642998665417)
--(axis cs:1,0.25642998665417)
--(axis cs:1,0.25642998665417)
--(axis cs:1,0.25642998665417)
--(axis cs:1,0.25642998665417)
--cycle;

\path [fill=mediumpurple148103189, fill opacity=0.2]
(axis cs:1,0.25642998665417)
--(axis cs:1,0.25614901334583)
--(axis cs:1,0.25614901334583)
--(axis cs:1,0.25614901334583)
--(axis cs:1,0.25614901334583)
--(axis cs:1,0.25614901334583)
--(axis cs:1,0.25614901334583)
--(axis cs:1,0.25614901334583)
--(axis cs:1.9,0.25614901334583)
--(axis cs:8,0.743559951173505)
--(axis cs:14,0.74356901334583)
--(axis cs:15.6,0.743569880171427)
--(axis cs:17,0.74357001334583)
--(axis cs:17,0.74357001334583)
--(axis cs:18,0.743569615663904)
--(axis cs:19.3,0.74357001334583)
--(axis cs:21.1,0.74357001334583)
--(axis cs:23.1,0.74357001334583)
--(axis cs:29.8,0.743786909262168)
--(axis cs:34.5,0.744634564198875)
--(axis cs:41.4,0.748062617172635)
--(axis cs:53.5,0.750203599791867)
--(axis cs:63.7,0.75211977364995)
--(axis cs:78.1,0.753900864825653)
--(axis cs:103.6,0.755711808727469)
--(axis cs:132.4,0.757523321716273)
--(axis cs:160.2,0.759250196722841)
--(axis cs:185.9,0.760752099602459)
--(axis cs:205.6,0.762216268953585)
--(axis cs:246.3,0.763569303084881)
--(axis cs:293.2,0.764637138183585)
--(axis cs:353.1,0.765610186785259)
--(axis cs:429.2,0.766577196653109)
--(axis cs:548.2,0.767678857460318)
--(axis cs:714.5,0.768665207882572)
--(axis cs:1003.9,0.769463613778252)
--(axis cs:1200.3,0.770290203941874)
--(axis cs:1632,0.771077622234343)
--(axis cs:2107.9,0.771850179972969)
--(axis cs:3123.8,0.772578021145613)
--(axis cs:3974.8,0.773353864819068)
--(axis cs:5561.2,0.774145548845851)
--(axis cs:8196.3,0.774977062886249)
--(axis cs:12493.9,0.775902597002334)
--(axis cs:20247.1,0.776849083399394)
--(axis cs:35235.4,0.777715488484595)
--(axis cs:65579.2,0.778536874482482)
--(axis cs:186338.8,0.779485185465728)
--(axis cs:377822.3,0.780407569807959)
--(axis cs:417787,0.78077451296348)
--(axis cs:417787,0.78109968703652)
--(axis cs:417787,0.78109968703652)
--(axis cs:377822.3,0.780685230192041)
--(axis cs:186338.8,0.779809414534272)
--(axis cs:65579.2,0.778869525517519)
--(axis cs:35235.4,0.778069711515405)
--(axis cs:20247.1,0.777223316600606)
--(axis cs:12493.9,0.776258802997666)
--(axis cs:8196.3,0.775297137113751)
--(axis cs:5561.2,0.774451451154149)
--(axis cs:3974.8,0.773644935180932)
--(axis cs:3123.8,0.772912978854387)
--(axis cs:2107.9,0.772164220027031)
--(axis cs:1632,0.771393377765657)
--(axis cs:1200.3,0.770638596058126)
--(axis cs:1003.9,0.769863586221748)
--(axis cs:714.5,0.768996592117428)
--(axis cs:548.2,0.767993142539682)
--(axis cs:429.2,0.766904203346891)
--(axis cs:353.1,0.765889013214741)
--(axis cs:293.2,0.764924861816415)
--(axis cs:246.3,0.763884096915119)
--(axis cs:205.6,0.762536931046415)
--(axis cs:185.9,0.761084700397541)
--(axis cs:160.2,0.759590603277159)
--(axis cs:132.4,0.757897878283727)
--(axis cs:103.6,0.756037591272531)
--(axis cs:78.1,0.754220135174347)
--(axis cs:63.7,0.75238802635005)
--(axis cs:53.5,0.750550600208133)
--(axis cs:41.4,0.748383782827365)
--(axis cs:34.5,0.744945235801125)
--(axis cs:29.8,0.744071090737832)
--(axis cs:23.1,0.74385098665417)
--(axis cs:21.1,0.74385098665417)
--(axis cs:19.3,0.74385098665417)
--(axis cs:18,0.743850784336096)
--(axis cs:17,0.74385098665417)
--(axis cs:17,0.74385098665417)
--(axis cs:15.6,0.743850919828573)
--(axis cs:14,0.74384998665417)
--(axis cs:8,0.743840448826495)
--(axis cs:1.9,0.25642998665417)
--(axis cs:1,0.25642998665417)
--(axis cs:1,0.25642998665417)
--(axis cs:1,0.25642998665417)
--(axis cs:1,0.25642998665417)
--(axis cs:1,0.25642998665417)
--(axis cs:1,0.25642998665417)
--(axis cs:1,0.25642998665417)
--cycle;

\addplot [semithick, steelblue31119180, line width=1.5]
table {%
4 0.7402555
4 0.7402555
5 0.7419757
6 0.7427602
6.4 0.7427602
7 0.7427602
7 0.7427602
8.4 0.7433071
9.5 0.7436395
10.9 0.7436992
12.1 0.7437089
14.9 0.7437094
16.5 0.745825
17 0.7488782
20.3 0.7495059
22.9 0.7506048
30.5 0.7513389
35.3 0.752083
39 0.7529987
44.2 0.7542881
48.1 0.7557299
53.4 0.7568167
59.8 0.7576658
61.8 0.7582776
67.4 0.7587221
75 0.7591172
85.4 0.7599558
98.1 0.7610207
113.6 0.761489
126.5 0.7625584
140.2 0.7633782
156.4 0.7640567
167.7 0.7646806
182.3 0.764995
202.9 0.7653924
220.7 0.7658612
240.1 0.7661472
265 0.7664532
286.6 0.7668155
320.9 0.7670614
358.9 0.767906
383.3 0.7683407
414.8 0.7687087
451 0.7689527
494.2 0.7694683
548.3 0.7698416
610.3 0.7702377
670.8 0.7704867
741.8 0.770852
827 0.7711817
919.1 0.7716761
1029.8 0.7721674
1148.7 0.7726495
1265.5 0.7728765
1405.3 0.7733153
1589.2 0.7736168
1793.9 0.7739634
2057.6 0.7740105
2382.7 0.7745302
2785.7 0.7742885
3256.3 0.7741767
3843.6 0.7745263
4528.1 0.7746339
5384 0.7745137
6435.3 0.7736154
7751.4 0.7732312
9351.5 0.770045
11297.5 0.7701849
13678.7 0.7677597
16572.8 0.7660937
20066.3 0.7629129
24460.2 0.7588615
29962.1 0.7556648
36548.6 0.7522242
45839.9 0.7478975
58173.5 0.7446656
70872 0.7402442
82961 0.7358286
99604.2 0.7314481
150376.8 0.7285472
};
\addplot [semithick, darkorange25512714]
table {%
2.1 0.6750402
4 0.7426524
4 0.7426524
4 0.7426524
4 0.7426524
4 0.7426524
6 0.7435618
6 0.7435618
6 0.7435618
6.1 0.7435717
7 0.7436622
7 0.7436622
7.2 0.7436622
9.1 0.7436951
13.1 0.7437105
14.7 0.7437105
16 0.7437105
16.1 0.7437105
16.8 0.7437105
17.4 0.7437105
18.3 0.7437105
19.1 0.7437105
22.8 0.7437105
25.3 0.7440986
29.4 0.7452357
33.4 0.7465244
39.4 0.7486776
43.3 0.7497838
49.2 0.7510409
52.1 0.7520914
58 0.752963
65.8 0.7539971
83.6 0.7551121
100.5 0.7560363
117.8 0.7569352
133.5 0.7578022
150.8 0.7587198
169.1 0.7596228
193.2 0.760488
224.7 0.761327
266.7 0.7621044
313.5 0.7628673
361.9 0.7635316
429.3 0.7641678
506 0.7647588
596.5 0.7653166
690.1 0.7658848
816.4 0.7664557
983.4 0.7670601
1178 0.7676835
1377.7 0.7682967
1607.8 0.7688979
1908.1 0.7695091
2285.2 0.7700932
2774.9 0.7707011
3371.2 0.7713112
4096.4 0.7719339
5018.7 0.7725638
6168.6 0.7732169
7565.1 0.7738781
9308.3 0.7745642
11446 0.775257
14073 0.7759508
17274.4 0.7766166
21232.1 0.7772554
26115.9 0.7778752
32277.9 0.778491
40967.7 0.7790789
49878.1 0.7796434
66407.4 0.780186
92515.9 0.7807084
116037 0.7811801
134654.2 0.7816054
153022.9 0.7820115
310360.5 0.7823833
444410.3 0.782735
516245.8 0.7830285
555577.6 0.7832681
578720.3 0.7834483
593668.8 0.7835647
};
\addplot [semithick, forestgreen4416044]
table {%
2 0.6679531
3.8 0.7419079
4 0.7426524
4 0.7426524
4 0.7426524
4.1 0.7426524
6 0.7435617
6 0.7435618
6 0.7435618
6 0.7435618
6.2 0.7424347
7 0.7436622
7 0.7397157
7.2 0.7436653
9.1 0.7123766
13 0.7437102
13.8 0.7437105
16.1 0.7450759
16.3 0.7443533
16.7 0.7451497
17.4 0.7451506
17.6 0.745151
18.9 0.7447017
21 0.7466984
23.3 0.7514093
27.3 0.7518119
29.9 0.7521801
33.7 0.7533187
39.8 0.7528142
43.3 0.7196979
48.8 0.7191939
51.4 0.7179226
56.7 0.7206452
63.5 0.727406
76.8 0.7587905
94.4 0.7308073
109.7 0.7339096
124.9 0.733857
140.2 0.7344802
156.7 0.7327795
174.3 0.7339246
200.4 0.7376904
229.7 0.7403424
269.1 0.7656342
313.5 0.745596
359 0.7481948
418.1 0.7503277
488.8 0.7516866
569.8 0.7532229
659.9 0.7547803
760.4 0.7558253
902.2 0.7691544
1073.5 0.7624047
1257.5 0.769667
1449.8 0.7698936
1680.2 0.77051
1974.9 0.7702136
2344.1 0.7707092
2808.3 0.7706215
3371.2 0.7710983
4047.9 0.7713727
4881.3 0.7716039
5928.3 0.7717031
7167.1 0.7720514
8672.8 0.771717
10549.5 0.7713528
12765.5 0.7722954
15471.9 0.7723958
18767.8 0.7714045
22776.4 0.7713745
27560.3 0.7703427
33851.2 0.767505
42228.9 0.7664213
50555.5 0.7650839
66407.4 0.7633705
90759.8 0.7600909
113202.9 0.7566578
131246 0.7536762
146113.4 0.7520726
249508.7 0.7489261
};
\addplot [semithick, crimson2143940]
table {%
1 0.2562895
1 0.2562895
1 0.2562895
1 0.2562895
1 0.2562895
1 0.2562895
1 0.2562895
6.4 0.7436664
8.9 0.7436983
11.4 0.7437076
12.5 0.7437071
14.7 0.7437105
16 0.7437105
16.8 0.7437105
22.9 0.7438677
30.8 0.744889
32.3 0.7492481
45.2 0.7516149
55.9 0.7536453
73.2 0.7556615
94.7 0.7579376
113.4 0.7599639
132.6 0.7617749
150.6 0.763495
179.7 0.7647581
232 0.7658883
310.3 0.7671276
406.9 0.7683644
578.9 0.7694123
784.8 0.7702646
1091 0.7712006
1447.3 0.7722406
1967.6 0.7731001
2703 0.7738744
3612.3 0.7749024
4811.5 0.7759382
6871.3 0.7765983
9889.8 0.7769246
15466.5 0.7770555
25401 0.7758079
33746.7 0.7760897
59603.8 0.767203
77059.3 0.7657525
90662.2 0.7599865
101728.7 0.7556496
111916.7 0.7526769
124374.8 0.7513255
137641.9 0.7507066
152559.1 0.7466194
};
\addplot [semithick, mediumpurple148103189]
table {%
1 0.2562895
1 0.2562895
1 0.2562895
1 0.2562895
1 0.2562895
1 0.2562895
1 0.2562895
1.9 0.2562895
8 0.7437002
14 0.7437095
15.6 0.7437104
17 0.7437105
17 0.7437105
18 0.7437102
19.3 0.7437105
21.1 0.7437105
23.1 0.7437105
29.8 0.743929
34.5 0.7447899
41.4 0.7482232
53.5 0.7503771
63.7 0.7522539
78.1 0.7540605
103.6 0.7558747
132.4 0.7577106
160.2 0.7594204
185.9 0.7609184
205.6 0.7623766
246.3 0.7637267
293.2 0.764781
353.1 0.7657496
429.2 0.7667407
548.2 0.767836
714.5 0.7688309
1003.9 0.7696636
1200.3 0.7704644
1632 0.7712355
2107.9 0.7720072
3123.8 0.7727455
3974.8 0.7734994
5561.2 0.7742985
8196.3 0.7751371
12493.9 0.7760807
20247.1 0.7770362
35235.4 0.7778926
65579.2 0.7787032
186338.8 0.7796473
377822.3 0.7805464
417787 0.7809371
};
\end{axis}

\end{tikzpicture}

%% file: sections/conclusion.tex
\section{Conclusion}
\label{sec:conclusion}
We presented a minimally interactive method, \newowa, for distributed logistic regression, which substantially improves on prior one- or few-shot distributed estimators. Our method scales to massive datasets, with better accuracy-to-sparsity ratio and similar runtimes than other methods, across multi-core and multi-node experiments.

%% file: sections/appendix.tex
\section{Proofs of Statements and Theorems}
\label{app:proofs}

\subsection{Proof of Lemma~\ref{lem:add_points_bound}}

\begin{proof}
By the results of~\citet{samadian2020unconditional},
we require
\begin{equation}
k = \frac{10S'}{\epsilon^2} ((d + 1) \log S' + \log \frac{1}{\delta})
\end{equation}
\noindent which is easily inverted to obtain
\begin{equation} \label{eqn:eps_k_relation}
\epsilon = \sqrt{\frac{10S'}{k} ((d + 1) \log S' + \log \frac{1}{\delta})}
\end{equation}

Similarly, with the coreset $\X^{(+2)}_i$ of size $k + 2$, we have
\begin{equation}
\epsilon^{(2+)} = \sqrt{\frac{10S'}{k + 2} ((d + 1) \log S' + \log \frac{1}{\delta})}
\end{equation}

To obtain the result, multiply the inside term by $(k / k)$ and substitute in Eq.~\ref{eqn:eps_k_relation}.
\end{proof}

\subsection{Proof of Lemma~\ref{lem:centroid_loss_bound}}

\begin{proof}
First, we can show that $\Lw(\X_i) - \barLw(\X_i) \ge 0$.
Observe that by the convexity of $\log(1 + e^{z_k})$ we have
\begin{equation}
\log(1 + e^{z_k}) \ge \log(1 + e^{\bar{z}}) + \sigma(\bar{z}) (z_k - \bar{z})
\end{equation}
\noindent for any $z_k$ where $\bar{z} \colonequals -w^\top \mu_i$.
Plugging this inequality into the original coreset loss:
\begin{align}
\Lw(\X_i) &\ge \sum_{z_k \in Z_i} \log(1 + e^{\bar{z}}) + \sigma(\bar{z}) (z_k - \bar{z}) \\
 &= |\X_i| \log(1 + e^{\bar{z}}) + \sigma(\bar{z}) \sum_{z_k \in Z_i} (z_k - \bar{z}) \\
 &= \barLw(\X_i)
\end{align}
\noindent where the last step follows because the sum of differences to the mean is trivially 0.
Then, using convexity combined with a second-order Taylor expansion of the logistic regression objective,
for any $z_k \in Z_i$,
\begin{equation}
\log(1 + e^{z_k}) - \log(1 + e^{\bar{z}}) \le \sigma(\bar{z}) (z_k - \bar{z}) + \frac{(z_k - \bar{z})^2}{8}
\label{eqn:pre_bound}
\end{equation}
\noindent where $1/4$ is the maximum value of the second derivative of $\log(1 + e^{z_k})$.
Consider now the expansion over $Z_i$ of the following:
\begin{equation}
\Lw(\X_i) - \barLw(\X_i) = \\
\sum_{z_k \in Z_i} \left( \log(1 + e^{z_k}) - \log(1 + e^{\bar{z}}) \right).
\end{equation}

Applying the convexity bound above, 
we can see that
\begin{equation}
\Lw(\X_i) - \barLw(\X_i) = \sum_{z_k \in Z_i}\!\sigma(\bar{z}) (z_k - \bar{z}) + \frac{(z_k - \bar{z})^2}{8}
\end{equation}
The first term zeroes out across all $z_k$.
The second term can then be expressed in terms of the variance.
\end{proof}

\subsection{Proof of Theorem~\ref{thm:aug_loss_bound}}

\begin{proof}
According to our scheme,
\begin{equation}
\!\Lw(\Xaug_i) = \Lw(\X_i)\!+\!\sum_{j \ne i}\!\left(\barLw(\X_j^+)\!+\!\barLw(\X_j^-)\right).\!\!
\end{equation}
Now considering the regularized objective $\regLw(\cdot)$,
since the total weight of points in $\regLw(\X)$ and $\regLw(\Xaug_i)$ are
equal, the regularization weight is equal and cancels, so that
\begin{equation}
\regLw(\X) - \regLw(\Xaug_i) = \Lw(\X) + \Lw(\Xaug_i)
\end{equation}
Expanding $\Lw(\X)$ into its components,
\begin{equation}
\Lw(\X) - \Lw(\Xaug_i) = \Lw(\X_i) - \Lw(\X_i)
 + \sum_{j \ne i} (\Lw(\X_j) - \barLw(\X_j^+) - \barLw(\X_j^-))
\end{equation}
\noindent and by Lemma~\ref{lem:centroid_loss_bound}, we obtain
\begin{equation}
0 \le \Lw(\X) - \Lw(\Xaug_i) \le \frac{p}{p - 1} \frac{n \Vmax}{4p}
\end{equation}
\noindent where all $\Var(Z_i)$ are bounded by $\Vmax$,
and all $|\X_j^+|$ and $|\X_j^-|$ are bounded by $n / p$.
Now considering the coreset loss bound, we have that
\begin{equation}
\frac{|\regLw(\X) - \regLw(\Xaug_i)|}{\regLw(\X)} \le \frac{1}{\regLw(\X)} \frac{p}{p - 1} \frac{n \Vmax}{4p}
\end{equation}
\noindent which simplifies to the final result.
\end{proof}

\subsection{Proof of Theorem~\ref{thm:navg_isoefficiency}}

\begin{proof}
    On average, each partition receives a portion of the dataset with $\mathcal{O}(z/p)$ zeros, so the Lasso solve takes $\mathcal{O}(k z/p)$.
    
    Each set of coefficients needs to be communicated back to the merge node and averaged. This can be performed efficently using an all-to-one reduction in $\log p$ steps with message size $s$. Letting $t_s$ and $t_w$ be startup and communication latency respectively, this merge takes runtime $\log p (t_s + t_w s)$. Dropping dominated terms, this simplifies to give $T_p = \mathcal{O}(k z / p) + t_w s \log p$.

    The isoefficiency curve has the property that $z \propto T_0$~\cite{grama1993isoefficiency}. (In brief we can rewrite the efficiency as $E = 1/(1 + T_0/T_1)$ and observe the conditions at which $E$ remains constant.)
    So we solve for $T_0$:
    \begin{align*}
        T_0 = p T_p  - T_1 &= t_w s p\log p + p \mathcal{O}(kz/p) - \mathcal{O}(kz) \\
            &= t_w s p\log p
    \end{align*}
    and deduce that $z$ should grow proportionally to $s p\log p$ to maintain efficiency.
    
    Since $s \leq d \leq D$, if $n\to\infty$ while $d$ stays constant or is bounded by $D$, then $s$ drops out of the relation.

    In the final regime where $d$ and $n$ grow at the same rate with $z$, then $s = \mathcal{O}(\sqrt{z})$ and we further simplify:
    \begin{align*}
        z &\propto T_0 = t_w \sqrt{z} p \log p
         \propto t_w^2 p^2 \log^2 p
    \end{align*}
    and the resulting isoefficiency function is $\Theta(p^2 \log^2 p)$.
\end{proof}

\subsection{Proof of Theorem~\ref{thm:owa_isoefficiency}}

\begin{proof}
    The distributed step starts with the same Lasso training as above. However, in order for the merge node to obtain the $\hat W$ matrix, this is followed by an all-to-one coefficient broadcast without reduction. Using a sequential broadcast strategy this takes time $p(t_s + t_w s)$. 

    The subsampled dataset $\mathcal{X}_C$ is then multiplied by $\hat{W}$ resulting in a dense $n_c \times p$ matrix. This involves $\mathcal{O}\big(\frac{n_c z p}{n}\big)$ time. To see this, observe that $\mathcal{X}_C \hat{w}_j$ only requires $\mathcal{O}\big(\frac{n_c z}{n}\big)$ operations, and this is repeated $p$ times. This is followed by a Lasso training over the resulting matrix which requires $\mathcal{O}(k n_c p)$ runtime. Since the cross-validation parameters are relatively small constants we assume the cross-validation does not affect runtime outside of constant factors.

    Thus $T_p = \mathcal{O}(k z / p) + pt_w s + \mathcal{O}\big(\frac{n_c z p}{n}\big) + \mathcal{O}(k n_c p)$. Solving for $T_0$:
    \begin{align*}
        T_0 = p T_p - T_1 &=  \mathcal{O}(s p^2) + p \bigg(\mathcal{O}\Big(\frac{n_c z p }{n}\Big) + \mathcal{O}(kn_c p) \bigg) \\
             &=  \mathcal{O}\big(s p^2 + p^2 \frac{n_c z}{n} + p^2 k n_c \Big)
    \end{align*}

    From this we can deduce interesting behaviors from the growth rates of the parameters. The first term implies $z = \Theta(p^2)$ at least. For the second term, if $n_c/n$ is constant (i.e. a fixed fraction subsample), we are forced to have $T_0 \propto p^2 z$ which is a contradiction. Similarly for the third term, if $n_c$ grows at a rate equal to $z$, the same issue applies. When such behavior occurs, isoefficiency is not attainable because the serial overhead grows too fast. 

    Thus to get isoefficiency we require $n_c/n = o(z)$ and $n_c = o(z)$. In the first growth regime, $n \propto z$ so $n_c = o(z)$ satisfies both. In the second regime, $n \propto \sqrt{z}$ and again the second condition is dominant.  

    Thus if $n_c \propto z^{\alpha}$, then the third relation yields $z \propto p^2 z^\alpha$ which simplifies to $z \propto p^{\frac{2}{1-\alpha}}$

    As for the second term, we have $z \propto p^2 z^{1+\alpha} / n$ which implies $z^{-\alpha} \propto p^2/n$ and thus
    $z \propto \Big(\frac{p^2}{n}\Big)^\alpha$. In the first regime on $n$ we get $z^{\alpha + 1} \propto p^{2\alpha}$ giving $z \propto p^{\frac{2\alpha}{\alpha + 1}}$,
    and in the second regime we get $z^{\alpha/2 + 1} = p^{2\alpha}$ which gives $z \propto p^{\frac{4\alpha}{2+\alpha}}$. In either case this term is dominated by the third relation $p^{\frac{2}{1-\alpha}}$ so the casework is complete. Therefore we conclude $$z = \Theta\Big(\max\big\{p^2, p^{\frac{2}{1-\alpha}}\big\}\Big)$$
\end{proof}

\subsection{Proof of Theorem~\ref{thm:acowa_isoefficiency}}

\begin{proof}
\newowa has the same steps as OWA but requires passing centroids and reweighting features. The centroids requires an additional all-to-all broadcast of $2d$ entries, which takes $t_s \log p + 2t_w s(p-1)$ using the optimal hypercube algorithm. The distributed solve on the augmented data is then $\mathcal{O}(z/p + 2pd)$. As for feature-weighting, the information is obtained for free from $\hat{W}$, and the OWA procedure is repeated. 

For the actual centroid-augmented distributed training step itself,
the isoefficiency result remains unchanged:
instead of each model taking $\mathcal{O}(knd/p)$ time to train,
with the augmented centroids it takes time $\mathcal{O}(k(n + 2p)d/p)$.
However, since that cost is identical in both the serial case and the parallel case,
the additional $2p$ points cancel out in the isoefficiency term $pT_p - T_1$
and the result is unchanged.

These are lighter than the most expensive parts of the OWA merge procedure, and so we conclude the asymptotics are the same.
\end{proof}

\subsection{Other things}

\begin{lemma} \label{lem:lasso_problem_size}
   {\it newGLMNET} solves sparse logistic regression with 
   $\mathcal{O}(k z)$ complexity,
   where $k$ a constant relating to the total number of iterations,
   and $z$ is the number of non-zero entries of $\mathcal{X}$ in sparse format or $nd$ if $\mathcal{X}$ is dense.
\end{lemma}
\begin{proof} The sparse logistic problem is solved with newGLMNET with at most $k_o$ outer and $k_i$ inner iterations. These can be specified by the user but generally default to $k_o=100$, $k_i=1000$. Each outer iteration is associated with gradient and Hessian computations which are $\mathcal{O}(nd)$ and each inner iteration involves a $\mathcal{O}(d)$ coordinate descent cycle which updates an $\mathcal{O}(1)$ coefficient update and an $\mathcal{O}(n)$ update to the Hessian-vector product. This gives complexity $\mathcal{O}(k_o nd + k_i \mathcal{O}(nd))$. Furthermore in a sparse dataset any such $nd$ iteration over the dataset is completed in $z$ time, where $z$ is the number of non-zero entries of $\mathcal{X}$. Note also that regardless, $z \leq nd$. This simplifies to $\mathcal{O}(k_o k_i z)$; let $k=k_o k_i$ to complete the proof. 
\end{proof}

\section{Runtime Breakdown}
\label{app:runtime-breakdown}

We also analyzed the time for each step of ACOWA.
Again tuning for 1000 nonzeros,
we ran ACOWA in the fully distributed setting
for 10 trials,
collecting the average runtime of each step
(specifically splitting out communication costs)
in Table~\ref{tab:breakdown}.

\begin{table}
    \centering
    \begin{tabular}{@{}lrrr@{}}
    \toprule
    {\bf step} & {\bf ember-100k} & {\bf ember-1M} & {\bf criteo} \\
    \midrule
    Centroids (4--5)       &  1.786s &   6.724s &   2.021s \\
    All-to-all (6)         &  7.658s &  58.268s & 109.464s \\
    Round 1 (7--8)         &  5.284s &  45.376s &   8.855s \\
    Model gather           &  0.167s &   1.540s &   1.099s \\
    Compute $\alpha_j$ (9) &  0.134s &   1.443s &   1.319s \\
    Round 2 (10--11)       &  4.281s &  29.875s &   7.068s \\
    Model gather           &  0.173s &   1.274s &   0.940s \\
    Round 3 (12--13)       &  0.310s &   0.551s &  19.617s \\
    {\bf Total}            & 19.793s & 145.051s & 150.383s \\
    \bottomrule
    \end{tabular}
    \label{tab:breakdown}
    \vspace*{0.5em}
    \caption{Runtime breakdown for \newowa.  Each step is associated with lines in Alg~\ref{alg:newowa}.}
\end{table}

We can see that the communication portion of \newowa
takes only a modest amount of time (roughly 20\% or less),
and because the communication being performed is only each (sparse) model $\hat{w}_i$,
adding more data (but preserving the sparsity of the solution)
does not affect the communication cost.
Although the centroid computation step is computationally intensive,
it is likely some of this burden could be alleviated by, e.g.,
the use of sparse centroids or other approximations. %
In addition, the use of sparsified centroids would reduce
the amount of data that need to be communicated between each node.

As mentioned by~\citet{izbicki2020distributed},
the last round of learning on the smaller set $(\X_C, \Y_C)$
takes a negligible amount of time compared to the rest of \newowa.

\newpage

\section{Comparing methods with full-data solution}
\label{app:liblinear-compare}

In the following figure we compare distributed methods also with a full-data logistic regression solution at varying sparsities. We fit the full-data model with LIBLINEAR. The other models are the same as in the main paper.

\begin{figure*}[h!]
    \subfigure[{\it newsgroups}, single-node multicore, 256 partitions.]{
        \input{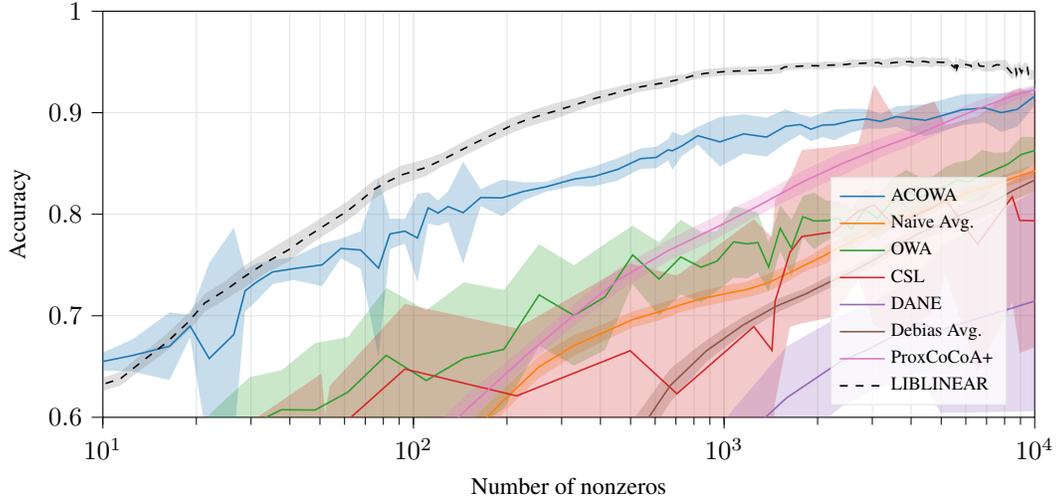}
    }
    \subfigure[{\it amazon7}, single-node multicore, 256 partitions.]{
        \input{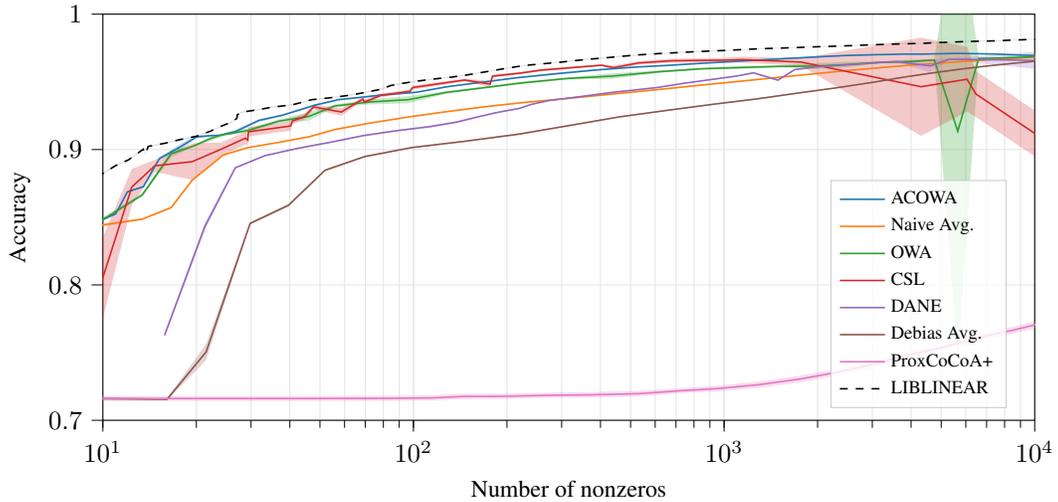}
    }
    \caption{Number of nonzeros vs. test set accuracy in the single-node parallel setting.  Note the significantly better performance of \newowa, especially for sparser solutions.  Although LIBLINEAR generally produces the best results (since it is not a distributed algorithm at all, and uses the full dataset), for the same reason it cannot scale to very large datasets that cannot fit in RAM.}
    \label{fig:nnz-sweep-single-node-liblinear}
\end{figure*}

\newpage

\section{Ablation of ACOWA techniques}
\label{app:ablation}

In the following figure we compare the two methodological improvements in ACOWA separately. Specifically, these are \textit{centroid augmentation} and \textit{feature reweighted second round}. Combined, they form the ACOWA method. With neither, we have the baseline OWA.
Our results show that the combination of both methods is necessary for best accuracy, and thus they are mutually beneficial.

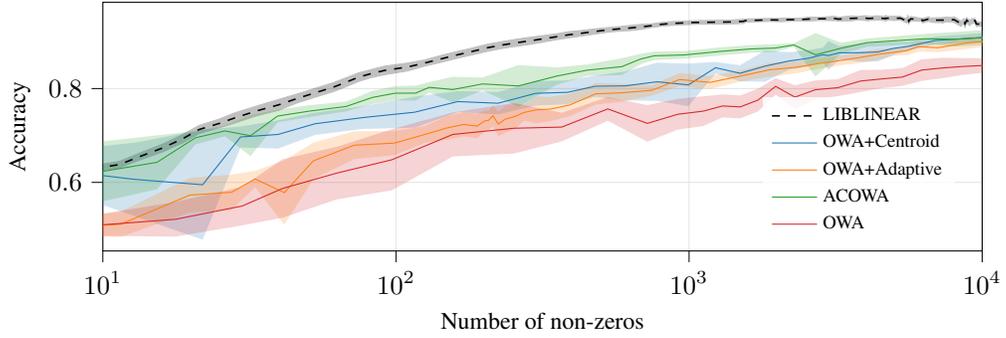
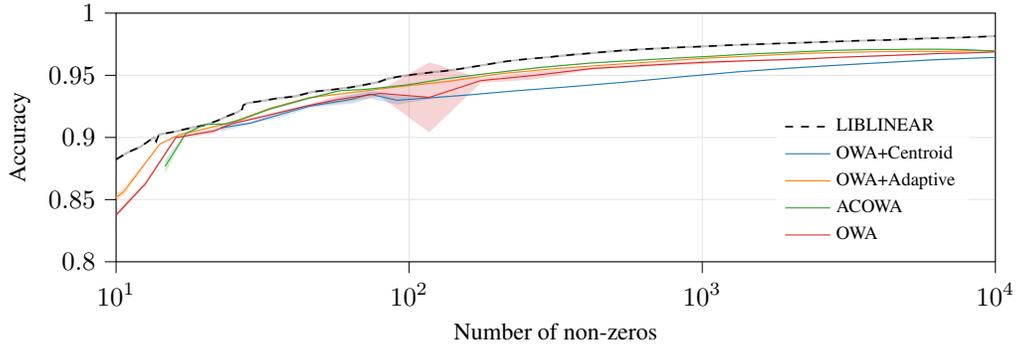
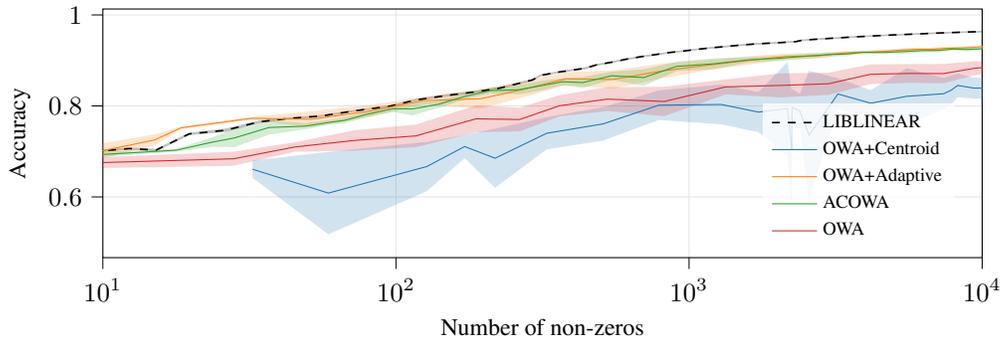
\begin{figure*}[h!]
    \subfigure[{\it newsgroups}, single-node multicore, 256 partitions.]{
        \input{figs/app/ablate_newsgroups.tex}
    }

    \subfigure[{\it amazon7}, single-node multicore, 256 partitions.]{
        \input{figs/app/ablate_amazon7.tex}
    }

    \subfigure[{\it ember100k}, single-node multicore, 256 partitions.]{
        \input{figs/app/ablate_ember100k.tex}
    }
    \caption{Number of nonzeros vs. test set accuracy in the single-node parallel setting. This experiment ablates the individual components of \newowa.}
    \label{fig:ablation}
\end{figure*}

\newpage

\section{Hyperparameter comparison of $\beta$}
\label{app:beta_comparison}

In this section we compare ACOWA at varying values of $\beta$, to assess the robustness of our method to hyperparameter differences. As seen, the feature reweighting is effective for a wide range of $\beta$.

\begin{figure*}[h!]
    \subfigure[{\it newsgroups}, single-node multicore, 256 partitions.]{
        \input{figs/app/beta_newsgroups.tex}
    }

    \subfigure[{\it amazon7}, single-node multicore, 256 partitions.]{
        \input{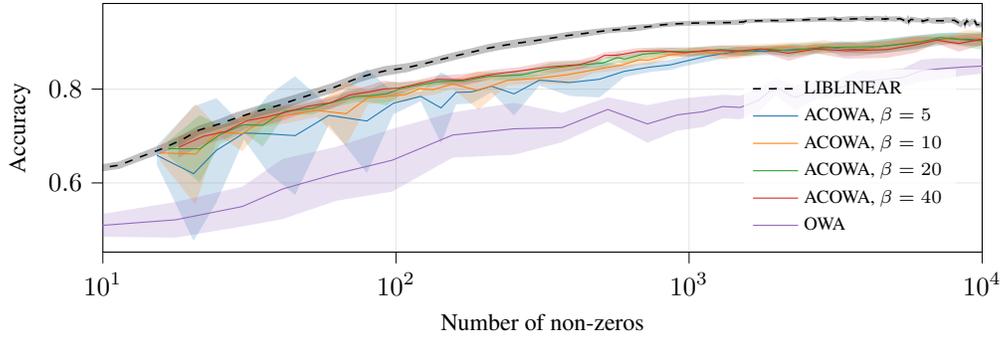}
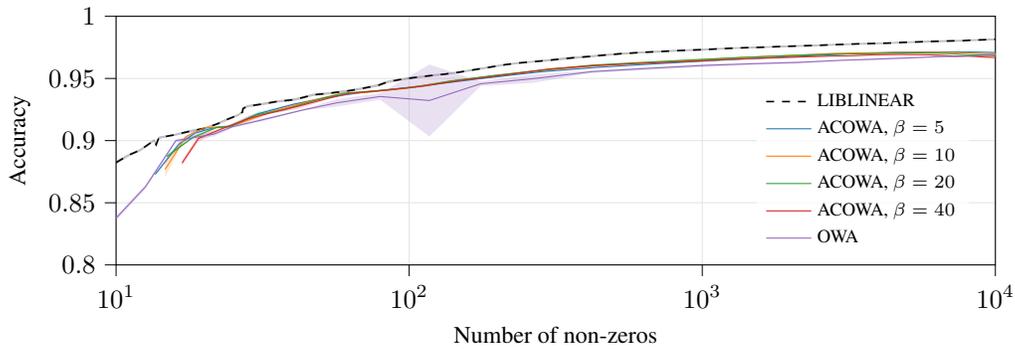
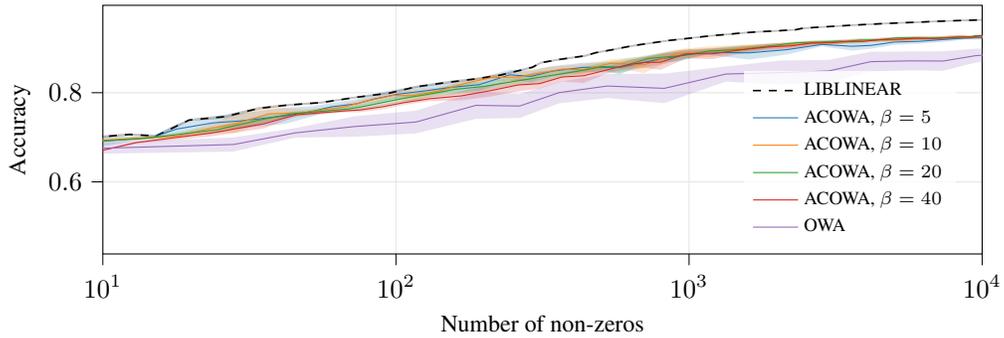
    }

    \subfigure[{\it ember100k}, single-node multicore, 256 partitions.]{
        \input{figs/app/beta_ember100k.tex}
    }
    \caption{Number of nonzeros vs. test set accuracy in the single-node parallel setting. Results compare the performance of ACOWA at varying values of $\beta$.}
    \label{fig:beta_sensitivity}
\end{figure*}

\newpage

\section{Other loss functions: elastic net}
\label{app:elasticnet}

Our general approach can be readily adapted for other loss functions besides $L_1$ penalized logistic regression. 
In this section we demonstrate the effectiveness of our method against baselines,
where all methods solve the Elastic Net regularized version of logistic regression.
We weigh the $L_1$ and $L_2$ components of the regularization equally.

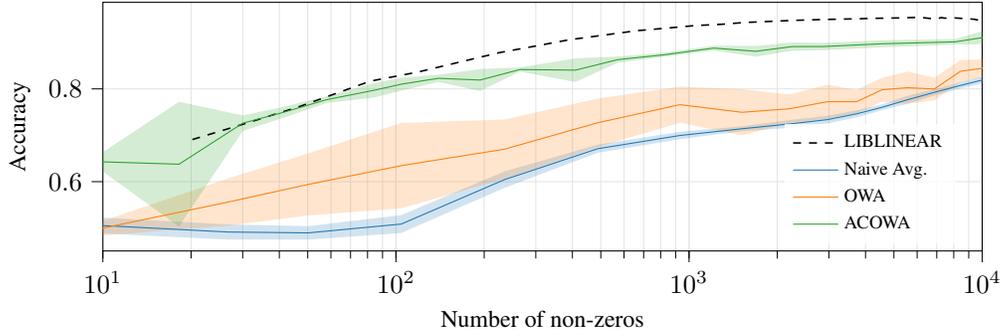
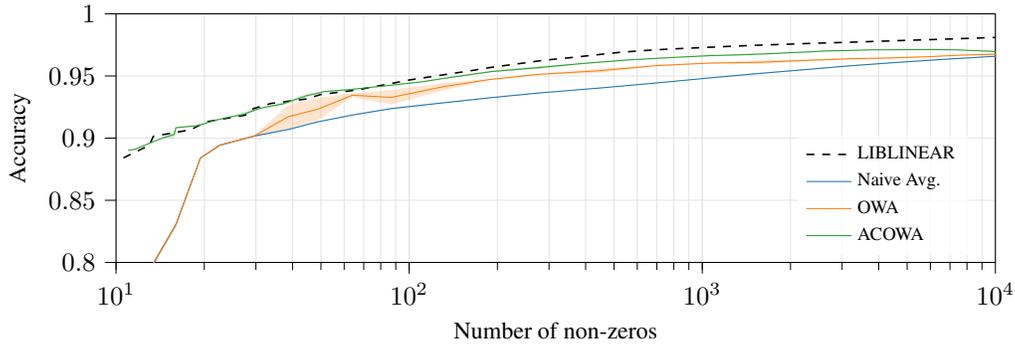
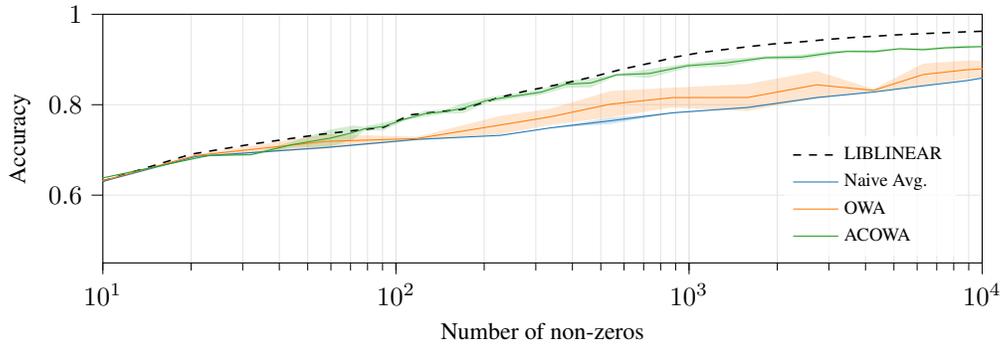
\begin{figure*}[h!]
    \subfigure[{\it newsgroups}, single-node multicore, 256 partitions.]{
        \input{figs/app/elnet_newsgroups_nnz_vs_test_acc.tex}
    }

    \subfigure[{\it amazon7}, single-node multicore, 256 partitions.]{
        \input{figs/app/elnet_amazon7_nnz_vs_test_acc.tex}
    }

    \subfigure[{\it ember100k}, single-node multicore, 256 partitions.]{
        \input{figs/app/elnet_ember100k_nnz_vs_test_acc.tex}
    }
    \caption{Number of nonzeros vs. test set accuracy in the single-node parallel setting, using the elastic net instead of L1-regularized loss. Results are very comparable.}
    \label{fig:elasticnet}
\end{figure*}

%% file: figs/app/ablate_newsgroups.tex
\begin{tikzpicture}[scale=1.0]

\definecolor{crimson2143940}{RGB}{214,39,40}
\definecolor{darkorange25512714}{RGB}{255,127,14}
\definecolor{forestgreen4416044}{RGB}{44,160,44}
\definecolor{gainsboro229}{RGB}{229,229,229}
\definecolor{steelblue31119180}{RGB}{31,119,180}

\begin{axis}[
width=0.95\textwidth,
height=0.35\textwidth,
legend cell align={left},
legend style={
  fill opacity=0.8,
  draw opacity=1,
  text opacity=1,
  at={(0.97,0.03)},
  anchor=south east,
  draw=none
},
log basis x={10},
tick align=outside,
tick pos=left,
x grid style={gainsboro229},
xlabel={\small Number of non-zeros},
xmajorgrids,
xmin=10, xmax=10000,
xminorgrids,
xmode=log,
xtick style={color=black},
xtick={1,10,100,1000,10000,100000},
xticklabels={
  \(\displaystyle {10^{0}}\),
  \(\displaystyle {10^{1}}\),
  \(\displaystyle {10^{2}}\),
  \(\displaystyle {10^{3}}\),
  \(\displaystyle {10^{4}}\),
  \(\displaystyle {10^{5}}\)
},
y grid style={gainsboro229},
ylabel={\small Accuracy},
ymajorgrids,
ymin=0.453803705002681, ymax=0.983827636559857,
yminorgrids,
ytick style={color=black}
]
\path [draw=black, fill=black, opacity=0.25]
(axis cs:7.3,0.624772282907598)
--(axis cs:7.3,0.613204717092402)
--(axis cs:11.4,0.631247485374618)
--(axis cs:13.5,0.648572938214047)
--(axis cs:16,0.664945327847899)
--(axis cs:18.7,0.68385420813806)
--(axis cs:21.3,0.703969053276345)
--(axis cs:24.9,0.716209587409845)
--(axis cs:28.2,0.726607875625531)
--(axis cs:33.4,0.741865913887423)
--(axis cs:39.8,0.753759780437761)
--(axis cs:45.9,0.769052438600199)
--(axis cs:53.5,0.782332362952856)
--(axis cs:61,0.793152448555292)
--(axis cs:67.5,0.80384760022996)
--(axis cs:74.6,0.81515832255594)
--(axis cs:82,0.823062481384222)
--(axis cs:91.4,0.829507132048915)
--(axis cs:109.4,0.837471686134735)
--(axis cs:124.7,0.847529532207815)
--(axis cs:140.9,0.855194840539648)
--(axis cs:160.4,0.863955764618935)
--(axis cs:182.5,0.872029004565359)
--(axis cs:206.1,0.879665349140664)
--(axis cs:229,0.885224251906663)
--(axis cs:256.2,0.88990423081424)
--(axis cs:286.3,0.894067745319329)
--(axis cs:317.6,0.899035476430673)
--(axis cs:350.9,0.903957154462824)
--(axis cs:387.7,0.907923454920331)
--(axis cs:429.3,0.91195119892004)
--(axis cs:476,0.915986056127355)
--(axis cs:522.7,0.919797135860345)
--(axis cs:577.6,0.922420242346947)
--(axis cs:629.6,0.923579017568873)
--(axis cs:687.7,0.926207682605642)
--(axis cs:745,0.930078154011115)
--(axis cs:801.4,0.932766993538885)
--(axis cs:873.1,0.934816417796579)
--(axis cs:1036,0.936936685905346)
--(axis cs:1401.2,0.93777411955527)
--(axis cs:1533.2,0.938835122161917)
--(axis cs:1562.1,0.941636342065576)
--(axis cs:1883.6,0.943173383697857)
--(axis cs:2046.6,0.942501893527464)
--(axis cs:2117.6,0.942920694981242)
--(axis cs:2374.8,0.944503198267039)
--(axis cs:2471.6,0.944015524007796)
--(axis cs:2551.1,0.943909795930695)
--(axis cs:2801.1,0.945370914482137)
--(axis cs:3146.6,0.94470525974192)
--(axis cs:3224.2,0.94399823836618)
--(axis cs:3582.5,0.944379817740586)
--(axis cs:3794.1,0.946515678141047)
--(axis cs:3969.8,0.94530419152968)
--(axis cs:4102.8,0.944337613319276)
--(axis cs:4122.7,0.945271965008471)
--(axis cs:4385.9,0.945871107294661)
--(axis cs:4553.2,0.94538690623264)
--(axis cs:4682.6,0.944352668307144)
--(axis cs:4683.1,0.943812204075351)
--(axis cs:4762.9,0.946027468100693)
--(axis cs:5062.3,0.945079835556285)
--(axis cs:5251.8,0.946304793657303)
--(axis cs:5525.2,0.939933952806881)
--(axis cs:5573.5,0.946729658490001)
--(axis cs:5608.7,0.943227733745367)
--(axis cs:5613.2,0.9379917308298)
--(axis cs:5710.4,0.943732737692214)
--(axis cs:6237.3,0.943334154053545)
--(axis cs:6252.9,0.939918146632955)
--(axis cs:6581,0.943651674567081)
--(axis cs:6639.7,0.942131828344086)
--(axis cs:6807.8,0.940874122961058)
--(axis cs:7319,0.940588779955173)
--(axis cs:7326.6,0.939884563147506)
--(axis cs:7536.8,0.942280720336347)
--(axis cs:7887,0.942011622332943)
--(axis cs:8036.3,0.940405923598633)
--(axis cs:8469.7,0.932698554443028)
--(axis cs:8680.1,0.939623127972319)
--(axis cs:8735.5,0.930962941020383)
--(axis cs:8857,0.940367672795517)
--(axis cs:9067.1,0.933889674144042)
--(axis cs:9213.9,0.941352171540595)
--(axis cs:9215.7,0.938372405782447)
--(axis cs:9465.2,0.940486275769612)
--(axis cs:9546.4,0.932346972754151)
--(axis cs:9997.5,0.932912867750874)
--(axis cs:10796.1,0.930227499809329)
--(axis cs:11340.7,0.930381047836809)
--(axis cs:12327,0.930050106556492)
--(axis cs:13121.2,0.929726655800776)
--(axis cs:13931.2,0.928631400798918)
--(axis cs:14631.9,0.927149742960196)
--(axis cs:15522.7,0.923902297336413)
--(axis cs:16460.8,0.924597029080089)
--(axis cs:17085.6,0.924032097983951)
--(axis cs:18776.2,0.921976617884333)
--(axis cs:20681.6,0.916404289540328)
--(axis cs:23180.2,0.913404379009665)
--(axis cs:24451.9,0.913195190242937)
--(axis cs:26376,0.909814785585069)
--(axis cs:28263.5,0.9071240769576)
--(axis cs:30205.2,0.905533100487785)
--(axis cs:32350.5,0.902434820964977)
--(axis cs:35605.1,0.900793163610903)
--(axis cs:36660.8,0.897215029781394)
--(axis cs:37040.5,0.898469125806124)
--(axis cs:38795,0.896057326736172)
--(axis cs:40263,0.893156866385775)
--(axis cs:41167.8,0.891780085798999)
--(axis cs:41431.4,0.890973259028528)
--(axis cs:42466.8,0.889847184153025)
--(axis cs:43649,0.889462684112021)
--(axis cs:45484.2,0.887977635682561)
--(axis cs:46192.1,0.886390901088182)
--(axis cs:46828.7,0.885790703809368)
--(axis cs:47407,0.885274581719501)
--(axis cs:47909.8,0.885143664391175)
--(axis cs:48345.3,0.884264455880051)
--(axis cs:48345.3,0.898521744119949)
--(axis cs:48345.3,0.898521744119949)
--(axis cs:47909.8,0.899683135608825)
--(axis cs:47407,0.900351018280499)
--(axis cs:46828.7,0.901343296190632)
--(axis cs:46192.1,0.902517698911818)
--(axis cs:45484.2,0.903237964317438)
--(axis cs:43649,0.906100715887979)
--(axis cs:42466.8,0.907491015846975)
--(axis cs:41431.4,0.908583140971472)
--(axis cs:41167.8,0.910793314201001)
--(axis cs:40263,0.911989733614225)
--(axis cs:38795,0.913348273263827)
--(axis cs:37040.5,0.914041674193876)
--(axis cs:36660.8,0.916893170218607)
--(axis cs:35605.1,0.916864436389097)
--(axis cs:32350.5,0.917440779035023)
--(axis cs:30205.2,0.919223099512215)
--(axis cs:28263.5,0.9194067230424)
--(axis cs:26376,0.921507614414931)
--(axis cs:24451.9,0.923007009757063)
--(axis cs:23180.2,0.924661220990335)
--(axis cs:20681.6,0.925299310459672)
--(axis cs:18776.2,0.926559382115667)
--(axis cs:17085.6,0.927875702016049)
--(axis cs:16460.8,0.931658570919911)
--(axis cs:15522.7,0.932708302663587)
--(axis cs:14631.9,0.933897057039804)
--(axis cs:13931.2,0.938094199201082)
--(axis cs:13121.2,0.938330144199224)
--(axis cs:12327,0.937474293443508)
--(axis cs:11340.7,0.937587352163191)
--(axis cs:10796.1,0.943508100190671)
--(axis cs:9997.5,0.941266132249127)
--(axis cs:9546.4,0.942897027245849)
--(axis cs:9465.2,0.949220924230388)
--(axis cs:9215.7,0.949116794217553)
--(axis cs:9213.9,0.948621228459405)
--(axis cs:9067.1,0.945258925855958)
--(axis cs:8857,0.951113927204483)
--(axis cs:8735.5,0.945434458979617)
--(axis cs:8680.1,0.949108072027681)
--(axis cs:8469.7,0.943787645556972)
--(axis cs:8036.3,0.952938876401367)
--(axis cs:7887,0.951599777667057)
--(axis cs:7536.8,0.952218079663653)
--(axis cs:7326.6,0.951508236852493)
--(axis cs:7319,0.951248020044827)
--(axis cs:6807.8,0.950785277038942)
--(axis cs:6639.7,0.950059771655914)
--(axis cs:6581,0.95324272543292)
--(axis cs:6252.9,0.949167853367045)
--(axis cs:6237.3,0.949567445946455)
--(axis cs:5710.4,0.953694262307786)
--(axis cs:5613.2,0.9489646691702)
--(axis cs:5608.7,0.953666466254633)
--(axis cs:5573.5,0.955932141509999)
--(axis cs:5525.2,0.948619847193119)
--(axis cs:5251.8,0.954139006342697)
--(axis cs:5062.3,0.954121364443715)
--(axis cs:4762.9,0.953884131899307)
--(axis cs:4683.1,0.956010195924649)
--(axis cs:4682.6,0.953872931692856)
--(axis cs:4553.2,0.95408109376736)
--(axis cs:4385.9,0.955282092705339)
--(axis cs:4122.7,0.953574834991529)
--(axis cs:4102.8,0.954331386680724)
--(axis cs:3969.8,0.95531680847032)
--(axis cs:3794.1,0.953750721858953)
--(axis cs:3582.5,0.954821582259414)
--(axis cs:3224.2,0.952009161633819)
--(axis cs:3146.6,0.95405274025808)
--(axis cs:2801.1,0.952144685517863)
--(axis cs:2551.1,0.952363604069306)
--(axis cs:2471.6,0.951193075992204)
--(axis cs:2374.8,0.949817801732961)
--(axis cs:2117.6,0.950513105018758)
--(axis cs:2046.6,0.949689706472536)
--(axis cs:1883.6,0.949550816302143)
--(axis cs:1562.1,0.948425857934424)
--(axis cs:1533.2,0.947855077838083)
--(axis cs:1401.2,0.94625428044473)
--(axis cs:1036,0.944784514094654)
--(axis cs:873.1,0.942823182203421)
--(axis cs:801.4,0.940879806461115)
--(axis cs:745,0.937889845988885)
--(axis cs:687.7,0.936880317394359)
--(axis cs:629.6,0.934628782431127)
--(axis cs:577.6,0.931971957653053)
--(axis cs:522.7,0.929714864139655)
--(axis cs:476,0.927137543872644)
--(axis cs:429.3,0.923807601079961)
--(axis cs:387.7,0.921091545079669)
--(axis cs:350.9,0.916451045537176)
--(axis cs:317.6,0.912943123569327)
--(axis cs:286.3,0.909392654680671)
--(axis cs:256.2,0.90494976918576)
--(axis cs:229,0.899868948093337)
--(axis cs:206.1,0.894780050859336)
--(axis cs:182.5,0.886533395434641)
--(axis cs:160.4,0.878546435381065)
--(axis cs:140.9,0.869738559460352)
--(axis cs:124.7,0.860722467792185)
--(axis cs:109.4,0.854364913865265)
--(axis cs:91.4,0.847778267951085)
--(axis cs:82,0.840202918615778)
--(axis cs:74.6,0.83355527744406)
--(axis cs:67.5,0.82286059977004)
--(axis cs:61,0.812348751444708)
--(axis cs:53.5,0.801252437047144)
--(axis cs:45.9,0.7882681613998)
--(axis cs:39.8,0.776852419562239)
--(axis cs:33.4,0.762570886112578)
--(axis cs:28.2,0.748281124374469)
--(axis cs:24.9,0.734722212590155)
--(axis cs:21.3,0.721407946723655)
--(axis cs:18.7,0.70185999186194)
--(axis cs:16,0.681816072152101)
--(axis cs:13.5,0.663583261785953)
--(axis cs:11.4,0.644883914625382)
--(axis cs:7.3,0.624772282907598)
--cycle;

\path [fill=steelblue31119180, fill opacity=0.2]
(axis cs:1,0.535106821687414)
--(axis cs:1,0.510766978312587)
--(axis cs:1,0.510766978312587)
--(axis cs:0.1,0.510766978312587)
--(axis cs:0.1,0.510766978312587)
--(axis cs:1.8,0.496480734682489)
--(axis cs:1.8,0.496480734682489)
--(axis cs:1.8,0.493812950177174)
--(axis cs:1.8,0.495492752932406)
--(axis cs:3.5,0.48949134140171)
--(axis cs:6,0.558050100865839)
--(axis cs:7.5,0.581893059515715)
--(axis cs:10,0.551544777562654)
--(axis cs:12.8,0.528500685860574)
--(axis cs:21.9,0.477895701891644)
--(axis cs:29.5,0.67433495275957)
--(axis cs:39.6,0.671998178545656)
--(axis cs:52.9,0.701083735372947)
--(axis cs:77.2,0.714012099149989)
--(axis cs:115.7,0.716569022570426)
--(axis cs:162.6,0.749734781083845)
--(axis cs:223,0.74487462248415)
--(axis cs:300,0.763777550958073)
--(axis cs:384.6,0.766294372461031)
--(axis cs:478.9,0.782351971377411)
--(axis cs:604.9,0.7748329762404)
--(axis cs:778.5,0.773707610576811)
--(axis cs:994.9,0.762417439918525)
--(axis cs:1231.3,0.83208811330626)
--(axis cs:1492.4,0.815347260195083)
--(axis cs:1814.4,0.814403811438079)
--(axis cs:2193.5,0.840209580104658)
--(axis cs:2545.1,0.846925553338102)
--(axis cs:2856.4,0.857051307089628)
--(axis cs:2985.5,0.839521393580699)
--(axis cs:3259.3,0.862012300612574)
--(axis cs:3812.4,0.857214236424975)
--(axis cs:4403,0.857870101926331)
--(axis cs:4966.1,0.871495449447745)
--(axis cs:5619.5,0.880821066139992)
--(axis cs:6924.6,0.894333917727003)
--(axis cs:9698.8,0.898337124509231)
--(axis cs:12231.1,0.899863502784943)
--(axis cs:14714.2,0.864050010828955)
--(axis cs:17644.7,0.894531386229478)
--(axis cs:21089.6,0.879302688496897)
--(axis cs:24893,0.888935464196173)
--(axis cs:28744.1,0.916026885062137)
--(axis cs:32476.5,0.902476557116311)
--(axis cs:36126.3,0.91613141395317)
--(axis cs:39759.5,0.911623133680167)
--(axis cs:43356,0.918547787550922)
--(axis cs:46312.9,0.926257173535837)
--(axis cs:49156.1,0.928877274426674)
--(axis cs:50571,0.940763142691269)
--(axis cs:51115.9,0.939102401232254)
--(axis cs:51186,0.944531509542746)
--(axis cs:51186.7,0.93990433551308)
--(axis cs:51186.7,0.938930481763779)
--(axis cs:51186.7,0.941778005954773)
--(axis cs:51186.7,0.936925224487837)
--(axis cs:51186.7,0.940268657459461)
--(axis cs:51186.7,0.937022266648942)
--(axis cs:51186.7,0.941544054348116)
--(axis cs:51186.7,0.937510069818059)
--(axis cs:51186.7,0.938934980350485)
--(axis cs:51186.7,0.942407229739929)
--(axis cs:51186.7,0.938013546606636)
--(axis cs:51186.7,0.936760936034155)
--(axis cs:51186.7,0.942343330261199)
--(axis cs:51186.7,0.93810196384485)
--(axis cs:51186.7,0.94167890281613)
--(axis cs:51186.7,0.937259057528691)
--(axis cs:51186.7,0.941000575205654)
--(axis cs:51186.7,0.931479760329105)
--(axis cs:51186.7,0.938142765178675)
--(axis cs:51186.7,0.937256414967294)
--(axis cs:51186.7,0.9404129308533)
--(axis cs:51186.7,0.94099194876567)
--(axis cs:51186.7,0.944718601000914)
--(axis cs:51186.7,0.935366073954398)
--(axis cs:51186.7,0.950880726045602)
--(axis cs:51186.7,0.950880726045602)
--(axis cs:51186.7,0.954394198999086)
--(axis cs:51186.7,0.95616865123433)
--(axis cs:51186.7,0.956481269146699)
--(axis cs:51186.7,0.958839385032706)
--(axis cs:51186.7,0.954581034821325)
--(axis cs:51186.7,0.959735639670895)
--(axis cs:51186.7,0.955361624794346)
--(axis cs:51186.7,0.955730942471309)
--(axis cs:51186.7,0.95486089718387)
--(axis cs:51186.7,0.95222643615515)
--(axis cs:51186.7,0.956148069738801)
--(axis cs:51186.7,0.957116463965844)
--(axis cs:51186.7,0.949475053393364)
--(axis cs:51186.7,0.957592770260071)
--(axis cs:51186.7,0.955120019649514)
--(axis cs:51186.7,0.953705730181941)
--(axis cs:51186.7,0.953043345651884)
--(axis cs:51186.7,0.954637133351058)
--(axis cs:51186.7,0.953608942540539)
--(axis cs:51186.7,0.956153975512164)
--(axis cs:51186.7,0.952543594045227)
--(axis cs:51186.7,0.953527518236221)
--(axis cs:51186.7,0.95361846448692)
--(axis cs:51186,0.956355490457254)
--(axis cs:51115.9,0.955751198767746)
--(axis cs:50571,0.950097457308731)
--(axis cs:49156.1,0.948673925573326)
--(axis cs:46312.9,0.951205026464163)
--(axis cs:43356,0.948443812449078)
--(axis cs:39759.5,0.939042666319834)
--(axis cs:36126.3,0.93888178604683)
--(axis cs:32476.5,0.939049842883689)
--(axis cs:28744.1,0.934461114937863)
--(axis cs:24893,0.925261535803827)
--(axis cs:21089.6,0.923092911503103)
--(axis cs:17644.7,0.923125813770522)
--(axis cs:14714.2,0.912879989171045)
--(axis cs:12231.1,0.920722097215057)
--(axis cs:9698.8,0.917812075490769)
--(axis cs:6924.6,0.910635082272997)
--(axis cs:5619.5,0.899658333860008)
--(axis cs:4966.1,0.899400750552255)
--(axis cs:4403,0.898030498073669)
--(axis cs:3812.4,0.895580763575025)
--(axis cs:3259.3,0.890427899387426)
--(axis cs:2985.5,0.901206206419301)
--(axis cs:2856.4,0.886870292910372)
--(axis cs:2545.1,0.881912246661898)
--(axis cs:2193.5,0.877448019895342)
--(axis cs:1814.4,0.880982188561921)
--(axis cs:1492.4,0.849781739804916)
--(axis cs:1231.3,0.85682048669374)
--(axis cs:994.9,0.853288160081475)
--(axis cs:778.5,0.855307389423189)
--(axis cs:604.9,0.836525023759599)
--(axis cs:478.9,0.827497228622589)
--(axis cs:384.6,0.817645027538969)
--(axis cs:300,0.815991449041927)
--(axis cs:223,0.792569977515851)
--(axis cs:162.6,0.794808618916155)
--(axis cs:115.7,0.782543977429574)
--(axis cs:77.2,0.762385300850012)
--(axis cs:52.9,0.748782864627053)
--(axis cs:39.6,0.732349821454344)
--(axis cs:29.5,0.71918744724043)
--(axis cs:21.9,0.712077698108356)
--(axis cs:12.8,0.684542714139425)
--(axis cs:10,0.677470222437346)
--(axis cs:7.5,0.636208140484285)
--(axis cs:6,0.598737699134162)
--(axis cs:3.5,0.56640945859829)
--(axis cs:1.8,0.519148047067594)
--(axis cs:1.8,0.518786849822826)
--(axis cs:1.8,0.522596665317511)
--(axis cs:1.8,0.522596665317511)
--(axis cs:0.1,0.535106821687414)
--(axis cs:0.1,0.535106821687414)
--(axis cs:1,0.535106821687414)
--(axis cs:1,0.535106821687414)
--cycle;

\path [fill=darkorange25512714, fill opacity=0.2]
(axis cs:1,0.519148047067594)
--(axis cs:1,0.495492752932406)
--(axis cs:2,0.496480734682489)
--(axis cs:2.1,0.495492752932406)
--(axis cs:2.4,0.495751202585388)
--(axis cs:3,0.495492752932406)
--(axis cs:3.6,0.491177526488012)
--(axis cs:6.9,0.488239940526827)
--(axis cs:11.5,0.483531413032034)
--(axis cs:19.8,0.536049997526389)
--(axis cs:27.5,0.541381132286338)
--(axis cs:33.1,0.589740999795921)
--(axis cs:41.6,0.509734334135352)
--(axis cs:52.3,0.607210297209478)
--(axis cs:72.1,0.64955196541818)
--(axis cs:99,0.658285444957347)
--(axis cs:125.7,0.684490235237351)
--(axis cs:149.2,0.690706085129743)
--(axis cs:170.1,0.697757708960655)
--(axis cs:189.1,0.693969943310869)
--(axis cs:200.6,0.708467759647437)
--(axis cs:207,0.711771151068154)
--(axis cs:214.2,0.719482129448878)
--(axis cs:223.5,0.698596268713617)
--(axis cs:235.6,0.705779512023399)
--(axis cs:256,0.714662511400689)
--(axis cs:273.3,0.731343213989961)
--(axis cs:302.4,0.72612559069747)
--(axis cs:336.9,0.730458936274282)
--(axis cs:391.7,0.740851739383182)
--(axis cs:479.8,0.765771363980176)
--(axis cs:588.3,0.771752821983621)
--(axis cs:746.2,0.784773926588896)
--(axis cs:928,0.805181866434597)
--(axis cs:1179.1,0.799141873661885)
--(axis cs:1500.1,0.814959893556884)
--(axis cs:1886,0.825286876579279)
--(axis cs:2333.8,0.83279809536288)
--(axis cs:2800.1,0.842616435433636)
--(axis cs:3305.9,0.849655397505025)
--(axis cs:3896,0.855247108316095)
--(axis cs:4456.2,0.861529268712865)
--(axis cs:5027,0.870209303172668)
--(axis cs:5544.4,0.871117471218624)
--(axis cs:6064.3,0.878787958303339)
--(axis cs:7040.6,0.87731454058492)
--(axis cs:8984,0.884832883902953)
--(axis cs:11622.9,0.894637972757538)
--(axis cs:15007.5,0.891130065405486)
--(axis cs:18425.1,0.908949398650519)
--(axis cs:21154.7,0.913329210605205)
--(axis cs:22465.6,0.909516444157577)
--(axis cs:23536.4,0.919700440291047)
--(axis cs:24602.9,0.915160401233179)
--(axis cs:25644.5,0.920085720500879)
--(axis cs:26720.4,0.917910863883709)
--(axis cs:27774.3,0.914756709174062)
--(axis cs:28825.9,0.907002042866208)
--(axis cs:29855.1,0.915006467054096)
--(axis cs:30889.5,0.917617663824008)
--(axis cs:31914.5,0.909154908084118)
--(axis cs:32923.5,0.918216175279446)
--(axis cs:33884,0.917753050268283)
--(axis cs:34815.8,0.91341958862951)
--(axis cs:35709.6,0.910107229010396)
--(axis cs:36561.4,0.91029836754872)
--(axis cs:37394.2,0.917786466997386)
--(axis cs:38160.6,0.924080982895294)
--(axis cs:38889.1,0.918219663847605)
--(axis cs:39548.2,0.910253793298241)
--(axis cs:40170.2,0.912250716869936)
--(axis cs:40758.8,0.919607372023013)
--(axis cs:41305.8,0.922541508599979)
--(axis cs:41823.8,0.914892254230268)
--(axis cs:42304.7,0.921450219519933)
--(axis cs:42746.4,0.920991369897834)
--(axis cs:43159.1,0.920481622179294)
--(axis cs:43561.7,0.918152329311966)
--(axis cs:43918.7,0.916181580773922)
--(axis cs:44255.5,0.921004652848174)
--(axis cs:44579.9,0.917015737274197)
--(axis cs:44886.4,0.926113676160954)
--(axis cs:45161.5,0.916565959129205)
--(axis cs:45425.9,0.923716107888297)
--(axis cs:45668.5,0.914143077467753)
--(axis cs:45904,0.91754857295288)
--(axis cs:46124.1,0.921907537502716)
--(axis cs:46333,0.918011225634917)
--(axis cs:46520.9,0.919306920776577)
--(axis cs:46710.9,0.921214832385215)
--(axis cs:46884.9,0.92250077355921)
--(axis cs:47054.9,0.921604151074105)
--(axis cs:47209.8,0.924262416379284)
--(axis cs:47360.1,0.919786467428157)
--(axis cs:47505.7,0.915601742818838)
--(axis cs:47635.5,0.918968303886782)
--(axis cs:47761,0.920747448803555)
--(axis cs:47891.9,0.920309987649447)
--(axis cs:48011.2,0.922719981904295)
--(axis cs:48113,0.917148975849216)
--(axis cs:48221.1,0.915215525477715)
--(axis cs:48221.1,0.936869674522285)
--(axis cs:48221.1,0.936869674522285)
--(axis cs:48113,0.936444624150784)
--(axis cs:48011.2,0.937794418095705)
--(axis cs:47891.9,0.937098812350553)
--(axis cs:47761,0.937992751196445)
--(axis cs:47635.5,0.940481096113218)
--(axis cs:47505.7,0.935241057181162)
--(axis cs:47360.1,0.939397132571843)
--(axis cs:47209.8,0.935009783620716)
--(axis cs:47054.9,0.938023248925895)
--(axis cs:46884.9,0.935706826440791)
--(axis cs:46710.9,0.938323567614785)
--(axis cs:46520.9,0.938190879223423)
--(axis cs:46333,0.933364174365083)
--(axis cs:46124.1,0.936743462497284)
--(axis cs:45904,0.93648842704712)
--(axis cs:45668.5,0.933860322532247)
--(axis cs:45425.9,0.936709892111703)
--(axis cs:45161.5,0.936761240870795)
--(axis cs:44886.4,0.936885523839046)
--(axis cs:44579.9,0.935335462725803)
--(axis cs:44255.5,0.937025147151826)
--(axis cs:43918.7,0.936081019226078)
--(axis cs:43561.7,0.937304470688034)
--(axis cs:43159.1,0.934886977820706)
--(axis cs:42746.4,0.939079430102166)
--(axis cs:42304.7,0.936402380480067)
--(axis cs:41823.8,0.932844945769731)
--(axis cs:41305.8,0.937440891400022)
--(axis cs:40758.8,0.936559427976987)
--(axis cs:40170.2,0.934510483130064)
--(axis cs:39548.2,0.935886206701759)
--(axis cs:38889.1,0.933865536152395)
--(axis cs:38160.6,0.937143617104706)
--(axis cs:37394.2,0.935629733002614)
--(axis cs:36561.4,0.93513183245128)
--(axis cs:35709.6,0.928668170989604)
--(axis cs:34815.8,0.93298721137049)
--(axis cs:33884,0.936018349731717)
--(axis cs:32923.5,0.932272024720554)
--(axis cs:31914.5,0.934500891915882)
--(axis cs:30889.5,0.932958936175993)
--(axis cs:29855.1,0.934505332945903)
--(axis cs:28825.9,0.932838357133792)
--(axis cs:27774.3,0.930318690825938)
--(axis cs:26720.4,0.931600936116291)
--(axis cs:25644.5,0.935016279499121)
--(axis cs:24602.9,0.932487798766821)
--(axis cs:23536.4,0.931231359708953)
--(axis cs:22465.6,0.931122355842423)
--(axis cs:21154.7,0.925002189394795)
--(axis cs:18425.1,0.920509401349481)
--(axis cs:15007.5,0.912507734594514)
--(axis cs:11622.9,0.911750827242463)
--(axis cs:8984,0.910464516097047)
--(axis cs:7040.6,0.89668745941508)
--(axis cs:6064.3,0.900271441696661)
--(axis cs:5544.4,0.890816928781376)
--(axis cs:5027,0.884804096827332)
--(axis cs:4456.2,0.885054731287135)
--(axis cs:3896,0.878736891683905)
--(axis cs:3305.9,0.870486602494975)
--(axis cs:2800.1,0.865280364566364)
--(axis cs:2333.8,0.85850610463712)
--(axis cs:1886,0.854926323420721)
--(axis cs:1500.1,0.838545106443116)
--(axis cs:1179.1,0.827122726338115)
--(axis cs:928,0.833238533565403)
--(axis cs:746.2,0.807861273411104)
--(axis cs:588.3,0.810145778016379)
--(axis cs:479.8,0.811690836019824)
--(axis cs:391.7,0.789139260616818)
--(axis cs:336.9,0.782052263725718)
--(axis cs:302.4,0.78336880930253)
--(axis cs:273.3,0.766704986010039)
--(axis cs:256,0.767502288599311)
--(axis cs:235.6,0.763341887976601)
--(axis cs:223.5,0.750028331286383)
--(axis cs:214.2,0.764812670551122)
--(axis cs:207,0.752914048931846)
--(axis cs:200.6,0.753200440352563)
--(axis cs:189.1,0.746136456689131)
--(axis cs:170.1,0.749092491039345)
--(axis cs:149.2,0.741325714870257)
--(axis cs:125.7,0.720212364762649)
--(axis cs:99,0.709327755042653)
--(axis cs:72.1,0.70901063458182)
--(axis cs:52.3,0.684803902790522)
--(axis cs:41.6,0.646609865864648)
--(axis cs:33.1,0.624722000204079)
--(axis cs:27.5,0.616560267713662)
--(axis cs:19.8,0.609646402473611)
--(axis cs:11.5,0.537852586967966)
--(axis cs:6.9,0.519657059473173)
--(axis cs:3.6,0.520446673511988)
--(axis cs:3,0.519148047067594)
--(axis cs:2.4,0.520486797414612)
--(axis cs:2.1,0.519148047067594)
--(axis cs:2,0.522596665317511)
--(axis cs:1,0.519148047067594)
--cycle;

\path [fill=forestgreen4416044, fill opacity=0.2]
(axis cs:1,0.535106821687414)
--(axis cs:1,0.510766978312587)
--(axis cs:1,0.510766978312587)
--(axis cs:0.1,0.510766978312587)
--(axis cs:0.1,0.510766978312587)
--(axis cs:1.4,0.495103562490797)
--(axis cs:1.8,0.495492752932406)
--(axis cs:1.8,0.497414658098345)
--(axis cs:1.8,0.495492752932406)
--(axis cs:3,0.492464557635324)
--(axis cs:5.3,0.54862397312391)
--(axis cs:6.5,0.532278216538844)
--(axis cs:7.8,0.599078090393079)
--(axis cs:9.9,0.558908313413439)
--(axis cs:15.3,0.585226728736278)
--(axis cs:20.7,0.680826551959415)
--(axis cs:26.1,0.698070087548791)
--(axis cs:31.9,0.655951364517091)
--(axis cs:39.5,0.727880473445652)
--(axis cs:50.4,0.739017687361329)
--(axis cs:67.6,0.751389133418129)
--(axis cs:83,0.766894613689569)
--(axis cs:99.3,0.775302630072984)
--(axis cs:116.7,0.774268321876912)
--(axis cs:130,0.791444433882135)
--(axis cs:156.3,0.770511459696345)
--(axis cs:197.4,0.796161664958543)
--(axis cs:261.5,0.761215953070824)
--(axis cs:353.7,0.808062485067055)
--(axis cs:449,0.827038067746259)
--(axis cs:589.2,0.828570213569905)
--(axis cs:684.3,0.851880244669479)
--(axis cs:760.4,0.860855756108767)
--(axis cs:958.3,0.863404276394848)
--(axis cs:1272.2,0.87036005983602)
--(axis cs:1624.9,0.876374117464698)
--(axis cs:1965.8,0.877991135134613)
--(axis cs:2278.2,0.889251512334422)
--(axis cs:2698.7,0.825637027480364)
--(axis cs:3293.3,0.87376089943203)
--(axis cs:4042.2,0.88903448115716)
--(axis cs:6356.1,0.895593710601104)
--(axis cs:8074.8,0.889523147191896)
--(axis cs:9912.5,0.892721609980502)
--(axis cs:12137.8,0.901950449750859)
--(axis cs:14900.7,0.89362731139774)
--(axis cs:18189.5,0.90445861418715)
--(axis cs:21675.1,0.920109913313616)
--(axis cs:24967.5,0.906428401430734)
--(axis cs:27795,0.917202099770978)
--(axis cs:30460.2,0.919443903972286)
--(axis cs:33331.9,0.923835233086738)
--(axis cs:36472.1,0.928517392106802)
--(axis cs:39769.6,0.922361908182448)
--(axis cs:42914.3,0.929669510942519)
--(axis cs:45510.7,0.931015954737134)
--(axis cs:48141.2,0.94068517342407)
--(axis cs:50294.1,0.931013730671656)
--(axis cs:51056.5,0.933288790372201)
--(axis cs:51179.2,0.92578481027045)
--(axis cs:51186.4,0.928873360421957)
--(axis cs:51186.7,0.940151706945696)
--(axis cs:51186.7,0.946340024353085)
--(axis cs:51186.7,0.941361167549773)
--(axis cs:51186.7,0.941947823609994)
--(axis cs:51186.7,0.941853647331409)
--(axis cs:51186.7,0.944535361603468)
--(axis cs:51186.7,0.942344457636739)
--(axis cs:51186.7,0.944081854028507)
--(axis cs:51186.7,0.941373259211236)
--(axis cs:51186.7,0.920519986246244)
--(axis cs:51186.7,0.934838865821617)
--(axis cs:51186.7,0.940263018627854)
--(axis cs:51186.7,0.937636835990594)
--(axis cs:51186.7,0.939778594989582)
--(axis cs:51186.7,0.938073937984021)
--(axis cs:51186.7,0.930315438499019)
--(axis cs:51186.7,0.92509195879878)
--(axis cs:51186.7,0.918424321040357)
--(axis cs:51186.7,0.944430956364883)
--(axis cs:51186.7,0.938448915506527)
--(axis cs:51186.7,0.953387484493474)
--(axis cs:51186.7,0.953387484493474)
--(axis cs:51186.7,0.955568843635117)
--(axis cs:51186.7,0.946615678959643)
--(axis cs:51186.7,0.94598124120122)
--(axis cs:51186.7,0.947679161500981)
--(axis cs:51186.7,0.947551862015979)
--(axis cs:51186.7,0.949130005010418)
--(axis cs:51186.7,0.948343764009405)
--(axis cs:51186.7,0.952460981372146)
--(axis cs:51186.7,0.954247334178383)
--(axis cs:51186.7,0.943544013753756)
--(axis cs:51186.7,0.956940740788764)
--(axis cs:51186.7,0.956893945971494)
--(axis cs:51186.7,0.958631542363261)
--(axis cs:51186.7,0.953068838396532)
--(axis cs:51186.7,0.954685952668591)
--(axis cs:51186.7,0.956366176390006)
--(axis cs:51186.7,0.952516232450227)
--(axis cs:51186.7,0.954192575646915)
--(axis cs:51186.7,0.953282293054304)
--(axis cs:51186.4,0.949121439578043)
--(axis cs:51179.2,0.94874938972955)
--(axis cs:51056.5,0.949852409627799)
--(axis cs:50294.1,0.951861069328344)
--(axis cs:48141.2,0.95292622657593)
--(axis cs:45510.7,0.952746445262865)
--(axis cs:42914.3,0.948946089057481)
--(axis cs:39769.6,0.945517291817553)
--(axis cs:36472.1,0.946726607893198)
--(axis cs:33331.9,0.949545366913262)
--(axis cs:30460.2,0.946927096027714)
--(axis cs:27795,0.941804300229022)
--(axis cs:24967.5,0.936339998569266)
--(axis cs:21675.1,0.942179286686384)
--(axis cs:18189.5,0.92810558581285)
--(axis cs:14900.7,0.92953168860226)
--(axis cs:12137.8,0.924934950249141)
--(axis cs:9912.5,0.924491990019498)
--(axis cs:8074.8,0.920591852808105)
--(axis cs:6356.1,0.916562689398896)
--(axis cs:4042.2,0.90661771884284)
--(axis cs:3293.3,0.897844700567971)
--(axis cs:2698.7,0.918284972519636)
--(axis cs:2278.2,0.896640487665578)
--(axis cs:1965.8,0.894591064865387)
--(axis cs:1624.9,0.892569882535302)
--(axis cs:1272.2,0.88775894016398)
--(axis cs:958.3,0.879718723605152)
--(axis cs:760.4,0.878895443891233)
--(axis cs:684.3,0.874206555330521)
--(axis cs:589.2,0.864685986430095)
--(axis cs:449,0.851400132253741)
--(axis cs:353.7,0.845176114932945)
--(axis cs:261.5,0.850052846929176)
--(axis cs:197.4,0.823536735041457)
--(axis cs:156.3,0.825140540303655)
--(axis cs:130,0.813879166117865)
--(axis cs:116.7,0.806299278123088)
--(axis cs:99.3,0.804910569927016)
--(axis cs:83,0.795128586310431)
--(axis cs:67.6,0.772302066581871)
--(axis cs:50.4,0.764354312638671)
--(axis cs:39.5,0.754994326554349)
--(axis cs:31.9,0.742806235482909)
--(axis cs:26.1,0.721273112451209)
--(axis cs:20.7,0.709413448040584)
--(axis cs:15.3,0.700310071263722)
--(axis cs:9.9,0.687143286586561)
--(axis cs:7.8,0.64031850960692)
--(axis cs:6.5,0.664438383461156)
--(axis cs:5.3,0.59458802687609)
--(axis cs:3,0.570623242364676)
--(axis cs:1.8,0.519148047067594)
--(axis cs:1.8,0.524413341901655)
--(axis cs:1.8,0.519148047067594)
--(axis cs:1.4,0.511551237509203)
--(axis cs:0.1,0.535106821687414)
--(axis cs:0.1,0.535106821687414)
--(axis cs:1,0.535106821687414)
--(axis cs:1,0.535106821687414)
--cycle;

\path [fill=crimson2143940, fill opacity=0.2]
(axis cs:1,0.518832586295629)
--(axis cs:1,0.495808213704371)
--(axis cs:2,0.495808213704371)
--(axis cs:2.1,0.495808213704371)
--(axis cs:2.9,0.492978926247414)
--(axis cs:5.2,0.484618498898933)
--(axis cs:9.7,0.485894553834489)
--(axis cs:17.7,0.484387514049115)
--(axis cs:29.9,0.508378877031672)
--(axis cs:41,0.524525160165317)
--(axis cs:62.7,0.563595723168981)
--(axis cs:97,0.583259620251776)
--(axis cs:156.3,0.655303657168394)
--(axis cs:252.7,0.661453986679824)
--(axis cs:368,0.687396915442286)
--(axis cs:528.1,0.731385817758191)
--(axis cs:721,0.686771195769093)
--(axis cs:914.9,0.712491253470475)
--(axis cs:1115,0.720002919915205)
--(axis cs:1301.5,0.73836934175404)
--(axis cs:1495.6,0.736484421616075)
--(axis cs:1717.9,0.750973503336582)
--(axis cs:1974.7,0.787262566333691)
--(axis cs:2296.8,0.756300312781396)
--(axis cs:2697.2,0.777669881989002)
--(axis cs:3213.5,0.783706897951725)
--(axis cs:3842.1,0.792351588849809)
--(axis cs:4559.1,0.801027946417791)
--(axis cs:5337.6,0.805540238581683)
--(axis cs:6211.6,0.816382353047734)
--(axis cs:7276.8,0.821865667454913)
--(axis cs:8564.9,0.829064295443569)
--(axis cs:10194.2,0.834024607006416)
--(axis cs:12430,0.843328220827342)
--(axis cs:15686.2,0.851328844734122)
--(axis cs:18649.3,0.865586231261125)
--(axis cs:20641.8,0.872453788591509)
--(axis cs:21860.1,0.865803724879153)
--(axis cs:22502.2,0.869005106542775)
--(axis cs:22663.2,0.874381516894749)
--(axis cs:24792.3,0.879172272570441)
--(axis cs:25912.6,0.878573086850859)
--(axis cs:27102.5,0.884762709529417)
--(axis cs:28340.4,0.895542501162263)
--(axis cs:29614.5,0.885662843442851)
--(axis cs:30895.5,0.887495262894273)
--(axis cs:32151.7,0.880167915603333)
--(axis cs:33349.7,0.885332672046915)
--(axis cs:34555,0.878553587284854)
--(axis cs:35664.6,0.886954430531477)
--(axis cs:36766.4,0.887081178592535)
--(axis cs:37774.1,0.882380990657779)
--(axis cs:38726.1,0.889515968881967)
--(axis cs:39612.4,0.885802548140295)
--(axis cs:40456.9,0.893837633841651)
--(axis cs:41211.1,0.882205223861797)
--(axis cs:41898.8,0.880951000976064)
--(axis cs:42511.6,0.884498904558222)
--(axis cs:43091,0.881964177722609)
--(axis cs:43622.6,0.880009222006561)
--(axis cs:44094.9,0.876193412547272)
--(axis cs:44529.1,0.879205799368849)
--(axis cs:44923.4,0.882466058013414)
--(axis cs:45284,0.885435912283279)
--(axis cs:45630.5,0.890399144431096)
--(axis cs:45942.8,0.886990980349165)
--(axis cs:46239.8,0.886126433249472)
--(axis cs:46505.7,0.880318783800646)
--(axis cs:46756.8,0.886182911247296)
--(axis cs:46987,0.888036481832649)
--(axis cs:47200.3,0.876362963277854)
--(axis cs:47399.7,0.883885965132227)
--(axis cs:47586.1,0.878486030454575)
--(axis cs:47761.3,0.884375347548612)
--(axis cs:47926.8,0.883740515385437)
--(axis cs:48069.1,0.887606277608909)
--(axis cs:48201.8,0.885721194857549)
--(axis cs:48327.8,0.890698448506988)
--(axis cs:48447.5,0.879883309864588)
--(axis cs:48563.5,0.89141900526005)
--(axis cs:48673.9,0.894155768004095)
--(axis cs:48673.9,0.907974031995905)
--(axis cs:48673.9,0.907974031995905)
--(axis cs:48563.5,0.91088799473995)
--(axis cs:48447.5,0.905742090135412)
--(axis cs:48327.8,0.906462151493012)
--(axis cs:48201.8,0.911883205142452)
--(axis cs:48069.1,0.911950122391091)
--(axis cs:47926.8,0.912177884614563)
--(axis cs:47761.3,0.905686852451388)
--(axis cs:47586.1,0.900129969545425)
--(axis cs:47399.7,0.909725634867773)
--(axis cs:47200.3,0.906068636722146)
--(axis cs:46987,0.907704518167351)
--(axis cs:46756.8,0.907250888752705)
--(axis cs:46505.7,0.906815016199354)
--(axis cs:46239.8,0.905177766750528)
--(axis cs:45942.8,0.910702219650835)
--(axis cs:45630.5,0.909689655568904)
--(axis cs:45284,0.903739087716721)
--(axis cs:44923.4,0.905555541986586)
--(axis cs:44529.1,0.903225400631151)
--(axis cs:44094.9,0.901002987452728)
--(axis cs:43622.6,0.906060377993439)
--(axis cs:43091,0.905258622277391)
--(axis cs:42511.6,0.909999495441778)
--(axis cs:41898.8,0.909731999023936)
--(axis cs:41211.1,0.913979376138203)
--(axis cs:40456.9,0.913615766158349)
--(axis cs:39612.4,0.906123051859705)
--(axis cs:38726.1,0.909153031118033)
--(axis cs:37774.1,0.902800409342221)
--(axis cs:36766.4,0.909103621407465)
--(axis cs:35664.6,0.913666969468523)
--(axis cs:34555,0.903522412715146)
--(axis cs:33349.7,0.907480127953086)
--(axis cs:32151.7,0.907409684396667)
--(axis cs:30895.5,0.905849737105727)
--(axis cs:29614.5,0.908746956557149)
--(axis cs:28340.4,0.910491098837736)
--(axis cs:27102.5,0.907073890470583)
--(axis cs:25912.6,0.902881913149141)
--(axis cs:24792.3,0.899443727429559)
--(axis cs:22663.2,0.899709083105251)
--(axis cs:22502.2,0.893106493457225)
--(axis cs:21860.1,0.898615275120847)
--(axis cs:20641.8,0.894893411408491)
--(axis cs:18649.3,0.893774968738875)
--(axis cs:15686.2,0.877686355265877)
--(axis cs:12430,0.878322179172658)
--(axis cs:10194.2,0.864732992993584)
--(axis cs:8564.9,0.864990504556431)
--(axis cs:7276.8,0.865711932545088)
--(axis cs:6211.6,0.862322046952266)
--(axis cs:5337.6,0.843350561418317)
--(axis cs:4559.1,0.840320853582209)
--(axis cs:3842.1,0.839591811150191)
--(axis cs:3213.5,0.819842302048276)
--(axis cs:2697.2,0.817449718010998)
--(axis cs:2296.8,0.807142287218604)
--(axis cs:1974.7,0.82223143366631)
--(axis cs:1717.9,0.799781096663418)
--(axis cs:1495.6,0.785432378383925)
--(axis cs:1301.5,0.78745145824596)
--(axis cs:1115,0.783546280084794)
--(axis cs:914.9,0.777925746529525)
--(axis cs:721,0.764781604230907)
--(axis cs:528.1,0.781923782241809)
--(axis cs:368,0.748450684557714)
--(axis cs:252.7,0.769335613320176)
--(axis cs:156.3,0.749132542831606)
--(axis cs:97,0.713457379748224)
--(axis cs:62.7,0.675712276831019)
--(axis cs:41,0.649388039834682)
--(axis cs:29.9,0.590911522968327)
--(axis cs:17.7,0.558292285950885)
--(axis cs:9.7,0.532028846165511)
--(axis cs:5.2,0.514494301101067)
--(axis cs:2.9,0.518023873752586)
--(axis cs:2.1,0.518832586295629)
--(axis cs:2,0.518832586295629)
--(axis cs:1,0.518832586295629)
--cycle;

\addplot [semithick, black, dashed]
table {%
7.3 0.6189885
11.4 0.6380657
13.5 0.6560781
16 0.6733807
18.7 0.6928571
21.3 0.7126885
24.9 0.7254659
28.2 0.7374445
33.4 0.7522184
39.8 0.7653061
45.9 0.7786603
53.5 0.7917924
61 0.8027506
67.5 0.8133541
74.6 0.8243568
82 0.8316327
91.4 0.8386427
109.4 0.8459183
124.7 0.854126
140.9 0.8624667
160.4 0.8712511
182.5 0.8792812
206.1 0.8872227
229 0.8925466
256.2 0.897427
286.3 0.9017302
317.6 0.9059893
350.9 0.9102041
387.7 0.9145075
429.3 0.9178794
476 0.9215618
522.7 0.924756
577.6 0.9271961
629.6 0.9291039
687.7 0.931544
745 0.933984
801.4 0.9368234
873.1 0.9388198
1036 0.9408606
1401.2 0.9420142
1533.2 0.9433451
1562.1 0.9450311
1883.6 0.9463621
2046.6 0.9460958
2117.6 0.9467169
2374.8 0.9471605
2471.6 0.9476043
2551.1 0.9481367
2801.1 0.9487578
3146.6 0.949379
3224.2 0.9480037
3582.5 0.9496007
3794.1 0.9501332
3969.8 0.9503105
4102.8 0.9493345
4122.7 0.9494234
4385.9 0.9505766
4553.2 0.949734
4682.6 0.9491128
4683.1 0.9499112
4762.9 0.9499558
5062.3 0.9496006
5251.8 0.9502219
5525.2 0.9442769
5573.5 0.9513309
5608.7 0.9484471
5613.2 0.9434782
5710.4 0.9487135
6237.3 0.9464508
6252.9 0.944543
6581 0.9484472
6639.7 0.9460958
6807.8 0.9458297
7319 0.9459184
7326.6 0.9456964
7536.8 0.9472494
7887 0.9468057
8036.3 0.9466724
8469.7 0.9382431
8680.1 0.9443656
8735.5 0.9381987
8857 0.9457408
9067.1 0.9395743
9213.9 0.9449867
9215.7 0.9437446
9465.2 0.9448536
9546.4 0.937622
9997.5 0.9370895
10796.1 0.9368678
11340.7 0.9339842
12327 0.9337622
13121.2 0.9340284
13931.2 0.9333628
14631.9 0.9305234
15522.7 0.9283053
16460.8 0.9281278
17085.6 0.9259539
18776.2 0.924268
20681.6 0.9208518
23180.2 0.9190328
24451.9 0.9181011
26376 0.9156612
28263.5 0.9132654
30205.2 0.9123781
32350.5 0.9099378
35605.1 0.9088288
36660.8 0.9070541
37040.5 0.9062554
38795 0.9047028
40263 0.9025733
41167.8 0.9012867
41431.4 0.8997782
42466.8 0.8986691
43649 0.8977817
45484.2 0.8956078
46192.1 0.8944543
46828.7 0.893567
47407 0.8928128
47909.8 0.8924134
48345.3 0.8913931
};
\addlegendentry{\scriptsize LIBLINEAR}
\addplot [steelblue31119180, opacity=1.0]
table {%
1 0.5229369
1 0.5229369
0.1 0.5229369
0.1 0.5229369
1.8 0.5095387
1.8 0.5095387
1.8 0.5062999
1.8 0.5073204
3.5 0.5279504
6 0.5783939
7.5 0.6090506
10 0.6145075
12.8 0.6065217
21.9 0.5949867
29.5 0.6967612
39.6 0.702174
52.9 0.7249333
77.2 0.7381987
115.7 0.7495565
162.6 0.7722717
223 0.7687223
300 0.7898845
384.6 0.7919697
478.9 0.8049246
604.9 0.805679
778.5 0.8145075
994.9 0.8078528
1231.3 0.8444543
1492.4 0.8325645
1814.4 0.847693
2193.5 0.8588288
2545.1 0.8644189
2856.4 0.8719608
2985.5 0.8703638
3259.3 0.8762201
3812.4 0.8763975
4403 0.8779503
4966.1 0.8854481
5619.5 0.8902397
6924.6 0.9024845
9698.8 0.9080746
12231.1 0.9102928
14714.2 0.888465
17644.7 0.9088286
21089.6 0.9011978
24893 0.9070985
28744.1 0.925244
32476.5 0.9207632
36126.3 0.9275066
39759.5 0.9253329
43356 0.9334958
46312.9 0.9387311
49156.1 0.9387756
50571 0.9454303
51115.9 0.9474268
51186 0.9504435
51186.7 0.9467614
51186.7 0.946229
51186.7 0.9471608
51186.7 0.9465396
51186.7 0.9469388
51186.7 0.9458297
51186.7 0.9472937
51186.7 0.9456079
51186.7 0.9470275
51186.7 0.95
51186.7 0.9437443
51186.7 0.9469387
51186.7 0.9492457
51186.7 0.9451642
51186.7 0.9482699
51186.7 0.946495
51186.7 0.9481811
51186.7 0.9456077
51186.7 0.9463619
51186.7 0.9480479
51186.7 0.9484471
51186.7 0.9485803
51186.7 0.9495564
51186.7 0.9431234
};
\addlegendentry{\scriptsize OWA+Centroid}
\addplot [darkorange25512714, opacity=1.0]
table {%
1 0.5073204
2 0.5095387
2.1 0.5073204
2.4 0.508119
3 0.5073204
3.6 0.5058121
6.9 0.5039485
11.5 0.510692
19.8 0.5728482
27.5 0.5789707
33.1 0.6072315
41.6 0.5781721
52.3 0.6460071
72.1 0.6792813
99 0.6838066
125.7 0.7023513
149.2 0.7160159
170.1 0.7234251
189.1 0.7200532
200.6 0.7308341
207 0.7323426
214.2 0.7421474
223.5 0.7243123
235.6 0.7345607
256 0.7410824
273.3 0.7490241
302.4 0.7547472
336.9 0.7562556
391.7 0.7649955
479.8 0.7887311
588.3 0.7909493
746.2 0.7963176
928 0.8192102
1179.1 0.8131323
1500.1 0.8267525
1886 0.8401066
2333.8 0.8456521
2800.1 0.8539484
3305.9 0.860071
3896 0.866992
4456.2 0.873292
5027 0.8775067
5544.4 0.8809672
6064.3 0.8895297
7040.6 0.887001
8984 0.8976487
11622.9 0.9031944
15007.5 0.9018189
18425.1 0.9147294
21154.7 0.9191657
22465.6 0.9203194
23536.4 0.9254659
24602.9 0.9238241
25644.5 0.927551
26720.4 0.9247559
27774.3 0.9225377
28825.9 0.9199202
29855.1 0.9247559
30889.5 0.9252883
31914.5 0.9218279
32923.5 0.9252441
33884 0.9268857
34815.8 0.9232034
35709.6 0.9193877
36561.4 0.9227151
37394.2 0.9267081
38160.6 0.9306123
38889.1 0.9260426
39548.2 0.92307
40170.2 0.9233806
40758.8 0.9280834
41305.8 0.9299912
41823.8 0.9238686
42304.7 0.9289263
42746.4 0.9300354
43159.1 0.9276843
43561.7 0.9277284
43918.7 0.9261313
44255.5 0.9290149
44579.9 0.9261756
44886.4 0.9314996
45161.5 0.9266636
45425.9 0.930213
45668.5 0.9240017
45904 0.9270185
46124.1 0.9293255
46333 0.9256877
46520.9 0.9287489
46710.9 0.9297692
46884.9 0.9291038
47054.9 0.9298137
47209.8 0.9296361
47360.1 0.9295918
47505.7 0.9254214
47635.5 0.9297247
47761 0.9293701
47891.9 0.9287044
48011.2 0.9302572
48113 0.9267968
48221.1 0.9260426
};
\addlegendentry{\scriptsize OWA+Adaptive}
\addplot [forestgreen4416044, opacity=1.0]
table {%
1 0.5229369
1 0.5229369
0.1 0.5229369
0.1 0.5229369
1.4 0.5033274
1.8 0.5073204
1.8 0.510914
1.8 0.5073204
3 0.5315439
5.3 0.571606
6.5 0.5983583
7.8 0.6196983
9.9 0.6230258
15.3 0.6427684
20.7 0.69512
26.1 0.7096716
31.9 0.6993788
39.5 0.7414374
50.4 0.751686
67.6 0.7618456
83 0.7810116
99.3 0.7901066
116.7 0.7902838
130 0.8026618
156.3 0.797826
197.4 0.8098492
261.5 0.8056344
353.7 0.8266193
449 0.8392191
589.2 0.8466281
684.3 0.8630434
760.4 0.8698756
958.3 0.8715615
1272.2 0.8790595
1624.9 0.884472
1965.8 0.8862911
2278.2 0.892946
2698.7 0.871961
3293.3 0.8858028
4042.2 0.8978261
6356.1 0.9060782
8074.8 0.9050575
9912.5 0.9086068
12137.8 0.9134427
14900.7 0.9115795
18189.5 0.9162821
21675.1 0.9311446
24967.5 0.9213842
27795 0.9295032
30460.2 0.9331855
33331.9 0.9366903
36472.1 0.937622
39769.6 0.9339396
42914.3 0.9393078
45510.7 0.9418812
48141.2 0.9468057
50294.1 0.9414374
51056.5 0.9415706
51179.2 0.9372671
51186.4 0.9389974
51186.7 0.946717
51186.7 0.9502663
51186.7 0.9469387
51186.7 0.949157
51186.7 0.9482698
51186.7 0.9488021
51186.7 0.950488
51186.7 0.9504879
51186.7 0.949157
51186.7 0.932032
51186.7 0.9445431
51186.7 0.946362
51186.7 0.9429903
51186.7 0.9444543
51186.7 0.9428129
51186.7 0.9389973
51186.7 0.9355366
51186.7 0.93252
51186.7 0.9499999
51186.7 0.9459182
};
\addlegendentry{\scriptsize ACOWA}
\addplot [crimson2143940, opacity=1.0]
table {%
1 0.5073204
2 0.5073204
2.1 0.5073204
2.9 0.5055014
5.2 0.4995564
9.7 0.5089617
17.7 0.5213399
29.9 0.5496452
41 0.5869566
62.7 0.619654
97 0.6483585
156.3 0.7022181
252.7 0.7153948
368 0.7179238
528.1 0.7566548
721 0.7257764
914.9 0.7452085
1115 0.7517746
1301.5 0.7629104
1495.6 0.7609584
1717.9 0.7753773
1974.7 0.804747
2296.8 0.7817213
2697.2 0.7975598
3213.5 0.8017746
3842.1 0.8159717
4559.1 0.8206744
5337.6 0.8244454
6211.6 0.8393522
7276.8 0.8437888
8564.9 0.8470274
10194.2 0.8493788
12430 0.8608252
15686.2 0.8645076
18649.3 0.8796806
20641.8 0.8836736
21860.1 0.8822095
22502.2 0.8810558
22663.2 0.8870453
24792.3 0.889308
25912.6 0.8907275
27102.5 0.8959183
28340.4 0.9030168
29614.5 0.8972049
30895.5 0.8966725
32151.7 0.8937888
33349.7 0.8964064
34555 0.891038
35664.6 0.9003107
36766.4 0.8980924
37774.1 0.8925907
38726.1 0.8993345
39612.4 0.8959628
40456.9 0.9037267
41211.1 0.8980923
41898.8 0.8953415
42511.6 0.8972492
43091 0.8936114
43622.6 0.8930348
44094.9 0.8885982
44529.1 0.8912156
44923.4 0.8940108
45284 0.8945875
45630.5 0.9000444
45942.8 0.8988466
46239.8 0.8956521
46505.7 0.8935669
46756.8 0.8967169
46987 0.8978705
47200.3 0.8912158
47399.7 0.8968058
47586.1 0.889308
47761.3 0.8950311
47926.8 0.8979592
48069.1 0.8997782
48201.8 0.8988022
48327.8 0.8985803
48447.5 0.8928127
48563.5 0.9011535
48673.9 0.9010649
};
\addlegendentry{\scriptsize OWA}
\end{axis}

\end{tikzpicture}

%% file: figs/app/ablate_amazon7.tex
\begin{tikzpicture}[scale=1.0]

\definecolor{crimson2143940}{RGB}{214,39,40}
\definecolor{darkorange25512714}{RGB}{255,127,14}
\definecolor{forestgreen4416044}{RGB}{44,160,44}
\definecolor{gainsboro229}{RGB}{229,229,229}
\definecolor{steelblue31119180}{RGB}{31,119,180}

\begin{axis}[
width=0.95\textwidth,
height=0.35\textwidth,
legend cell align={left},
legend style={
  fill opacity=0.8,
  draw opacity=1,
  text opacity=1,
  at={(0.97,0.03)},
  anchor=south east,
  draw=none
},
log basis x={10},
tick align=outside,
tick pos=left,
x grid style={gainsboro229},
xlabel={\small Number of non-zeros},
xmajorgrids,
xmin=10, xmax=10000,
xminorgrids,
xmode=log,
xtick style={color=black},
xtick={1,10,100,1000,10000,100000},
xticklabels={
  \(\displaystyle {10^{0}}\),
  \(\displaystyle {10^{1}}\),
  \(\displaystyle {10^{2}}\),
  \(\displaystyle {10^{3}}\),
  \(\displaystyle {10^{4}}\),
  \(\displaystyle {10^{5}}\)
},
y grid style={gainsboro229},
ylabel={\small Accuracy},
ymajorgrids,
ymin=0.8, ymax=1,
yminorgrids,
ytick style={color=black}
]
\path [draw=black, fill=black, opacity=0.25]
(axis cs:2.6,0.716699550848701)
--(axis cs:2.6,0.715128849151299)
--(axis cs:3.4,0.81952570481517)
--(axis cs:6,0.847445013473959)
--(axis cs:6,0.845808769933628)
--(axis cs:8,0.854536920596075)
--(axis cs:9.4,0.869431274221341)
--(axis cs:9.7,0.879563730167319)
--(axis cs:11,0.887548242846798)
--(axis cs:12.1,0.891904437391186)
--(axis cs:13.5,0.899759869013071)
--(axis cs:13.7,0.896183606024433)
--(axis cs:14,0.901778790225654)
--(axis cs:15.9,0.904308942728128)
--(axis cs:17.7,0.906656842521443)
--(axis cs:20.1,0.909441796219294)
--(axis cs:22.2,0.913063249001839)
--(axis cs:24,0.916631812749045)
--(axis cs:26,0.919858926567373)
--(axis cs:27,0.922604096617512)
--(axis cs:27.1,0.925192701831859)
--(axis cs:28,0.927070910923437)
--(axis cs:31.8,0.929006893841048)
--(axis cs:35,0.930723355863999)
--(axis cs:40.8,0.932504480168738)
--(axis cs:43.9,0.934357213910798)
--(axis cs:46.9,0.936269795486476)
--(axis cs:53.4,0.937623877873884)
--(axis cs:62.5,0.939655360628122)
--(axis cs:70.1,0.941739634154137)
--(axis cs:78.7,0.943990194326653)
--(axis cs:84.5,0.946683964365662)
--(axis cs:91.7,0.948319538934575)
--(axis cs:102.9,0.950105539262445)
--(axis cs:117.1,0.951716983898598)
--(axis cs:132.7,0.953066591352035)
--(axis cs:149.1,0.954726211930287)
--(axis cs:163.3,0.956232613646038)
--(axis cs:181,0.957985681117577)
--(axis cs:198,0.959490679711198)
--(axis cs:214.3,0.960905720098627)
--(axis cs:236.4,0.962089369254567)
--(axis cs:264.5,0.963181417228027)
--(axis cs:299.5,0.964196324003525)
--(axis cs:328.1,0.96507300788429)
--(axis cs:365.6,0.966183024248989)
--(axis cs:410.1,0.967172468568898)
--(axis cs:447.3,0.968032645441353)
--(axis cs:481.3,0.968661406186122)
--(axis cs:530.5,0.969466247481581)
--(axis cs:574,0.970044585902803)
--(axis cs:632.8,0.970709026191011)
--(axis cs:711.5,0.971211509761905)
--(axis cs:784.5,0.971764300274969)
--(axis cs:878.4,0.972313376821374)
--(axis cs:996.5,0.972764137971201)
--(axis cs:1100.9,0.973330519369006)
--(axis cs:1231.1,0.973943435707243)
--(axis cs:1379.6,0.97439651388584)
--(axis cs:1565.3,0.974780448035859)
--(axis cs:1748.1,0.97518469668337)
--(axis cs:1956.4,0.975642333080425)
--(axis cs:2213.9,0.975892716984715)
--(axis cs:2492.8,0.976418614182033)
--(axis cs:2785.7,0.976809215621662)
--(axis cs:3118.3,0.977208230862408)
--(axis cs:3418.6,0.977542682084095)
--(axis cs:3897,0.97780486545608)
--(axis cs:4414.3,0.978215552786547)
--(axis cs:4809.1,0.978656222089594)
--(axis cs:5327.3,0.979009218564834)
--(axis cs:5902,0.97934372737591)
--(axis cs:6739.6,0.979749529741929)
--(axis cs:7728.6,0.980046330962645)
--(axis cs:8257.9,0.980472836877965)
--(axis cs:9120.7,0.980795708880521)
--(axis cs:9918.5,0.981120002476664)
--(axis cs:11064.9,0.98148144954128)
--(axis cs:12058.4,0.98177414380778)
--(axis cs:13497.3,0.982090306136808)
--(axis cs:14368.5,0.982456994088787)
--(axis cs:15724.7,0.982784194055584)
--(axis cs:17438.8,0.98303916833735)
--(axis cs:20057.4,0.983250585984078)
--(axis cs:20905.3,0.983650966892635)
--(axis cs:24531.9,0.983822003577431)
--(axis cs:24893.6,0.983777585346397)
--(axis cs:25972.2,0.984096626217094)
--(axis cs:28043.6,0.984374791051014)
--(axis cs:30764.7,0.984475670421598)
--(axis cs:32627.8,0.984550763307318)
--(axis cs:35673.4,0.984625772200411)
--(axis cs:38077.9,0.985147117865666)
--(axis cs:38625.6,0.985143536075608)
--(axis cs:39702.6,0.98528137886924)
--(axis cs:44319.7,0.985291643262306)
--(axis cs:46249,0.985125682252571)
--(axis cs:51316.9,0.985072692561562)
--(axis cs:54318.4,0.984695366225206)
--(axis cs:56613.4,0.984961513834496)
--(axis cs:58392.8,0.984577824509245)
--(axis cs:65683.6,0.983785757086811)
--(axis cs:67808.1,0.984171536633663)
--(axis cs:70386.5,0.984014928563905)
--(axis cs:73010.4,0.984485404650545)
--(axis cs:73554.2,0.983504254698873)
--(axis cs:73576.3,0.983311828745323)
--(axis cs:75957.2,0.98323617091623)
--(axis cs:82428.4,0.983002308486928)
--(axis cs:85144.2,0.982651191180779)
--(axis cs:90873.8,0.982249982934818)
--(axis cs:96637,0.981694028453485)
--(axis cs:99187.9,0.981304009569564)
--(axis cs:102451.6,0.982080432629551)
--(axis cs:108631.6,0.981579049543343)
--(axis cs:110207.7,0.979630126330863)
--(axis cs:112839.3,0.980523907745537)
--(axis cs:114599.4,0.980225359322637)
--(axis cs:116891.4,0.979762583989588)
--(axis cs:118013.4,0.980762995217109)
--(axis cs:129856.2,0.979366541827896)
--(axis cs:129856.2,0.980066458172104)
--(axis cs:129856.2,0.980066458172104)
--(axis cs:118013.4,0.981367004782891)
--(axis cs:116891.4,0.980596016010412)
--(axis cs:114599.4,0.980876840677363)
--(axis cs:112839.3,0.981390492254463)
--(axis cs:110207.7,0.980455873669137)
--(axis cs:108631.6,0.981894750456656)
--(axis cs:102451.6,0.982576167370449)
--(axis cs:99187.9,0.981767390430436)
--(axis cs:96637,0.982132571546515)
--(axis cs:90873.8,0.982829217065182)
--(axis cs:85144.2,0.983124408819221)
--(axis cs:82428.4,0.983328291513071)
--(axis cs:75957.2,0.98361502908377)
--(axis cs:73576.3,0.983776571254677)
--(axis cs:73554.2,0.983935145301127)
--(axis cs:73010.4,0.984743195349455)
--(axis cs:70386.5,0.984349471436095)
--(axis cs:67808.1,0.984629063366337)
--(axis cs:65683.6,0.984251442913189)
--(axis cs:58392.8,0.984905975490756)
--(axis cs:56613.4,0.985192886165503)
--(axis cs:54318.4,0.985062833774794)
--(axis cs:51316.9,0.985333507438438)
--(axis cs:46249,0.985472917747429)
--(axis cs:44319.7,0.985477956737694)
--(axis cs:39702.6,0.98564322113076)
--(axis cs:38625.6,0.985451063924392)
--(axis cs:38077.9,0.985547682134334)
--(axis cs:35673.4,0.985099827799588)
--(axis cs:32627.8,0.984977436692682)
--(axis cs:30764.7,0.984861529578402)
--(axis cs:28043.6,0.984802408948987)
--(axis cs:25972.2,0.984647173782906)
--(axis cs:24893.6,0.984341814653603)
--(axis cs:24531.9,0.984220796422569)
--(axis cs:20905.3,0.984014833107365)
--(axis cs:20057.4,0.983651414015921)
--(axis cs:17438.8,0.98345483166265)
--(axis cs:15724.7,0.983247005944416)
--(axis cs:14368.5,0.982988005911213)
--(axis cs:13497.3,0.982640893863192)
--(axis cs:12058.4,0.98231205619222)
--(axis cs:11064.9,0.982081550458719)
--(axis cs:9918.5,0.981776797523336)
--(axis cs:9120.7,0.981370491119479)
--(axis cs:8257.9,0.981076563122034)
--(axis cs:7728.6,0.980665469037355)
--(axis cs:6739.6,0.980400470258071)
--(axis cs:5902,0.98006067262409)
--(axis cs:5327.3,0.979592381435166)
--(axis cs:4809.1,0.979297577910406)
--(axis cs:4414.3,0.978786647213454)
--(axis cs:3897,0.97850333454392)
--(axis cs:3418.6,0.978109917915905)
--(axis cs:3118.3,0.977779769137592)
--(axis cs:2785.7,0.977380984378338)
--(axis cs:2492.8,0.976883585817967)
--(axis cs:2213.9,0.976599283015285)
--(axis cs:1956.4,0.976081666919575)
--(axis cs:1748.1,0.97569090331663)
--(axis cs:1565.3,0.975261151964142)
--(axis cs:1379.6,0.97487988611416)
--(axis cs:1231.1,0.974483564292757)
--(axis cs:1100.9,0.973951480630994)
--(axis cs:996.5,0.973501062028799)
--(axis cs:878.4,0.973000623178626)
--(axis cs:784.5,0.972515899725031)
--(axis cs:711.5,0.971983890238095)
--(axis cs:632.8,0.971397573808989)
--(axis cs:574,0.970782414097198)
--(axis cs:530.5,0.970387352518419)
--(axis cs:481.3,0.969556593813878)
--(axis cs:447.3,0.968917354558647)
--(axis cs:410.1,0.968074331431102)
--(axis cs:365.6,0.967211175751011)
--(axis cs:328.1,0.96614179211571)
--(axis cs:299.5,0.965028075996476)
--(axis cs:264.5,0.963787582771973)
--(axis cs:236.4,0.962772030745433)
--(axis cs:214.3,0.961482079901373)
--(axis cs:198,0.960340320288802)
--(axis cs:181,0.958796918882423)
--(axis cs:163.3,0.957107986353962)
--(axis cs:149.1,0.955540188069713)
--(axis cs:132.7,0.953940408647965)
--(axis cs:117.1,0.952602016101402)
--(axis cs:102.9,0.951007860737555)
--(axis cs:91.7,0.949325461065425)
--(axis cs:84.5,0.947855435634338)
--(axis cs:78.7,0.945216205673347)
--(axis cs:70.1,0.942706165845863)
--(axis cs:62.5,0.940641439371878)
--(axis cs:53.4,0.938611722126116)
--(axis cs:46.9,0.937276404513524)
--(axis cs:43.9,0.935557786089202)
--(axis cs:40.8,0.933622319831262)
--(axis cs:35,0.931792444136001)
--(axis cs:31.8,0.930067506158952)
--(axis cs:28,0.928124089076563)
--(axis cs:27.1,0.926192898168141)
--(axis cs:27,0.923390503382488)
--(axis cs:26,0.920858873432628)
--(axis cs:24,0.917666387250954)
--(axis cs:22.2,0.914042550998161)
--(axis cs:20.1,0.910408403780706)
--(axis cs:17.7,0.907963957478557)
--(axis cs:15.9,0.905120257271872)
--(axis cs:14,0.902842809774346)
--(axis cs:13.7,0.897363793975567)
--(axis cs:13.5,0.900742930986929)
--(axis cs:12.1,0.892904962608814)
--(axis cs:11,0.888720957153202)
--(axis cs:9.7,0.881332869832681)
--(axis cs:9.4,0.870846725778659)
--(axis cs:8,0.856058279403925)
--(axis cs:6,0.847757230066371)
--(axis cs:6,0.849100786526041)
--(axis cs:3.4,0.82418089518483)
--(axis cs:2.6,0.716699550848701)
--cycle;

\path [fill=steelblue31119180, fill opacity=0.2]
(axis cs:22.8,0.910552410601488)
--(axis cs:22.8,0.904828989398512)
--(axis cs:28.7,0.910630812479438)
--(axis cs:37.3,0.916870903868158)
--(axis cs:45.7,0.923690860110147)
--(axis cs:55.4,0.925993966959024)
--(axis cs:64.3,0.927303633597181)
--(axis cs:73.3,0.931586029172351)
--(axis cs:91,0.92636698148273)
--(axis cs:123.2,0.931595908906414)
--(axis cs:164.3,0.934030849956622)
--(axis cs:225.8,0.937052654563623)
--(axis cs:346.4,0.940378517233484)
--(axis cs:548.9,0.944191736816856)
--(axis cs:870.5,0.948644108754176)
--(axis cs:1331.6,0.952428216284295)
--(axis cs:1993,0.955120218692922)
--(axis cs:2954.7,0.957809284363479)
--(axis cs:4320.6,0.959960498414436)
--(axis cs:6094.8,0.961778860876723)
--(axis cs:8300.8,0.963331776142032)
--(axis cs:10784.7,0.964332844970918)
--(axis cs:14448.9,0.965074948051724)
--(axis cs:16618.9,0.965055757989094)
--(axis cs:20019.2,0.965017684526578)
--(axis cs:25005.1,0.964612977256538)
--(axis cs:30941.6,0.963936839632583)
--(axis cs:37669.6,0.963190852718184)
--(axis cs:45289.3,0.961995468018447)
--(axis cs:54001.8,0.961112731151522)
--(axis cs:63066.9,0.959750856149082)
--(axis cs:73139.2,0.958693990608953)
--(axis cs:83657.4,0.957933781222297)
--(axis cs:92516,0.957409280964266)
--(axis cs:100085.3,0.95652021907491)
--(axis cs:109289.7,0.955294103235149)
--(axis cs:118537.3,0.953811268388218)
--(axis cs:127041.5,0.952490254787723)
--(axis cs:136105.2,0.951120381755585)
--(axis cs:145647.9,0.950479992337184)
--(axis cs:156697.5,0.947497165980524)
--(axis cs:164222,0.946044825796338)
--(axis cs:172899.1,0.945507569353722)
--(axis cs:182352.1,0.945748260823148)
--(axis cs:192171,0.944198340042383)
--(axis cs:201226.6,0.94392554921411)
--(axis cs:209124.5,0.942965013814313)
--(axis cs:215737.7,0.942448544069946)
--(axis cs:220906.7,0.94251432032204)
--(axis cs:224919.9,0.941774152036265)
--(axis cs:227972.9,0.941561083227139)
--(axis cs:230256,0.941076518885641)
--(axis cs:232046.1,0.940762583526057)
--(axis cs:233400.9,0.939910868337941)
--(axis cs:234394.5,0.940287344279953)
--(axis cs:235470.1,0.940444903058163)
--(axis cs:236050,0.940113825104215)
--(axis cs:236394.7,0.940380410587055)
--(axis cs:236623.7,0.939750332010516)
--(axis cs:236748.5,0.940198141827466)
--(axis cs:236813,0.940122324986622)
--(axis cs:236844.6,0.93979589550979)
--(axis cs:236860.7,0.939028332941484)
--(axis cs:236867.6,0.938853632157327)
--(axis cs:236871,0.938884973728766)
--(axis cs:236871.3,0.938679883745838)
--(axis cs:236871.4,0.935229850225168)
--(axis cs:236871.4,0.935653186127023)
--(axis cs:236871.4,0.935607311530486)
--(axis cs:236871.4,0.936036370603032)
--(axis cs:236871.4,0.935734611326136)
--(axis cs:236871.4,0.936184941409124)
--(axis cs:236871.4,0.936284324727026)
--(axis cs:236871.4,0.93673754520683)
--(axis cs:236871.4,0.936967891253704)
--(axis cs:236871.4,0.937433451531338)
--(axis cs:236871.4,0.937632066756708)
--(axis cs:236871.4,0.93793189785284)
--(axis cs:236871.4,0.938070882138441)
--(axis cs:236871.4,0.936544227610263)
--(axis cs:236871.4,0.935564889693131)
--(axis cs:236871.4,0.936773910306869)
--(axis cs:236871.4,0.936773910306869)
--(axis cs:236871.4,0.937295772389737)
--(axis cs:236871.4,0.939454317861559)
--(axis cs:236871.4,0.939015502147159)
--(axis cs:236871.4,0.938606933243292)
--(axis cs:236871.4,0.938514748468662)
--(axis cs:236871.4,0.938143508746296)
--(axis cs:236871.4,0.93768745479317)
--(axis cs:236871.4,0.937637475272974)
--(axis cs:236871.4,0.937252458590875)
--(axis cs:236871.4,0.937001188673864)
--(axis cs:236871.4,0.936965029396968)
--(axis cs:236871.4,0.936640888469514)
--(axis cs:236871.4,0.936992013872976)
--(axis cs:236871.4,0.936810149774832)
--(axis cs:236871.3,0.939871116254162)
--(axis cs:236871,0.940213026271234)
--(axis cs:236867.6,0.940378567842673)
--(axis cs:236860.7,0.940729667058516)
--(axis cs:236844.6,0.94107250449021)
--(axis cs:236813,0.940947675013378)
--(axis cs:236748.5,0.941511458172534)
--(axis cs:236623.7,0.941867867989484)
--(axis cs:236394.7,0.941703189412945)
--(axis cs:236050,0.941079374895785)
--(axis cs:235470.1,0.941641696941837)
--(axis cs:234394.5,0.942051855720047)
--(axis cs:233400.9,0.941599531662059)
--(axis cs:232046.1,0.942493616473943)
--(axis cs:230256,0.943019481114359)
--(axis cs:227972.9,0.943525516772861)
--(axis cs:224919.9,0.943640447963735)
--(axis cs:220906.7,0.94397707967796)
--(axis cs:215737.7,0.944885055930054)
--(axis cs:209124.5,0.945033386185687)
--(axis cs:201226.6,0.94542385078589)
--(axis cs:192171,0.946758059957617)
--(axis cs:182352.1,0.947703139176852)
--(axis cs:172899.1,0.948197830646278)
--(axis cs:164222,0.948980574203663)
--(axis cs:156697.5,0.950915634019476)
--(axis cs:145647.9,0.953496007662816)
--(axis cs:136105.2,0.954245618244415)
--(axis cs:127041.5,0.955077945212278)
--(axis cs:118537.3,0.956447131611782)
--(axis cs:109289.7,0.957693096764851)
--(axis cs:100085.3,0.95877278092509)
--(axis cs:92516,0.959870519035734)
--(axis cs:83657.4,0.959380018777703)
--(axis cs:73139.2,0.960177209391047)
--(axis cs:63066.9,0.961270543850918)
--(axis cs:54001.8,0.961846068848478)
--(axis cs:45289.3,0.963539931981553)
--(axis cs:37669.6,0.964076347281816)
--(axis cs:30941.6,0.964769560367417)
--(axis cs:25005.1,0.965304822743462)
--(axis cs:20019.2,0.965591515473422)
--(axis cs:16618.9,0.965717242010905)
--(axis cs:14448.9,0.965719251948276)
--(axis cs:10784.7,0.964992955029082)
--(axis cs:8300.8,0.964122823857968)
--(axis cs:6094.8,0.963052339123277)
--(axis cs:4320.6,0.960875701585565)
--(axis cs:2954.7,0.958668115636521)
--(axis cs:1993,0.956379381307078)
--(axis cs:1331.6,0.953270183715705)
--(axis cs:870.5,0.949400691245823)
--(axis cs:548.9,0.944913863183144)
--(axis cs:346.4,0.940947482766516)
--(axis cs:225.8,0.937666745436377)
--(axis cs:164.3,0.934895350043378)
--(axis cs:123.2,0.932779691093586)
--(axis cs:91,0.93346241851727)
--(axis cs:73.3,0.93818677082765)
--(axis cs:64.3,0.934584766402819)
--(axis cs:55.4,0.930801633040976)
--(axis cs:45.7,0.926628739889853)
--(axis cs:37.3,0.921758096131842)
--(axis cs:28.7,0.912352387520562)
--(axis cs:22.8,0.910552410601488)
--cycle;

\path [fill=darkorange25512714, fill opacity=0.2]
(axis cs:9.4,0.850188624299473)
--(axis cs:9.4,0.844200175700527)
--(axis cs:10.6,0.853352394371446)
--(axis cs:14.1,0.894082085256609)
--(axis cs:16.4,0.901449471918562)
--(axis cs:23,0.909540765400189)
--(axis cs:25.6,0.912450645663362)
--(axis cs:33.7,0.923192634948575)
--(axis cs:45.8,0.931457459915957)
--(axis cs:65,0.935718075098771)
--(axis cs:95,0.940729855084135)
--(axis cs:135.7,0.94436204701783)
--(axis cs:206.5,0.950814575042869)
--(axis cs:312,0.95486375682335)
--(axis cs:482.8,0.958135139124832)
--(axis cs:704.5,0.960594278396919)
--(axis cs:979.4,0.962961975309792)
--(axis cs:1274,0.964454078286758)
--(axis cs:1634.4,0.965623745793163)
--(axis cs:1964,0.966596854793135)
--(axis cs:2342.8,0.96732869478667)
--(axis cs:2754.3,0.967731950182277)
--(axis cs:3247.3,0.968220711428677)
--(axis cs:3833.8,0.968530169851589)
--(axis cs:4592,0.968756276775854)
--(axis cs:5652.5,0.968868426531291)
--(axis cs:6451.9,0.968795535540713)
--(axis cs:7218.5,0.969041175225246)
--(axis cs:8587.8,0.969357344101291)
--(axis cs:10453.9,0.969222861534563)
--(axis cs:13024,0.968315766937967)
--(axis cs:15120.3,0.967572210704341)
--(axis cs:15261.6,0.967650007379015)
--(axis cs:17745.1,0.966880622678738)
--(axis cs:21740.4,0.965881465129631)
--(axis cs:26749.1,0.964766962034195)
--(axis cs:32489.9,0.963549647682832)
--(axis cs:33200.6,0.963094226714264)
--(axis cs:44453,0.962051206753415)
--(axis cs:66461.3,0.959647462844093)
--(axis cs:76682.5,0.959133385926857)
--(axis cs:88138,0.958433270045869)
--(axis cs:97629.2,0.957975560650852)
--(axis cs:107901.1,0.957590920625008)
--(axis cs:121561,0.957325665089125)
--(axis cs:142716.8,0.956998966827771)
--(axis cs:163725.4,0.956189217701046)
--(axis cs:181242.6,0.955758684508314)
--(axis cs:194679.4,0.955210431923522)
--(axis cs:204673.8,0.954429437230937)
--(axis cs:212687.1,0.953696007529025)
--(axis cs:219008,0.953082557403157)
--(axis cs:223866,0.952737798131131)
--(axis cs:227430.1,0.952413083173863)
--(axis cs:230052.7,0.952251847008839)
--(axis cs:231910.6,0.952063461508271)
--(axis cs:233290.4,0.951862050444317)
--(axis cs:234272.8,0.951726018872314)
--(axis cs:234964.7,0.951643734334853)
--(axis cs:235465.8,0.951655359486715)
--(axis cs:235834.3,0.951555542898474)
--(axis cs:236106,0.951540434678477)
--(axis cs:236295.5,0.951583032683278)
--(axis cs:236438,0.951478363701139)
--(axis cs:236538.2,0.951447387335646)
--(axis cs:236617.3,0.951468528495491)
--(axis cs:236677.4,0.951456179643747)
--(axis cs:236718.9,0.95147721155826)
--(axis cs:236751.4,0.951466232188506)
--(axis cs:236776.6,0.951444239355949)
--(axis cs:236791.6,0.951497436225028)
--(axis cs:236805.6,0.951450144250437)
--(axis cs:236815,0.951431534525711)
--(axis cs:236821.8,0.951519967976612)
--(axis cs:236827.4,0.951449044790311)
--(axis cs:236831.4,0.951430063772536)
--(axis cs:236834.7,0.951451275424309)
--(axis cs:236836.7,0.9514528404692)
--(axis cs:236839.2,0.951423955183004)
--(axis cs:236840.9,0.951454428608851)
--(axis cs:236841.6,0.951459085349031)
--(axis cs:236841.6,0.952353314650969)
--(axis cs:236841.6,0.952353314650969)
--(axis cs:236840.9,0.952345771391149)
--(axis cs:236839.2,0.952337244816996)
--(axis cs:236836.7,0.9523537595308)
--(axis cs:236834.7,0.952362724575691)
--(axis cs:236831.4,0.952349936227464)
--(axis cs:236827.4,0.952376755209689)
--(axis cs:236821.8,0.952390232023388)
--(axis cs:236815,0.952392265474289)
--(axis cs:236805.6,0.952344855749563)
--(axis cs:236791.6,0.952376963774972)
--(axis cs:236776.6,0.952341960644051)
--(axis cs:236751.4,0.952358167811494)
--(axis cs:236718.9,0.95236458844174)
--(axis cs:236677.4,0.952430020356253)
--(axis cs:236617.3,0.952404671504509)
--(axis cs:236538.2,0.952477412664353)
--(axis cs:236438,0.952379836298861)
--(axis cs:236295.5,0.952401767316722)
--(axis cs:236106,0.952439965321523)
--(axis cs:235834.3,0.952498657101526)
--(axis cs:235465.8,0.952501240513285)
--(axis cs:234964.7,0.952565665665147)
--(axis cs:234272.8,0.952642981127686)
--(axis cs:233290.4,0.952817949555683)
--(axis cs:231910.6,0.952934538491729)
--(axis cs:230052.7,0.953087552991161)
--(axis cs:227430.1,0.953317116826137)
--(axis cs:223866,0.953608201868869)
--(axis cs:219008,0.954050042596843)
--(axis cs:212687.1,0.954658792470975)
--(axis cs:204673.8,0.955353362769063)
--(axis cs:194679.4,0.956117568076478)
--(axis cs:181242.6,0.956562715491686)
--(axis cs:163725.4,0.957308782298954)
--(axis cs:142716.8,0.957730033172229)
--(axis cs:121561,0.958458934910875)
--(axis cs:107901.1,0.958668879374992)
--(axis cs:97629.2,0.959164839349148)
--(axis cs:88138,0.959431529954131)
--(axis cs:76682.5,0.959982614073143)
--(axis cs:66461.3,0.960626737155907)
--(axis cs:44453,0.962648593246585)
--(axis cs:33200.6,0.964154973285736)
--(axis cs:32489.9,0.964550752317168)
--(axis cs:26749.1,0.965563037965805)
--(axis cs:21740.4,0.966518334870369)
--(axis cs:17745.1,0.967777377321262)
--(axis cs:15261.6,0.968678592620985)
--(axis cs:15120.3,0.968773789295658)
--(axis cs:13024,0.969073433062032)
--(axis cs:10453.9,0.969829938465437)
--(axis cs:8587.8,0.970125455898709)
--(axis cs:7218.5,0.969685224774754)
--(axis cs:6451.9,0.969399064459287)
--(axis cs:5652.5,0.969637973468709)
--(axis cs:4592,0.969538723224146)
--(axis cs:3833.8,0.969356630148411)
--(axis cs:3247.3,0.968857888571323)
--(axis cs:2754.3,0.968623449817723)
--(axis cs:2342.8,0.96810150521333)
--(axis cs:1964,0.967373945206865)
--(axis cs:1634.4,0.966401254206837)
--(axis cs:1274,0.965088121713242)
--(axis cs:979.4,0.963921624690208)
--(axis cs:704.5,0.961510321603081)
--(axis cs:482.8,0.958810260875168)
--(axis cs:312,0.95575224317665)
--(axis cs:206.5,0.951891824957131)
--(axis cs:135.7,0.94575095298217)
--(axis cs:95,0.941805144915865)
--(axis cs:65,0.937029324901229)
--(axis cs:45.8,0.932607140084043)
--(axis cs:33.7,0.924082765051425)
--(axis cs:25.6,0.913958154336638)
--(axis cs:23,0.910869034599812)
--(axis cs:16.4,0.902859328081438)
--(axis cs:14.1,0.895174114743391)
--(axis cs:10.6,0.859014205628554)
--(axis cs:9.4,0.850188624299473)
--cycle;

\path [fill=forestgreen4416044, fill opacity=0.2]
(axis cs:14.7,0.882722457923316)
--(axis cs:14.7,0.870724142076684)
--(axis cs:17.3,0.902501343886115)
--(axis cs:20.1,0.909942665471852)
--(axis cs:23.5,0.910263461853932)
--(axis cs:26.6,0.913453199572071)
--(axis cs:33.5,0.922033300976574)
--(axis cs:44.9,0.929951944898231)
--(axis cs:57.9,0.936956030407229)
--(axis cs:71.6,0.938163540794995)
--(axis cs:98.1,0.941620492731152)
--(axis cs:133.2,0.946731555279746)
--(axis cs:187.3,0.95107003498869)
--(axis cs:283.1,0.955714033398992)
--(axis cs:418.7,0.9593327891507)
--(axis cs:598.5,0.961578849367408)
--(axis cs:830.4,0.963502640783644)
--(axis cs:1109.1,0.965082961354035)
--(axis cs:1382,0.96649128880845)
--(axis cs:1681.5,0.967299965593445)
--(axis cs:1979.3,0.967946360571527)
--(axis cs:2320.3,0.968787740884114)
--(axis cs:2814.1,0.969802052422111)
--(axis cs:3259.2,0.969947424483039)
--(axis cs:3810.6,0.97038720855003)
--(axis cs:4502,0.970451339666696)
--(axis cs:5235.6,0.970604588591039)
--(axis cs:6538.4,0.970612842436237)
--(axis cs:8014.4,0.970256200350792)
--(axis cs:9564.2,0.969139942381569)
--(axis cs:10226.5,0.969324084914331)
--(axis cs:11125.7,0.968752064109217)
--(axis cs:13221.5,0.967850080685046)
--(axis cs:16099.8,0.967233004887608)
--(axis cs:19219.6,0.966512439241118)
--(axis cs:22429.2,0.96541634930478)
--(axis cs:28605.5,0.96462288727673)
--(axis cs:37216,0.96343466483781)
--(axis cs:47662.8,0.962006970549528)
--(axis cs:55932.7,0.961257850767043)
--(axis cs:62966.5,0.960309799186776)
--(axis cs:71603.7,0.95906042469023)
--(axis cs:81596,0.957658872467637)
--(axis cs:92995.2,0.955979408150546)
--(axis cs:107866.7,0.954554238691556)
--(axis cs:129571.9,0.952221649474865)
--(axis cs:150922.9,0.950125019485556)
--(axis cs:169751.2,0.948089314271991)
--(axis cs:185427.4,0.946759774170098)
--(axis cs:198217.1,0.945987093623219)
--(axis cs:208122.2,0.94525021963296)
--(axis cs:215699.2,0.944794217462992)
--(axis cs:221413.8,0.944247652067465)
--(axis cs:225623.1,0.943576714031445)
--(axis cs:228684.5,0.942793412802226)
--(axis cs:230928.9,0.94197099127499)
--(axis cs:232543.2,0.940880247371586)
--(axis cs:233696.1,0.940208671819385)
--(axis cs:234524.6,0.939078357327033)
--(axis cs:235111,0.937872696465323)
--(axis cs:235537.8,0.937125558520636)
--(axis cs:235847.9,0.935800210352293)
--(axis cs:236083.4,0.935192058869585)
--(axis cs:236261,0.933599848101509)
--(axis cs:236402.3,0.933309276915662)
--(axis cs:236506.7,0.93271200088876)
--(axis cs:236599.9,0.932451699865521)
--(axis cs:236666.7,0.931792131656945)
--(axis cs:236726.9,0.930856514322121)
--(axis cs:236766.6,0.930415865430364)
--(axis cs:236796.6,0.929120309872901)
--(axis cs:236819,0.928690460659668)
--(axis cs:236834.3,0.926012816286943)
--(axis cs:236844.8,0.926399172740828)
--(axis cs:236854.1,0.924916928661368)
--(axis cs:236860.8,0.923884802471435)
--(axis cs:236865.4,0.922721876427848)
--(axis cs:236867.5,0.92122258546938)
--(axis cs:236869.3,0.920455348463182)
--(axis cs:236869.3,0.919270941234242)
--(axis cs:236870.4,0.918031551583221)
--(axis cs:236870.4,0.919324448416779)
--(axis cs:236870.4,0.919324448416779)
--(axis cs:236869.3,0.920493858765758)
--(axis cs:236869.3,0.921489051536818)
--(axis cs:236867.5,0.922741614530619)
--(axis cs:236865.4,0.924126523572152)
--(axis cs:236860.8,0.925337397528565)
--(axis cs:236854.1,0.926187871338632)
--(axis cs:236844.8,0.927609027259171)
--(axis cs:236834.3,0.929049383713057)
--(axis cs:236819,0.929853739340332)
--(axis cs:236796.6,0.930923890127099)
--(axis cs:236766.6,0.931861934569636)
--(axis cs:236726.9,0.932649685677879)
--(axis cs:236666.7,0.933494868343055)
--(axis cs:236599.9,0.933988500134479)
--(axis cs:236506.7,0.93443099911124)
--(axis cs:236402.3,0.935012523084338)
--(axis cs:236261,0.936003951898492)
--(axis cs:236083.4,0.936508941130415)
--(axis cs:235847.9,0.937379989647708)
--(axis cs:235537.8,0.938253041479364)
--(axis cs:235111,0.939384703534677)
--(axis cs:234524.6,0.940115642672967)
--(axis cs:233696.1,0.941238328180615)
--(axis cs:232543.2,0.941940352628414)
--(axis cs:230928.9,0.94321900872501)
--(axis cs:228684.5,0.943681787197774)
--(axis cs:225623.1,0.944758885968555)
--(axis cs:221413.8,0.945800947932535)
--(axis cs:215699.2,0.945857782537008)
--(axis cs:208122.2,0.94652238036704)
--(axis cs:198217.1,0.947004506376781)
--(axis cs:185427.4,0.948217025829902)
--(axis cs:169751.2,0.949716885728009)
--(axis cs:150922.9,0.951323380514444)
--(axis cs:129571.9,0.953105550525134)
--(axis cs:107866.7,0.955830361308444)
--(axis cs:92995.2,0.957088391849454)
--(axis cs:81596,0.958866327532363)
--(axis cs:71603.7,0.96027077530977)
--(axis cs:62966.5,0.961185600813224)
--(axis cs:55932.7,0.962330149232957)
--(axis cs:47662.8,0.962766429450472)
--(axis cs:37216,0.96410053516219)
--(axis cs:28605.5,0.96580411272327)
--(axis cs:22429.2,0.96657945069522)
--(axis cs:19219.6,0.967374760758882)
--(axis cs:16099.8,0.968151195112392)
--(axis cs:13221.5,0.968818319314953)
--(axis cs:11125.7,0.969690935890783)
--(axis cs:10226.5,0.969942715085669)
--(axis cs:9564.2,0.970096657618431)
--(axis cs:8014.4,0.970937199649208)
--(axis cs:6538.4,0.971335557563763)
--(axis cs:5235.6,0.971331611408961)
--(axis cs:4502,0.971053260333304)
--(axis cs:3810.6,0.97097719144997)
--(axis cs:3259.2,0.970774775516962)
--(axis cs:2814.1,0.970389347577889)
--(axis cs:2320.3,0.969610459115886)
--(axis cs:1979.3,0.968969039428473)
--(axis cs:1681.5,0.968175834406555)
--(axis cs:1382,0.96729091119155)
--(axis cs:1109.1,0.965921638645965)
--(axis cs:830.4,0.964395959216356)
--(axis cs:598.5,0.962516350632592)
--(axis cs:418.7,0.9604876108493)
--(axis cs:283.1,0.957084766601008)
--(axis cs:187.3,0.95203256501131)
--(axis cs:133.2,0.947804044720254)
--(axis cs:98.1,0.942837107268848)
--(axis cs:71.6,0.939354659205005)
--(axis cs:57.9,0.93817076959277)
--(axis cs:44.9,0.932446455101769)
--(axis cs:33.5,0.923642099023426)
--(axis cs:26.6,0.914930600427929)
--(axis cs:23.5,0.911390338146068)
--(axis cs:20.1,0.910985134528148)
--(axis cs:17.3,0.903769456113885)
--(axis cs:14.7,0.882722457923316)
--cycle;

\path [fill=crimson2143940, fill opacity=0.2]
(axis cs:7.5,0.808411538169715)
--(axis cs:7.5,0.803948661830284)
--(axis cs:12.6,0.861155651790486)
--(axis cs:16,0.898675551624732)
--(axis cs:21.6,0.903493600059747)
--(axis cs:24.9,0.910104445725045)
--(axis cs:31.2,0.915252566333817)
--(axis cs:43.8,0.924242518246004)
--(axis cs:56.9,0.92755891976532)
--(axis cs:79.6,0.932955914367734)
--(axis cs:117.2,0.904059327503128)
--(axis cs:175.3,0.943833037105532)
--(axis cs:264,0.946541685841693)
--(axis cs:421.4,0.954469596949273)
--(axis cs:670.1,0.95705111490522)
--(axis cs:1017.2,0.959189190122439)
--(axis cs:1498.4,0.961181352260508)
--(axis cs:2127.8,0.962040260976097)
--(axis cs:2957.7,0.964197366571638)
--(axis cs:3935.3,0.964921821640611)
--(axis cs:5146.9,0.966023466034207)
--(axis cs:6424.7,0.967109925318948)
--(axis cs:7939.8,0.967414484621613)
--(axis cs:9572.1,0.968269791204998)
--(axis cs:11466.4,0.969027487296011)
--(axis cs:13822.3,0.969469619688835)
--(axis cs:16163.4,0.969818396465791)
--(axis cs:18869.5,0.970311624477549)
--(axis cs:22178.3,0.970536032771367)
--(axis cs:25993.6,0.970474099816014)
--(axis cs:30090.4,0.970219824097813)
--(axis cs:34025,0.969809806400767)
--(axis cs:37185,0.969883699879858)
--(axis cs:37196.6,0.969485430969594)
--(axis cs:40026.9,0.969690561607058)
--(axis cs:46320.1,0.968779554983335)
--(axis cs:51862.5,0.967720260065999)
--(axis cs:59933.7,0.966694342909202)
--(axis cs:74694.1,0.964963587564426)
--(axis cs:98506.4,0.961191292212985)
--(axis cs:105518.7,0.959887802375462)
--(axis cs:106588.8,0.959865786095696)
--(axis cs:119687.6,0.958297327352244)
--(axis cs:130264.1,0.957231790775149)
--(axis cs:140222.2,0.956297002152485)
--(axis cs:151145.8,0.95527990955345)
--(axis cs:162958.1,0.954154583312522)
--(axis cs:175814.4,0.952995302586279)
--(axis cs:190607.1,0.951663553200752)
--(axis cs:208589.2,0.949802376790598)
--(axis cs:219732.4,0.948311849300512)
--(axis cs:225301.7,0.94756849560582)
--(axis cs:228610.5,0.947148836143881)
--(axis cs:230849.5,0.946918521426219)
--(axis cs:232487.8,0.946820038598339)
--(axis cs:233650.8,0.946551881544506)
--(axis cs:234506.1,0.946454144363976)
--(axis cs:235127.1,0.946393168654211)
--(axis cs:235591.7,0.94633512348868)
--(axis cs:235912.6,0.946252490242829)
--(axis cs:236158.7,0.946230677383403)
--(axis cs:236335.5,0.946167760029513)
--(axis cs:236462.4,0.946190587939312)
--(axis cs:236557.7,0.9461663364364)
--(axis cs:236634.4,0.946165184204547)
--(axis cs:236685.9,0.946133025805624)
--(axis cs:236726.5,0.946115871746543)
--(axis cs:236755.5,0.946173278127168)
--(axis cs:236774.8,0.946129364141538)
--(axis cs:236791.4,0.946079227220023)
--(axis cs:236802.7,0.946128271779448)
--(axis cs:236812.4,0.946093645506816)
--(axis cs:236820.4,0.946157290093617)
--(axis cs:236826.3,0.946123868731528)
--(axis cs:236830.1,0.946141427981942)
--(axis cs:236832.6,0.946109009119205)
--(axis cs:236835.9,0.946107013232049)
--(axis cs:236837.6,0.946156934383926)
--(axis cs:236839.2,0.946150029933621)
--(axis cs:236839.9,0.946106939955417)
--(axis cs:236840.7,0.946105560662995)
--(axis cs:236840.7,0.947064239337005)
--(axis cs:236840.7,0.947064239337005)
--(axis cs:236839.9,0.947065060044583)
--(axis cs:236839.2,0.947084770066379)
--(axis cs:236837.6,0.947074265616074)
--(axis cs:236835.9,0.94714558676795)
--(axis cs:236832.6,0.947050390880795)
--(axis cs:236830.1,0.947108772018058)
--(axis cs:236826.3,0.947096731268472)
--(axis cs:236820.4,0.947087109906382)
--(axis cs:236812.4,0.947180354493184)
--(axis cs:236802.7,0.947092528220552)
--(axis cs:236791.4,0.947183572779977)
--(axis cs:236774.8,0.947064435858462)
--(axis cs:236755.5,0.947228321872832)
--(axis cs:236726.5,0.947087928253456)
--(axis cs:236685.9,0.947112774194376)
--(axis cs:236634.4,0.947091815795452)
--(axis cs:236557.7,0.9471320635636)
--(axis cs:236462.4,0.947129012060688)
--(axis cs:236335.5,0.947135039970487)
--(axis cs:236158.7,0.947150122616597)
--(axis cs:235912.6,0.947216109757171)
--(axis cs:235591.7,0.94723667651132)
--(axis cs:235127.1,0.947271031345789)
--(axis cs:234506.1,0.947363455636024)
--(axis cs:233650.8,0.947458918455494)
--(axis cs:232487.8,0.947626361401661)
--(axis cs:230849.5,0.947810078573781)
--(axis cs:228610.5,0.948075563856119)
--(axis cs:225301.7,0.94846310439418)
--(axis cs:219732.4,0.949167550699488)
--(axis cs:208589.2,0.950544023209403)
--(axis cs:190607.1,0.952436846799248)
--(axis cs:175814.4,0.953715497413721)
--(axis cs:162958.1,0.954984216687478)
--(axis cs:151145.8,0.95598309044655)
--(axis cs:140222.2,0.957054997847515)
--(axis cs:130264.1,0.958034209224851)
--(axis cs:119687.6,0.959214072647756)
--(axis cs:106588.8,0.960725813904304)
--(axis cs:105518.7,0.960763197624538)
--(axis cs:98506.4,0.962217907787015)
--(axis cs:74694.1,0.965670612435574)
--(axis cs:59933.7,0.967363657090797)
--(axis cs:51862.5,0.968361739934001)
--(axis cs:46320.1,0.969201245016665)
--(axis cs:40026.9,0.970042638392942)
--(axis cs:37196.6,0.969922569030406)
--(axis cs:37185,0.970339300120142)
--(axis cs:34025,0.970274793599233)
--(axis cs:30090.4,0.970684375902187)
--(axis cs:25993.6,0.971106300183986)
--(axis cs:22178.3,0.971320767228633)
--(axis cs:18869.5,0.971174775522451)
--(axis cs:16163.4,0.970749003534209)
--(axis cs:13822.3,0.970217980311164)
--(axis cs:11466.4,0.969897512703989)
--(axis cs:9572.1,0.968973608795002)
--(axis cs:7939.8,0.968362515378387)
--(axis cs:6424.7,0.968019474681052)
--(axis cs:5146.9,0.967103533965793)
--(axis cs:3935.3,0.966189178359389)
--(axis cs:2957.7,0.964879833428362)
--(axis cs:2127.8,0.963768539023903)
--(axis cs:1498.4,0.962514247739492)
--(axis cs:1017.2,0.961755409877561)
--(axis cs:670.1,0.95940488509478)
--(axis cs:421.4,0.956486203050727)
--(axis cs:264,0.953004114158307)
--(axis cs:175.3,0.947580362894468)
--(axis cs:117.2,0.960435872496872)
--(axis cs:79.6,0.937973685632266)
--(axis cs:56.9,0.933375680234679)
--(axis cs:43.8,0.925455281753996)
--(axis cs:31.2,0.917575233666183)
--(axis cs:24.9,0.912172954274955)
--(axis cs:21.6,0.906981999940253)
--(axis cs:16,0.901225848375268)
--(axis cs:12.6,0.864534548209514)
--(axis cs:7.5,0.808411538169715)
--cycle;

\addplot [semithick, black, dashed]
table {%
2.6 0.7159142
3.4 0.8218533
6 0.8482729
6 0.846783
8 0.8552976
9.4 0.870139
9.7 0.8804483
11 0.8881346
12.1 0.8924047
13.5 0.9002514
13.7 0.8967737
14 0.9023108
15.9 0.9047146
17.7 0.9073104
20.1 0.9099251
22.2 0.9135529
24 0.9171491
26 0.9203589
27 0.9229973
27.1 0.9256928
28 0.9275975
31.8 0.9295372
35 0.9312579
40.8 0.9330634
43.9 0.9349575
46.9 0.9367731
53.4 0.9381178
62.5 0.9401484
70.1 0.9422229
78.7 0.9446032
84.5 0.9472697
91.7 0.9488225
102.9 0.9505567
117.1 0.9521595
132.7 0.9535035
149.1 0.9551332
163.3 0.9566703
181 0.9583913
198 0.9599155
214.3 0.9611939
236.4 0.9624307
264.5 0.9634845
299.5 0.9646122
328.1 0.9656074
365.6 0.9666971
410.1 0.9676234
447.3 0.968475
481.3 0.969109
530.5 0.9699268
574 0.9704135
632.8 0.9710533
711.5 0.9715977
784.5 0.9721401
878.4 0.972657
996.5 0.9731326
1100.9 0.973641
1231.1 0.9742135
1379.6 0.9746382
1565.3 0.9750208
1748.1 0.9754378
1956.4 0.975862
2213.9 0.976246
2492.8 0.9766511
2785.7 0.9770951
3118.3 0.977494
3418.6 0.9778263
3897 0.9781541
4414.3 0.9785011
4809.1 0.9789769
5327.3 0.9793008
5902 0.9797022
6739.6 0.980075
7728.6 0.9803559
8257.9 0.9807747
9120.7 0.9810831
9918.5 0.9814484
11064.9 0.9817815
12058.4 0.9820431
13497.3 0.9823656
14368.5 0.9827225
15724.7 0.9830156
17438.8 0.983247
20057.4 0.983451
20905.3 0.9838329
24531.9 0.9840214
24893.6 0.9840597
25972.2 0.9843719
28043.6 0.9845886
30764.7 0.9846686
32627.8 0.9847641
35673.4 0.9848628
38077.9 0.9853474
38625.6 0.9852973
39702.6 0.9854623
44319.7 0.9853848
46249 0.9852993
51316.9 0.9852031
54318.4 0.9848791
56613.4 0.9850772
58392.8 0.9847419
65683.6 0.9840186
67808.1 0.9844003
70386.5 0.9841822
73010.4 0.9846143
73554.2 0.9837197
73576.3 0.9835442
75957.2 0.9834256
82428.4 0.9831653
85144.2 0.9828878
90873.8 0.9825396
96637 0.9819133
99187.9 0.9815357
102451.6 0.9823283
108631.6 0.9817369
110207.7 0.980043
112839.3 0.9809572
114599.4 0.9805511
116891.4 0.9801793
118013.4 0.981065
129856.2 0.9797165
};
\addlegendentry{\scriptsize LIBLINEAR}
\addplot [steelblue31119180, opacity=1.0]
table {%
22.8 0.9076907
28.7 0.9114916
37.3 0.9193145
45.7 0.9251598
55.4 0.9283978
64.3 0.9309442
73.3 0.9348864
91 0.9299147
123.2 0.9321878
164.3 0.9344631
225.8 0.9373597
346.4 0.940663
548.9 0.9445528
870.5 0.9490224
1331.6 0.9528492
1993 0.9557498
2954.7 0.9582387
4320.6 0.9604181
6094.8 0.9624156
8300.8 0.9637273
10784.7 0.9646629
14448.9 0.9653971
16618.9 0.9653865
20019.2 0.9653046
25005.1 0.9649589
30941.6 0.9643532
37669.6 0.9636336
45289.3 0.9627677
54001.8 0.9614794
63066.9 0.9605107
73139.2 0.9594356
83657.4 0.9586569
92516 0.9586399
100085.3 0.9576465
109289.7 0.9564936
118537.3 0.9551292
127041.5 0.9537841
136105.2 0.952683
145647.9 0.951988
156697.5 0.9492064
164222 0.9475127
172899.1 0.9468527
182352.1 0.9467257
192171 0.9454782
201226.6 0.9446747
209124.5 0.9439992
215737.7 0.9436668
220906.7 0.9432457
224919.9 0.9427073
227972.9 0.9425433
230256 0.942048
232046.1 0.9416281
233400.9 0.9407552
234394.5 0.9411696
235470.1 0.9410433
236050 0.9405966
236394.7 0.9410418
236623.7 0.9408091
236748.5 0.9408548
236813 0.940535
236844.6 0.9404342
236860.7 0.939879
236867.6 0.9396161
236871 0.939549
236871.3 0.9392755
236871.4 0.93602
236871.4 0.9363226
236871.4 0.9361241
236871.4 0.9365007
236871.4 0.9363679
236871.4 0.9367187
236871.4 0.9369609
236871.4 0.9372125
236871.4 0.9375557
236871.4 0.9379741
236871.4 0.9381195
236871.4 0.9384737
236871.4 0.9387626
236871.4 0.93692
236871.4 0.9361694
};
\addlegendentry{\scriptsize OWA+Centroid}
\addplot [darkorange25512714, opacity=1.0]
table {%
9.4 0.8471944
10.6 0.8561833
14.1 0.8946281
16.4 0.9021544
23 0.9102049
25.6 0.9132044
33.7 0.9236377
45.8 0.9320323
65 0.9363737
95 0.9412675
135.7 0.9450565
206.5 0.9513532
312 0.955308
482.8 0.9584727
704.5 0.9610523
979.4 0.9634418
1274 0.9647711
1634.4 0.9660125
1964 0.9669854
2342.8 0.9677151
2754.3 0.9681777
3247.3 0.9685393
3833.8 0.9689434
4592 0.9691475
5652.5 0.9692532
6451.9 0.9690973
7218.5 0.9693632
8587.8 0.9697414
10453.9 0.9695264
13024 0.9686946
15120.3 0.968173
15261.6 0.9681643
17745.1 0.967329
21740.4 0.9661999
26749.1 0.965165
32489.9 0.9640502
33200.6 0.9636246
44453 0.9623499
66461.3 0.9601371
76682.5 0.959558
88138 0.9589324
97629.2 0.9585702
107901.1 0.9581299
121561 0.9578923
142716.8 0.9573645
163725.4 0.956749
181242.6 0.9561607
194679.4 0.955664
204673.8 0.9548914
212687.1 0.9541774
219008 0.9535663
223866 0.953173
227430.1 0.9528651
230052.7 0.9526697
231910.6 0.952499
233290.4 0.95234
234272.8 0.9521845
234964.7 0.9521047
235465.8 0.9520783
235834.3 0.9520271
236106 0.9519902
236295.5 0.9519924
236438 0.9519291
236538.2 0.9519624
236617.3 0.9519366
236677.4 0.9519431
236718.9 0.9519209
236751.4 0.9519122
236776.6 0.9518931
236791.6 0.9519372
236805.6 0.9518975
236815 0.9519119
236821.8 0.9519551
236827.4 0.9519129
236831.4 0.95189
236834.7 0.951907
236836.7 0.9519033
236839.2 0.9518806
236840.9 0.9519001
236841.6 0.9519062
};
\addlegendentry{\scriptsize OWA+Adaptive}
\addplot [forestgreen4416044, opacity=1.0]
table {%
14.7 0.8767233
17.3 0.9031354
20.1 0.9104639
23.5 0.9108269
26.6 0.9141919
33.5 0.9228377
44.9 0.9311992
57.9 0.9375634
71.6 0.9387591
98.1 0.9422288
133.2 0.9472678
187.3 0.9515513
283.1 0.9563994
418.7 0.9599102
598.5 0.9620476
830.4 0.9639493
1109.1 0.9655023
1382 0.9668911
1681.5 0.9677379
1979.3 0.9684577
2320.3 0.9691991
2814.1 0.9700957
3259.2 0.9703611
3810.6 0.9706822
4502 0.9707523
5235.6 0.9709681
6538.4 0.9709742
8014.4 0.9705967
9564.2 0.9696183
10226.5 0.9696334
11125.7 0.9692215
13221.5 0.9683342
16099.8 0.9676921
19219.6 0.9669436
22429.2 0.9659979
28605.5 0.9652135
37216 0.9637676
47662.8 0.9623867
55932.7 0.961794
62966.5 0.9607477
71603.7 0.9596656
81596 0.9582626
92995.2 0.9565339
107866.7 0.9551923
129571.9 0.9526636
150922.9 0.9507242
169751.2 0.9489031
185427.4 0.9474884
198217.1 0.9464958
208122.2 0.9458863
215699.2 0.945326
221413.8 0.9450243
225623.1 0.9441678
228684.5 0.9432376
230928.9 0.942595
232543.2 0.9414103
233696.1 0.9407235
234524.6 0.939597
235111 0.9386287
235537.8 0.9376893
235847.9 0.9365901
236083.4 0.9358505
236261 0.9348019
236402.3 0.9341609
236506.7 0.9335715
236599.9 0.9332201
236666.7 0.9326435
236726.9 0.9317531
236766.6 0.9311389
236796.6 0.9300221
236819 0.9292721
236834.3 0.9275311
236844.8 0.9270041
236854.1 0.9255524
236860.8 0.9246111
236865.4 0.9234242
236867.5 0.9219821
236869.3 0.9209722
236869.3 0.9198824
236870.4 0.918678
};
\addlegendentry{\scriptsize ACOWA}
\addplot [crimson2143940, opacity=1.0]
table {%
7.5 0.8061801
12.6 0.8628451
16 0.8999507
21.6 0.9052378
24.9 0.9111387
31.2 0.9164139
43.8 0.9248489
56.9 0.9304673
79.6 0.9354648
117.2 0.9322476
175.3 0.9457067
264 0.9497729
421.4 0.9554779
670.1 0.958228
1017.2 0.9604723
1498.4 0.9618478
2127.8 0.9629044
2957.7 0.9645386
3935.3 0.9655555
5146.9 0.9665635
6424.7 0.9675647
7939.8 0.9678885
9572.1 0.9686217
11466.4 0.9694625
13822.3 0.9698438
16163.4 0.9702837
18869.5 0.9707432
22178.3 0.9709284
25993.6 0.9707902
30090.4 0.9704521
34025 0.9700423
37185 0.9701115
37196.6 0.969704
40026.9 0.9698666
46320.1 0.9689904
51862.5 0.968041
59933.7 0.967029
74694.1 0.9653171
98506.4 0.9617046
105518.7 0.9603255
106588.8 0.9602958
119687.6 0.9587557
130264.1 0.957633
140222.2 0.956676
151145.8 0.9556315
162958.1 0.9545694
175814.4 0.9533554
190607.1 0.9520502
208589.2 0.9501732
219732.4 0.9487397
225301.7 0.9480158
228610.5 0.9476122
230849.5 0.9473643
232487.8 0.9472232
233650.8 0.9470054
234506.1 0.9469088
235127.1 0.9468321
235591.7 0.9467859
235912.6 0.9467343
236158.7 0.9466904
236335.5 0.9466514
236462.4 0.9466598
236557.7 0.9466492
236634.4 0.9466285
236685.9 0.9466229
236726.5 0.9466019
236755.5 0.9467008
236774.8 0.9465969
236791.4 0.9466314
236802.7 0.9466104
236812.4 0.946637
236820.4 0.9466222
236826.3 0.9466103
236830.1 0.9466251
236832.6 0.9465797
236835.9 0.9466263
236837.6 0.9466156
236839.2 0.9466174
236839.9 0.946586
236840.7 0.9465849
};
\addlegendentry{\scriptsize OWA}
\end{axis}

\end{tikzpicture}

%% file: figs/app/ablate_ember100k.tex
\begin{tikzpicture}[scale=1.0]

\definecolor{crimson2143940}{RGB}{214,39,40}
\definecolor{darkorange25512714}{RGB}{255,127,14}
\definecolor{forestgreen4416044}{RGB}{44,160,44}
\definecolor{gainsboro229}{RGB}{229,229,229}
\definecolor{steelblue31119180}{RGB}{31,119,180}

\begin{axis}[
width=0.95\textwidth,
height=0.35\textwidth,
legend cell align={left},
legend style={
  fill opacity=0.8,
  draw opacity=1,
  text opacity=1,
  at={(0.97,0.03)},
  anchor=south east,
  draw=none
},
log basis x={10},
tick align=outside,
tick pos=left,
x grid style={gainsboro229},
xlabel={\small Number of non-zeros},
xmajorgrids,
xmin=10, xmax=10000,
xminorgrids,
xmode=log,
xtick style={color=black},
xtick={1,10,100,1000,10000,100000},
xticklabels={
  \(\displaystyle {10^{0}}\),
  \(\displaystyle {10^{1}}\),
  \(\displaystyle {10^{2}}\),
  \(\displaystyle {10^{3}}\),
  \(\displaystyle {10^{4}}\),
  \(\displaystyle {10^{5}}\)
},
y grid style={gainsboro229},
ylabel={\small Accuracy},
ymajorgrids,
ymin=0.466553014712844, ymax=1.01306780701849,
yminorgrids,
ytick style={color=black}
]
\path [draw=black, fill=black, opacity=0.25]
(axis cs:1,0.500981347719775)
--(axis cs:1,0.498865452280225)
--(axis cs:1,0.498865452280225)
--(axis cs:4.6,0.689452662709201)
--(axis cs:5.8,0.689445789362906)
--(axis cs:7.5,0.692916890316155)
--(axis cs:12.3,0.705101937576124)
--(axis cs:15.2,0.701505585499548)
--(axis cs:19.7,0.737047495326584)
--(axis cs:26.2,0.743844371574276)
--(axis cs:34.1,0.763523122122849)
--(axis cs:42.5,0.77087733509491)
--(axis cs:55.7,0.776812474375393)
--(axis cs:72.7,0.788527149384978)
--(axis cs:95.2,0.798110882528294)
--(axis cs:116.8,0.811734882846242)
--(axis cs:147.6,0.820895147552563)
--(axis cs:189.9,0.830296647842886)
--(axis cs:225.7,0.837811820348729)
--(axis cs:257.1,0.846829749213158)
--(axis cs:292.2,0.855802912053159)
--(axis cs:311.7,0.866239987207497)
--(axis cs:360.7,0.873588211657017)
--(axis cs:441.5,0.881607939416193)
--(axis cs:480.8,0.88888521004167)
--(axis cs:548.7,0.895606301475125)
--(axis cs:610.5,0.901766748688719)
--(axis cs:679.9,0.906763027019709)
--(axis cs:770.7,0.91221695497956)
--(axis cs:857.8,0.916561706025472)
--(axis cs:962.1,0.920042898261189)
--(axis cs:1067.1,0.923671394915627)
--(axis cs:1197.7,0.927105641614923)
--(axis cs:1356.9,0.930110047382186)
--(axis cs:1509.8,0.93306206258712)
--(axis cs:1700.7,0.935640379362934)
--(axis cs:2013,0.938117999358734)
--(axis cs:2313.8,0.94063958091648)
--(axis cs:2423,0.943368218767725)
--(axis cs:2656.3,0.945516406497594)
--(axis cs:2979.9,0.947576444864234)
--(axis cs:3338.7,0.949485604096682)
--(axis cs:3709.4,0.95126722478933)
--(axis cs:4101,0.952573024846842)
--(axis cs:4529.6,0.954026417428042)
--(axis cs:4994,0.955383030922103)
--(axis cs:5533.2,0.956555777581136)
--(axis cs:6100.8,0.957680483968925)
--(axis cs:6738.1,0.958852395189902)
--(axis cs:7440.2,0.960071800851666)
--(axis cs:8217,0.961030896974574)
--(axis cs:9043.1,0.961944087597913)
--(axis cs:10008.2,0.96278700207505)
--(axis cs:10967.9,0.963575928308512)
--(axis cs:11806.5,0.964282186211603)
--(axis cs:13154,0.965032713495285)
--(axis cs:14409,0.965735570226938)
--(axis cs:15679.8,0.966206477143022)
--(axis cs:17075.7,0.966782303454616)
--(axis cs:18546.3,0.967319580993747)
--(axis cs:20025.7,0.967769191999057)
--(axis cs:21636.8,0.968192760928746)
--(axis cs:23496.5,0.968468099263431)
--(axis cs:25376.7,0.968679206337064)
--(axis cs:27315.6,0.968911525005651)
--(axis cs:29182.3,0.969086626485915)
--(axis cs:31006.2,0.969366805165719)
--(axis cs:33135.9,0.9694540026789)
--(axis cs:35352.8,0.969589564178704)
--(axis cs:37493.6,0.96972152140011)
--(axis cs:39483.8,0.969708293298294)
--(axis cs:40868.4,0.969703876299993)
--(axis cs:43308.7,0.9697482662152)
--(axis cs:45580.9,0.969673437816846)
--(axis cs:48173.1,0.969613694030613)
--(axis cs:51057.1,0.969501052971487)
--(axis cs:54538.4,0.969471849720971)
--(axis cs:57527.6,0.969307946072128)
--(axis cs:60246.6,0.969188645058644)
--(axis cs:63510.6,0.969193589861986)
--(axis cs:66424.1,0.969040115244682)
--(axis cs:66424.1,0.970241684755318)
--(axis cs:66424.1,0.970241684755318)
--(axis cs:63510.6,0.970326410138013)
--(axis cs:60246.6,0.970371354941357)
--(axis cs:57527.6,0.970402053927872)
--(axis cs:54538.4,0.970426550279029)
--(axis cs:51057.1,0.970490547028513)
--(axis cs:48173.1,0.970596305969387)
--(axis cs:45580.9,0.970716762183154)
--(axis cs:43308.7,0.9706933337848)
--(axis cs:40868.4,0.970697523700007)
--(axis cs:39483.8,0.970728106701706)
--(axis cs:37493.6,0.97065347859989)
--(axis cs:35352.8,0.970485435821296)
--(axis cs:33135.9,0.9704427973211)
--(axis cs:31006.2,0.970369994834281)
--(axis cs:29182.3,0.970130173514084)
--(axis cs:27315.6,0.969906874994349)
--(axis cs:25376.7,0.969773993662936)
--(axis cs:23496.5,0.969480300736569)
--(axis cs:21636.8,0.969220239071254)
--(axis cs:20025.7,0.968832408000943)
--(axis cs:18546.3,0.968448819006253)
--(axis cs:17075.7,0.967980696545384)
--(axis cs:15679.8,0.967448522856978)
--(axis cs:14409,0.966979229773062)
--(axis cs:13154,0.966370286504715)
--(axis cs:11806.5,0.965649613788397)
--(axis cs:10967.9,0.964779071691488)
--(axis cs:10008.2,0.96412779792495)
--(axis cs:9043.1,0.963267712402087)
--(axis cs:8217,0.962379103025426)
--(axis cs:7440.2,0.961423399148334)
--(axis cs:6738.1,0.960459204810098)
--(axis cs:6100.8,0.959009516031075)
--(axis cs:5533.2,0.957937622418864)
--(axis cs:4994,0.956808769077897)
--(axis cs:4529.6,0.955634982571958)
--(axis cs:4101,0.954164975153158)
--(axis cs:3709.4,0.95260917521067)
--(axis cs:3338.7,0.950842795903318)
--(axis cs:2979.9,0.948870155135766)
--(axis cs:2656.3,0.946967193502406)
--(axis cs:2423,0.944986581232275)
--(axis cs:2313.8,0.94211061908352)
--(axis cs:2013,0.939713800641266)
--(axis cs:1700.7,0.937291420637066)
--(axis cs:1509.8,0.93480953741288)
--(axis cs:1356.9,0.931968152617814)
--(axis cs:1197.7,0.928757758385077)
--(axis cs:1067.1,0.925105205084373)
--(axis cs:962.1,0.921713901738811)
--(axis cs:857.8,0.918164893974528)
--(axis cs:770.7,0.91374324502044)
--(axis cs:679.9,0.908866772980291)
--(axis cs:610.5,0.903583251311281)
--(axis cs:548.7,0.897493898524875)
--(axis cs:480.8,0.89079678995833)
--(axis cs:441.5,0.883865460583807)
--(axis cs:360.7,0.875623788342982)
--(axis cs:311.7,0.868153012792503)
--(axis cs:292.2,0.858338887946841)
--(axis cs:257.1,0.849283650786843)
--(axis cs:225.7,0.840428179651271)
--(axis cs:189.9,0.832891552157114)
--(axis cs:147.6,0.824208252447437)
--(axis cs:116.8,0.814366717153758)
--(axis cs:95.2,0.800509317471706)
--(axis cs:72.7,0.790369450615022)
--(axis cs:55.7,0.779049325624607)
--(axis cs:42.5,0.77337446490509)
--(axis cs:34.1,0.766537077877151)
--(axis cs:26.2,0.749032228425724)
--(axis cs:19.7,0.740007504673416)
--(axis cs:15.2,0.704509414500452)
--(axis cs:12.3,0.707521462423876)
--(axis cs:7.5,0.696639709683845)
--(axis cs:5.8,0.692052610637094)
--(axis cs:4.6,0.691935737290799)
--(axis cs:1,0.500981347719775)
--(axis cs:1,0.500981347719775)
--cycle;

\path [fill=steelblue31119180, fill opacity=0.2]
(axis cs:32.4,0.679847625331549)
--(axis cs:32.4,0.642095774668451)
--(axis cs:58.8,0.517837518062928)
--(axis cs:127.5,0.612698761529931)
--(axis cs:171.6,0.685356921070344)
--(axis cs:217.9,0.620083359699243)
--(axis cs:326.9,0.70458594348935)
--(axis cs:509.4,0.723755500738939)
--(axis cs:788.7,0.769130945483603)
--(axis cs:1289,0.75946927574984)
--(axis cs:1725.2,0.742223262156284)
--(axis cs:2166.5,0.692164699982658)
--(axis cs:2240.8,0.574857455822285)
--(axis cs:2245.6,0.756240849479995)
--(axis cs:2387.4,0.751807882759305)
--(axis cs:2566,0.59427969532723)
--(axis cs:3221.1,0.791658165635112)
--(axis cs:4167.6,0.777222658598795)
--(axis cs:5525.8,0.759292700193437)
--(axis cs:7395.5,0.792037116362187)
--(axis cs:7828.6,0.802475674649737)
--(axis cs:8253.4,0.819230241588053)
--(axis cs:9385.6,0.816178775153507)
--(axis cs:10880.7,0.815775533071974)
--(axis cs:10937.9,0.807049092884071)
--(axis cs:11114.5,0.778141175356328)
--(axis cs:14048.3,0.841523289545244)
--(axis cs:17864,0.844317801267512)
--(axis cs:22418.9,0.845279488818191)
--(axis cs:27844.9,0.851687983244529)
--(axis cs:32950.1,0.881306243075926)
--(axis cs:34305.1,0.862091392268651)
--(axis cs:36925.7,0.871637861539409)
--(axis cs:40353.4,0.868542783495379)
--(axis cs:40403,0.880281778898155)
--(axis cs:48736.7,0.886042271882971)
--(axis cs:57236.4,0.897222226706102)
--(axis cs:64772.3,0.895743305752447)
--(axis cs:70835.6,0.900410576631051)
--(axis cs:73835.5,0.900748035696426)
--(axis cs:73897.5,0.902559799829986)
--(axis cs:75936.4,0.894734944826268)
--(axis cs:78130.4,0.897727657799747)
--(axis cs:78380.3,0.903599174080796)
--(axis cs:82358.5,0.907497377872251)
--(axis cs:85825.1,0.90922886927831)
--(axis cs:88807.5,0.899836540925545)
--(axis cs:91221.2,0.909426800071798)
--(axis cs:93274.1,0.911674587328014)
--(axis cs:94236.4,0.915655255277879)
--(axis cs:94659.6,0.883361861201733)
--(axis cs:94979.7,0.913845992627475)
--(axis cs:95824.2,0.909174852474879)
--(axis cs:96030.7,0.917885495511901)
--(axis cs:96183.6,0.913978160309475)
--(axis cs:97208.1,0.919907122185328)
--(axis cs:98125.7,0.914575158067415)
--(axis cs:98850.9,0.916932664473153)
--(axis cs:99302.8,0.91654338638085)
--(axis cs:99597,0.917369802615626)
--(axis cs:99796.5,0.920574525183732)
--(axis cs:99901.9,0.917562009578199)
--(axis cs:99960,0.917594947552212)
--(axis cs:99986.5,0.917070734577971)
--(axis cs:99996.4,0.922327842372778)
--(axis cs:99999.3,0.920254708291022)
--(axis cs:100000,0.924173025525705)
--(axis cs:100000,0.918872478004465)
--(axis cs:100000,0.920529994073732)
--(axis cs:100000,0.80002217444995)
--(axis cs:100000,0.919075448464504)
--(axis cs:100000,0.920785303344261)
--(axis cs:100000,0.920703720392312)
--(axis cs:100000,0.920142946352755)
--(axis cs:100000,0.921122688779361)
--(axis cs:100000,0.918026156103591)
--(axis cs:100000,0.919579752184695)
--(axis cs:100000,0.921079614103)
--(axis cs:100000,0.913655751209837)
--(axis cs:100000,0.922363064663177)
--(axis cs:100000,0.913688290707069)
--(axis cs:100000,0.929766709292931)
--(axis cs:100000,0.929766709292931)
--(axis cs:100000,0.926946735336823)
--(axis cs:100000,0.927257448790163)
--(axis cs:100000,0.925846985897)
--(axis cs:100000,0.929017247815305)
--(axis cs:100000,0.926088843896409)
--(axis cs:100000,0.926690911220639)
--(axis cs:100000,0.926465453647245)
--(axis cs:100000,0.926934479607688)
--(axis cs:100000,0.928961496655739)
--(axis cs:100000,0.927037951535496)
--(axis cs:100000,0.98822622555005)
--(axis cs:100000,0.927351605926268)
--(axis cs:100000,0.928177721995535)
--(axis cs:100000,0.927798974474295)
--(axis cs:99999.3,0.925176891708978)
--(axis cs:99996.4,0.927017157627222)
--(axis cs:99986.5,0.926322865422029)
--(axis cs:99960,0.925438652447788)
--(axis cs:99901.9,0.925632990421801)
--(axis cs:99796.5,0.926153674816268)
--(axis cs:99597,0.925133397384374)
--(axis cs:99302.8,0.92316161361915)
--(axis cs:98850.9,0.925189335526846)
--(axis cs:98125.7,0.921216441932585)
--(axis cs:97208.1,0.923567877814672)
--(axis cs:96183.6,0.921890639690526)
--(axis cs:96030.7,0.924232904488099)
--(axis cs:95824.2,0.921606547525121)
--(axis cs:94979.7,0.922352207372525)
--(axis cs:94659.6,0.929994738798267)
--(axis cs:94236.4,0.922782744722121)
--(axis cs:93274.1,0.920384012671986)
--(axis cs:91221.2,0.918696599928202)
--(axis cs:88807.5,0.918293459074455)
--(axis cs:85825.1,0.91846273072169)
--(axis cs:82358.5,0.919122622127749)
--(axis cs:78380.3,0.914444225919204)
--(axis cs:78130.4,0.908965942200253)
--(axis cs:75936.4,0.903473055173732)
--(axis cs:73897.5,0.914905200170014)
--(axis cs:73835.5,0.914790364303574)
--(axis cs:70835.6,0.910266223368949)
--(axis cs:64772.3,0.908215094247553)
--(axis cs:57236.4,0.907496173293898)
--(axis cs:48736.7,0.905612528117029)
--(axis cs:40403,0.896541621101845)
--(axis cs:40353.4,0.888237216504621)
--(axis cs:36925.7,0.896847138460591)
--(axis cs:34305.1,0.890398807731349)
--(axis cs:32950.1,0.901265556924074)
--(axis cs:27844.9,0.882138816755471)
--(axis cs:22418.9,0.883153911181809)
--(axis cs:17864,0.882740398732488)
--(axis cs:14048.3,0.886673510454756)
--(axis cs:11114.5,0.872742224643673)
--(axis cs:10937.9,0.868944507115929)
--(axis cs:10880.7,0.862475866928026)
--(axis cs:9385.6,0.862159624846493)
--(axis cs:8253.4,0.870881358411947)
--(axis cs:7828.6,0.866394125350263)
--(axis cs:7395.5,0.861736083637813)
--(axis cs:5525.8,0.882939099806563)
--(axis cs:4167.6,0.834713941401205)
--(axis cs:3221.1,0.861018434364888)
--(axis cs:2566,0.87738350467277)
--(axis cs:2387.4,0.826530717240695)
--(axis cs:2245.6,0.837730950520005)
--(axis cs:2240.8,0.831895944177715)
--(axis cs:2166.5,0.896608500017342)
--(axis cs:1725.2,0.830641737843716)
--(axis cs:1289,0.84699912425016)
--(axis cs:788.7,0.834089254516397)
--(axis cs:509.4,0.797359499261061)
--(axis cs:326.9,0.77534245651065)
--(axis cs:217.9,0.750058240300757)
--(axis cs:171.6,0.736236678929656)
--(axis cs:127.5,0.721001438470069)
--(axis cs:58.8,0.699043881937072)
--(axis cs:32.4,0.679847625331549)
--cycle;

\path [fill=darkorange25512714, fill opacity=0.2]
(axis cs:1,0.502302935100388)
--(axis cs:1,0.499510464899612)
--(axis cs:1,0.499510464899612)
--(axis cs:1,0.499510464899612)
--(axis cs:1,0.499510464899612)
--(axis cs:1,0.499510464899612)
--(axis cs:1,0.499510464899612)
--(axis cs:1,0.499510464899612)
--(axis cs:1,0.499510464899612)
--(axis cs:1,0.499510464899612)
--(axis cs:1,0.499510464899612)
--(axis cs:1,0.499510464899612)
--(axis cs:1,0.499510464899612)
--(axis cs:1,0.499510464899612)
--(axis cs:4.1,0.657999504087865)
--(axis cs:5,0.665456562991632)
--(axis cs:6.8,0.658599008637676)
--(axis cs:15,0.712850541051502)
--(axis cs:18.5,0.749199064485072)
--(axis cs:32.2,0.769645360140584)
--(axis cs:51.2,0.75306263738105)
--(axis cs:80.4,0.774596956068723)
--(axis cs:124,0.803243696319096)
--(axis cs:194.5,0.793906145211854)
--(axis cs:271.1,0.818576300835856)
--(axis cs:379.2,0.845498955293003)
--(axis cs:549.7,0.838200038277957)
--(axis cs:805.2,0.860344842199475)
--(axis cs:1166.5,0.878666111738313)
--(axis cs:1624.3,0.891346399203995)
--(axis cs:2139.3,0.902301576072228)
--(axis cs:2727.9,0.903784457980579)
--(axis cs:3402.5,0.912624297725985)
--(axis cs:4072.6,0.913567593728978)
--(axis cs:4744.8,0.916473272813576)
--(axis cs:5416.8,0.919775436194057)
--(axis cs:6045.5,0.920605361755409)
--(axis cs:6679.2,0.920995941793493)
--(axis cs:7427.1,0.922713494018542)
--(axis cs:8257.6,0.924450095222597)
--(axis cs:9167.8,0.925284852336385)
--(axis cs:10202.2,0.92845463404764)
--(axis cs:11497.8,0.92700169249811)
--(axis cs:12978.6,0.92697461152914)
--(axis cs:14687.8,0.926499081925589)
--(axis cs:16251.6,0.927417696269791)
--(axis cs:17915.3,0.926605877757749)
--(axis cs:19809.4,0.929003519406544)
--(axis cs:21719,0.927164719246717)
--(axis cs:23808.3,0.928214136106452)
--(axis cs:26041.2,0.925682732468533)
--(axis cs:28546.5,0.92819703382725)
--(axis cs:31325.1,0.928211676373298)
--(axis cs:34215,0.92821532486526)
--(axis cs:37232.5,0.926478044466933)
--(axis cs:40157.2,0.925369018898146)
--(axis cs:43337,0.92601473794117)
--(axis cs:47025.5,0.92470226934651)
--(axis cs:50630.5,0.924231708533777)
--(axis cs:54228.7,0.922422622245446)
--(axis cs:55971.3,0.926218343744272)
--(axis cs:56343.6,0.926525872643653)
--(axis cs:57439.2,0.92884724440793)
--(axis cs:58731.4,0.929724263849298)
--(axis cs:59290.8,0.928247432282999)
--(axis cs:59314.8,0.928260130260092)
--(axis cs:62702.6,0.925385040628663)
--(axis cs:66328.5,0.923783938870771)
--(axis cs:70231.8,0.9225363239414)
--(axis cs:74403.4,0.922824002112486)
--(axis cs:78528.1,0.920794826020941)
--(axis cs:82647.4,0.91988780390917)
--(axis cs:86473.3,0.920014191770404)
--(axis cs:90081.9,0.923568498183388)
--(axis cs:93280,0.91891983259507)
--(axis cs:96047.2,0.920442938392507)
--(axis cs:98120.6,0.922648084320852)
--(axis cs:99313.6,0.920844261377909)
--(axis cs:99847.9,0.921912515288832)
--(axis cs:99983.5,0.920217377157389)
--(axis cs:99999.5,0.923765309245597)
--(axis cs:99999.5,0.931132890754403)
--(axis cs:99999.5,0.931132890754403)
--(axis cs:99983.5,0.927604422842611)
--(axis cs:99847.9,0.928241084711168)
--(axis cs:99313.6,0.930053738622091)
--(axis cs:98120.6,0.931055515679148)
--(axis cs:96047.2,0.930051661607493)
--(axis cs:93280,0.92822996740493)
--(axis cs:90081.9,0.931871501816612)
--(axis cs:86473.3,0.928744208229596)
--(axis cs:82647.4,0.92960739609083)
--(axis cs:78528.1,0.930063573979059)
--(axis cs:74403.4,0.931847597887514)
--(axis cs:70231.8,0.9287236760586)
--(axis cs:66328.5,0.931672861129229)
--(axis cs:62702.6,0.932508159371337)
--(axis cs:59314.8,0.934999669739908)
--(axis cs:59290.8,0.933975767717001)
--(axis cs:58731.4,0.934097536150702)
--(axis cs:57439.2,0.93548755559207)
--(axis cs:56343.6,0.932449327356347)
--(axis cs:55971.3,0.933061456255728)
--(axis cs:54228.7,0.930242377754554)
--(axis cs:50630.5,0.931964891466223)
--(axis cs:47025.5,0.931647530653489)
--(axis cs:43337,0.93352826205883)
--(axis cs:40157.2,0.931601181101854)
--(axis cs:37232.5,0.932050155533067)
--(axis cs:34215,0.93418927513474)
--(axis cs:31325.1,0.934251523626702)
--(axis cs:28546.5,0.93433616617275)
--(axis cs:26041.2,0.932940667531467)
--(axis cs:23808.3,0.934730663893548)
--(axis cs:21719,0.935055280753283)
--(axis cs:19809.4,0.934299880593456)
--(axis cs:17915.3,0.932960922242251)
--(axis cs:16251.6,0.933160303730209)
--(axis cs:14687.8,0.930017518074411)
--(axis cs:12978.6,0.93325038847086)
--(axis cs:11497.8,0.93211670750189)
--(axis cs:10202.2,0.93292516595236)
--(axis cs:9167.8,0.931833547663615)
--(axis cs:8257.6,0.930081704777403)
--(axis cs:7427.1,0.930174705981458)
--(axis cs:6679.2,0.927764258206507)
--(axis cs:6045.5,0.928561238244591)
--(axis cs:5416.8,0.926596163805943)
--(axis cs:4744.8,0.924414927186424)
--(axis cs:4072.6,0.922584206271022)
--(axis cs:3402.5,0.920699102274015)
--(axis cs:2727.9,0.916344142019421)
--(axis cs:2139.3,0.915120223927772)
--(axis cs:1624.3,0.909951800796005)
--(axis cs:1166.5,0.901035488261687)
--(axis cs:805.2,0.893041757800525)
--(axis cs:549.7,0.878078361722043)
--(axis cs:379.2,0.873879444706997)
--(axis cs:271.1,0.853830899164144)
--(axis cs:194.5,0.837585254788146)
--(axis cs:124,0.820040903680904)
--(axis cs:80.4,0.801464443931277)
--(axis cs:51.2,0.78826876261895)
--(axis cs:32.2,0.776041239859416)
--(axis cs:18.5,0.754732735514928)
--(axis cs:15,0.737389458948498)
--(axis cs:6.8,0.699782791362325)
--(axis cs:5,0.691044837008368)
--(axis cs:4.1,0.691120295912135)
--(axis cs:1,0.502302935100388)
--(axis cs:1,0.502302935100388)
--(axis cs:1,0.502302935100388)
--(axis cs:1,0.502302935100388)
--(axis cs:1,0.502302935100388)
--(axis cs:1,0.502302935100388)
--(axis cs:1,0.502302935100388)
--(axis cs:1,0.502302935100388)
--(axis cs:1,0.502302935100388)
--(axis cs:1,0.502302935100388)
--(axis cs:1,0.502302935100388)
--(axis cs:1,0.502302935100388)
--(axis cs:1,0.502302935100388)
--cycle;

\path [fill=forestgreen4416044, fill opacity=0.2]
(axis cs:2.6,0.667393229579635)
--(axis cs:2.6,0.664738370420365)
--(axis cs:3.1,0.664058455703161)
--(axis cs:3.6,0.636420238144391)
--(axis cs:4.3,0.663769914881702)
--(axis cs:5.1,0.663558864496318)
--(axis cs:7.5,0.681934111353935)
--(axis cs:17.8,0.699619486206494)
--(axis cs:23.7,0.709971234954948)
--(axis cs:28.1,0.709333005203958)
--(axis cs:36.9,0.736835748200972)
--(axis cs:48.8,0.748568418993425)
--(axis cs:58.7,0.758066468591131)
--(axis cs:66.8,0.761805300827823)
--(axis cs:79.4,0.775395296704201)
--(axis cs:97.5,0.787015071805367)
--(axis cs:113.8,0.779151048176671)
--(axis cs:139.6,0.795509012373085)
--(axis cs:172.4,0.814886614574186)
--(axis cs:210.6,0.827356405842786)
--(axis cs:262.2,0.82547376713972)
--(axis cs:312.8,0.833852043243155)
--(axis cs:370.4,0.845548148648617)
--(axis cs:433.5,0.839290833058382)
--(axis cs:544.3,0.857126851605461)
--(axis cs:700.6,0.844147436090329)
--(axis cs:907.2,0.876540621526905)
--(axis cs:1202.5,0.880194973900716)
--(axis cs:1596.8,0.895509807227289)
--(axis cs:2047.9,0.898926835351809)
--(axis cs:2568.4,0.904347506936906)
--(axis cs:3118.9,0.908959440827872)
--(axis cs:3441.2,0.909625914273092)
--(axis cs:3905.5,0.915519848742537)
--(axis cs:4626,0.914454511575382)
--(axis cs:5415.9,0.914566378540255)
--(axis cs:6247,0.917700756265332)
--(axis cs:7084.9,0.917720948742731)
--(axis cs:7986.1,0.921592976251956)
--(axis cs:8870.5,0.920252999914763)
--(axis cs:9829,0.922934917679001)
--(axis cs:10719.6,0.924077724687239)
--(axis cs:12003.4,0.92134133066968)
--(axis cs:13726.8,0.924333050351039)
--(axis cs:15532.3,0.923084000532938)
--(axis cs:17246.6,0.928353030789355)
--(axis cs:18993.4,0.924537301457057)
--(axis cs:20802.3,0.924623847858332)
--(axis cs:22797.5,0.925593234707414)
--(axis cs:24941.3,0.928436480464283)
--(axis cs:27217,0.925275749650366)
--(axis cs:29266.9,0.926321348061081)
--(axis cs:31296.7,0.926440514167977)
--(axis cs:33935.8,0.922248026987961)
--(axis cs:36856.7,0.924632385807316)
--(axis cs:39926.9,0.922458286078849)
--(axis cs:43094.4,0.924642029530055)
--(axis cs:46383.7,0.922888079096121)
--(axis cs:49598.7,0.929267696323772)
--(axis cs:52890,0.925054002276341)
--(axis cs:53166.1,0.926410884213882)
--(axis cs:54630.2,0.926181482342982)
--(axis cs:54726.7,0.92740512067715)
--(axis cs:55496,0.923284428832588)
--(axis cs:56491.1,0.929095217141267)
--(axis cs:57884.7,0.925373759426729)
--(axis cs:61427,0.926251497297622)
--(axis cs:65286.2,0.926786391809014)
--(axis cs:69319.4,0.928999815645071)
--(axis cs:73597.2,0.928443539842579)
--(axis cs:77945.7,0.92594072451695)
--(axis cs:82333.8,0.926691227867928)
--(axis cs:86683.6,0.928451383161616)
--(axis cs:90999.8,0.926828306941715)
--(axis cs:94866.9,0.926755945152722)
--(axis cs:97781.8,0.928802484551548)
--(axis cs:99386.1,0.929434686252015)
--(axis cs:99913.6,0.927252663643488)
--(axis cs:99996.2,0.928885238938284)
--(axis cs:100000,0.928807448890116)
--(axis cs:100000,0.927769309937575)
--(axis cs:100000,0.933050490062425)
--(axis cs:100000,0.933050490062425)
--(axis cs:100000,0.933650551109883)
--(axis cs:99996.2,0.933414761061716)
--(axis cs:99913.6,0.931904136356512)
--(axis cs:99386.1,0.933485313747985)
--(axis cs:97781.8,0.933317515448452)
--(axis cs:94866.9,0.932094054847278)
--(axis cs:90999.8,0.931798093058285)
--(axis cs:86683.6,0.933936816838384)
--(axis cs:82333.8,0.932295372132072)
--(axis cs:77945.7,0.92993947548305)
--(axis cs:73597.2,0.933536460157421)
--(axis cs:69319.4,0.934426784354929)
--(axis cs:65286.2,0.932098608190986)
--(axis cs:61427,0.930525102702378)
--(axis cs:57884.7,0.932299240573271)
--(axis cs:56491.1,0.932724582858732)
--(axis cs:55496,0.933180571167412)
--(axis cs:54726.7,0.93340667932285)
--(axis cs:54630.2,0.932876917657018)
--(axis cs:53166.1,0.933212515786118)
--(axis cs:52890,0.932449397723659)
--(axis cs:49598.7,0.931815903676228)
--(axis cs:46383.7,0.931190120903879)
--(axis cs:43094.4,0.931994170469945)
--(axis cs:39926.9,0.930115313921151)
--(axis cs:36856.7,0.932824414192684)
--(axis cs:33935.8,0.928400373012039)
--(axis cs:31296.7,0.932616085832023)
--(axis cs:29266.9,0.931867251938919)
--(axis cs:27217,0.932171050349634)
--(axis cs:24941.3,0.932723519535717)
--(axis cs:22797.5,0.931908565292586)
--(axis cs:20802.3,0.932034352141668)
--(axis cs:18993.4,0.931106098542943)
--(axis cs:17246.6,0.931538769210645)
--(axis cs:15532.3,0.932349199467062)
--(axis cs:13726.8,0.931620149648961)
--(axis cs:12003.4,0.930472069330319)
--(axis cs:10719.6,0.93109387531276)
--(axis cs:9829,0.928049882320999)
--(axis cs:8870.5,0.928022200085237)
--(axis cs:7986.1,0.928810623748044)
--(axis cs:7084.9,0.926252451257269)
--(axis cs:6247,0.925679443734668)
--(axis cs:5415.9,0.924480621459745)
--(axis cs:4626,0.920787288424618)
--(axis cs:3905.5,0.920538551257463)
--(axis cs:3441.2,0.918500485726908)
--(axis cs:3118.9,0.917900759172128)
--(axis cs:2568.4,0.915542693063094)
--(axis cs:2047.9,0.911132964648191)
--(axis cs:1596.8,0.907230192772711)
--(axis cs:1202.5,0.904480426099284)
--(axis cs:907.2,0.897332978473095)
--(axis cs:700.6,0.880760963909671)
--(axis cs:544.3,0.875454748394539)
--(axis cs:433.5,0.863480966941618)
--(axis cs:370.4,0.860905051351383)
--(axis cs:312.8,0.855206156756845)
--(axis cs:262.2,0.84569463286028)
--(axis cs:210.6,0.839843594157214)
--(axis cs:172.4,0.827449985425814)
--(axis cs:139.6,0.811601187626915)
--(axis cs:113.8,0.807578751823329)
--(axis cs:97.5,0.800756328194633)
--(axis cs:79.4,0.787461503295799)
--(axis cs:66.8,0.774811299172177)
--(axis cs:58.7,0.768791531408869)
--(axis cs:48.8,0.762359781006575)
--(axis cs:36.9,0.768437851799028)
--(axis cs:28.1,0.749865394796042)
--(axis cs:23.7,0.731235365045052)
--(axis cs:17.8,0.705807113793506)
--(axis cs:7.5,0.695419488646065)
--(axis cs:5.1,0.666894735503683)
--(axis cs:4.3,0.666287085118298)
--(axis cs:3.6,0.643169961855609)
--(axis cs:3.1,0.667158144296839)
--(axis cs:2.6,0.667393229579635)
--cycle;

\path [fill=crimson2143940, fill opacity=0.2]
(axis cs:1,0.502265695453306)
--(axis cs:1,0.499547704546694)
--(axis cs:1,0.499547704546694)
--(axis cs:1,0.499547704546694)
--(axis cs:1,0.491394596181282)
--(axis cs:1,0.499547704546694)
--(axis cs:1,0.499547704546694)
--(axis cs:1,0.499547704546694)
--(axis cs:1,0.499547704546694)
--(axis cs:1,0.499547704546694)
--(axis cs:1,0.492214609207389)
--(axis cs:1,0.491658679587431)
--(axis cs:1,0.499547704546694)
--(axis cs:1,0.492214609207389)
--(axis cs:7.7,0.662357009408556)
--(axis cs:27.9,0.668332103248928)
--(axis cs:45.3,0.699489169034073)
--(axis cs:72,0.701694158207165)
--(axis cs:117,0.709227945977742)
--(axis cs:187.5,0.742261640349936)
--(axis cs:264.1,0.743970363532853)
--(axis cs:360.8,0.776928529144277)
--(axis cs:526.1,0.788730281312072)
--(axis cs:823.3,0.77746419186788)
--(axis cs:1334.9,0.8254278084035)
--(axis cs:2053.2,0.818638270536976)
--(axis cs:2993.2,0.824436926128271)
--(axis cs:4168.3,0.847972272776711)
--(axis cs:5665.8,0.850855173690689)
--(axis cs:7372.3,0.849795340610195)
--(axis cs:9439,0.86756354541727)
--(axis cs:11923.9,0.876295222031054)
--(axis cs:14917.8,0.88660787650813)
--(axis cs:18302.6,0.892706544602254)
--(axis cs:21750.3,0.891387906573298)
--(axis cs:24976.3,0.901725078938666)
--(axis cs:26979.4,0.906685146462261)
--(axis cs:28801.6,0.907517676359088)
--(axis cs:31742.1,0.9109643337167)
--(axis cs:35352.3,0.819572504426107)
--(axis cs:38972.2,0.914315678237984)
--(axis cs:42735.6,0.916596627267413)
--(axis cs:46285.3,0.919170493637499)
--(axis cs:49578.1,0.919243799945402)
--(axis cs:50161.6,0.920428131456255)
--(axis cs:50668.8,0.921600828308508)
--(axis cs:51619.1,0.91791324056901)
--(axis cs:52836.6,0.92328206591517)
--(axis cs:56301.4,0.922992221938099)
--(axis cs:59957.7,0.924854835793834)
--(axis cs:63842.4,0.908124147304685)
--(axis cs:68150.9,0.924441296348172)
--(axis cs:71898,0.922284837434891)
--(axis cs:72696.8,0.923437180483301)
--(axis cs:74261.8,0.923539110480473)
--(axis cs:75746,0.906513321098572)
--(axis cs:76646.7,0.924963810375303)
--(axis cs:78564.7,0.926253433544166)
--(axis cs:79099.9,0.922393065963621)
--(axis cs:83017,0.926515232802819)
--(axis cs:87165.8,0.925886910659992)
--(axis cs:90929.4,0.925206375809933)
--(axis cs:94103.3,0.921152470059134)
--(axis cs:96605.6,0.925445043928816)
--(axis cs:97058.3,0.923908353395578)
--(axis cs:98258.1,0.92540836022835)
--(axis cs:98302.5,0.869495115497381)
--(axis cs:98345.9,0.923092175814972)
--(axis cs:99192.2,0.926115901361826)
--(axis cs:99332.3,0.924061020334221)
--(axis cs:99334.4,0.92490355891427)
--(axis cs:99815.8,0.924816426592763)
--(axis cs:99968.8,0.92455652869213)
--(axis cs:99996.3,0.92316049152345)
--(axis cs:100000,0.924163064027227)
--(axis cs:100000,0.923726403875583)
--(axis cs:100000,0.921375431767808)
--(axis cs:100000,0.92384324680242)
--(axis cs:100000,0.923687181034543)
--(axis cs:100000,0.921734145411212)
--(axis cs:100000,0.924062229143759)
--(axis cs:100000,0.930015970856241)
--(axis cs:100000,0.930015970856241)
--(axis cs:100000,0.929725854588788)
--(axis cs:100000,0.929626418965457)
--(axis cs:100000,0.93066835319758)
--(axis cs:100000,0.932001168232192)
--(axis cs:100000,0.929188396124417)
--(axis cs:100000,0.930783735972773)
--(axis cs:99996.3,0.92985270847655)
--(axis cs:99968.8,0.93220347130787)
--(axis cs:99815.8,0.929941973407237)
--(axis cs:99334.4,0.930908041085729)
--(axis cs:99332.3,0.930648779665779)
--(axis cs:99192.2,0.930597498638174)
--(axis cs:98345.9,0.931342824185028)
--(axis cs:98302.5,0.95672668450262)
--(axis cs:98258.1,0.93016523977165)
--(axis cs:97058.3,0.932705046604422)
--(axis cs:96605.6,0.932574956071184)
--(axis cs:94103.3,0.931239329940866)
--(axis cs:90929.4,0.927257024190067)
--(axis cs:87165.8,0.932691289340008)
--(axis cs:83017,0.931791367197181)
--(axis cs:79099.9,0.928776734036379)
--(axis cs:78564.7,0.932881566455834)
--(axis cs:76646.7,0.930389589624697)
--(axis cs:75746,0.940388478901428)
--(axis cs:74261.8,0.932951089519527)
--(axis cs:72696.8,0.929854619516698)
--(axis cs:71898,0.932266762565109)
--(axis cs:68150.9,0.929935303651828)
--(axis cs:63842.4,0.939757452695315)
--(axis cs:59957.7,0.930653564206167)
--(axis cs:56301.4,0.930937978061901)
--(axis cs:52836.6,0.92938793408483)
--(axis cs:51619.1,0.92522995943099)
--(axis cs:50668.8,0.929592371691492)
--(axis cs:50161.6,0.927768268543746)
--(axis cs:49578.1,0.928507600054598)
--(axis cs:46285.3,0.927246106362501)
--(axis cs:42735.6,0.927486772732587)
--(axis cs:38972.2,0.923506121762016)
--(axis cs:35352.3,0.962145695573893)
--(axis cs:31742.1,0.9212340662833)
--(axis cs:28801.6,0.920464123640912)
--(axis cs:26979.4,0.920480053537739)
--(axis cs:24976.3,0.917639921061334)
--(axis cs:21750.3,0.904283693426702)
--(axis cs:18302.6,0.913331655397746)
--(axis cs:14917.8,0.912097323491869)
--(axis cs:11923.9,0.900039577968946)
--(axis cs:9439,0.89820665458273)
--(axis cs:7372.3,0.892339659389805)
--(axis cs:5665.8,0.891988226309311)
--(axis cs:4168.3,0.891364327223289)
--(axis cs:2993.2,0.873133073871729)
--(axis cs:2053.2,0.871558529463024)
--(axis cs:1334.9,0.8584139915965)
--(axis cs:823.3,0.84208960813212)
--(axis cs:526.1,0.841306318687928)
--(axis cs:360.8,0.823601470855723)
--(axis cs:264.1,0.796246436467147)
--(axis cs:187.5,0.801130159650064)
--(axis cs:117,0.759108454022258)
--(axis cs:72,0.746415841792835)
--(axis cs:45.3,0.720277430965927)
--(axis cs:27.9,0.699294696751072)
--(axis cs:7.7,0.684504590591444)
--(axis cs:1,0.505103790792611)
--(axis cs:1,0.502265695453306)
--(axis cs:1,0.505654720412569)
--(axis cs:1,0.505103790792611)
--(axis cs:1,0.502265695453306)
--(axis cs:1,0.502265695453306)
--(axis cs:1,0.502265695453306)
--(axis cs:1,0.502265695453306)
--(axis cs:1,0.502265695453306)
--(axis cs:1,0.506057003818718)
--(axis cs:1,0.502265695453306)
--(axis cs:1,0.502265695453306)
--(axis cs:1,0.502265695453306)
--cycle;

\addplot [semithick, black, dashed]
table {%
1 0.4999234
1 0.4999234
4.6 0.6906942
5.8 0.6907492
7.5 0.6947783
12.3 0.7063117
15.2 0.7030075
19.7 0.7385275
26.2 0.7464383
34.1 0.7650301
42.5 0.7721259
55.7 0.7779309
72.7 0.7894483
95.2 0.7993101
116.8 0.8130508
147.6 0.8225517
189.9 0.8315941
225.7 0.83912
257.1 0.8480567
292.2 0.8570709
311.7 0.8671965
360.7 0.874606
441.5 0.8827367
480.8 0.889841
548.7 0.8965501
610.5 0.902675
679.9 0.9078149
770.7 0.9129801
857.8 0.9173633
962.1 0.9208784
1067.1 0.9243883
1197.7 0.9279317
1356.9 0.9310391
1509.8 0.9339358
1700.7 0.9364659
2013 0.9389159
2313.8 0.9413751
2423 0.9441774
2656.3 0.9462418
2979.9 0.9482233
3338.7 0.9501642
3709.4 0.9519382
4101 0.953369
4529.6 0.9548307
4994 0.9560959
5533.2 0.9572467
6100.8 0.958345
6738.1 0.9596558
7440.2 0.9607476
8217 0.961705
9043.1 0.9626059
10008.2 0.9634574
10967.9 0.9641775
11806.5 0.9649659
13154 0.9657015
14409 0.9663574
15679.8 0.9668275
17075.7 0.9673815
18546.3 0.9678842
20025.7 0.9683008
21636.8 0.9687065
23496.5 0.9689742
25376.7 0.9692266
27315.6 0.9694092
29182.3 0.9696084
31006.2 0.9698684
33135.9 0.9699484
35352.8 0.9700375
37493.6 0.9701875
39483.8 0.9702182
40868.4 0.9702007
43308.7 0.9702208
45580.9 0.9701951
48173.1 0.970105
51057.1 0.9699958
54538.4 0.9699492
57527.6 0.969855
60246.6 0.96978
63510.6 0.96976
66424.1 0.9696409
};
\addlegendentry{\scriptsize LIBLINEAR}
\addplot [steelblue31119180, opacity=1.0]
table {%
32.4 0.6609717
58.8 0.6084407
127.5 0.6668501
171.6 0.7107968
217.9 0.6850708
326.9 0.7399642
509.4 0.7605575
788.7 0.8016101
1289 0.8032342
1725.2 0.7864325
2166.5 0.7943866
2240.8 0.7033767
2245.6 0.7969859
2387.4 0.7891693
2566 0.7358316
3221.1 0.8263383
4167.6 0.8059683
5525.8 0.8211159
7395.5 0.8268866
7828.6 0.8344349
8253.4 0.8450558
9385.6 0.8391692
10880.7 0.8391257
10937.9 0.8379968
11114.5 0.8254417
14048.3 0.8640984
17864 0.8635291
22418.9 0.8642167
27844.9 0.8669134
32950.1 0.8912859
34305.1 0.8762451
36925.7 0.8842425
40353.4 0.87839
40403 0.8884117
48736.7 0.8958274
57236.4 0.9023592
64772.3 0.9019792
70835.6 0.9053384
73835.5 0.9077692
73897.5 0.9087325
75936.4 0.899104
78130.4 0.9033468
78380.3 0.9090217
82358.5 0.91331
85825.1 0.9138458
88807.5 0.909065
91221.2 0.9140617
93274.1 0.9160293
94236.4 0.919219
94659.6 0.9066783
94979.7 0.9180991
95824.2 0.9153907
96030.7 0.9210592
96183.6 0.9179344
97208.1 0.9217375
98125.7 0.9178958
98850.9 0.921061
99302.8 0.9198525
99597 0.9212516
99796.5 0.9233641
99901.9 0.9215975
99960 0.9215168
99986.5 0.9216968
99996.4 0.9246725
99999.3 0.9227158
100000 0.925986
100000 0.9235251
100000 0.9239408
100000 0.8941242
100000 0.9230567
100000 0.9248734
100000 0.9238191
100000 0.9233042
100000 0.9239068
100000 0.9220575
100000 0.9242985
100000 0.9234633
100000 0.9204566
100000 0.9246549
100000 0.9217275
};
\addlegendentry{\scriptsize OWA+Centroid}
\addplot [darkorange25512714, opacity=1.0]
table {%
1 0.5009067
1 0.5009067
1 0.5009067
1 0.5009067
1 0.5009067
1 0.5009067
1 0.5009067
1 0.5009067
1 0.5009067
1 0.5009067
1 0.5009067
1 0.5009067
1 0.5009067
4.1 0.6745599
5 0.6782507
6.8 0.6791909
15 0.72512
18.5 0.7519659
32.2 0.7728433
51.2 0.7706657
80.4 0.7880307
124 0.8116423
194.5 0.8157457
271.1 0.8362036
379.2 0.8596892
549.7 0.8581392
805.2 0.8766933
1166.5 0.8898508
1624.3 0.9006491
2139.3 0.9087109
2727.9 0.9100643
3402.5 0.9166617
4072.6 0.9180759
4744.8 0.9204441
5416.8 0.9231858
6045.5 0.9245833
6679.2 0.9243801
7427.1 0.9264441
8257.6 0.9272659
9167.8 0.9285592
10202.2 0.9306899
11497.8 0.9295592
12978.6 0.9301125
14687.8 0.9282583
16251.6 0.930289
17915.3 0.9297834
19809.4 0.9316517
21719 0.93111
23808.3 0.9314724
26041.2 0.9293117
28546.5 0.9312666
31325.1 0.9312316
34215 0.9312023
37232.5 0.9292641
40157.2 0.9284851
43337 0.9297715
47025.5 0.9281749
50630.5 0.9280983
54228.7 0.9263325
55971.3 0.9296399
56343.6 0.9294876
57439.2 0.9321674
58731.4 0.9319109
59290.8 0.9311116
59314.8 0.9316299
62702.6 0.9289466
66328.5 0.9277284
70231.8 0.92563
74403.4 0.9273358
78528.1 0.9254292
82647.4 0.9247476
86473.3 0.9243792
90081.9 0.92772
93280 0.9235749
96047.2 0.9252473
98120.6 0.9268518
99313.6 0.925449
99847.9 0.9250768
99983.5 0.9239109
99999.5 0.9274491
};
\addlegendentry{\scriptsize OWA+Adaptive}
\addplot [forestgreen4416044, opacity=1.0]
table {%
2.6 0.6660658
3.1 0.6656083
3.6 0.6397951
4.3 0.6650285
5.1 0.6652268
7.5 0.6886768
17.8 0.7027133
23.7 0.7206033
28.1 0.7295992
36.9 0.7526368
48.8 0.7554641
58.7 0.763429
66.8 0.7683083
79.4 0.7814284
97.5 0.7938857
113.8 0.7933649
139.6 0.8035551
172.4 0.8211683
210.6 0.8336
262.2 0.8355842
312.8 0.8445291
370.4 0.8532266
433.5 0.8513859
544.3 0.8662908
700.6 0.8624542
907.2 0.8869368
1202.5 0.8923377
1596.8 0.90137
2047.9 0.9050299
2568.4 0.9099451
3118.9 0.9134301
3441.2 0.9140632
3905.5 0.9180292
4626 0.9176209
5415.9 0.9195235
6247 0.9216901
7084.9 0.9219867
7986.1 0.9252018
8870.5 0.9241376
9829 0.9254924
10719.6 0.9275858
12003.4 0.9259067
13726.8 0.9279766
15532.3 0.9277166
17246.6 0.9299459
18993.4 0.9278217
20802.3 0.9283291
22797.5 0.9287509
24941.3 0.93058
27217 0.9287234
29266.9 0.9290943
31296.7 0.9295283
33935.8 0.9253242
36856.7 0.9287284
39926.9 0.9262868
43094.4 0.9283181
46383.7 0.9270391
49598.7 0.9305418
52890 0.9287517
53166.1 0.9298117
54630.2 0.9295292
54726.7 0.9304059
55496 0.9282325
56491.1 0.9309099
57884.7 0.9288365
61427 0.9283883
65286.2 0.9294425
69319.4 0.9317133
73597.2 0.93099
77945.7 0.9279401
82333.8 0.9294933
86683.6 0.9311941
90999.8 0.9293132
94866.9 0.929425
97781.8 0.93106
99386.1 0.93146
99913.6 0.9295784
99996.2 0.93115
100000 0.931229
100000 0.9304099
};
\addlegendentry{\scriptsize ACOWA}
\addplot [crimson2143940, opacity=1.0]
table {%
1 0.5009067
1 0.5009067
1 0.5009067
1 0.4987258
1 0.5009067
1 0.5009067
1 0.5009067
1 0.5009067
1 0.5009067
1 0.4986592
1 0.4986567
1 0.5009067
1 0.4986592
7.7 0.6734308
27.9 0.6838134
45.3 0.7098833
72 0.724055
117 0.7341682
187.5 0.7716959
264.1 0.7701084
360.8 0.800265
526.1 0.8150183
823.3 0.8097769
1334.9 0.8419209
2053.2 0.8450984
2993.2 0.848785
4168.3 0.8696683
5665.8 0.8714217
7372.3 0.8710675
9439 0.8828851
11923.9 0.8881674
14917.8 0.8993526
18302.6 0.9030191
21750.3 0.8978358
24976.3 0.9096825
26979.4 0.9135826
28801.6 0.9139909
31742.1 0.9160992
35352.3 0.8908591
38972.2 0.9189109
42735.6 0.9220417
46285.3 0.9232083
49578.1 0.9238757
50161.6 0.9240982
50668.8 0.9255966
51619.1 0.9215716
52836.6 0.926335
56301.4 0.9269651
59957.7 0.9277542
63842.4 0.9239408
68150.9 0.9271883
71898 0.9272758
72696.8 0.9266459
74261.8 0.9282451
75746 0.9234509
76646.7 0.9276767
78564.7 0.9295675
79099.9 0.9255849
83017 0.9291533
87165.8 0.9292891
90929.4 0.9262317
94103.3 0.9261959
96605.6 0.92901
97058.3 0.9283067
98258.1 0.9277868
98302.5 0.9131109
98345.9 0.9272175
99192.2 0.9283567
99332.3 0.9273549
99334.4 0.9279058
99815.8 0.9273792
99968.8 0.92838
99996.3 0.9265066
100000 0.9274734
100000 0.9264574
100000 0.9266883
100000 0.9272558
100000 0.9266568
100000 0.92573
100000 0.9270391
};
\addlegendentry{\scriptsize OWA}
\end{axis}

\end{tikzpicture}

%% file: figs/app/beta_newsgroups.tex
\begin{tikzpicture}[scale=1.0]

\definecolor{crimson2143940}{RGB}{214,39,40}
\definecolor{darkorange25512714}{RGB}{255,127,14}
\definecolor{forestgreen4416044}{RGB}{44,160,44}
\definecolor{gainsboro229}{RGB}{229,229,229}
\definecolor{mediumpurple148103189}{RGB}{148,103,189}
\definecolor{steelblue31119180}{RGB}{31,119,180}

\begin{axis}[
width=0.95\textwidth,
height=0.35\textwidth,
legend cell align={left},
legend style={
  fill opacity=0.8,
  draw opacity=1,
  text opacity=1,
  at={(0.97,0.03)},
  anchor=south east,
  draw=none
},
log basis x={10},
tick align=outside,
tick pos=left,
x grid style={gainsboro229},
xlabel={\small Number of non-zeros},
xmajorgrids,
xmin=10, xmax=10000,
xminorgrids,
xmode=log,
xtick style={color=black},
xtick={1,10,100,1000,10000,100000},
xticklabels={
  \(\displaystyle {10^{0}}\),
  \(\displaystyle {10^{1}}\),
  \(\displaystyle {10^{2}}\),
  \(\displaystyle {10^{3}}\),
  \(\displaystyle {10^{4}}\),
  \(\displaystyle {10^{5}}\)
},
y grid style={gainsboro229},
ylabel={\small Accuracy},
ymajorgrids,
ymin=0.452470554815425, ymax=0.982630200562852,
yminorgrids,
ytick style={color=black}
]
\path [draw=black, fill=black, opacity=0.25]
(axis cs:7.3,0.624772282907598)
--(axis cs:7.3,0.613204717092402)
--(axis cs:11.4,0.631247485374618)
--(axis cs:13.5,0.648572938214047)
--(axis cs:16,0.664945327847899)
--(axis cs:18.7,0.68385420813806)
--(axis cs:21.3,0.703969053276345)
--(axis cs:24.9,0.716209587409845)
--(axis cs:28.2,0.726607875625531)
--(axis cs:33.4,0.741865913887423)
--(axis cs:39.8,0.753759780437761)
--(axis cs:45.9,0.769052438600199)
--(axis cs:53.5,0.782332362952856)
--(axis cs:61,0.793152448555292)
--(axis cs:67.5,0.80384760022996)
--(axis cs:74.6,0.81515832255594)
--(axis cs:82,0.823062481384222)
--(axis cs:91.4,0.829507132048915)
--(axis cs:109.4,0.837471686134735)
--(axis cs:124.7,0.847529532207815)
--(axis cs:140.9,0.855194840539648)
--(axis cs:160.4,0.863955764618935)
--(axis cs:182.5,0.872029004565359)
--(axis cs:206.1,0.879665349140664)
--(axis cs:229,0.885224251906663)
--(axis cs:256.2,0.88990423081424)
--(axis cs:286.3,0.894067745319329)
--(axis cs:317.6,0.899035476430673)
--(axis cs:350.9,0.903957154462824)
--(axis cs:387.7,0.907923454920331)
--(axis cs:429.3,0.91195119892004)
--(axis cs:476,0.915986056127355)
--(axis cs:522.7,0.919797135860345)
--(axis cs:577.6,0.922420242346947)
--(axis cs:629.6,0.923579017568873)
--(axis cs:687.7,0.926207682605642)
--(axis cs:745,0.930078154011115)
--(axis cs:801.4,0.932766993538885)
--(axis cs:873.1,0.934816417796579)
--(axis cs:1036,0.936936685905346)
--(axis cs:1401.2,0.93777411955527)
--(axis cs:1533.2,0.938835122161917)
--(axis cs:1562.1,0.941636342065576)
--(axis cs:1883.6,0.943173383697857)
--(axis cs:2046.6,0.942501893527464)
--(axis cs:2117.6,0.942920694981242)
--(axis cs:2374.8,0.944503198267039)
--(axis cs:2471.6,0.944015524007796)
--(axis cs:2551.1,0.943909795930695)
--(axis cs:2801.1,0.945370914482137)
--(axis cs:3146.6,0.94470525974192)
--(axis cs:3224.2,0.94399823836618)
--(axis cs:3582.5,0.944379817740586)
--(axis cs:3794.1,0.946515678141047)
--(axis cs:3969.8,0.94530419152968)
--(axis cs:4102.8,0.944337613319276)
--(axis cs:4122.7,0.945271965008471)
--(axis cs:4385.9,0.945871107294661)
--(axis cs:4553.2,0.94538690623264)
--(axis cs:4682.6,0.944352668307144)
--(axis cs:4683.1,0.943812204075351)
--(axis cs:4762.9,0.946027468100693)
--(axis cs:5062.3,0.945079835556285)
--(axis cs:5251.8,0.946304793657303)
--(axis cs:5525.2,0.939933952806881)
--(axis cs:5573.5,0.946729658490001)
--(axis cs:5608.7,0.943227733745367)
--(axis cs:5613.2,0.9379917308298)
--(axis cs:5710.4,0.943732737692214)
--(axis cs:6237.3,0.943334154053545)
--(axis cs:6252.9,0.939918146632955)
--(axis cs:6581,0.943651674567081)
--(axis cs:6639.7,0.942131828344086)
--(axis cs:6807.8,0.940874122961058)
--(axis cs:7319,0.940588779955173)
--(axis cs:7326.6,0.939884563147506)
--(axis cs:7536.8,0.942280720336347)
--(axis cs:7887,0.942011622332943)
--(axis cs:8036.3,0.940405923598633)
--(axis cs:8469.7,0.932698554443028)
--(axis cs:8680.1,0.939623127972319)
--(axis cs:8735.5,0.930962941020383)
--(axis cs:8857,0.940367672795517)
--(axis cs:9067.1,0.933889674144042)
--(axis cs:9213.9,0.941352171540595)
--(axis cs:9215.7,0.938372405782447)
--(axis cs:9465.2,0.940486275769612)
--(axis cs:9546.4,0.932346972754151)
--(axis cs:9997.5,0.932912867750874)
--(axis cs:10796.1,0.930227499809329)
--(axis cs:11340.7,0.930381047836809)
--(axis cs:12327,0.930050106556492)
--(axis cs:13121.2,0.929726655800776)
--(axis cs:13931.2,0.928631400798918)
--(axis cs:14631.9,0.927149742960196)
--(axis cs:15522.7,0.923902297336413)
--(axis cs:16460.8,0.924597029080089)
--(axis cs:17085.6,0.924032097983951)
--(axis cs:18776.2,0.921976617884333)
--(axis cs:20681.6,0.916404289540328)
--(axis cs:23180.2,0.913404379009665)
--(axis cs:24451.9,0.913195190242937)
--(axis cs:26376,0.909814785585069)
--(axis cs:28263.5,0.9071240769576)
--(axis cs:30205.2,0.905533100487785)
--(axis cs:32350.5,0.902434820964977)
--(axis cs:35605.1,0.900793163610903)
--(axis cs:36660.8,0.897215029781394)
--(axis cs:37040.5,0.898469125806124)
--(axis cs:38795,0.896057326736172)
--(axis cs:40263,0.893156866385775)
--(axis cs:41167.8,0.891780085798999)
--(axis cs:41431.4,0.890973259028528)
--(axis cs:42466.8,0.889847184153025)
--(axis cs:43649,0.889462684112021)
--(axis cs:45484.2,0.887977635682561)
--(axis cs:46192.1,0.886390901088182)
--(axis cs:46828.7,0.885790703809368)
--(axis cs:47407,0.885274581719501)
--(axis cs:47909.8,0.885143664391175)
--(axis cs:48345.3,0.884264455880051)
--(axis cs:48345.3,0.898521744119949)
--(axis cs:48345.3,0.898521744119949)
--(axis cs:47909.8,0.899683135608825)
--(axis cs:47407,0.900351018280499)
--(axis cs:46828.7,0.901343296190632)
--(axis cs:46192.1,0.902517698911818)
--(axis cs:45484.2,0.903237964317438)
--(axis cs:43649,0.906100715887979)
--(axis cs:42466.8,0.907491015846975)
--(axis cs:41431.4,0.908583140971472)
--(axis cs:41167.8,0.910793314201001)
--(axis cs:40263,0.911989733614225)
--(axis cs:38795,0.913348273263827)
--(axis cs:37040.5,0.914041674193876)
--(axis cs:36660.8,0.916893170218607)
--(axis cs:35605.1,0.916864436389097)
--(axis cs:32350.5,0.917440779035023)
--(axis cs:30205.2,0.919223099512215)
--(axis cs:28263.5,0.9194067230424)
--(axis cs:26376,0.921507614414931)
--(axis cs:24451.9,0.923007009757063)
--(axis cs:23180.2,0.924661220990335)
--(axis cs:20681.6,0.925299310459672)
--(axis cs:18776.2,0.926559382115667)
--(axis cs:17085.6,0.927875702016049)
--(axis cs:16460.8,0.931658570919911)
--(axis cs:15522.7,0.932708302663587)
--(axis cs:14631.9,0.933897057039804)
--(axis cs:13931.2,0.938094199201082)
--(axis cs:13121.2,0.938330144199224)
--(axis cs:12327,0.937474293443508)
--(axis cs:11340.7,0.937587352163191)
--(axis cs:10796.1,0.943508100190671)
--(axis cs:9997.5,0.941266132249127)
--(axis cs:9546.4,0.942897027245849)
--(axis cs:9465.2,0.949220924230388)
--(axis cs:9215.7,0.949116794217553)
--(axis cs:9213.9,0.948621228459405)
--(axis cs:9067.1,0.945258925855958)
--(axis cs:8857,0.951113927204483)
--(axis cs:8735.5,0.945434458979617)
--(axis cs:8680.1,0.949108072027681)
--(axis cs:8469.7,0.943787645556972)
--(axis cs:8036.3,0.952938876401367)
--(axis cs:7887,0.951599777667057)
--(axis cs:7536.8,0.952218079663653)
--(axis cs:7326.6,0.951508236852493)
--(axis cs:7319,0.951248020044827)
--(axis cs:6807.8,0.950785277038942)
--(axis cs:6639.7,0.950059771655914)
--(axis cs:6581,0.95324272543292)
--(axis cs:6252.9,0.949167853367045)
--(axis cs:6237.3,0.949567445946455)
--(axis cs:5710.4,0.953694262307786)
--(axis cs:5613.2,0.9489646691702)
--(axis cs:5608.7,0.953666466254633)
--(axis cs:5573.5,0.955932141509999)
--(axis cs:5525.2,0.948619847193119)
--(axis cs:5251.8,0.954139006342697)
--(axis cs:5062.3,0.954121364443715)
--(axis cs:4762.9,0.953884131899307)
--(axis cs:4683.1,0.956010195924649)
--(axis cs:4682.6,0.953872931692856)
--(axis cs:4553.2,0.95408109376736)
--(axis cs:4385.9,0.955282092705339)
--(axis cs:4122.7,0.953574834991529)
--(axis cs:4102.8,0.954331386680724)
--(axis cs:3969.8,0.95531680847032)
--(axis cs:3794.1,0.953750721858953)
--(axis cs:3582.5,0.954821582259414)
--(axis cs:3224.2,0.952009161633819)
--(axis cs:3146.6,0.95405274025808)
--(axis cs:2801.1,0.952144685517863)
--(axis cs:2551.1,0.952363604069306)
--(axis cs:2471.6,0.951193075992204)
--(axis cs:2374.8,0.949817801732961)
--(axis cs:2117.6,0.950513105018758)
--(axis cs:2046.6,0.949689706472536)
--(axis cs:1883.6,0.949550816302143)
--(axis cs:1562.1,0.948425857934424)
--(axis cs:1533.2,0.947855077838083)
--(axis cs:1401.2,0.94625428044473)
--(axis cs:1036,0.944784514094654)
--(axis cs:873.1,0.942823182203421)
--(axis cs:801.4,0.940879806461115)
--(axis cs:745,0.937889845988885)
--(axis cs:687.7,0.936880317394359)
--(axis cs:629.6,0.934628782431127)
--(axis cs:577.6,0.931971957653053)
--(axis cs:522.7,0.929714864139655)
--(axis cs:476,0.927137543872644)
--(axis cs:429.3,0.923807601079961)
--(axis cs:387.7,0.921091545079669)
--(axis cs:350.9,0.916451045537176)
--(axis cs:317.6,0.912943123569327)
--(axis cs:286.3,0.909392654680671)
--(axis cs:256.2,0.90494976918576)
--(axis cs:229,0.899868948093337)
--(axis cs:206.1,0.894780050859336)
--(axis cs:182.5,0.886533395434641)
--(axis cs:160.4,0.878546435381065)
--(axis cs:140.9,0.869738559460352)
--(axis cs:124.7,0.860722467792185)
--(axis cs:109.4,0.854364913865265)
--(axis cs:91.4,0.847778267951085)
--(axis cs:82,0.840202918615778)
--(axis cs:74.6,0.83355527744406)
--(axis cs:67.5,0.82286059977004)
--(axis cs:61,0.812348751444708)
--(axis cs:53.5,0.801252437047144)
--(axis cs:45.9,0.7882681613998)
--(axis cs:39.8,0.776852419562239)
--(axis cs:33.4,0.762570886112578)
--(axis cs:28.2,0.748281124374469)
--(axis cs:24.9,0.734722212590155)
--(axis cs:21.3,0.721407946723655)
--(axis cs:18.7,0.70185999186194)
--(axis cs:16,0.681816072152101)
--(axis cs:13.5,0.663583261785953)
--(axis cs:11.4,0.644883914625382)
--(axis cs:7.3,0.624772282907598)
--cycle;

\path [fill=steelblue31119180, fill opacity=0.2]
(axis cs:15.3,0.67963607933462)
--(axis cs:15.3,0.63775532066538)
--(axis cs:20.4,0.476568720531217)
--(axis cs:24.4,0.557931386600591)
--(axis cs:30.3,0.692159194737832)
--(axis cs:36.4,0.623512406365208)
--(axis cs:45.3,0.574303886092865)
--(axis cs:59,0.723851876769112)
--(axis cs:79.6,0.620804702381559)
--(axis cs:98,0.74626567585919)
--(axis cs:121.5,0.774410903865758)
--(axis cs:142.5,0.683955604320484)
--(axis cs:160.4,0.767378269014767)
--(axis cs:181.3,0.769578652885162)
--(axis cs:211.8,0.793160594258842)
--(axis cs:252.9,0.743911559807903)
--(axis cs:308.7,0.807392810442297)
--(axis cs:393,0.785246176724789)
--(axis cs:494.4,0.782276853384888)
--(axis cs:601.8,0.825696306284057)
--(axis cs:739.5,0.836730333424467)
--(axis cs:864.2,0.840624583510516)
--(axis cs:957.6,0.848557122219704)
--(axis cs:1146.9,0.858304968834794)
--(axis cs:1460.2,0.870913742958926)
--(axis cs:1848.4,0.872290607203579)
--(axis cs:2244.4,0.872572519647601)
--(axis cs:2499.8,0.879352779795843)
--(axis cs:2676.2,0.878123105075384)
--(axis cs:2903.6,0.88264309236216)
--(axis cs:3332.6,0.869985672910326)
--(axis cs:3814.3,0.871086391521064)
--(axis cs:4231,0.874344064801596)
--(axis cs:5635.8,0.883099728581345)
--(axis cs:7860.3,0.896835811094194)
--(axis cs:9580.4,0.891225318615747)
--(axis cs:11342.3,0.894577852452242)
--(axis cs:13348.5,0.897860047228231)
--(axis cs:15747,0.898917224021649)
--(axis cs:18490.3,0.902147451580099)
--(axis cs:21363.7,0.911126923569655)
--(axis cs:24178.4,0.909064513414396)
--(axis cs:26670.7,0.909462621490625)
--(axis cs:28956.7,0.916754904299935)
--(axis cs:31225.5,0.918556813680955)
--(axis cs:33679.9,0.917905249416485)
--(axis cs:36288,0.928485010078066)
--(axis cs:39032.9,0.926499246714593)
--(axis cs:41794.3,0.932123838652834)
--(axis cs:44241.4,0.933878814227857)
--(axis cs:46346.6,0.933373984463407)
--(axis cs:48438.8,0.939198718751726)
--(axis cs:50120.6,0.936071178823143)
--(axis cs:50885.3,0.932224277457616)
--(axis cs:51139.1,0.938582666945027)
--(axis cs:51184.1,0.937540972111965)
--(axis cs:51186.4,0.933487897220195)
--(axis cs:51186.7,0.936001137358866)
--(axis cs:51186.7,0.942283760372734)
--(axis cs:51186.7,0.9385019953183)
--(axis cs:51186.7,0.939342632361632)
--(axis cs:51186.7,0.94590674333834)
--(axis cs:51186.7,0.945808263543772)
--(axis cs:51186.7,0.938133648021586)
--(axis cs:51186.7,0.941666521626939)
--(axis cs:51186.7,0.945521595645903)
--(axis cs:51186.7,0.941413677964716)
--(axis cs:51186.7,0.933781749891356)
--(axis cs:51186.7,0.924380988284864)
--(axis cs:51186.7,0.940649293452971)
--(axis cs:51186.7,0.940946579524064)
--(axis cs:51186.7,0.940800133649098)
--(axis cs:51186.7,0.93306501965266)
--(axis cs:51186.7,0.931338272676771)
--(axis cs:51186.7,0.920977613882316)
--(axis cs:51186.7,0.931149460199834)
--(axis cs:51186.7,0.929382901318017)
--(axis cs:51186.7,0.926166495744466)
--(axis cs:51186.7,0.938263952109948)
--(axis cs:51186.7,0.944051087209885)
--(axis cs:51186.7,0.938935902867604)
--(axis cs:51186.7,0.955740097132396)
--(axis cs:51186.7,0.955740097132396)
--(axis cs:51186.7,0.952665912790115)
--(axis cs:51186.7,0.953484047890052)
--(axis cs:51186.7,0.950053304255534)
--(axis cs:51186.7,0.946659698681983)
--(axis cs:51186.7,0.951192939800166)
--(axis cs:51186.7,0.948587386117684)
--(axis cs:51186.7,0.947898527323229)
--(axis cs:51186.7,0.95389198034734)
--(axis cs:51186.7,0.951835066350902)
--(axis cs:51186.7,0.952310020475936)
--(axis cs:51186.7,0.950477506547029)
--(axis cs:51186.7,0.949177411715136)
--(axis cs:51186.7,0.954683250108645)
--(axis cs:51186.7,0.958231122035284)
--(axis cs:51186.7,0.956341804354097)
--(axis cs:51186.7,0.957534878373062)
--(axis cs:51186.7,0.957429551978414)
--(axis cs:51186.7,0.957652336456228)
--(axis cs:51186.7,0.95400445666166)
--(axis cs:51186.7,0.954534767638368)
--(axis cs:51186.7,0.9564404046817)
--(axis cs:51186.7,0.954166639627266)
--(axis cs:51186.7,0.953173462641134)
--(axis cs:51186.4,0.952669902779805)
--(axis cs:51184.1,0.952609827888035)
--(axis cs:51139.1,0.952277933054973)
--(axis cs:50885.3,0.950029722542384)
--(axis cs:50120.6,0.948489421176857)
--(axis cs:48438.8,0.954856281248275)
--(axis cs:46346.6,0.953227415536593)
--(axis cs:44241.4,0.953255185772143)
--(axis cs:41794.3,0.952348161347166)
--(axis cs:39032.9,0.949543153285407)
--(axis cs:36288,0.951372989921934)
--(axis cs:33679.9,0.938172750583515)
--(axis cs:31225.5,0.943288586319045)
--(axis cs:28956.7,0.938169695700065)
--(axis cs:26670.7,0.940492978509375)
--(axis cs:24178.4,0.940891086585604)
--(axis cs:21363.7,0.933593876430345)
--(axis cs:18490.3,0.931924948419901)
--(axis cs:15747,0.93151737597835)
--(axis cs:13348.5,0.926984552771769)
--(axis cs:11342.3,0.922813347547758)
--(axis cs:9580.4,0.918446281384254)
--(axis cs:7860.3,0.920998988905806)
--(axis cs:5635.8,0.913705871418655)
--(axis cs:4231,0.902940535198404)
--(axis cs:3814.3,0.897680408478936)
--(axis cs:3332.6,0.902507727089674)
--(axis cs:2903.6,0.89739290763784)
--(axis cs:2676.2,0.897653094924616)
--(axis cs:2499.8,0.891454820204156)
--(axis cs:2244.4,0.895750280352399)
--(axis cs:1848.4,0.889732392796421)
--(axis cs:1460.2,0.889778257041074)
--(axis cs:1146.9,0.881801431165206)
--(axis cs:957.6,0.868656677780296)
--(axis cs:864.2,0.863013616489484)
--(axis cs:739.5,0.857591066575533)
--(axis cs:601.8,0.850435093715943)
--(axis cs:494.4,0.859781546615112)
--(axis cs:393,0.842881623275212)
--(axis cs:308.7,0.830850389557703)
--(axis cs:252.9,0.835857840192097)
--(axis cs:211.8,0.819261605741158)
--(axis cs:181.3,0.818442547114838)
--(axis cs:160.4,0.818691130985233)
--(axis cs:142.5,0.835831595679516)
--(axis cs:121.5,0.794710496134242)
--(axis cs:98,0.79357452414081)
--(axis cs:79.6,0.843258897618441)
--(axis cs:59,0.764347123230888)
--(axis cs:45.3,0.827559513907135)
--(axis cs:36.4,0.785804193634791)
--(axis cs:30.3,0.720707005262168)
--(axis cs:24.4,0.780932613399409)
--(axis cs:20.4,0.762916879468783)
--(axis cs:15.3,0.67963607933462)
--cycle;

\path [fill=darkorange25512714, fill opacity=0.2]
(axis cs:15.7,0.681336621654244)
--(axis cs:15.7,0.646702178345756)
--(axis cs:20.7,0.554247289746892)
--(axis cs:24.8,0.700807732792523)
--(axis cs:29.9,0.667463530230574)
--(axis cs:35.1,0.719255315295266)
--(axis cs:42.9,0.727623511869949)
--(axis cs:53.8,0.740770454299727)
--(axis cs:67.6,0.679346162355913)
--(axis cs:80.2,0.767417151312804)
--(axis cs:94.2,0.773616479381718)
--(axis cs:107.9,0.773507685400143)
--(axis cs:120.5,0.784370946643498)
--(axis cs:133.5,0.787134404549673)
--(axis cs:156.3,0.800569318922348)
--(axis cs:191.2,0.756225541386923)
--(axis cs:240.3,0.807651930469998)
--(axis cs:305,0.813123744285079)
--(axis cs:382.5,0.809781371154776)
--(axis cs:470.5,0.83409432629809)
--(axis cs:589.2,0.841238404358246)
--(axis cs:675.8,0.851437614805192)
--(axis cs:727,0.850990325172736)
--(axis cs:847.9,0.865507537136265)
--(axis cs:1055.1,0.872711897412219)
--(axis cs:1332.3,0.86299766044286)
--(axis cs:1624.9,0.868645174057)
--(axis cs:1900.4,0.877540519702307)
--(axis cs:2166.5,0.878630314262451)
--(axis cs:2447.9,0.873484194035148)
--(axis cs:2883.3,0.879130281554182)
--(axis cs:3398.2,0.875878923278804)
--(axis cs:4042.2,0.874216908571054)
--(axis cs:5980.7,0.889029626934089)
--(axis cs:7564.4,0.896388775907877)
--(axis cs:8958.7,0.892859912723246)
--(axis cs:10587.2,0.892855767355352)
--(axis cs:12545.2,0.901189393980082)
--(axis cs:14900.7,0.904235023261221)
--(axis cs:17605.9,0.897768611184405)
--(axis cs:20527.4,0.915479970683992)
--(axis cs:23365.7,0.917494388044468)
--(axis cs:25966.9,0.924563934116923)
--(axis cs:28221.5,0.920049450023514)
--(axis cs:30460.2,0.914763544337182)
--(axis cs:32836.4,0.939490127179138)
--(axis cs:35401.2,0.934611862123565)
--(axis cs:38114,0.933364428098567)
--(axis cs:40849.2,0.933228594964111)
--(axis cs:43376,0.936372849923647)
--(axis cs:45510.7,0.935573185405524)
--(axis cs:47701.6,0.939648223238675)
--(axis cs:49738.2,0.93575194975314)
--(axis cs:50808.3,0.930292503315859)
--(axis cs:51129,0.931981883103089)
--(axis cs:51182.2,0.931558872313635)
--(axis cs:51186.4,0.928724230927198)
--(axis cs:51186.7,0.940155044126262)
--(axis cs:51186.7,0.941318345047785)
--(axis cs:51186.7,0.945307194963999)
--(axis cs:51186.7,0.942065955257631)
--(axis cs:51186.7,0.944926319545143)
--(axis cs:51186.7,0.944312393599625)
--(axis cs:51186.7,0.942798965152941)
--(axis cs:51186.7,0.943957545124107)
--(axis cs:51186.7,0.946308738687669)
--(axis cs:51186.7,0.943103341115007)
--(axis cs:51186.7,0.936389273688494)
--(axis cs:51186.7,0.9267070634692)
--(axis cs:51186.7,0.942026772585987)
--(axis cs:51186.7,0.940427019643131)
--(axis cs:51186.7,0.942786058476215)
--(axis cs:51186.7,0.931584213265162)
--(axis cs:51186.7,0.931738796483362)
--(axis cs:51186.7,0.920748385040116)
--(axis cs:51186.7,0.933006448892429)
--(axis cs:51186.7,0.925616049883813)
--(axis cs:51186.7,0.916532266196946)
--(axis cs:51186.7,0.932423750760065)
--(axis cs:51186.7,0.945689352372218)
--(axis cs:51186.7,0.938448915506527)
--(axis cs:51186.7,0.953387484493474)
--(axis cs:51186.7,0.953387484493474)
--(axis cs:51186.7,0.951826247627783)
--(axis cs:51186.7,0.947966649239935)
--(axis cs:51186.7,0.939723333803055)
--(axis cs:51186.7,0.943860150116187)
--(axis cs:51186.7,0.948182551107571)
--(axis cs:51186.7,0.946509814959884)
--(axis cs:51186.7,0.948740203516638)
--(axis cs:51186.7,0.952444186734838)
--(axis cs:51186.7,0.949583141523785)
--(axis cs:51186.7,0.952119580356869)
--(axis cs:51186.7,0.949011427414013)
--(axis cs:51186.7,0.9468511365308)
--(axis cs:51186.7,0.954382526311506)
--(axis cs:51186.7,0.958138658884993)
--(axis cs:51186.7,0.955909261312331)
--(axis cs:51186.7,0.955865054875893)
--(axis cs:51186.7,0.958532034847059)
--(axis cs:51186.7,0.956841006400375)
--(axis cs:51186.7,0.953387680454857)
--(axis cs:51186.7,0.954739644742369)
--(axis cs:51186.7,0.956556205036001)
--(axis cs:51186.7,0.952559054952215)
--(axis cs:51186.7,0.953278955873738)
--(axis cs:51186.4,0.947318369072802)
--(axis cs:51182.2,0.949629527686365)
--(axis cs:51129,0.946988916896911)
--(axis cs:50808.3,0.947968496684141)
--(axis cs:49738.2,0.94792145024686)
--(axis cs:47701.6,0.953075576761325)
--(axis cs:45510.7,0.954400414594476)
--(axis cs:43376,0.949873750076353)
--(axis cs:40849.2,0.952574405035889)
--(axis cs:38114,0.953857971901433)
--(axis cs:35401.2,0.948528937876435)
--(axis cs:32836.4,0.948442472820862)
--(axis cs:30460.2,0.944153655662819)
--(axis cs:28221.5,0.943215749976486)
--(axis cs:25966.9,0.942339665883076)
--(axis cs:23365.7,0.939914611955532)
--(axis cs:20527.4,0.935895429316008)
--(axis cs:17605.9,0.936215388815595)
--(axis cs:14900.7,0.928861776738779)
--(axis cs:12545.2,0.930931406019918)
--(axis cs:10587.2,0.921874032644648)
--(axis cs:8958.7,0.926040287276754)
--(axis cs:7564.4,0.915235024092123)
--(axis cs:5980.7,0.914164373065911)
--(axis cs:4042.2,0.898365491428946)
--(axis cs:3398.2,0.898389076721196)
--(axis cs:2883.3,0.904454318445817)
--(axis cs:2447.9,0.899452805964852)
--(axis cs:2166.5,0.899186685737549)
--(axis cs:1900.4,0.896727480297693)
--(axis cs:1624.9,0.895330025943)
--(axis cs:1332.3,0.88607053955714)
--(axis cs:1055.1,0.888601102587781)
--(axis cs:847.9,0.878503062863735)
--(axis cs:727,0.872967074827264)
--(axis cs:675.8,0.873584785194808)
--(axis cs:589.2,0.861246195641754)
--(axis cs:470.5,0.84744947370191)
--(axis cs:382.5,0.851354428845224)
--(axis cs:305,0.833016655714921)
--(axis cs:240.3,0.829880869530002)
--(axis cs:191.2,0.838095858613077)
--(axis cs:156.3,0.820903281077652)
--(axis cs:133.5,0.810558795450327)
--(axis cs:120.5,0.816782253356502)
--(axis cs:107.9,0.803333714599857)
--(axis cs:94.2,0.795859920618282)
--(axis cs:80.2,0.796735448687196)
--(axis cs:67.6,0.815685037644087)
--(axis cs:53.8,0.768457545700273)
--(axis cs:42.9,0.761551288130051)
--(axis cs:35.1,0.748446684704734)
--(axis cs:29.9,0.742119669769426)
--(axis cs:24.8,0.717737067207477)
--(axis cs:20.7,0.768733910253108)
--(axis cs:15.7,0.681336621654244)
--cycle;

\path [fill=forestgreen4416044, fill opacity=0.2]
(axis cs:16.7,0.688899480485175)
--(axis cs:16.7,0.657240719514825)
--(axis cs:21.4,0.600555420224268)
--(axis cs:25.1,0.683076075546978)
--(axis cs:30.2,0.707435345435295)
--(axis cs:35.1,0.677130762415193)
--(axis cs:41.8,0.744316957584877)
--(axis cs:51.8,0.746111536517223)
--(axis cs:60.9,0.761130546465675)
--(axis cs:72.5,0.772217348186041)
--(axis cs:80.8,0.772782289143529)
--(axis cs:92.9,0.765244407051118)
--(axis cs:104.3,0.788935370371676)
--(axis cs:117.5,0.798335750841875)
--(axis cs:138.6,0.807707492315678)
--(axis cs:170.7,0.811978884670775)
--(axis cs:207.6,0.818736664056687)
--(axis cs:266.2,0.820161528564487)
--(axis cs:339.4,0.833066963076521)
--(axis cs:421.7,0.843390440070543)
--(axis cs:508.3,0.845898192999978)
--(axis cs:570.3,0.857215193646862)
--(axis cs:602.1,0.854037055340054)
--(axis cs:697.3,0.864576922238522)
--(axis cs:863.8,0.871877154528817)
--(axis cs:1049.7,0.869546945983856)
--(axis cs:1281,0.874809779657298)
--(axis cs:1574.1,0.870317843200127)
--(axis cs:1839,0.877938894254391)
--(axis cs:2179.3,0.867741355542054)
--(axis cs:2670.7,0.874375772573956)
--(axis cs:3257.5,0.879037770933114)
--(axis cs:4409.7,0.875610301312473)
--(axis cs:6052.6,0.885176704344751)
--(axis cs:7184,0.900256880214144)
--(axis cs:8503.2,0.891552085083294)
--(axis cs:10072.2,0.883474661063062)
--(axis cs:12029.9,0.906478712052998)
--(axis cs:14354.6,0.909422993580352)
--(axis cs:17041.9,0.910908509091264)
--(axis cs:19957.7,0.913753344435439)
--(axis cs:22843.1,0.916003969203645)
--(axis cs:25575.2,0.915350021769481)
--(axis cs:27895.1,0.919314434660817)
--(axis cs:30136.5,0.923331500246685)
--(axis cs:32560.4,0.928535837312732)
--(axis cs:35149.3,0.932306190084216)
--(axis cs:37894.2,0.928131296509083)
--(axis cs:40579.2,0.928723025173162)
--(axis cs:43102.6,0.93730335939443)
--(axis cs:45207.4,0.929715919832537)
--(axis cs:47470.8,0.93974378575156)
--(axis cs:49710.3,0.937164827371134)
--(axis cs:50858.7,0.931950250091107)
--(axis cs:51148.3,0.928183680435025)
--(axis cs:51184.4,0.932923593757711)
--(axis cs:51186.7,0.942375854924909)
--(axis cs:51186.7,0.943161424143847)
--(axis cs:51186.7,0.946430252602828)
--(axis cs:51186.7,0.948373031824119)
--(axis cs:51186.7,0.945325281298946)
--(axis cs:51186.7,0.94443060497104)
--(axis cs:51186.7,0.945932723903707)
--(axis cs:51186.7,0.946226199480975)
--(axis cs:51186.7,0.9469575696074)
--(axis cs:51186.7,0.940596314016009)
--(axis cs:51186.7,0.939200945360351)
--(axis cs:51186.7,0.925741485958172)
--(axis cs:51186.7,0.943767276833498)
--(axis cs:51186.7,0.942416703022589)
--(axis cs:51186.7,0.941434833574501)
--(axis cs:51186.7,0.935511899819755)
--(axis cs:51186.7,0.935366015797805)
--(axis cs:51186.7,0.922739868334699)
--(axis cs:51186.7,0.936792632499989)
--(axis cs:51186.7,0.930922790703218)
--(axis cs:51186.7,0.9230367433086)
--(axis cs:51186.7,0.930099182948583)
--(axis cs:51186.7,0.929198152319828)
--(axis cs:51186.7,0.944670755335304)
--(axis cs:51186.7,0.941629765743126)
--(axis cs:51186.7,0.955087634256874)
--(axis cs:51186.7,0.955087634256874)
--(axis cs:51186.7,0.954796644664696)
--(axis cs:51186.7,0.944981047680172)
--(axis cs:51186.7,0.944878617051417)
--(axis cs:51186.7,0.9387198566914)
--(axis cs:51186.7,0.941304409296782)
--(axis cs:51186.7,0.947147167500011)
--(axis cs:51186.7,0.945760331665301)
--(axis cs:51186.7,0.947153784202196)
--(axis cs:51186.7,0.951533300180245)
--(axis cs:51186.7,0.949958166425499)
--(axis cs:51186.7,0.952969096977411)
--(axis cs:51186.7,0.952505923166502)
--(axis cs:51186.7,0.947106714041828)
--(axis cs:51186.7,0.953966854639649)
--(axis cs:51186.7,0.954700885983991)
--(axis cs:51186.7,0.9557932303926)
--(axis cs:51186.7,0.955637000519025)
--(axis cs:51186.7,0.956551276096293)
--(axis cs:51186.7,0.95663439502896)
--(axis cs:51186.7,0.956892918701054)
--(axis cs:51186.7,0.956063568175881)
--(axis cs:51186.7,0.956941147397172)
--(axis cs:51186.7,0.953733175856153)
--(axis cs:51186.7,0.955583145075091)
--(axis cs:51184.4,0.944006606242289)
--(axis cs:51148.3,0.944930519564975)
--(axis cs:50858.7,0.947197949908893)
--(axis cs:49710.3,0.944201572628866)
--(axis cs:47470.8,0.94969701424844)
--(axis cs:45207.4,0.951117880167463)
--(axis cs:43102.6,0.94787844060557)
--(axis cs:40579.2,0.949981574826838)
--(axis cs:37894.2,0.953413103490916)
--(axis cs:35149.3,0.946398209915785)
--(axis cs:32560.4,0.946264362687268)
--(axis cs:30136.5,0.939933899753315)
--(axis cs:27895.1,0.939692165339183)
--(axis cs:25575.2,0.94001837823052)
--(axis cs:22843.1,0.936879630796355)
--(axis cs:19957.7,0.936113855564561)
--(axis cs:17041.9,0.929907690908736)
--(axis cs:14354.6,0.932280606419648)
--(axis cs:12029.9,0.926351487947002)
--(axis cs:10072.2,0.922292738936937)
--(axis cs:8503.2,0.919095714916706)
--(axis cs:7184,0.919707719785856)
--(axis cs:6052.6,0.913492095655249)
--(axis cs:4409.7,0.905756098687527)
--(axis cs:3257.5,0.899666429066886)
--(axis cs:2670.7,0.899093627426044)
--(axis cs:2179.3,0.896233844457946)
--(axis cs:1839,0.893489905745609)
--(axis cs:1574.1,0.890019156799873)
--(axis cs:1281,0.893779420342702)
--(axis cs:1049.7,0.889903054016144)
--(axis cs:863.8,0.888637045471183)
--(axis cs:697.3,0.884580277761478)
--(axis cs:602.1,0.874001944659946)
--(axis cs:570.3,0.877390006353138)
--(axis cs:508.3,0.865459407000023)
--(axis cs:421.7,0.859715359929458)
--(axis cs:339.4,0.856196636923479)
--(axis cs:266.2,0.839732071435513)
--(axis cs:207.6,0.836010535943313)
--(axis cs:170.7,0.823069915329225)
--(axis cs:138.6,0.824058107684322)
--(axis cs:117.5,0.816393649158125)
--(axis cs:104.3,0.813726429628324)
--(axis cs:92.9,0.815056992948883)
--(axis cs:80.8,0.798291510856471)
--(axis cs:72.5,0.796371851813959)
--(axis cs:60.9,0.783057653534325)
--(axis cs:51.8,0.771723463482777)
--(axis cs:41.8,0.761628042415123)
--(axis cs:35.1,0.768832037584807)
--(axis cs:30.2,0.740568254564705)
--(axis cs:25.1,0.723134724453022)
--(axis cs:21.4,0.744520179775732)
--(axis cs:16.7,0.688899480485175)
--cycle;

\path [fill=crimson2143940, fill opacity=0.2]
(axis cs:18.1,0.698436062015446)
--(axis cs:18.1,0.653383137984554)
--(axis cs:22,0.681786035318947)
--(axis cs:26.2,0.69402968986421)
--(axis cs:31.1,0.717034199277445)
--(axis cs:35.9,0.723194900031569)
--(axis cs:41.8,0.734537475045164)
--(axis cs:50.5,0.753079575531301)
--(axis cs:59.7,0.756766219219327)
--(axis cs:69.8,0.777234569784225)
--(axis cs:79.2,0.780157053923963)
--(axis cs:91.2,0.782585845016086)
--(axis cs:101.9,0.789360006721187)
--(axis cs:117.8,0.801943792726913)
--(axis cs:138.9,0.811973556896535)
--(axis cs:168.8,0.807046154239721)
--(axis cs:212.3,0.82175180986304)
--(axis cs:263.8,0.83051506180294)
--(axis cs:334.3,0.835690259664752)
--(axis cs:422.4,0.842126417644586)
--(axis cs:478.8,0.852360408248698)
--(axis cs:514.6,0.865497510860739)
--(axis cs:568.6,0.863544905369335)
--(axis cs:668.9,0.872858083479786)
--(axis cs:808.5,0.873971177007834)
--(axis cs:969.6,0.867910383238469)
--(axis cs:1165.8,0.875085943860788)
--(axis cs:1460.3,0.867464403888299)
--(axis cs:1754.5,0.879451912561593)
--(axis cs:2185,0.860171817321845)
--(axis cs:2708.2,0.876069084813621)
--(axis cs:3447.6,0.864633613773272)
--(axis cs:4739.1,0.866702938686128)
--(axis cs:5918.1,0.877434634847759)
--(axis cs:6983.3,0.897271425965919)
--(axis cs:8314.8,0.879920417251504)
--(axis cs:9887.1,0.892974212461893)
--(axis cs:11838.6,0.889076826787942)
--(axis cs:14155.1,0.909716476891442)
--(axis cs:16847.1,0.904761986346153)
--(axis cs:19737.1,0.922578982594979)
--(axis cs:22663,0.909703483809562)
--(axis cs:25491.4,0.917832670805325)
--(axis cs:27952.5,0.924705254711041)
--(axis cs:30308.3,0.923281114400619)
--(axis cs:32781.1,0.926279895313393)
--(axis cs:35465.9,0.930143690133602)
--(axis cs:38267,0.930480239934592)
--(axis cs:40922.4,0.924525445172675)
--(axis cs:43347,0.925242336007614)
--(axis cs:45405.3,0.932122981949715)
--(axis cs:47879.4,0.932982267427307)
--(axis cs:50141.9,0.93223980623294)
--(axis cs:51002.8,0.926923988001949)
--(axis cs:51171.4,0.926294926790623)
--(axis cs:51185.8,0.929714228270043)
--(axis cs:51186.7,0.947311456630741)
--(axis cs:51186.7,0.945889450333067)
--(axis cs:51186.7,0.947735601541526)
--(axis cs:51186.7,0.949699518369926)
--(axis cs:51186.7,0.945854681815706)
--(axis cs:51186.7,0.944830975238685)
--(axis cs:51186.7,0.945222404689938)
--(axis cs:51186.7,0.949285761573226)
--(axis cs:51186.7,0.944978682970883)
--(axis cs:51186.7,0.944626810184873)
--(axis cs:51186.7,0.94176200908081)
--(axis cs:51186.7,0.929876582640832)
--(axis cs:51186.7,0.945202507169202)
--(axis cs:51186.7,0.942668231669418)
--(axis cs:51186.7,0.942861585365072)
--(axis cs:51186.7,0.939420515989694)
--(axis cs:51186.7,0.932402613793435)
--(axis cs:51186.7,0.92375707631292)
--(axis cs:51186.7,0.937313203830805)
--(axis cs:51186.7,0.928459798928172)
--(axis cs:51186.7,0.922159916228019)
--(axis cs:51186.7,0.929553115618686)
--(axis cs:51186.7,0.927696859823134)
--(axis cs:51186.7,0.945906330241845)
--(axis cs:51186.7,0.94477611954916)
--(axis cs:51186.7,0.95637728045084)
--(axis cs:51186.7,0.95637728045084)
--(axis cs:51186.7,0.952496469758156)
--(axis cs:51186.7,0.942844340176865)
--(axis cs:51186.7,0.942053084381314)
--(axis cs:51186.7,0.939242083771981)
--(axis cs:51186.7,0.939242001071828)
--(axis cs:51186.7,0.944851796169195)
--(axis cs:51186.7,0.94492072368708)
--(axis cs:51186.7,0.946834586206565)
--(axis cs:51186.7,0.949133284010306)
--(axis cs:51186.7,0.950217214634928)
--(axis cs:51186.7,0.953693768330582)
--(axis cs:51186.7,0.950627092830798)
--(axis cs:51186.7,0.946254817359168)
--(axis cs:51186.7,0.95415639091919)
--(axis cs:51186.7,0.956171589815127)
--(axis cs:51186.7,0.956884517029117)
--(axis cs:51186.7,0.955949238426774)
--(axis cs:51186.7,0.956197195310062)
--(axis cs:51186.7,0.957120824761315)
--(axis cs:51186.7,0.955919918184294)
--(axis cs:51186.7,0.954737081630074)
--(axis cs:51186.7,0.955724798458474)
--(axis cs:51186.7,0.955884949666933)
--(axis cs:51186.7,0.954817943369259)
--(axis cs:51185.8,0.940649371729957)
--(axis cs:51171.4,0.939188673209377)
--(axis cs:51002.8,0.940156611998051)
--(axis cs:50141.9,0.94078579376706)
--(axis cs:47879.4,0.946875532572693)
--(axis cs:45405.3,0.947469018050285)
--(axis cs:43347,0.944500263992386)
--(axis cs:40922.4,0.945394554827325)
--(axis cs:38267,0.946804560065408)
--(axis cs:35465.9,0.942882109866398)
--(axis cs:32781.1,0.940534704686607)
--(axis cs:30308.3,0.944066085599381)
--(axis cs:27952.5,0.941310745288959)
--(axis cs:25491.4,0.942149729194676)
--(axis cs:22663,0.931556316190438)
--(axis cs:19737.1,0.930837217405021)
--(axis cs:16847.1,0.923631813653847)
--(axis cs:14155.1,0.927018123108558)
--(axis cs:11838.6,0.913585173212058)
--(axis cs:9887.1,0.918737987538107)
--(axis cs:8314.8,0.914844182748496)
--(axis cs:6983.3,0.911512974034081)
--(axis cs:5918.1,0.911385165152241)
--(axis cs:4739.1,0.901975061313872)
--(axis cs:3447.6,0.899075186226728)
--(axis cs:2708.2,0.894738315186379)
--(axis cs:2185,0.892889382678155)
--(axis cs:1754.5,0.892154287438407)
--(axis cs:1460.3,0.894558796111701)
--(axis cs:1165.8,0.890841456139212)
--(axis cs:969.6,0.887280216761531)
--(axis cs:808.5,0.889205422992166)
--(axis cs:668.9,0.887745316520214)
--(axis cs:568.6,0.881086694630665)
--(axis cs:514.6,0.878335689139261)
--(axis cs:478.8,0.870798191751302)
--(axis cs:422.4,0.862487382355414)
--(axis cs:334.3,0.859340940335248)
--(axis cs:263.8,0.84694713819706)
--(axis cs:212.3,0.84098139013696)
--(axis cs:168.8,0.831995445760279)
--(axis cs:138.9,0.829375243103465)
--(axis cs:117.8,0.819263207273087)
--(axis cs:101.9,0.815076193278813)
--(axis cs:91.2,0.818034954983914)
--(axis cs:79.2,0.806267346076037)
--(axis cs:69.8,0.797920630215775)
--(axis cs:59.7,0.784759780780673)
--(axis cs:50.5,0.774604424468699)
--(axis cs:41.8,0.765906124954836)
--(axis cs:35.9,0.757018299968431)
--(axis cs:31.1,0.748804200722554)
--(axis cs:26.2,0.72504751013579)
--(axis cs:22,0.716793964681053)
--(axis cs:18.1,0.698436062015446)
--cycle;

\path [fill=mediumpurple148103189, fill opacity=0.2]
(axis cs:1,0.519148047067594)
--(axis cs:1,0.495492752932406)
--(axis cs:2,0.495492752932406)
--(axis cs:2.1,0.495492752932406)
--(axis cs:2.9,0.492635781241291)
--(axis cs:5.2,0.484209165546269)
--(axis cs:9.7,0.485262460194799)
--(axis cs:17.7,0.483374932433542)
--(axis cs:29.9,0.507248083473834)
--(axis cs:41,0.522814393009595)
--(axis cs:62.7,0.562059595553426)
--(axis cs:97,0.581475759017076)
--(axis cs:156.3,0.65401809194456)
--(axis cs:252.7,0.659975882506262)
--(axis cs:368,0.686560407560902)
--(axis cs:528.1,0.730693388671303)
--(axis cs:721,0.685702362141722)
--(axis cs:914.9,0.711594724559135)
--(axis cs:1115,0.71913230173124)
--(axis cs:1301.5,0.737696859483488)
--(axis cs:1495.6,0.735813777490295)
--(axis cs:1717.9,0.750304782353736)
--(axis cs:1974.7,0.78678345204557)
--(axis cs:2296.8,0.75560371840339)
--(axis cs:2697.2,0.77712485181585)
--(axis cs:3213.5,0.78321180074869)
--(axis cs:3842.1,0.791704342679013)
--(axis cs:4559.1,0.800489587738192)
--(axis cs:5337.6,0.805022193037423)
--(axis cs:6211.6,0.815752925633566)
--(axis cs:7276.8,0.821264922458353)
--(axis cs:8564.9,0.828572064455924)
--(axis cs:10194.2,0.833603866284344)
--(axis cs:12430,0.842848762763069)
--(axis cs:15686.2,0.850967716082887)
--(axis cs:18649.3,0.865200012663845)
--(axis cs:20641.8,0.872146339573804)
--(axis cs:21860.1,0.865354168352582)
--(axis cs:22502.2,0.8686748894178)
--(axis cs:22663.2,0.874034499682571)
--(axis cs:24792.3,0.878894529983825)
--(axis cs:25912.6,0.878240027564261)
--(axis cs:27102.5,0.884457020315291)
--(axis cs:28340.4,0.89533768793038)
--(axis cs:29614.5,0.885346564156009)
--(axis cs:30895.5,0.887243785179481)
--(axis cs:32151.7,0.879794671582061)
--(axis cs:33349.7,0.885029226056822)
--(axis cs:34555,0.878211485239985)
--(axis cs:35664.6,0.886588437576802)
--(axis cs:36766.4,0.886779445427672)
--(axis cs:37774.1,0.882101220794095)
--(axis cs:38726.1,0.889246918214568)
--(axis cs:39612.4,0.885524133527058)
--(axis cs:40456.9,0.893566650349531)
--(axis cs:41211.1,0.881769880897905)
--(axis cs:41898.8,0.880556667717857)
--(axis cs:42511.6,0.884149516706119)
--(axis cs:43091,0.881645016649593)
--(axis cs:43622.6,0.87965229077008)
--(axis cs:44094.9,0.875853492420425)
--(axis cs:44529.1,0.878876702802744)
--(axis cs:44923.4,0.882149705139537)
--(axis cs:45284,0.885185137421607)
--(axis cs:45630.5,0.890134841918139)
--(axis cs:45942.8,0.886666108701753)
--(axis cs:46239.8,0.885865407749239)
--(axis cs:46505.7,0.879955754497957)
--(axis cs:46756.8,0.885894255370952)
--(axis cs:46987,0.887767006783955)
--(axis cs:47200.3,0.875955960887698)
--(axis cs:47399.7,0.883531931504175)
--(axis cs:47586.1,0.878189483232413)
--(axis cs:47761.3,0.884083355062749)
--(axis cs:47926.8,0.883350890243031)
--(axis cs:48069.1,0.887272738528218)
--(axis cs:48201.8,0.885362744787827)
--(axis cs:48327.8,0.890482467381193)
--(axis cs:48447.5,0.879529014399903)
--(axis cs:48563.5,0.891152257385538)
--(axis cs:48673.9,0.89396644166268)
--(axis cs:48673.9,0.90816335833732)
--(axis cs:48673.9,0.90816335833732)
--(axis cs:48563.5,0.911154742614462)
--(axis cs:48447.5,0.906096385600097)
--(axis cs:48327.8,0.906678132618808)
--(axis cs:48201.8,0.912241655212173)
--(axis cs:48069.1,0.912283661471782)
--(axis cs:47926.8,0.912567509756969)
--(axis cs:47761.3,0.905978844937251)
--(axis cs:47586.1,0.900426516767587)
--(axis cs:47399.7,0.910079668495824)
--(axis cs:47200.3,0.906475639112302)
--(axis cs:46987,0.907973993216045)
--(axis cs:46756.8,0.907539544629048)
--(axis cs:46505.7,0.907178045502043)
--(axis cs:46239.8,0.905438792250761)
--(axis cs:45942.8,0.911027091298247)
--(axis cs:45630.5,0.909953958081861)
--(axis cs:45284,0.903989862578392)
--(axis cs:44923.4,0.905871894860462)
--(axis cs:44529.1,0.903554497197256)
--(axis cs:44094.9,0.901342907579575)
--(axis cs:43622.6,0.90641730922992)
--(axis cs:43091,0.905577783350407)
--(axis cs:42511.6,0.910348883293882)
--(axis cs:41898.8,0.910126332282143)
--(axis cs:41211.1,0.914414719102095)
--(axis cs:40456.9,0.913886749650469)
--(axis cs:39612.4,0.906401466472943)
--(axis cs:38726.1,0.909422081785432)
--(axis cs:37774.1,0.903080179205905)
--(axis cs:36766.4,0.909405354572328)
--(axis cs:35664.6,0.914032962423198)
--(axis cs:34555,0.903864514760015)
--(axis cs:33349.7,0.907783573943178)
--(axis cs:32151.7,0.907782928417939)
--(axis cs:30895.5,0.906101214820519)
--(axis cs:29614.5,0.909063235843991)
--(axis cs:28340.4,0.91069591206962)
--(axis cs:27102.5,0.907379579684708)
--(axis cs:25912.6,0.903214972435739)
--(axis cs:24792.3,0.899721470016175)
--(axis cs:22663.2,0.900056100317429)
--(axis cs:22502.2,0.893436710582201)
--(axis cs:21860.1,0.899064831647418)
--(axis cs:20641.8,0.895200860426196)
--(axis cs:18649.3,0.894161187336155)
--(axis cs:15686.2,0.878047483917113)
--(axis cs:12430,0.878801637236931)
--(axis cs:10194.2,0.865153733715656)
--(axis cs:8564.9,0.865482735544076)
--(axis cs:7276.8,0.866312677541647)
--(axis cs:6211.6,0.862951474366434)
--(axis cs:5337.6,0.843868606962577)
--(axis cs:4559.1,0.840859212261808)
--(axis cs:3842.1,0.840239057320987)
--(axis cs:3213.5,0.82033739925131)
--(axis cs:2697.2,0.81799474818415)
--(axis cs:2296.8,0.807838881596609)
--(axis cs:1974.7,0.82271054795443)
--(axis cs:1717.9,0.800449817646264)
--(axis cs:1495.6,0.786103022509705)
--(axis cs:1301.5,0.788123940516512)
--(axis cs:1115,0.784416898268759)
--(axis cs:914.9,0.778822275440865)
--(axis cs:721,0.765850437858278)
--(axis cs:528.1,0.782616211328697)
--(axis cs:368,0.749287192439098)
--(axis cs:252.7,0.770813717493738)
--(axis cs:156.3,0.75041810805544)
--(axis cs:97,0.715241240982924)
--(axis cs:62.7,0.677248404446574)
--(axis cs:41,0.651098806990405)
--(axis cs:29.9,0.592042316526166)
--(axis cs:17.7,0.559304867566458)
--(axis cs:9.7,0.532660939805201)
--(axis cs:5.2,0.514903634453731)
--(axis cs:2.9,0.518367018758709)
--(axis cs:2.1,0.519148047067594)
--(axis cs:2,0.519148047067594)
--(axis cs:1,0.519148047067594)
--cycle;

\addplot [semithick, black, dashed]
table {%
7.3 0.6189885
11.4 0.6380657
13.5 0.6560781
16 0.6733807
18.7 0.6928571
21.3 0.7126885
24.9 0.7254659
28.2 0.7374445
33.4 0.7522184
39.8 0.7653061
45.9 0.7786603
53.5 0.7917924
61 0.8027506
67.5 0.8133541
74.6 0.8243568
82 0.8316327
91.4 0.8386427
109.4 0.8459183
124.7 0.854126
140.9 0.8624667
160.4 0.8712511
182.5 0.8792812
206.1 0.8872227
229 0.8925466
256.2 0.897427
286.3 0.9017302
317.6 0.9059893
350.9 0.9102041
387.7 0.9145075
429.3 0.9178794
476 0.9215618
522.7 0.924756
577.6 0.9271961
629.6 0.9291039
687.7 0.931544
745 0.933984
801.4 0.9368234
873.1 0.9388198
1036 0.9408606
1401.2 0.9420142
1533.2 0.9433451
1562.1 0.9450311
1883.6 0.9463621
2046.6 0.9460958
2117.6 0.9467169
2374.8 0.9471605
2471.6 0.9476043
2551.1 0.9481367
2801.1 0.9487578
3146.6 0.949379
3224.2 0.9480037
3582.5 0.9496007
3794.1 0.9501332
3969.8 0.9503105
4102.8 0.9493345
4122.7 0.9494234
4385.9 0.9505766
4553.2 0.949734
4682.6 0.9491128
4683.1 0.9499112
4762.9 0.9499558
5062.3 0.9496006
5251.8 0.9502219
5525.2 0.9442769
5573.5 0.9513309
5608.7 0.9484471
5613.2 0.9434782
5710.4 0.9487135
6237.3 0.9464508
6252.9 0.944543
6581 0.9484472
6639.7 0.9460958
6807.8 0.9458297
7319 0.9459184
7326.6 0.9456964
7536.8 0.9472494
7887 0.9468057
8036.3 0.9466724
8469.7 0.9382431
8680.1 0.9443656
8735.5 0.9381987
8857 0.9457408
9067.1 0.9395743
9213.9 0.9449867
9215.7 0.9437446
9465.2 0.9448536
9546.4 0.937622
9997.5 0.9370895
10796.1 0.9368678
11340.7 0.9339842
12327 0.9337622
13121.2 0.9340284
13931.2 0.9333628
14631.9 0.9305234
15522.7 0.9283053
16460.8 0.9281278
17085.6 0.9259539
18776.2 0.924268
20681.6 0.9208518
23180.2 0.9190328
24451.9 0.9181011
26376 0.9156612
28263.5 0.9132654
30205.2 0.9123781
32350.5 0.9099378
35605.1 0.9088288
36660.8 0.9070541
37040.5 0.9062554
38795 0.9047028
40263 0.9025733
41167.8 0.9012867
41431.4 0.8997782
42466.8 0.8986691
43649 0.8977817
45484.2 0.8956078
46192.1 0.8944543
46828.7 0.893567
47407 0.8928128
47909.8 0.8924134
48345.3 0.8913931
};
\addlegendentry{\scriptsize LIBLINEAR}
\addplot [steelblue31119180, opacity=1.0]
table {%
15.3 0.6586957
20.4 0.6197428
24.4 0.669432
30.3 0.7064331
36.4 0.7046583
45.3 0.7009317
59 0.7440995
79.6 0.7320318
98 0.7699201
121.5 0.7845607
142.5 0.7598936
160.4 0.7930347
181.3 0.7940106
211.8 0.8062111
252.9 0.7898847
308.7 0.8191216
393 0.8140639
494.4 0.8210292
601.8 0.8380657
739.5 0.8471607
864.2 0.8518191
957.6 0.8586069
1146.9 0.8700532
1460.2 0.880346
1848.4 0.8810115
2244.4 0.8841614
2499.8 0.8854038
2676.2 0.8878881
2903.6 0.890018
3332.6 0.8862467
3814.3 0.8843834
4231 0.8886423
5635.8 0.8984028
7860.3 0.9089174
9580.4 0.9048358
11342.3 0.9086956
13348.5 0.9124223
15747 0.9152173
18490.3 0.9170362
21363.7 0.9223604
24178.4 0.9249778
26670.7 0.9249778
28956.7 0.9274623
31225.5 0.9309227
33679.9 0.928039
36288 0.939929
39032.9 0.9380212
41794.3 0.942236
44241.4 0.943567
46346.6 0.9433007
48438.8 0.9470275
50120.6 0.9422803
50885.3 0.941127
51139.1 0.9454303
51184.1 0.9450754
51186.4 0.9430789
51186.7 0.9445873
51186.7 0.9482252
51186.7 0.9474712
51186.7 0.9469387
51186.7 0.9499556
51186.7 0.9517303
51186.7 0.9477816
51186.7 0.9496007
51186.7 0.9509317
51186.7 0.9498224
51186.7 0.9442325
51186.7 0.9367792
51186.7 0.9455634
51186.7 0.9466283
51186.7 0.9463176
51186.7 0.9434785
51186.7 0.9396184
51186.7 0.9347825
51186.7 0.9411712
51186.7 0.9380213
51186.7 0.9381099
51186.7 0.945874
51186.7 0.9483585
51186.7 0.947338
};
\addlegendentry{\scriptsize ACOWA, $\beta=5$}
\addplot [darkorange25512714, opacity=1.0]
table {%
15.7 0.6640194
20.7 0.6614906
24.8 0.7092724
29.9 0.7047916
35.1 0.733851
42.9 0.7445874
53.8 0.754614
67.6 0.7475156
80.2 0.7820763
94.2 0.7847382
107.9 0.7884207
120.5 0.8005766
133.5 0.7988466
156.3 0.8107363
191.2 0.7971607
240.3 0.8187664
305 0.8230702
382.5 0.8305679
470.5 0.8407719
589.2 0.8512423
675.8 0.8625112
727 0.8619787
847.9 0.8720053
1055.1 0.8806565
1332.3 0.8745341
1624.9 0.8819876
1900.4 0.887134
2166.5 0.8889085
2447.9 0.8864685
2883.3 0.8917923
3398.2 0.887134
4042.2 0.8862912
5980.7 0.901597
7564.4 0.9058119
8958.7 0.9094501
10587.2 0.9073649
12545.2 0.9160604
14900.7 0.9165484
17605.9 0.916992
20527.4 0.9256877
23365.7 0.9287045
25966.9 0.9334518
28221.5 0.9316326
30460.2 0.9294586
32836.4 0.9439663
35401.2 0.9415704
38114 0.9436112
40849.2 0.9429015
43376 0.9431233
45510.7 0.9449868
47701.6 0.9463619
49738.2 0.9418367
50808.3 0.9391305
51129 0.9394854
51182.2 0.9405942
51186.4 0.9380213
51186.7 0.946717
51186.7 0.9469387
51186.7 0.9509317
51186.7 0.9484028
51186.7 0.949157
51186.7 0.9505767
51186.7 0.9506655
51186.7 0.9499113
51186.7 0.951109
51186.7 0.950621
51186.7 0.9453859
51186.7 0.9367791
51186.7 0.9455191
51186.7 0.9462733
51186.7 0.9461846
51186.7 0.9420142
51186.7 0.9402395
51186.7 0.9336291
51186.7 0.9405945
51186.7 0.9347381
51186.7 0.9281278
51186.7 0.9401952
51186.7 0.9487578
51186.7 0.9459182
};
\addlegendentry{\scriptsize ACOWA, $\beta=10$}
\addplot [forestgreen4416044, opacity=1.0]
table {%
16.7 0.6730701
21.4 0.6725378
25.1 0.7031054
30.2 0.7240018
35.1 0.7229814
41.8 0.7529725
51.8 0.7589175
60.9 0.7720941
72.5 0.7842946
80.8 0.7855369
92.9 0.7901507
104.3 0.8013309
117.5 0.8073647
138.6 0.8158828
170.7 0.8175244
207.6 0.8273736
266.2 0.8299468
339.4 0.8446318
421.7 0.8515529
508.3 0.8556788
570.3 0.8673026
602.1 0.8640195
697.3 0.8745786
863.8 0.8802571
1049.7 0.879725
1281 0.8842946
1574.1 0.8801685
1839 0.8857144
2179.3 0.8819876
2670.7 0.8867347
3257.5 0.8893521
4409.7 0.8906832
6052.6 0.8993344
7184 0.9099823
8503.2 0.9053239
10072.2 0.9028837
12029.9 0.9164151
14354.6 0.9208518
17041.9 0.9204081
19957.7 0.9249336
22843.1 0.9264418
25575.2 0.9276842
27895.1 0.9295033
30136.5 0.9316327
32560.4 0.9374001
35149.3 0.9393522
37894.2 0.9407722
40579.2 0.9393523
43102.6 0.9425909
45207.4 0.9404169
47470.8 0.9447204
49710.3 0.9406832
50858.7 0.9395741
51148.3 0.9365571
51184.4 0.9384651
51186.7 0.9489795
51186.7 0.9484473
51186.7 0.9516857
51186.7 0.9522183
51186.7 0.9511091
51186.7 0.9505325
51186.7 0.951242
51186.7 0.9509316
51186.7 0.9513754
51186.7 0.9476486
51186.7 0.9465839
51186.7 0.9364241
51186.7 0.9481366
51186.7 0.9476929
51186.7 0.9456965
51186.7 0.9435226
51186.7 0.9412599
51186.7 0.9342501
51186.7 0.9419699
51186.7 0.9361136
51186.7 0.9308783
51186.7 0.9374889
51186.7 0.9370896
51186.7 0.9497337
51186.7 0.9483587
};
\addlegendentry{\scriptsize ACOWA, $\beta=20$}
\addplot [crimson2143940, opacity=1.0]
table {%
18.1 0.6759096
22 0.69929
26.2 0.7095386
31.1 0.7329192
35.9 0.7401066
41.8 0.7502218
50.5 0.763842
59.7 0.770763
69.8 0.7875776
79.2 0.7932122
91.2 0.8003104
101.9 0.8022181
117.8 0.8106035
138.9 0.8206744
168.8 0.8195208
212.3 0.8313666
263.8 0.8387311
334.3 0.8475156
422.4 0.8523069
478.8 0.8615793
514.6 0.8719166
568.6 0.8723158
668.9 0.8803017
808.5 0.8815883
969.6 0.8775953
1165.8 0.8829637
1460.3 0.8810116
1754.5 0.8858031
2185 0.8765306
2708.2 0.8854037
3447.6 0.8818544
4739.1 0.884339
5918.1 0.8944099
6983.3 0.9043922
8314.8 0.8973823
9887.1 0.9058561
11838.6 0.901331
14155.1 0.9183673
16847.1 0.9141969
19737.1 0.9267081
22663 0.9206299
25491.4 0.9299912
27952.5 0.933008
30308.3 0.9336736
32781.1 0.9334073
35465.9 0.9365129
38267 0.9386424
40922.4 0.93496
43347 0.9348713
45405.3 0.939796
47879.4 0.9399289
50141.9 0.9365128
51002.8 0.9335403
51171.4 0.9327418
51185.8 0.9351818
51186.7 0.9510647
51186.7 0.9508872
51186.7 0.9517302
51186.7 0.9522183
51186.7 0.9508873
51186.7 0.9509759
51186.7 0.9507098
51186.7 0.9526175
51186.7 0.9509316
51186.7 0.9503992
51186.7 0.9479592
51186.7 0.9380657
51186.7 0.9479148
51186.7 0.948181
51186.7 0.9465394
51186.7 0.9442769
51186.7 0.9396186
51186.7 0.9343389
51186.7 0.9410825
51186.7 0.9338509
51186.7 0.930701
51186.7 0.9358031
51186.7 0.9352706
51186.7 0.9492014
51186.7 0.9505767
};
\addlegendentry{\scriptsize ACOWA, $\beta=40$}
\addplot [mediumpurple148103189, opacity=1.0]
table {%
1 0.5073204
2 0.5073204
2.1 0.5073204
2.9 0.5055014
5.2 0.4995564
9.7 0.5089617
17.7 0.5213399
29.9 0.5496452
41 0.5869566
62.7 0.619654
97 0.6483585
156.3 0.7022181
252.7 0.7153948
368 0.7179238
528.1 0.7566548
721 0.7257764
914.9 0.7452085
1115 0.7517746
1301.5 0.7629104
1495.6 0.7609584
1717.9 0.7753773
1974.7 0.804747
2296.8 0.7817213
2697.2 0.7975598
3213.5 0.8017746
3842.1 0.8159717
4559.1 0.8206744
5337.6 0.8244454
6211.6 0.8393522
7276.8 0.8437888
8564.9 0.8470274
10194.2 0.8493788
12430 0.8608252
15686.2 0.8645076
18649.3 0.8796806
20641.8 0.8836736
21860.1 0.8822095
22502.2 0.8810558
22663.2 0.8870453
24792.3 0.889308
25912.6 0.8907275
27102.5 0.8959183
28340.4 0.9030168
29614.5 0.8972049
30895.5 0.8966725
32151.7 0.8937888
33349.7 0.8964064
34555 0.891038
35664.6 0.9003107
36766.4 0.8980924
37774.1 0.8925907
38726.1 0.8993345
39612.4 0.8959628
40456.9 0.9037267
41211.1 0.8980923
41898.8 0.8953415
42511.6 0.8972492
43091 0.8936114
43622.6 0.8930348
44094.9 0.8885982
44529.1 0.8912156
44923.4 0.8940108
45284 0.8945875
45630.5 0.9000444
45942.8 0.8988466
46239.8 0.8956521
46505.7 0.8935669
46756.8 0.8967169
46987 0.8978705
47200.3 0.8912158
47399.7 0.8968058
47586.1 0.889308
47761.3 0.8950311
47926.8 0.8979592
48069.1 0.8997782
48201.8 0.8988022
48327.8 0.8985803
48447.5 0.8928127
48563.5 0.9011535
48673.9 0.9010649
};
\addlegendentry{\scriptsize OWA}
\end{axis}

\end{tikzpicture}

%% file: figs/app/beta_ember100k.tex
\begin{tikzpicture}[scale=1.0]
 
\definecolor{crimson2143940}{RGB}{214,39,40}
\definecolor{darkorange25512714}{RGB}{255,127,14}
\definecolor{forestgreen4416044}{RGB}{44,160,44}
\definecolor{gainsboro229}{RGB}{229,229,229}
\definecolor{mediumpurple148103189}{RGB}{148,103,189}
\definecolor{steelblue31119180}{RGB}{31,119,180}

\begin{axis}[
width=0.95\textwidth,
height=0.35\textwidth,
legend cell align={left},
legend style={
  fill opacity=0.8,
  draw opacity=1,
  text opacity=1,
  at={(0.97,0.03)},
  anchor=south east,
  draw=none
},
log basis x={10},
tick align=outside,
tick pos=left,
x grid style={gainsboro229},
xlabel={\small Number of non-zeros},
xmajorgrids,
xmin=10, xmax=10000,
xminorgrids,
xmode=log,
xtick style={color=black},
xtick={1,10,100,1000,10000,100000},
xticklabels={
  \(\displaystyle {10^{0}}\),
  \(\displaystyle {10^{1}}\),
  \(\displaystyle {10^{2}}\),
  \(\displaystyle {10^{3}}\),
  \(\displaystyle {10^{4}}\),
  \(\displaystyle {10^{5}}\)
},
y grid style={gainsboro229},
ylabel={\small Accuracy},
ymajorgrids,
ymin=0.438301798309981, ymax=0.996081740434645,
yminorgrids,
ytick style={color=black}
]
\path [draw=black, fill=black, opacity=0.25]
(axis cs:1,0.500981347719775)
--(axis cs:1,0.498865452280225)
--(axis cs:1,0.498865452280225)
--(axis cs:4.6,0.689452662709201)
--(axis cs:5.8,0.689445789362906)
--(axis cs:7.5,0.692916890316155)
--(axis cs:12.3,0.705101937576124)
--(axis cs:15.2,0.701505585499548)
--(axis cs:19.7,0.737047495326584)
--(axis cs:26.2,0.743844371574276)
--(axis cs:34.1,0.763523122122849)
--(axis cs:42.5,0.77087733509491)
--(axis cs:55.7,0.776812474375393)
--(axis cs:72.7,0.788527149384978)
--(axis cs:95.2,0.798110882528294)
--(axis cs:116.8,0.811734882846242)
--(axis cs:147.6,0.820895147552563)
--(axis cs:189.9,0.830296647842886)
--(axis cs:225.7,0.837811820348729)
--(axis cs:257.1,0.846829749213158)
--(axis cs:292.2,0.855802912053159)
--(axis cs:311.7,0.866239987207497)
--(axis cs:360.7,0.873588211657017)
--(axis cs:441.5,0.881607939416193)
--(axis cs:480.8,0.88888521004167)
--(axis cs:548.7,0.895606301475125)
--(axis cs:610.5,0.901766748688719)
--(axis cs:679.9,0.906763027019709)
--(axis cs:770.7,0.91221695497956)
--(axis cs:857.8,0.916561706025472)
--(axis cs:962.1,0.920042898261189)
--(axis cs:1067.1,0.923671394915627)
--(axis cs:1197.7,0.927105641614923)
--(axis cs:1356.9,0.930110047382186)
--(axis cs:1509.8,0.93306206258712)
--(axis cs:1700.7,0.935640379362934)
--(axis cs:2013,0.938117999358734)
--(axis cs:2313.8,0.94063958091648)
--(axis cs:2423,0.943368218767725)
--(axis cs:2656.3,0.945516406497594)
--(axis cs:2979.9,0.947576444864234)
--(axis cs:3338.7,0.949485604096682)
--(axis cs:3709.4,0.95126722478933)
--(axis cs:4101,0.952573024846842)
--(axis cs:4529.6,0.954026417428042)
--(axis cs:4994,0.955383030922103)
--(axis cs:5533.2,0.956555777581136)
--(axis cs:6100.8,0.957680483968925)
--(axis cs:6738.1,0.958852395189902)
--(axis cs:7440.2,0.960071800851666)
--(axis cs:8217,0.961030896974574)
--(axis cs:9043.1,0.961944087597913)
--(axis cs:10008.2,0.96278700207505)
--(axis cs:10967.9,0.963575928308512)
--(axis cs:11806.5,0.964282186211603)
--(axis cs:13154,0.965032713495285)
--(axis cs:14409,0.965735570226938)
--(axis cs:15679.8,0.966206477143022)
--(axis cs:17075.7,0.966782303454616)
--(axis cs:18546.3,0.967319580993747)
--(axis cs:20025.7,0.967769191999057)
--(axis cs:21636.8,0.968192760928746)
--(axis cs:23496.5,0.968468099263431)
--(axis cs:25376.7,0.968679206337064)
--(axis cs:27315.6,0.968911525005651)
--(axis cs:29182.3,0.969086626485915)
--(axis cs:31006.2,0.969366805165719)
--(axis cs:33135.9,0.9694540026789)
--(axis cs:35352.8,0.969589564178704)
--(axis cs:37493.6,0.96972152140011)
--(axis cs:39483.8,0.969708293298294)
--(axis cs:40868.4,0.969703876299993)
--(axis cs:43308.7,0.9697482662152)
--(axis cs:45580.9,0.969673437816846)
--(axis cs:48173.1,0.969613694030613)
--(axis cs:51057.1,0.969501052971487)
--(axis cs:54538.4,0.969471849720971)
--(axis cs:57527.6,0.969307946072128)
--(axis cs:60246.6,0.969188645058644)
--(axis cs:63510.6,0.969193589861986)
--(axis cs:66424.1,0.969040115244682)
--(axis cs:66424.1,0.970241684755318)
--(axis cs:66424.1,0.970241684755318)
--(axis cs:63510.6,0.970326410138013)
--(axis cs:60246.6,0.970371354941357)
--(axis cs:57527.6,0.970402053927872)
--(axis cs:54538.4,0.970426550279029)
--(axis cs:51057.1,0.970490547028513)
--(axis cs:48173.1,0.970596305969387)
--(axis cs:45580.9,0.970716762183154)
--(axis cs:43308.7,0.9706933337848)
--(axis cs:40868.4,0.970697523700007)
--(axis cs:39483.8,0.970728106701706)
--(axis cs:37493.6,0.97065347859989)
--(axis cs:35352.8,0.970485435821296)
--(axis cs:33135.9,0.9704427973211)
--(axis cs:31006.2,0.970369994834281)
--(axis cs:29182.3,0.970130173514084)
--(axis cs:27315.6,0.969906874994349)
--(axis cs:25376.7,0.969773993662936)
--(axis cs:23496.5,0.969480300736569)
--(axis cs:21636.8,0.969220239071254)
--(axis cs:20025.7,0.968832408000943)
--(axis cs:18546.3,0.968448819006253)
--(axis cs:17075.7,0.967980696545384)
--(axis cs:15679.8,0.967448522856978)
--(axis cs:14409,0.966979229773062)
--(axis cs:13154,0.966370286504715)
--(axis cs:11806.5,0.965649613788397)
--(axis cs:10967.9,0.964779071691488)
--(axis cs:10008.2,0.96412779792495)
--(axis cs:9043.1,0.963267712402087)
--(axis cs:8217,0.962379103025426)
--(axis cs:7440.2,0.961423399148334)
--(axis cs:6738.1,0.960459204810098)
--(axis cs:6100.8,0.959009516031075)
--(axis cs:5533.2,0.957937622418864)
--(axis cs:4994,0.956808769077897)
--(axis cs:4529.6,0.955634982571958)
--(axis cs:4101,0.954164975153158)
--(axis cs:3709.4,0.95260917521067)
--(axis cs:3338.7,0.950842795903318)
--(axis cs:2979.9,0.948870155135766)
--(axis cs:2656.3,0.946967193502406)
--(axis cs:2423,0.944986581232275)
--(axis cs:2313.8,0.94211061908352)
--(axis cs:2013,0.939713800641266)
--(axis cs:1700.7,0.937291420637066)
--(axis cs:1509.8,0.93480953741288)
--(axis cs:1356.9,0.931968152617814)
--(axis cs:1197.7,0.928757758385077)
--(axis cs:1067.1,0.925105205084373)
--(axis cs:962.1,0.921713901738811)
--(axis cs:857.8,0.918164893974528)
--(axis cs:770.7,0.91374324502044)
--(axis cs:679.9,0.908866772980291)
--(axis cs:610.5,0.903583251311281)
--(axis cs:548.7,0.897493898524875)
--(axis cs:480.8,0.89079678995833)
--(axis cs:441.5,0.883865460583807)
--(axis cs:360.7,0.875623788342982)
--(axis cs:311.7,0.868153012792503)
--(axis cs:292.2,0.858338887946841)
--(axis cs:257.1,0.849283650786843)
--(axis cs:225.7,0.840428179651271)
--(axis cs:189.9,0.832891552157114)
--(axis cs:147.6,0.824208252447437)
--(axis cs:116.8,0.814366717153758)
--(axis cs:95.2,0.800509317471706)
--(axis cs:72.7,0.790369450615022)
--(axis cs:55.7,0.779049325624607)
--(axis cs:42.5,0.77337446490509)
--(axis cs:34.1,0.766537077877151)
--(axis cs:26.2,0.749032228425724)
--(axis cs:19.7,0.740007504673416)
--(axis cs:15.2,0.704509414500452)
--(axis cs:12.3,0.707521462423876)
--(axis cs:7.5,0.696639709683845)
--(axis cs:5.8,0.692052610637094)
--(axis cs:4.6,0.691935737290799)
--(axis cs:1,0.500981347719775)
--(axis cs:1,0.500981347719775)
--cycle;

\path [fill=steelblue31119180, fill opacity=0.2]
(axis cs:1,0.502302935100388)
--(axis cs:1,0.499510464899612)
--(axis cs:1,0.499510464899612)
--(axis cs:1,0.499510464899612)
--(axis cs:1.2,0.46365543204292)
--(axis cs:3.2,0.664320235986722)
--(axis cs:5.5,0.667067124672846)
--(axis cs:9.7,0.678529015079988)
--(axis cs:14.8,0.697694736174791)
--(axis cs:17.8,0.706232918151105)
--(axis cs:23.7,0.720851005031539)
--(axis cs:29.6,0.719798945690941)
--(axis cs:42.1,0.730445964200377)
--(axis cs:51.7,0.74654516445029)
--(axis cs:60.3,0.760405192629944)
--(axis cs:73.3,0.764899521087144)
--(axis cs:83.6,0.773397969549857)
--(axis cs:101.4,0.784905390211354)
--(axis cs:125.6,0.792187042556656)
--(axis cs:153.8,0.800865255820708)
--(axis cs:192.3,0.803747292446876)
--(axis cs:244.7,0.828378146308741)
--(axis cs:289.8,0.823570590279029)
--(axis cs:340.5,0.83667387571156)
--(axis cs:435.6,0.846835661435959)
--(axis cs:583.9,0.844937397199833)
--(axis cs:812.6,0.864721679472537)
--(axis cs:1146.7,0.885389750791758)
--(axis cs:1603.2,0.874329421874658)
--(axis cs:2161.8,0.885758369530332)
--(axis cs:2836,0.903872411460017)
--(axis cs:3594.6,0.894700983949573)
--(axis cs:4175.9,0.898717762436594)
--(axis cs:5009.1,0.90730342641361)
--(axis cs:6111.6,0.909985495014043)
--(axis cs:7423.9,0.913933813799143)
--(axis cs:8778.5,0.91952610410937)
--(axis cs:10221.5,0.919148847039702)
--(axis cs:11716.4,0.920857738257473)
--(axis cs:13296.1,0.919449949025626)
--(axis cs:14907.2,0.92157480209891)
--(axis cs:16476.1,0.923727844588276)
--(axis cs:18359.2,0.923393561240201)
--(axis cs:20244.5,0.922544147181932)
--(axis cs:22002,0.924306418367833)
--(axis cs:23720.5,0.926578585473497)
--(axis cs:24795.8,0.923949094318455)
--(axis cs:25398,0.926454506165414)
--(axis cs:26247.1,0.927041927565048)
--(axis cs:27722.1,0.928459317690102)
--(axis cs:29889.9,0.927484282707477)
--(axis cs:32306.4,0.925014820439439)
--(axis cs:35047.9,0.92764823137633)
--(axis cs:37921,0.925933306548706)
--(axis cs:40845.5,0.92648375643097)
--(axis cs:43887.3,0.924557577867182)
--(axis cs:47014.8,0.925460969747458)
--(axis cs:50282.7,0.924567591601108)
--(axis cs:53622.6,0.928001092795778)
--(axis cs:56948.2,0.924936912861186)
--(axis cs:60475.1,0.92742805412062)
--(axis cs:60972.4,0.926225206804554)
--(axis cs:63730,0.926559027285261)
--(axis cs:64100.1,0.927122438334547)
--(axis cs:64477.7,0.925682680922806)
--(axis cs:67337.1,0.928240954020246)
--(axis cs:67576.8,0.928865184039529)
--(axis cs:68316.7,0.927697657178471)
--(axis cs:71701.6,0.927643262438045)
--(axis cs:75977.4,0.928841933614037)
--(axis cs:80352.6,0.926220311853728)
--(axis cs:84686.7,0.928598475208263)
--(axis cs:88912.9,0.927847476588069)
--(axis cs:92918.8,0.927961163397533)
--(axis cs:96322.9,0.927497477305621)
--(axis cs:98584,0.928859006190842)
--(axis cs:99673.5,0.928340891754678)
--(axis cs:99963.6,0.927916273602336)
--(axis cs:99998.7,0.928446472745642)
--(axis cs:100000,0.928616316301001)
--(axis cs:100000,0.928224267254131)
--(axis cs:100000,0.933073732745869)
--(axis cs:100000,0.933073732745869)
--(axis cs:100000,0.933380083698999)
--(axis cs:99998.7,0.932693527254358)
--(axis cs:99963.6,0.932711926397664)
--(axis cs:99673.5,0.932570908245321)
--(axis cs:98584,0.933997593809158)
--(axis cs:96322.9,0.932960722694379)
--(axis cs:92918.8,0.933118836602467)
--(axis cs:88912.9,0.933732723411931)
--(axis cs:84686.7,0.933696524791737)
--(axis cs:80352.6,0.932693088146272)
--(axis cs:75977.4,0.933766066385963)
--(axis cs:71701.6,0.933381937561955)
--(axis cs:68316.7,0.932892542821529)
--(axis cs:67576.8,0.932846215960471)
--(axis cs:67337.1,0.932555845979754)
--(axis cs:64477.7,0.932379119077195)
--(axis cs:64100.1,0.932270961665453)
--(axis cs:63730,0.931195772714739)
--(axis cs:60972.4,0.932388393195446)
--(axis cs:60475.1,0.93314354587938)
--(axis cs:56948.2,0.932468087138814)
--(axis cs:53622.6,0.933380507204222)
--(axis cs:50282.7,0.930744208398892)
--(axis cs:47014.8,0.931659230252542)
--(axis cs:43887.3,0.930565822132818)
--(axis cs:40845.5,0.93278144356903)
--(axis cs:37921,0.932068693451294)
--(axis cs:35047.9,0.93273196862367)
--(axis cs:32306.4,0.930436779560561)
--(axis cs:29889.9,0.933527117292523)
--(axis cs:27722.1,0.932982482309898)
--(axis cs:26247.1,0.932501672434952)
--(axis cs:25398,0.932783893834586)
--(axis cs:24795.8,0.931550905681545)
--(axis cs:23720.5,0.931851414526503)
--(axis cs:22002,0.930969981632167)
--(axis cs:20244.5,0.930265652818068)
--(axis cs:18359.2,0.931410038759799)
--(axis cs:16476.1,0.931312155411724)
--(axis cs:14907.2,0.92910039790109)
--(axis cs:13296.1,0.928570050974374)
--(axis cs:11716.4,0.927760461742527)
--(axis cs:10221.5,0.927427952960298)
--(axis cs:8778.5,0.92569249589063)
--(axis cs:7423.9,0.924876186200857)
--(axis cs:6111.6,0.920566104985957)
--(axis cs:5009.1,0.91982657358639)
--(axis cs:4175.9,0.913665237563406)
--(axis cs:3594.6,0.913727216050427)
--(axis cs:2836,0.912652588539983)
--(axis cs:2161.8,0.907496630469668)
--(axis cs:1603.2,0.904368778125343)
--(axis cs:1146.7,0.896855449208242)
--(axis cs:812.6,0.888776520527463)
--(axis cs:583.9,0.869204202800168)
--(axis cs:435.6,0.867164338564041)
--(axis cs:340.5,0.85981952428844)
--(axis cs:289.8,0.849851009720971)
--(axis cs:244.7,0.851823853691259)
--(axis cs:192.3,0.832249707553124)
--(axis cs:153.8,0.823069544179292)
--(axis cs:125.6,0.816026357443344)
--(axis cs:101.4,0.808602809788646)
--(axis cs:83.6,0.797492030450143)
--(axis cs:73.3,0.785242278912856)
--(axis cs:60.3,0.776519807370056)
--(axis cs:51.7,0.76972163554971)
--(axis cs:42.1,0.766165835799623)
--(axis cs:29.6,0.753817654309059)
--(axis cs:23.7,0.743649194968461)
--(axis cs:17.8,0.730447281848896)
--(axis cs:14.8,0.702033663825209)
--(axis cs:9.7,0.702849384920012)
--(axis cs:5.5,0.691936275327154)
--(axis cs:3.2,0.667168164013278)
--(axis cs:1.2,0.60440796795708)
--(axis cs:1,0.502302935100388)
--(axis cs:1,0.502302935100388)
--(axis cs:1,0.502302935100388)
--cycle;

\path [fill=darkorange25512714, fill opacity=0.2]
(axis cs:2.6,0.667393229579635)
--(axis cs:2.6,0.664738370420365)
--(axis cs:3.1,0.664058455703161)
--(axis cs:3.6,0.636420238144391)
--(axis cs:4.3,0.663769914881702)
--(axis cs:5.1,0.663558864496318)
--(axis cs:7.5,0.681934111353935)
--(axis cs:17.8,0.699619486206494)
--(axis cs:23.7,0.709971234954948)
--(axis cs:28.1,0.709333005203958)
--(axis cs:36.9,0.736835748200972)
--(axis cs:48.8,0.748568418993425)
--(axis cs:58.7,0.758066468591131)
--(axis cs:66.8,0.761805300827823)
--(axis cs:79.4,0.775395296704201)
--(axis cs:97.5,0.787015071805367)
--(axis cs:113.8,0.779151048176671)
--(axis cs:139.6,0.795509012373085)
--(axis cs:172.4,0.814886614574186)
--(axis cs:210.6,0.827356405842786)
--(axis cs:262.2,0.82547376713972)
--(axis cs:312.8,0.833852043243155)
--(axis cs:370.4,0.845548148648617)
--(axis cs:433.5,0.839290833058382)
--(axis cs:544.3,0.857126851605461)
--(axis cs:700.6,0.844147436090329)
--(axis cs:907.2,0.876540621526905)
--(axis cs:1202.5,0.880194973900716)
--(axis cs:1596.8,0.895509807227289)
--(axis cs:2047.9,0.898926835351809)
--(axis cs:2568.4,0.904347506936906)
--(axis cs:3118.9,0.908959440827872)
--(axis cs:3441.2,0.909625914273092)
--(axis cs:3905.5,0.915519848742537)
--(axis cs:4626,0.914454511575382)
--(axis cs:5415.9,0.914566378540255)
--(axis cs:6247,0.917700756265332)
--(axis cs:7084.9,0.917720948742731)
--(axis cs:7986.1,0.921592976251956)
--(axis cs:8870.5,0.920252999914763)
--(axis cs:9829,0.922934917679001)
--(axis cs:10719.6,0.924077724687239)
--(axis cs:12003.4,0.92134133066968)
--(axis cs:13726.8,0.924333050351039)
--(axis cs:15532.3,0.923084000532938)
--(axis cs:17246.6,0.928353030789355)
--(axis cs:18993.4,0.924537301457057)
--(axis cs:20802.3,0.924623847858332)
--(axis cs:22797.5,0.925593234707414)
--(axis cs:24941.3,0.928436480464283)
--(axis cs:27217,0.925275749650366)
--(axis cs:29266.9,0.926321348061081)
--(axis cs:31296.7,0.926440514167977)
--(axis cs:33935.8,0.922248026987961)
--(axis cs:36856.7,0.924632385807316)
--(axis cs:39926.9,0.922458286078849)
--(axis cs:43094.4,0.924642029530055)
--(axis cs:46383.7,0.922888079096121)
--(axis cs:49598.7,0.929267696323772)
--(axis cs:52890,0.925054002276341)
--(axis cs:53166.1,0.926410884213882)
--(axis cs:54630.2,0.926181482342982)
--(axis cs:54726.7,0.92740512067715)
--(axis cs:55496,0.923284428832588)
--(axis cs:56491.1,0.929095217141267)
--(axis cs:57884.7,0.925373759426729)
--(axis cs:61427,0.926251497297622)
--(axis cs:65286.2,0.926786391809014)
--(axis cs:69319.4,0.928999815645071)
--(axis cs:73597.2,0.928443539842579)
--(axis cs:77945.7,0.92594072451695)
--(axis cs:82333.8,0.926691227867928)
--(axis cs:86683.6,0.928451383161616)
--(axis cs:90999.8,0.926828306941715)
--(axis cs:94866.9,0.926755945152722)
--(axis cs:97781.8,0.928802484551548)
--(axis cs:99386.1,0.929434686252015)
--(axis cs:99913.6,0.927252663643488)
--(axis cs:99996.2,0.928885238938284)
--(axis cs:100000,0.928807448890116)
--(axis cs:100000,0.927769309937575)
--(axis cs:100000,0.933050490062425)
--(axis cs:100000,0.933050490062425)
--(axis cs:100000,0.933650551109883)
--(axis cs:99996.2,0.933414761061716)
--(axis cs:99913.6,0.931904136356512)
--(axis cs:99386.1,0.933485313747985)
--(axis cs:97781.8,0.933317515448452)
--(axis cs:94866.9,0.932094054847278)
--(axis cs:90999.8,0.931798093058285)
--(axis cs:86683.6,0.933936816838384)
--(axis cs:82333.8,0.932295372132072)
--(axis cs:77945.7,0.92993947548305)
--(axis cs:73597.2,0.933536460157421)
--(axis cs:69319.4,0.934426784354929)
--(axis cs:65286.2,0.932098608190986)
--(axis cs:61427,0.930525102702378)
--(axis cs:57884.7,0.932299240573271)
--(axis cs:56491.1,0.932724582858732)
--(axis cs:55496,0.933180571167412)
--(axis cs:54726.7,0.93340667932285)
--(axis cs:54630.2,0.932876917657018)
--(axis cs:53166.1,0.933212515786118)
--(axis cs:52890,0.932449397723659)
--(axis cs:49598.7,0.931815903676228)
--(axis cs:46383.7,0.931190120903879)
--(axis cs:43094.4,0.931994170469945)
--(axis cs:39926.9,0.930115313921151)
--(axis cs:36856.7,0.932824414192684)
--(axis cs:33935.8,0.928400373012039)
--(axis cs:31296.7,0.932616085832023)
--(axis cs:29266.9,0.931867251938919)
--(axis cs:27217,0.932171050349634)
--(axis cs:24941.3,0.932723519535717)
--(axis cs:22797.5,0.931908565292586)
--(axis cs:20802.3,0.932034352141668)
--(axis cs:18993.4,0.931106098542943)
--(axis cs:17246.6,0.931538769210645)
--(axis cs:15532.3,0.932349199467062)
--(axis cs:13726.8,0.931620149648961)
--(axis cs:12003.4,0.930472069330319)
--(axis cs:10719.6,0.93109387531276)
--(axis cs:9829,0.928049882320999)
--(axis cs:8870.5,0.928022200085237)
--(axis cs:7986.1,0.928810623748044)
--(axis cs:7084.9,0.926252451257269)
--(axis cs:6247,0.925679443734668)
--(axis cs:5415.9,0.924480621459745)
--(axis cs:4626,0.920787288424618)
--(axis cs:3905.5,0.920538551257463)
--(axis cs:3441.2,0.918500485726908)
--(axis cs:3118.9,0.917900759172128)
--(axis cs:2568.4,0.915542693063094)
--(axis cs:2047.9,0.911132964648191)
--(axis cs:1596.8,0.907230192772711)
--(axis cs:1202.5,0.904480426099284)
--(axis cs:907.2,0.897332978473095)
--(axis cs:700.6,0.880760963909671)
--(axis cs:544.3,0.875454748394539)
--(axis cs:433.5,0.863480966941618)
--(axis cs:370.4,0.860905051351383)
--(axis cs:312.8,0.855206156756845)
--(axis cs:262.2,0.84569463286028)
--(axis cs:210.6,0.839843594157214)
--(axis cs:172.4,0.827449985425814)
--(axis cs:139.6,0.811601187626915)
--(axis cs:113.8,0.807578751823329)
--(axis cs:97.5,0.800756328194633)
--(axis cs:79.4,0.787461503295799)
--(axis cs:66.8,0.774811299172177)
--(axis cs:58.7,0.768791531408869)
--(axis cs:48.8,0.762359781006575)
--(axis cs:36.9,0.768437851799028)
--(axis cs:28.1,0.749865394796042)
--(axis cs:23.7,0.731235365045052)
--(axis cs:17.8,0.705807113793506)
--(axis cs:7.5,0.695419488646065)
--(axis cs:5.1,0.666894735503683)
--(axis cs:4.3,0.666287085118298)
--(axis cs:3.6,0.643169961855609)
--(axis cs:3.1,0.667158144296839)
--(axis cs:2.6,0.667393229579635)
--cycle;

\path [fill=forestgreen4416044, fill opacity=0.2]
(axis cs:3.5,0.666732416426936)
--(axis cs:3.5,0.664219183573064)
--(axis cs:4.2,0.663958391945568)
--(axis cs:5,0.647294041378968)
--(axis cs:5.9,0.663790135996221)
--(axis cs:9.2,0.689319595162708)
--(axis cs:14.1,0.695520226919595)
--(axis cs:24.6,0.70914500575455)
--(axis cs:34.1,0.730118736818841)
--(axis cs:44.4,0.748531860749978)
--(axis cs:55.9,0.753290603066884)
--(axis cs:68.3,0.757782444181591)
--(axis cs:75.7,0.762009789017808)
--(axis cs:89.7,0.771489935175143)
--(axis cs:109.6,0.780625011134006)
--(axis cs:127.4,0.78908340054843)
--(axis cs:159.9,0.795949221115734)
--(axis cs:201,0.804031014732468)
--(axis cs:254.5,0.816567052518537)
--(axis cs:310.5,0.820853633256898)
--(axis cs:363.2,0.833700217181062)
--(axis cs:436.3,0.843376533126446)
--(axis cs:508.5,0.849976472093279)
--(axis cs:580.3,0.840656114247513)
--(axis cs:725.2,0.869405676083328)
--(axis cs:912.5,0.869589959127246)
--(axis cs:1179.5,0.888171531835179)
--(axis cs:1509.5,0.886790507390368)
--(axis cs:1889.7,0.900980537048332)
--(axis cs:2321.6,0.908331650564453)
--(axis cs:2790.3,0.909948279234005)
--(axis cs:3260.7,0.912097451393729)
--(axis cs:3449.4,0.910533025934762)
--(axis cs:3799.5,0.913100770732989)
--(axis cs:4402,0.914228378326212)
--(axis cs:5092.6,0.92026598211752)
--(axis cs:5813.6,0.919244239666654)
--(axis cs:6596,0.918005606382272)
--(axis cs:7448.1,0.919234377820428)
--(axis cs:8351.6,0.920926413173564)
--(axis cs:9075.5,0.920741553142355)
--(axis cs:9449.3,0.924764633703574)
--(axis cs:10099.2,0.923781870632262)
--(axis cs:11264.4,0.926008087879932)
--(axis cs:13294.5,0.92279824677004)
--(axis cs:15820.2,0.925809106519873)
--(axis cs:18140.7,0.923071946432357)
--(axis cs:20227.5,0.927295314593837)
--(axis cs:22235.1,0.92566153091234)
--(axis cs:24512.3,0.925876939440653)
--(axis cs:26777.2,0.926384905876232)
--(axis cs:28484.7,0.927001039438879)
--(axis cs:30027.7,0.926170235174794)
--(axis cs:32398.4,0.921463703406566)
--(axis cs:35192.2,0.920879667556302)
--(axis cs:37955,0.920896292401323)
--(axis cs:40416.2,0.923081330835511)
--(axis cs:42353.3,0.925282500515826)
--(axis cs:43681.1,0.928605592906994)
--(axis cs:43905.2,0.925072391243093)
--(axis cs:44074.2,0.926561655788945)
--(axis cs:45208.4,0.927441215347642)
--(axis cs:47960.5,0.924724651329838)
--(axis cs:50928.1,0.923577449068692)
--(axis cs:54023.4,0.924667847916183)
--(axis cs:57345.4,0.925487094422959)
--(axis cs:60736.4,0.92410832674464)
--(axis cs:64416.3,0.923483647938859)
--(axis cs:68398.8,0.927737330179864)
--(axis cs:72464.7,0.92628225231867)
--(axis cs:76719.1,0.925656382855829)
--(axis cs:81029.8,0.926818783992062)
--(axis cs:85463.5,0.926506905961943)
--(axis cs:89804.6,0.926270148073427)
--(axis cs:93957.1,0.9271357905102)
--(axis cs:97191.2,0.927454297769832)
--(axis cs:99169.7,0.928806759322411)
--(axis cs:99867.6,0.92711734499112)
--(axis cs:99991,0.92781658537803)
--(axis cs:99999.7,0.928469400034068)
--(axis cs:100000,0.927198630612725)
--(axis cs:100000,0.932219569387275)
--(axis cs:100000,0.932219569387275)
--(axis cs:99999.7,0.932872399965932)
--(axis cs:99991,0.932196614621971)
--(axis cs:99867.6,0.93201925500888)
--(axis cs:99169.7,0.933196440677589)
--(axis cs:97191.2,0.932515702230168)
--(axis cs:93957.1,0.9326458094898)
--(axis cs:89804.6,0.931304851926573)
--(axis cs:85463.5,0.932494694038057)
--(axis cs:81029.8,0.932878016007938)
--(axis cs:76719.1,0.932003617144171)
--(axis cs:72464.7,0.93347614768133)
--(axis cs:68398.8,0.934164469820136)
--(axis cs:64416.3,0.930064552061141)
--(axis cs:60736.4,0.93161007325536)
--(axis cs:57345.4,0.931174305577041)
--(axis cs:54023.4,0.931976952083817)
--(axis cs:50928.1,0.932350950931308)
--(axis cs:47960.5,0.931853548670162)
--(axis cs:45208.4,0.932501984652358)
--(axis cs:44074.2,0.933168544211055)
--(axis cs:43905.2,0.934089408756907)
--(axis cs:43681.1,0.935656207093006)
--(axis cs:42353.3,0.934262699484174)
--(axis cs:40416.2,0.932778669164489)
--(axis cs:37955,0.931850107598677)
--(axis cs:35192.2,0.932225132443697)
--(axis cs:32398.4,0.931981296593434)
--(axis cs:30027.7,0.932823164825206)
--(axis cs:28484.7,0.933508960561121)
--(axis cs:26777.2,0.932753494123769)
--(axis cs:24512.3,0.932851260559347)
--(axis cs:22235.1,0.93273346908766)
--(axis cs:20227.5,0.932419685406163)
--(axis cs:18140.7,0.931041653567643)
--(axis cs:15820.2,0.932259293480127)
--(axis cs:13294.5,0.93126335322996)
--(axis cs:11264.4,0.931343512120067)
--(axis cs:10099.2,0.930622929367738)
--(axis cs:9449.3,0.930868766296426)
--(axis cs:9075.5,0.928046846857645)
--(axis cs:8351.6,0.928826986826436)
--(axis cs:7448.1,0.927582422179571)
--(axis cs:6596,0.926244393617728)
--(axis cs:5813.6,0.926589160333346)
--(axis cs:5092.6,0.92582581788248)
--(axis cs:4402,0.922728221673788)
--(axis cs:3799.5,0.922545829267011)
--(axis cs:3449.4,0.919620374065237)
--(axis cs:3260.7,0.919529348606271)
--(axis cs:2790.3,0.917466520765995)
--(axis cs:2321.6,0.915450149435547)
--(axis cs:1889.7,0.909725862951668)
--(axis cs:1509.5,0.906827692609632)
--(axis cs:1179.5,0.899581868164821)
--(axis cs:912.5,0.895009840872753)
--(axis cs:725.2,0.884426323916672)
--(axis cs:580.3,0.873338685752487)
--(axis cs:508.5,0.867773327906721)
--(axis cs:436.3,0.857388466873554)
--(axis cs:363.2,0.850108382818938)
--(axis cs:310.5,0.848749566743101)
--(axis cs:254.5,0.834806147481464)
--(axis cs:201,0.824587185267532)
--(axis cs:159.9,0.819502778884266)
--(axis cs:127.4,0.802728399451569)
--(axis cs:109.6,0.797373588865994)
--(axis cs:89.7,0.783980064824857)
--(axis cs:75.7,0.773775210982192)
--(axis cs:68.3,0.768972555818409)
--(axis cs:55.9,0.763805796933116)
--(axis cs:44.4,0.757696339250022)
--(axis cs:34.1,0.74725126318116)
--(axis cs:24.6,0.72494679424545)
--(axis cs:14.1,0.700369373080404)
--(axis cs:9.2,0.691763804837292)
--(axis cs:5.9,0.666293264003779)
--(axis cs:5,0.671827558621032)
--(axis cs:4.2,0.666488408054432)
--(axis cs:3.5,0.666732416426936)
--cycle;

\path [fill=crimson2143940, fill opacity=0.2]
(axis cs:6.2,0.66920136870823)
--(axis cs:6.2,0.63856363129177)
--(axis cs:6.2,0.653913735507975)
--(axis cs:7.5,0.66334355147683)
--(axis cs:9.1,0.663242613791943)
--(axis cs:13,0.685792180248631)
--(axis cs:23.3,0.701852321743864)
--(axis cs:35.3,0.71970456264914)
--(axis cs:45.8,0.746633295629962)
--(axis cs:67.3,0.755998675657985)
--(axis cs:75.6,0.756477347180978)
--(axis cs:90.1,0.762916740423529)
--(axis cs:106,0.772550871510496)
--(axis cs:130.7,0.781189759120064)
--(axis cs:166.7,0.782816210000359)
--(axis cs:208.5,0.790914777867528)
--(axis cs:261.8,0.803257925824693)
--(axis cs:309.3,0.809345415283653)
--(axis cs:368.3,0.825051066598139)
--(axis cs:436.9,0.82186179551002)
--(axis cs:526.5,0.843498857708783)
--(axis cs:624,0.854113845350683)
--(axis cs:711.7,0.855899162064415)
--(axis cs:798.1,0.850992355475303)
--(axis cs:972.6,0.87674858188758)
--(axis cs:1208.2,0.875287421858181)
--(axis cs:1513,0.888602800201259)
--(axis cs:1857.5,0.894217742712047)
--(axis cs:2231.6,0.898308940439457)
--(axis cs:2628,0.906701039740139)
--(axis cs:3054.8,0.90686314835956)
--(axis cs:3475.4,0.91140758222638)
--(axis cs:3616.5,0.909989683094227)
--(axis cs:3919.5,0.910919091146516)
--(axis cs:4502.9,0.915301273058275)
--(axis cs:5206.3,0.915085025013486)
--(axis cs:5955.6,0.917750474438937)
--(axis cs:6800.9,0.917458293123979)
--(axis cs:7667.4,0.919169746824657)
--(axis cs:8545.4,0.920537164535006)
--(axis cs:9324.8,0.921776119816304)
--(axis cs:9717.5,0.923975430355775)
--(axis cs:10359.4,0.924261292003221)
--(axis cs:11384.5,0.922453622209352)
--(axis cs:12811.4,0.924308147151437)
--(axis cs:15251.9,0.922913634921444)
--(axis cs:17741.3,0.923197512094722)
--(axis cs:19904.9,0.925085785551238)
--(axis cs:21948.1,0.926933025444723)
--(axis cs:24236.7,0.926946461431716)
--(axis cs:26274.9,0.926963102420694)
--(axis cs:26468.3,0.926524424101462)
--(axis cs:27069,0.927262358969616)
--(axis cs:27278.5,0.926392857145985)
--(axis cs:28233.1,0.927548103040116)
--(axis cs:29228.5,0.926366722646366)
--(axis cs:29409.9,0.928005402543377)
--(axis cs:32893.4,0.928217718914936)
--(axis cs:36948.3,0.931647397520951)
--(axis cs:40163.3,0.925347219567878)
--(axis cs:43039.6,0.925950270020998)
--(axis cs:45828,0.927382549801321)
--(axis cs:48607.1,0.925408209078013)
--(axis cs:51487.4,0.925452542583318)
--(axis cs:54450.8,0.924350576644557)
--(axis cs:57551.6,0.923038718047842)
--(axis cs:60782.9,0.923909185450083)
--(axis cs:64276.9,0.923119981855284)
--(axis cs:68066.8,0.926019624728794)
--(axis cs:72024.7,0.926096481517583)
--(axis cs:76132.7,0.925567750476665)
--(axis cs:80386.1,0.925798173010778)
--(axis cs:84793.9,0.927162773595099)
--(axis cs:89203.8,0.925700413119981)
--(axis cs:93410,0.926355258592026)
--(axis cs:96890.4,0.928429090790082)
--(axis cs:99007.3,0.928455233631469)
--(axis cs:99828.1,0.926911030439629)
--(axis cs:99988.3,0.927167817507171)
--(axis cs:99999.7,0.928761234470126)
--(axis cs:100000,0.927232732388646)
--(axis cs:100000,0.933108867611354)
--(axis cs:100000,0.933108867611354)
--(axis cs:99999.7,0.933531965529874)
--(axis cs:99988.3,0.932323782492829)
--(axis cs:99828.1,0.931647369560371)
--(axis cs:99007.3,0.932871566368531)
--(axis cs:96890.4,0.932844509209918)
--(axis cs:93410,0.931956341407974)
--(axis cs:89203.8,0.932103186880019)
--(axis cs:84793.9,0.932809226404902)
--(axis cs:80386.1,0.932540026989222)
--(axis cs:76132.7,0.932342249523335)
--(axis cs:72024.7,0.933266918482417)
--(axis cs:68066.8,0.933340575271206)
--(axis cs:64276.9,0.930381618144716)
--(axis cs:60782.9,0.931214014549917)
--(axis cs:57551.6,0.930594481952159)
--(axis cs:54450.8,0.931924423355443)
--(axis cs:51487.4,0.932844257416682)
--(axis cs:48607.1,0.932678390921987)
--(axis cs:45828,0.932067650198679)
--(axis cs:43039.6,0.931559929979003)
--(axis cs:40163.3,0.932934180432121)
--(axis cs:36948.3,0.933191002479049)
--(axis cs:32893.4,0.934800681085064)
--(axis cs:29409.9,0.933712597456623)
--(axis cs:29228.5,0.933296877353634)
--(axis cs:28233.1,0.934240096959884)
--(axis cs:27278.5,0.932917142854015)
--(axis cs:27069,0.933394441030384)
--(axis cs:26468.3,0.933442375898538)
--(axis cs:26274.9,0.934392097579305)
--(axis cs:24236.7,0.933820138568284)
--(axis cs:21948.1,0.932517174555276)
--(axis cs:19904.9,0.933582414448761)
--(axis cs:17741.3,0.930959287905278)
--(axis cs:15251.9,0.930112965078556)
--(axis cs:12811.4,0.931960252848563)
--(axis cs:11384.5,0.930357977790648)
--(axis cs:10359.4,0.930765307996778)
--(axis cs:9717.5,0.930352769644225)
--(axis cs:9324.8,0.927660480183696)
--(axis cs:8545.4,0.929748035464994)
--(axis cs:7667.4,0.926933653175343)
--(axis cs:6800.9,0.925765106876021)
--(axis cs:5955.6,0.926589725561063)
--(axis cs:5206.3,0.924811974986514)
--(axis cs:4502.9,0.922962526941725)
--(axis cs:3919.5,0.918867508853484)
--(axis cs:3616.5,0.919970116905773)
--(axis cs:3475.4,0.91918041777362)
--(axis cs:3054.8,0.91673025164044)
--(axis cs:2628,0.916547160259861)
--(axis cs:2231.6,0.912256059560543)
--(axis cs:1857.5,0.910603657287953)
--(axis cs:1513,0.904232199798741)
--(axis cs:1208.2,0.899495778141819)
--(axis cs:972.6,0.89752801811242)
--(axis cs:798.1,0.885359244524697)
--(axis cs:711.7,0.880167237935585)
--(axis cs:624,0.872727754649317)
--(axis cs:526.5,0.859274542291217)
--(axis cs:436.9,0.85430960448998)
--(axis cs:368.3,0.845302333401861)
--(axis cs:309.3,0.831462784716346)
--(axis cs:261.8,0.831696874175307)
--(axis cs:208.5,0.817413622132472)
--(axis cs:166.7,0.802925589999641)
--(axis cs:130.7,0.792015240879936)
--(axis cs:106,0.782469328489504)
--(axis cs:90.1,0.774460059576471)
--(axis cs:75.6,0.766054252819022)
--(axis cs:67.3,0.762907924342015)
--(axis cs:45.8,0.754616704370038)
--(axis cs:35.3,0.74058363735086)
--(axis cs:23.3,0.716779678256136)
--(axis cs:13,0.69054961975137)
--(axis cs:9.1,0.666117386208056)
--(axis cs:7.5,0.66618164852317)
--(axis cs:6.2,0.671163064492025)
--(axis cs:6.2,0.66920136870823)
--cycle;

\path [fill=mediumpurple148103189, fill opacity=0.2]
(axis cs:1,0.502302935100388)
--(axis cs:1,0.499510464899612)
--(axis cs:1,0.499510464899612)
--(axis cs:1,0.499510464899612)
--(axis cs:1,0.49119370408688)
--(axis cs:1,0.499510464899612)
--(axis cs:1,0.499510464899612)
--(axis cs:1,0.499510464899612)
--(axis cs:1,0.499510464899612)
--(axis cs:1,0.499510464899612)
--(axis cs:1,0.492038012379104)
--(axis cs:1,0.491466917495949)
--(axis cs:1,0.499510464899612)
--(axis cs:1,0.492038012379104)
--(axis cs:7.7,0.662053561702026)
--(axis cs:27.9,0.667907879587263)
--(axis cs:45.3,0.699204345587491)
--(axis cs:72,0.701081418955679)
--(axis cs:117,0.708544524811265)
--(axis cs:187.5,0.74145507294102)
--(axis cs:264.1,0.743254120331968)
--(axis cs:360.8,0.776289055379517)
--(axis cs:526.1,0.788009928248305)
--(axis cs:823.3,0.776578748252752)
--(axis cs:1334.9,0.824975859201715)
--(axis cs:2053.2,0.817913201236294)
--(axis cs:2993.2,0.823769732079965)
--(axis cs:4168.3,0.847377750995994)
--(axis cs:5665.8,0.850291602871076)
--(axis cs:7372.3,0.84921243379735)
--(axis cs:9439,0.867143699063827)
--(axis cs:11923.9,0.875969896647082)
--(axis cs:14917.8,0.886258641340459)
--(axis cs:18302.6,0.892423956516624)
--(axis cs:21750.3,0.891211219245131)
--(axis cs:24976.3,0.901507027030332)
--(axis cs:26979.4,0.906496140137856)
--(axis cs:28801.6,0.907340294923935)
--(axis cs:31742.1,0.910823626396642)
--(axis cs:35352.3,0.817619085336718)
--(axis cs:38972.2,0.914189758437247)
--(axis cs:42735.6,0.916447419566676)
--(axis cs:46285.3,0.91905984831962)
--(axis cs:49578.1,0.919116875073847)
--(axis cs:50161.6,0.920327563012847)
--(axis cs:50668.8,0.921491334838717)
--(axis cs:51619.1,0.917812992982618)
--(axis cs:52836.6,0.923198408396222)
--(axis cs:56301.4,0.922883355807188)
--(axis cs:59957.7,0.924775386447967)
--(axis cs:63842.4,0.907690734107484)
--(axis cs:68150.9,0.924366022037077)
--(axis cs:71898,0.922148073412558)
--(axis cs:72696.8,0.923349254079944)
--(axis cs:74261.8,0.923410155384667)
--(axis cs:75746,0.906049191907274)
--(axis cs:76646.7,0.924889470868168)
--(axis cs:78564.7,0.926162620388814)
--(axis cs:79099.9,0.922305602261832)
--(axis cs:83017,0.926442943604821)
--(axis cs:87165.8,0.925793682731951)
--(axis cs:90929.4,0.925178279534195)
--(axis cs:94103.3,0.921014268308256)
--(axis cs:96605.6,0.92534735581247)
--(axis cs:97058.3,0.923787828433631)
--(axis cs:98258.1,0.925343185427737)
--(axis cs:98302.5,0.868299941210263)
--(axis cs:98345.9,0.922979132304504)
--(axis cs:99192.2,0.926054498249501)
--(axis cs:99332.3,0.923970760344029)
--(axis cs:99334.4,0.924821290501807)
--(axis cs:99815.8,0.924746200620332)
--(axis cs:99968.8,0.924451756654969)
--(axis cs:99996.3,0.92306880034195)
--(axis cs:100000,0.924072353095818)
--(axis cs:100000,0.923651568208099)
--(axis cs:100000,0.921229846778925)
--(axis cs:100000,0.923749734880493)
--(axis cs:100000,0.923605806544309)
--(axis cs:100000,0.921624649669841)
--(axis cs:100000,0.923980655934793)
--(axis cs:100000,0.930097544065207)
--(axis cs:100000,0.930097544065207)
--(axis cs:100000,0.929835350330159)
--(axis cs:100000,0.929707793455691)
--(axis cs:100000,0.930761865119507)
--(axis cs:100000,0.932146753221075)
--(axis cs:100000,0.929263231791901)
--(axis cs:100000,0.930874446904182)
--(axis cs:99996.3,0.92994439965805)
--(axis cs:99968.8,0.932308243345031)
--(axis cs:99815.8,0.930012199379668)
--(axis cs:99334.4,0.930990309498193)
--(axis cs:99332.3,0.93073903965597)
--(axis cs:99192.2,0.930658901750499)
--(axis cs:98345.9,0.931455867695496)
--(axis cs:98302.5,0.957921858789738)
--(axis cs:98258.1,0.930230414572263)
--(axis cs:97058.3,0.932825571566368)
--(axis cs:96605.6,0.93267264418753)
--(axis cs:94103.3,0.931377531691744)
--(axis cs:90929.4,0.927285120465805)
--(axis cs:87165.8,0.932784517268049)
--(axis cs:83017,0.931863656395179)
--(axis cs:79099.9,0.928864197738168)
--(axis cs:78564.7,0.932972379611185)
--(axis cs:76646.7,0.930463929131832)
--(axis cs:75746,0.940852608092726)
--(axis cs:74261.8,0.933080044615333)
--(axis cs:72696.8,0.929942545920056)
--(axis cs:71898,0.932403526587442)
--(axis cs:68150.9,0.930010577962923)
--(axis cs:63842.4,0.940190865892516)
--(axis cs:59957.7,0.930733013552033)
--(axis cs:56301.4,0.931046844192812)
--(axis cs:52836.6,0.929471591603778)
--(axis cs:51619.1,0.925330207017382)
--(axis cs:50668.8,0.929701865161283)
--(axis cs:50161.6,0.927868836987153)
--(axis cs:49578.1,0.928634524926153)
--(axis cs:46285.3,0.92735675168038)
--(axis cs:42735.6,0.927635980433324)
--(axis cs:38972.2,0.923632041562753)
--(axis cs:35352.3,0.964099114663282)
--(axis cs:31742.1,0.921374773603358)
--(axis cs:28801.6,0.920641505076065)
--(axis cs:26979.4,0.920669059862144)
--(axis cs:24976.3,0.917857972969668)
--(axis cs:21750.3,0.904460380754869)
--(axis cs:18302.6,0.913614243483376)
--(axis cs:14917.8,0.912446558659541)
--(axis cs:11923.9,0.900364903352918)
--(axis cs:9439,0.898626500936172)
--(axis cs:7372.3,0.89292256620265)
--(axis cs:5665.8,0.892551797128924)
--(axis cs:4168.3,0.891958849004006)
--(axis cs:2993.2,0.873800267920035)
--(axis cs:2053.2,0.872283598763706)
--(axis cs:1334.9,0.858865940798285)
--(axis cs:823.3,0.842975051747248)
--(axis cs:526.1,0.842026671751695)
--(axis cs:360.8,0.824240944620482)
--(axis cs:264.1,0.796962679668032)
--(axis cs:187.5,0.80193672705898)
--(axis cs:117,0.759791875188735)
--(axis cs:72,0.747028581044321)
--(axis cs:45.3,0.720562254412509)
--(axis cs:27.9,0.699718920412737)
--(axis cs:7.7,0.684808038297974)
--(axis cs:1,0.505280387620896)
--(axis cs:1,0.502302935100388)
--(axis cs:1,0.505846482504051)
--(axis cs:1,0.505280387620896)
--(axis cs:1,0.502302935100388)
--(axis cs:1,0.502302935100388)
--(axis cs:1,0.502302935100388)
--(axis cs:1,0.502302935100388)
--(axis cs:1,0.502302935100388)
--(axis cs:1,0.50625789591312)
--(axis cs:1,0.502302935100388)
--(axis cs:1,0.502302935100388)
--(axis cs:1,0.502302935100388)
--cycle;

\addplot [semithick, black, dashed]
table {%
1 0.4999234
1 0.4999234
4.6 0.6906942
5.8 0.6907492
7.5 0.6947783
12.3 0.7063117
15.2 0.7030075
19.7 0.7385275
26.2 0.7464383
34.1 0.7650301
42.5 0.7721259
55.7 0.7779309
72.7 0.7894483
95.2 0.7993101
116.8 0.8130508
147.6 0.8225517
189.9 0.8315941
225.7 0.83912
257.1 0.8480567
292.2 0.8570709
311.7 0.8671965
360.7 0.874606
441.5 0.8827367
480.8 0.889841
548.7 0.8965501
610.5 0.902675
679.9 0.9078149
770.7 0.9129801
857.8 0.9173633
962.1 0.9208784
1067.1 0.9243883
1197.7 0.9279317
1356.9 0.9310391
1509.8 0.9339358
1700.7 0.9364659
2013 0.9389159
2313.8 0.9413751
2423 0.9441774
2656.3 0.9462418
2979.9 0.9482233
3338.7 0.9501642
3709.4 0.9519382
4101 0.953369
4529.6 0.9548307
4994 0.9560959
5533.2 0.9572467
6100.8 0.958345
6738.1 0.9596558
7440.2 0.9607476
8217 0.961705
9043.1 0.9626059
10008.2 0.9634574
10967.9 0.9641775
11806.5 0.9649659
13154 0.9657015
14409 0.9663574
15679.8 0.9668275
17075.7 0.9673815
18546.3 0.9678842
20025.7 0.9683008
21636.8 0.9687065
23496.5 0.9689742
25376.7 0.9692266
27315.6 0.9694092
29182.3 0.9696084
31006.2 0.9698684
33135.9 0.9699484
35352.8 0.9700375
37493.6 0.9701875
39483.8 0.9702182
40868.4 0.9702007
43308.7 0.9702208
45580.9 0.9701951
48173.1 0.970105
51057.1 0.9699958
54538.4 0.9699492
57527.6 0.969855
60246.6 0.96978
63510.6 0.96976
66424.1 0.9696409
};
\addlegendentry{\scriptsize LIBLINEAR}
\addplot [steelblue31119180, opacity=1.0]
table {%
1 0.5009067
1 0.5009067
1 0.5009067
1.2 0.5340317
3.2 0.6657442
5.5 0.6795017
9.7 0.6906892
14.8 0.6998642
17.8 0.7183401
23.7 0.7322501
29.6 0.7368083
42.1 0.7483059
51.7 0.7581334
60.3 0.7684625
73.3 0.7750709
83.6 0.785445
101.4 0.7967541
125.6 0.8041067
153.8 0.8119674
192.3 0.8179985
244.7 0.840101
289.8 0.8367108
340.5 0.8482467
435.6 0.857
583.9 0.8570708
812.6 0.8767491
1146.7 0.8911226
1603.2 0.8893491
2161.8 0.8966275
2836 0.9082625
3594.6 0.9042141
4175.9 0.9061915
5009.1 0.913565
6111.6 0.9152758
7423.9 0.919405
8778.5 0.9226093
10221.5 0.9232884
11716.4 0.9243091
13296.1 0.92401
14907.2 0.9253376
16476.1 0.92752
18359.2 0.9274018
20244.5 0.9264049
22002 0.9276382
23720.5 0.929215
24795.8 0.92775
25398 0.9296192
26247.1 0.9297718
27722.1 0.9307209
29889.9 0.9305057
32306.4 0.9277258
35047.9 0.9301901
37921 0.929001
40845.5 0.9296326
43887.3 0.9275617
47014.8 0.9285601
50282.7 0.9276559
53622.6 0.9306908
56948.2 0.9287025
60475.1 0.9302858
60972.4 0.9293068
63730 0.9288774
64100.1 0.9296967
64477.7 0.9290309
67337.1 0.9303984
67576.8 0.9308557
68316.7 0.9302951
71701.6 0.9305126
75977.4 0.931304
80352.6 0.9294567
84686.7 0.9311475
88912.9 0.9307901
92918.8 0.93054
96322.9 0.9302291
98584 0.9314283
99673.5 0.9304559
99963.6 0.9303141
99998.7 0.93057
100000 0.9309982
100000 0.930649
};
\addlegendentry{\scriptsize ACOWA, $\beta=5$}
\addplot [darkorange25512714, opacity=1.0]
table {%
2.6 0.6660658
3.1 0.6656083
3.6 0.6397951
4.3 0.6650285
5.1 0.6652268
7.5 0.6886768
17.8 0.7027133
23.7 0.7206033
28.1 0.7295992
36.9 0.7526368
48.8 0.7554641
58.7 0.763429
66.8 0.7683083
79.4 0.7814284
97.5 0.7938857
113.8 0.7933649
139.6 0.8035551
172.4 0.8211683
210.6 0.8336
262.2 0.8355842
312.8 0.8445291
370.4 0.8532266
433.5 0.8513859
544.3 0.8662908
700.6 0.8624542
907.2 0.8869368
1202.5 0.8923377
1596.8 0.90137
2047.9 0.9050299
2568.4 0.9099451
3118.9 0.9134301
3441.2 0.9140632
3905.5 0.9180292
4626 0.9176209
5415.9 0.9195235
6247 0.9216901
7084.9 0.9219867
7986.1 0.9252018
8870.5 0.9241376
9829 0.9254924
10719.6 0.9275858
12003.4 0.9259067
13726.8 0.9279766
15532.3 0.9277166
17246.6 0.9299459
18993.4 0.9278217
20802.3 0.9283291
22797.5 0.9287509
24941.3 0.93058
27217 0.9287234
29266.9 0.9290943
31296.7 0.9295283
33935.8 0.9253242
36856.7 0.9287284
39926.9 0.9262868
43094.4 0.9283181
46383.7 0.9270391
49598.7 0.9305418
52890 0.9287517
53166.1 0.9298117
54630.2 0.9295292
54726.7 0.9304059
55496 0.9282325
56491.1 0.9309099
57884.7 0.9288365
61427 0.9283883
65286.2 0.9294425
69319.4 0.9317133
73597.2 0.93099
77945.7 0.9279401
82333.8 0.9294933
86683.6 0.9311941
90999.8 0.9293132
94866.9 0.929425
97781.8 0.93106
99386.1 0.93146
99913.6 0.9295784
99996.2 0.93115
100000 0.931229
100000 0.9304099
};
\addlegendentry{\scriptsize ACOWA, $\beta=10$}
\addplot [forestgreen4416044, opacity=1.0]
table {%
3.5 0.6654758
4.2 0.6652234
5 0.6595608
5.9 0.6650417
9.2 0.6905417
14.1 0.6979448
24.6 0.7170459
34.1 0.738685
44.4 0.7531141
55.9 0.7585482
68.3 0.7633775
75.7 0.7678925
89.7 0.777735
109.6 0.7889993
127.4 0.7959059
159.9 0.807726
201 0.8143091
254.5 0.8256866
310.5 0.8348016
363.2 0.8419043
436.3 0.8503825
508.5 0.8588749
580.3 0.8569974
725.2 0.876916
912.5 0.8822999
1179.5 0.8938767
1509.5 0.8968091
1889.7 0.9053532
2321.6 0.9118909
2790.3 0.9137074
3260.7 0.9158134
3449.4 0.9150767
3799.5 0.9178233
4402 0.9184783
5092.6 0.9230459
5813.6 0.9229167
6596 0.922125
7448.1 0.9234084
8351.6 0.9248767
9075.5 0.9243942
9449.3 0.9278167
10099.2 0.9272024
11264.4 0.9286758
13294.5 0.9270308
15820.2 0.9290342
18140.7 0.9270568
20227.5 0.9298575
22235.1 0.9291975
24512.3 0.9293641
26777.2 0.9295692
28484.7 0.930255
30027.7 0.9294967
32398.4 0.9267225
35192.2 0.9265524
37955 0.9263732
40416.2 0.92793
42353.3 0.9297726
43681.1 0.9321309
43905.2 0.9295809
44074.2 0.9298651
45208.4 0.9299716
47960.5 0.9282891
50928.1 0.9279642
54023.4 0.9283224
57345.4 0.9283307
60736.4 0.9278592
64416.3 0.9267741
68398.8 0.9309509
72464.7 0.9298792
76719.1 0.92883
81029.8 0.9298484
85463.5 0.9295008
89804.6 0.9287875
93957.1 0.9298908
97191.2 0.929985
99169.7 0.9310016
99867.6 0.9295683
99991 0.9300066
99999.7 0.9306709
100000 0.9297091
};
\addlegendentry{\scriptsize ACOWA, $\beta=20$}
\addplot [crimson2143940, opacity=1.0]
table {%
6.2 0.6538825
6.2 0.6625384
7.5 0.6647626
9.1 0.66468
13 0.6881709
23.3 0.709316
35.3 0.7301441
45.8 0.750625
67.3 0.7594533
75.6 0.7612658
90.1 0.7686884
106 0.7775101
130.7 0.7866025
166.7 0.7928709
208.5 0.8041642
261.8 0.8174774
309.3 0.8204041
368.3 0.8351767
436.9 0.8380857
526.5 0.8513867
624 0.8634208
711.7 0.8680332
798.1 0.8681758
972.6 0.8871383
1208.2 0.8873916
1513 0.8964175
1857.5 0.9024107
2231.6 0.9052825
2628 0.9116241
3054.8 0.9117967
3475.4 0.915294
3616.5 0.9149799
3919.5 0.9148933
4502.9 0.9191319
5206.3 0.9199485
5955.6 0.9221701
6800.9 0.9216117
7667.4 0.9230517
8545.4 0.9251426
9324.8 0.9247183
9717.5 0.9271641
10359.4 0.9275133
11384.5 0.9264058
12811.4 0.9281342
15251.9 0.9265133
17741.3 0.9270784
19904.9 0.9293341
21948.1 0.9297251
24236.7 0.9303833
26274.9 0.9306776
26468.3 0.9299834
27069 0.9303284
27278.5 0.929655
28233.1 0.9308941
29228.5 0.9298318
29409.9 0.930859
32893.4 0.9315092
36948.3 0.9324192
40163.3 0.9291407
43039.6 0.9287551
45828 0.9297251
48607.1 0.9290433
51487.4 0.9291484
54450.8 0.9281375
57551.6 0.9268166
60782.9 0.9275616
64276.9 0.9267508
68066.8 0.9296801
72024.7 0.9296817
76132.7 0.928955
80386.1 0.9291691
84793.9 0.929986
89203.8 0.9289018
93410 0.9291558
96890.4 0.9306368
99007.3 0.9306634
99828.1 0.9292792
99988.3 0.9297458
99999.7 0.9311466
100000 0.9301708
};
\addlegendentry{\scriptsize ACOWA, $\beta=40$}
\addplot [mediumpurple148103189, opacity=1.0]
table {%
1 0.5009067
1 0.5009067
1 0.5009067
1 0.4987258
1 0.5009067
1 0.5009067
1 0.5009067
1 0.5009067
1 0.5009067
1 0.4986592
1 0.4986567
1 0.5009067
1 0.4986592
7.7 0.6734308
27.9 0.6838134
45.3 0.7098833
72 0.724055
117 0.7341682
187.5 0.7716959
264.1 0.7701084
360.8 0.800265
526.1 0.8150183
823.3 0.8097769
1334.9 0.8419209
2053.2 0.8450984
2993.2 0.848785
4168.3 0.8696683
5665.8 0.8714217
7372.3 0.8710675
9439 0.8828851
11923.9 0.8881674
14917.8 0.8993526
18302.6 0.9030191
21750.3 0.8978358
24976.3 0.9096825
26979.4 0.9135826
28801.6 0.9139909
31742.1 0.9160992
35352.3 0.8908591
38972.2 0.9189109
42735.6 0.9220417
46285.3 0.9232083
49578.1 0.9238757
50161.6 0.9240982
50668.8 0.9255966
51619.1 0.9215716
52836.6 0.926335
56301.4 0.9269651
59957.7 0.9277542
63842.4 0.9239408
68150.9 0.9271883
71898 0.9272758
72696.8 0.9266459
74261.8 0.9282451
75746 0.9234509
76646.7 0.9276767
78564.7 0.9295675
79099.9 0.9255849
83017 0.9291533
87165.8 0.9292891
90929.4 0.9262317
94103.3 0.9261959
96605.6 0.92901
97058.3 0.9283067
98258.1 0.9277868
98302.5 0.9131109
98345.9 0.9272175
99192.2 0.9283567
99332.3 0.9273549
99334.4 0.9279058
99815.8 0.9273792
99968.8 0.92838
99996.3 0.9265066
100000 0.9274734
100000 0.9264574
100000 0.9266883
100000 0.9272558
100000 0.9266568
100000 0.92573
100000 0.9270391
};
\addlegendentry{\scriptsize OWA}
\end{axis}

\end{tikzpicture}

%% file: figs/app/elnet_newsgroups_nnz_vs_test_acc.tex
\begin{tikzpicture}[scale=1.0]

\definecolor{darkgray153}{RGB}{153,153,153}
\definecolor{darkorange25512714}{RGB}{255,127,14}
\definecolor{forestgreen4416044}{RGB}{44,160,44}
\definecolor{gainsboro229}{RGB}{229,229,229}
\definecolor{steelblue31119180}{RGB}{31,119,180}

\begin{axis}[
width=0.95\textwidth,
height=0.35\textwidth,
legend cell align={left},
legend style={
  fill opacity=0.8,
  draw opacity=1,
  text opacity=1,
  at={(0.97,0.03)},
  anchor=south east,
  draw=none
},
log basis x={10},
tick align=outside,
tick pos=left,
x grid style={gainsboro229},
xlabel={\small Number of non-zeros},
xmajorgrids,
xmin=10, xmax=10000,
xminorgrids,
xmode=log,
xtick style={color=black},
y grid style={gainsboro229},
ylabel={\small Accuracy},
ymajorgrids,
ymin=0.450573709414845, ymax=0.986651143296915,
yminorgrids,
ytick style={color=black}
]
\path [fill=steelblue31119180, fill opacity=0.2]
(axis cs:2.4,0.52568935924037)
--(axis cs:2.4,0.49826864075963)
--(axis cs:4.4,0.49236508813712)
--(axis cs:10.4,0.487707925476356)
--(axis cs:26.8,0.47507810356449)
--(axis cs:50.2,0.474940865500394)
--(axis cs:104.2,0.489237244529843)
--(axis cs:234.6,0.586553361790267)
--(axis cs:487.6,0.66226481265741)
--(axis cs:931.2,0.691488313337435)
--(axis cs:1517.6,0.708872244129516)
--(axis cs:2202.2,0.715818565846451)
--(axis cs:2962.8,0.725457380985665)
--(axis cs:3727.4,0.73842808599576)
--(axis cs:4556.4,0.753958930195421)
--(axis cs:5557.6,0.767773355504659)
--(axis cs:6868.8,0.785683553060047)
--(axis cs:8438.4,0.80073199596352)
--(axis cs:10228,0.81311825215756)
--(axis cs:12588.2,0.825524628038652)
--(axis cs:15643.2,0.838278312870367)
--(axis cs:20459,0.853926860923087)
--(axis cs:23608.4,0.861988387948071)
--(axis cs:24088,0.861462273025151)
--(axis cs:28730.4,0.872935882258772)
--(axis cs:32598.8,0.891847821230472)
--(axis cs:34451,0.89191043628871)
--(axis cs:36173.6,0.894164224579017)
--(axis cs:37779.8,0.892505857054653)
--(axis cs:39233,0.889970924311553)
--(axis cs:40534.2,0.890388546823407)
--(axis cs:41725.2,0.890946182900143)
--(axis cs:42819.2,0.891243720589994)
--(axis cs:43779.8,0.891131574971858)
--(axis cs:44619.8,0.890973598041436)
--(axis cs:45333.2,0.891158434253002)
--(axis cs:45946.2,0.891069834253002)
--(axis cs:46473,0.891069834253002)
--(axis cs:46916.4,0.891069834253002)
--(axis cs:47307.2,0.891069834253002)
--(axis cs:47648.8,0.891057280258681)
--(axis cs:47936.4,0.891057280258681)
--(axis cs:48196.6,0.891057280258681)
--(axis cs:48443.4,0.891057280258681)
--(axis cs:48635.2,0.891057280258681)
--(axis cs:48818.2,0.891057280258681)
--(axis cs:48953,0.891057280258681)
--(axis cs:49087.2,0.891057280258681)
--(axis cs:49217.2,0.891057280258681)
--(axis cs:49329.4,0.891057280258681)
--(axis cs:49425,0.891057280258681)
--(axis cs:49425,0.901844319741319)
--(axis cs:49425,0.901844319741319)
--(axis cs:49329.4,0.901844319741319)
--(axis cs:49217.2,0.901844319741319)
--(axis cs:49087.2,0.901844319741319)
--(axis cs:48953,0.901844319741319)
--(axis cs:48818.2,0.901844319741319)
--(axis cs:48635.2,0.901844319741319)
--(axis cs:48443.4,0.901844319741319)
--(axis cs:48196.6,0.901844319741319)
--(axis cs:47936.4,0.901844319741319)
--(axis cs:47648.8,0.901844319741319)
--(axis cs:47307.2,0.901654565746998)
--(axis cs:46916.4,0.901654565746998)
--(axis cs:46473,0.901654565746998)
--(axis cs:45946.2,0.901654565746998)
--(axis cs:45333.2,0.901743165746998)
--(axis cs:44619.8,0.901573201958564)
--(axis cs:43779.8,0.902124825028142)
--(axis cs:42819.2,0.902190279410006)
--(axis cs:41725.2,0.902487817099857)
--(axis cs:40534.2,0.902690253176593)
--(axis cs:39233,0.903108275688447)
--(axis cs:37779.8,0.901992142945348)
--(axis cs:36173.6,0.902286575420983)
--(axis cs:34451,0.90329796371129)
--(axis cs:32598.8,0.899633778769529)
--(axis cs:28730.4,0.883408117741228)
--(axis cs:24088,0.870214526974849)
--(axis cs:23608.4,0.869688412051929)
--(axis cs:20459,0.861778339076913)
--(axis cs:15643.2,0.851340087129632)
--(axis cs:12588.2,0.840136571961348)
--(axis cs:10228,0.82805294784244)
--(axis cs:8438.4,0.81417440403648)
--(axis cs:6868.8,0.799409646939953)
--(axis cs:5557.6,0.785199044495341)
--(axis cs:4556.4,0.766715469804579)
--(axis cs:3727.4,0.75385271400424)
--(axis cs:2962.8,0.741978619014335)
--(axis cs:2202.2,0.732983434153549)
--(axis cs:1517.6,0.719344555870484)
--(axis cs:931.2,0.707624086662565)
--(axis cs:487.6,0.67970438734259)
--(axis cs:234.6,0.622142238209733)
--(axis cs:104.2,0.527798755470157)
--(axis cs:50.2,0.504118334499606)
--(axis cs:26.8,0.50770829643551)
--(axis cs:10.4,0.521874874523644)
--(axis cs:4.4,0.52609131186288)
--(axis cs:2.4,0.52568935924037)
--cycle;

\path [fill=darkorange25512714, fill opacity=0.2]
(axis cs:1.8,0.532352431319055)
--(axis cs:1.8,0.500833568680945)
--(axis cs:4,0.488815508674739)
--(axis cs:10,0.484062164498056)
--(axis cs:26.6,0.504919120313814)
--(axis cs:50.2,0.527028290141168)
--(axis cs:104.2,0.542460952951285)
--(axis cs:234.6,0.606026582879147)
--(axis cs:487.6,0.674800980952795)
--(axis cs:931.2,0.727064060523255)
--(axis cs:1517.6,0.699625513547601)
--(axis cs:2202.2,0.726572262591961)
--(axis cs:2962.8,0.735695550250715)
--(axis cs:3727.4,0.746822393306675)
--(axis cs:4556.4,0.772525269679222)
--(axis cs:5557.6,0.767632891699461)
--(axis cs:6868.8,0.775469466530345)
--(axis cs:8438.4,0.813885374457214)
--(axis cs:10228,0.826562451370891)
--(axis cs:12588.2,0.836908706668059)
--(axis cs:15643.2,0.834843346716296)
--(axis cs:20459,0.845399726559514)
--(axis cs:23608.4,0.851769448782475)
--(axis cs:24088,0.871693325367132)
--(axis cs:28730.4,0.878167951106805)
--(axis cs:32598.8,0.877746932068067)
--(axis cs:34451,0.880459886149814)
--(axis cs:36173.6,0.891884430355805)
--(axis cs:37779.8,0.896650806442479)
--(axis cs:39233,0.896717566694674)
--(axis cs:40534.2,0.882169223096463)
--(axis cs:41725.2,0.892051223359218)
--(axis cs:42819.2,0.889688743308431)
--(axis cs:43779.8,0.894794188880998)
--(axis cs:44619.8,0.892281574820662)
--(axis cs:45333.2,0.880460584396421)
--(axis cs:45946.2,0.88564080356156)
--(axis cs:46473,0.89710582758928)
--(axis cs:46916.4,0.890192833156474)
--(axis cs:47307.2,0.880900172714258)
--(axis cs:47648.8,0.884423443812026)
--(axis cs:47936.4,0.884448335617967)
--(axis cs:48196.6,0.883982279720772)
--(axis cs:48443.4,0.889512739136993)
--(axis cs:48635.2,0.895125389457145)
--(axis cs:48818.2,0.873538066095309)
--(axis cs:48953,0.891004628339506)
--(axis cs:49087.2,0.894043263507134)
--(axis cs:49217.2,0.893015407482968)
--(axis cs:49329.4,0.902238632556797)
--(axis cs:49425,0.883086091871062)
--(axis cs:49425,0.910525108128938)
--(axis cs:49425,0.910525108128938)
--(axis cs:49329.4,0.914264967443203)
--(axis cs:49217.2,0.907162192517032)
--(axis cs:49087.2,0.911457936492866)
--(axis cs:48953,0.901718971660494)
--(axis cs:48818.2,0.898068333904691)
--(axis cs:48635.2,0.915167410542854)
--(axis cs:48443.4,0.901082060863007)
--(axis cs:48196.6,0.911936520279228)
--(axis cs:47936.4,0.905080864382033)
--(axis cs:47648.8,0.905816156187974)
--(axis cs:47307.2,0.920519427285742)
--(axis cs:46916.4,0.911936366843526)
--(axis cs:46473,0.90857257241072)
--(axis cs:45946.2,0.91169679643844)
--(axis cs:45333.2,0.900195815603579)
--(axis cs:44619.8,0.910734825179338)
--(axis cs:43779.8,0.908577811119002)
--(axis cs:42819.2,0.903212856691569)
--(axis cs:41725.2,0.908658776640782)
--(axis cs:40534.2,0.902746376903537)
--(axis cs:39233,0.909848833305326)
--(axis cs:37779.8,0.904768793557521)
--(axis cs:36173.6,0.915214369644195)
--(axis cs:34451,0.910135313850186)
--(axis cs:32598.8,0.908055867931934)
--(axis cs:28730.4,0.892373248893195)
--(axis cs:24088,0.894588674632868)
--(axis cs:23608.4,0.888070551217525)
--(axis cs:20459,0.899054673440486)
--(axis cs:15643.2,0.881217053283704)
--(axis cs:12588.2,0.87666689333194)
--(axis cs:10228,0.863766348629109)
--(axis cs:8438.4,0.862245825542786)
--(axis cs:6868.8,0.823998533469655)
--(axis cs:5557.6,0.837513908300539)
--(axis cs:4556.4,0.823747930320778)
--(axis cs:3727.4,0.797986806693325)
--(axis cs:2962.8,0.808936049749285)
--(axis cs:2202.2,0.787890937408039)
--(axis cs:1517.6,0.799398886452399)
--(axis cs:931.2,0.804967939476745)
--(axis cs:487.6,0.779147019047205)
--(axis cs:234.6,0.733991017120853)
--(axis cs:104.2,0.726571447048715)
--(axis cs:50.2,0.661081709858832)
--(axis cs:26.6,0.606350079686187)
--(axis cs:10,0.515405835501944)
--(axis cs:4,0.522719691325261)
--(axis cs:1.8,0.532352431319055)
--cycle;

\path [fill=forestgreen4416044, fill opacity=0.2]
(axis cs:1,0.535068890768472)
--(axis cs:1,0.504327509231528)
--(axis cs:1,0.504327509231528)
--(axis cs:1.6,0.49826864075963)
--(axis cs:1.6,0.494626472846138)
--(axis cs:2.6,0.536908800026052)
--(axis cs:6.6,0.612767430312943)
--(axis cs:9.6,0.629890174858519)
--(axis cs:18.2,0.5029481937991)
--(axis cs:30,0.709149294881269)
--(axis cs:40.6,0.736126555492591)
--(axis cs:58,0.769452766598246)
--(axis cs:84,0.781409859275661)
--(axis cs:104.4,0.796222242493985)
--(axis cs:139.6,0.815154277259926)
--(axis cs:193.6,0.795422942654713)
--(axis cs:264.8,0.837240768328591)
--(axis cs:409.8,0.814927583066632)
--(axis cs:571.2,0.856435846336492)
--(axis cs:819,0.868948450431205)
--(axis cs:1214.2,0.882696571228116)
--(axis cs:1690.8,0.869495929576616)
--(axis cs:2245.2,0.88225222553442)
--(axis cs:2862,0.883575462350818)
--(axis cs:4547.6,0.889663964330244)
--(axis cs:8100.8,0.895447126622131)
--(axis cs:10087.6,0.896725912902015)
--(axis cs:13798.6,0.893616649323182)
--(axis cs:19188.6,0.917521277812866)
--(axis cs:24900,0.924159854053399)
--(axis cs:29554.6,0.929963952383607)
--(axis cs:33695.4,0.923126884380173)
--(axis cs:38245.4,0.933186927743292)
--(axis cs:42851.2,0.951534987115683)
--(axis cs:46810.8,0.937876077020721)
--(axis cs:50068.6,0.943661212788634)
--(axis cs:51038.8,0.943053774347761)
--(axis cs:51079.2,0.948536099994365)
--(axis cs:51079.2,0.947920052558811)
--(axis cs:51079.2,0.93664522908218)
--(axis cs:51079.2,0.944119277581688)
--(axis cs:51079.2,0.948351720973611)
--(axis cs:51079.2,0.941016491686589)
--(axis cs:51079.2,0.944143732636245)
--(axis cs:51079.2,0.932990464536993)
--(axis cs:51079.2,0.938772046534281)
--(axis cs:51079.2,0.943955250979901)
--(axis cs:51079.2,0.940022516172582)
--(axis cs:51079.2,0.941507507217436)
--(axis cs:51079.2,0.940564168243582)
--(axis cs:51079.2,0.935803919017838)
--(axis cs:51079.2,0.955411280982162)
--(axis cs:51079.2,0.955411280982162)
--(axis cs:51079.2,0.956329831756419)
--(axis cs:51079.2,0.955031692782564)
--(axis cs:51079.2,0.952435483827418)
--(axis cs:51079.2,0.957020349020099)
--(axis cs:51079.2,0.959009553465719)
--(axis cs:51079.2,0.955918335463007)
--(axis cs:51079.2,0.950621467363756)
--(axis cs:51079.2,0.949134708313411)
--(axis cs:51079.2,0.957770679026389)
--(axis cs:51079.2,0.957744322418312)
--(axis cs:51079.2,0.96166877091782)
--(axis cs:51079.2,0.956604747441189)
--(axis cs:51079.2,0.957409100005635)
--(axis cs:51038.8,0.957035425652239)
--(axis cs:50068.6,0.962283987211367)
--(axis cs:46810.8,0.954937122979278)
--(axis cs:42851.2,0.954409812884317)
--(axis cs:38245.4,0.953591472256708)
--(axis cs:33695.4,0.949632315619827)
--(axis cs:29554.6,0.939068847616393)
--(axis cs:24900,0.943630545946601)
--(axis cs:19188.6,0.937136722187134)
--(axis cs:13798.6,0.933357750676818)
--(axis cs:10087.6,0.925102087097985)
--(axis cs:8100.8,0.907391673377869)
--(axis cs:4547.6,0.904834835669756)
--(axis cs:2862,0.899033737649182)
--(axis cs:2245.2,0.90017937446558)
--(axis cs:1690.8,0.892527270423384)
--(axis cs:1214.2,0.893168628771884)
--(axis cs:819,0.877813149568795)
--(axis cs:571.2,0.869739353663509)
--(axis cs:409.8,0.865994816933368)
--(axis cs:264.8,0.846166831671409)
--(axis cs:193.6,0.842731857345287)
--(axis cs:139.6,0.830453722740074)
--(axis cs:104.4,0.824186157506015)
--(axis cs:84,0.812379340724339)
--(axis cs:58,0.784052033401754)
--(axis cs:40.6,0.760235844507409)
--(axis cs:30,0.742847105118731)
--(axis cs:18.2,0.7722962062009)
--(axis cs:9.6,0.656179025141481)
--(axis cs:6.6,0.626983769687057)
--(axis cs:2.6,0.567616799973948)
--(axis cs:1.6,0.514601927153862)
--(axis cs:1.6,0.52568935924037)
--(axis cs:1,0.535068890768472)
--(axis cs:1,0.535068890768472)
--cycle;

\addplot [semithick, black, dashed]
table {%
20.2 0.690772
30.2 0.7241348
46.6 0.7613132
64 0.7926354
81.8 0.8167702
118.4 0.8361134
164.6 0.857675
224.8 0.8780832
300.4 0.8925466
389.4 0.9058562
508.4 0.9157056
661.2 0.9253772
849.2 0.931322
1048.6 0.9365572
1287.2 0.9392192
1540.8 0.9431234
2315.8 0.9479148
2984 0.949867
3848.8 0.9512868
4292 0.952174
5290.8 0.9530614
6152.2 0.9543034
6951.2 0.9503992
7244.8 0.953771
7455.4 0.9525286
8045.4 0.9530614
9358 0.9503104
10229.8 0.9455188
10501.2 0.9456076
11480 0.9492458
11823.6 0.9440994
13140.6 0.9440106
13495.2 0.9463178
14875.4 0.9445432
15318 0.9368234
15840.6 0.9417036
16129.2 0.936291
17403.4 0.9366458
20222.6 0.9360248
23187.8 0.9359362
25847.8 0.9311448
29827.2 0.9268854
35197.2 0.9203192
38721 0.9144634
41783.2 0.9114462
46046.8 0.9057676
47575.4 0.902041
48697 0.8992014
49500.6 0.896362
50042.2 0.8946762
};
\addlegendentry{\scriptsize LIBLINEAR}
\addplot [steelblue31119180, opacity=1.0, mark=*, mark size=0, mark options={solid}]
table {%
2.4 0.511979
4.4 0.5092282
10.4 0.5047914
26.8 0.4913932
50.2 0.4895296
104.2 0.508518
234.6 0.6043478
487.6 0.6709846
931.2 0.6995562
1517.6 0.7141084
2202.2 0.724401
2962.8 0.733718
3727.4 0.7461404
4556.4 0.7603372
5557.6 0.7764862
6868.8 0.7925466
8438.4 0.8074532
10228 0.8205856
12588.2 0.8328306
15643.2 0.8448092
20459 0.8578526
23608.4 0.8658384
24088 0.8658384
28730.4 0.878172
32598.8 0.8957408
34451 0.8976042
36173.6 0.8982254
37779.8 0.897249
39233 0.8965396
40534.2 0.8965394
41725.2 0.896717
42819.2 0.896717
43779.8 0.8966282
44619.8 0.8962734
45333.2 0.8964508
45946.2 0.8963622
46473 0.8963622
46916.4 0.8963622
47307.2 0.8963622
47648.8 0.8964508
47936.4 0.8964508
48196.6 0.8964508
48443.4 0.8964508
48635.2 0.8964508
48818.2 0.8964508
48953 0.8964508
49087.2 0.8964508
49217.2 0.8964508
49329.4 0.8964508
49425 0.8964508
};
\addlegendentry{\scriptsize Naive Avg.}
\addplot [darkorange25512714, opacity=1.0, mark=*, mark size=0, mark options={solid}]
table {%
1.8 0.516593
4 0.5057676
10 0.499734
26.6 0.5556346
50.2 0.594055
104.2 0.6345162
234.6 0.6700088
487.6 0.726974
931.2 0.766016
1517.6 0.7495122
2202.2 0.7572316
2962.8 0.7723158
3727.4 0.7724046
4556.4 0.7981366
5557.6 0.8025734
6868.8 0.799734
8438.4 0.8380656
10228 0.8451644
12588.2 0.8567878
15643.2 0.8580302
20459 0.8722272
23608.4 0.86992
24088 0.883141
28730.4 0.8852706
32598.8 0.8929014
34451 0.8952976
36173.6 0.9035494
37779.8 0.9007098
39233 0.9032832
40534.2 0.8924578
41725.2 0.900355
42819.2 0.8964508
43779.8 0.901686
44619.8 0.9015082
45333.2 0.8903282
45946.2 0.8986688
46473 0.9028392
46916.4 0.9010646
47307.2 0.9007098
47648.8 0.8951198
47936.4 0.8947646
48196.6 0.8979594
48443.4 0.8952974
48635.2 0.9051464
48818.2 0.8858032
48953 0.8963618
49087.2 0.9027506
49217.2 0.9000888
49329.4 0.9082518
49425 0.8968056
};
\addlegendentry{\scriptsize OWA}
\addplot [forestgreen4416044, opacity=1.0, mark=*, mark size=0, mark options={solid}]
table {%
1 0.5196982
1 0.5196982
1.6 0.511979
1.6 0.5046142
2.6 0.5522628
6.6 0.6198756
9.6 0.6430346
18.2 0.6376222
30 0.7259982
40.6 0.7481812
58 0.7767524
84 0.7968946
104.4 0.8102042
139.6 0.822804
193.6 0.8190774
264.8 0.8417038
409.8 0.8404612
571.2 0.8630876
819 0.8733808
1214.2 0.8879326
1690.8 0.8810116
2245.2 0.8912158
2862 0.8913046
4547.6 0.8972494
8100.8 0.9014194
10087.6 0.910914
13798.6 0.9134872
19188.6 0.927329
24900 0.9338952
29554.6 0.9345164
33695.4 0.9363796
38245.4 0.9433892
42851.2 0.9529724
46810.8 0.9464066
50068.6 0.9529726
51038.8 0.9500446
51079.2 0.9529726
51079.2 0.9522624
51079.2 0.949157
51079.2 0.9509318
51079.2 0.9530612
51079.2 0.9450756
51079.2 0.9473826
51079.2 0.9444544
51079.2 0.9488908
51079.2 0.9504878
51079.2 0.946229
51079.2 0.9482696
51079.2 0.948447
51079.2 0.9456076
};
\addlegendentry{\scriptsize ACOWA}
\end{axis}

\end{tikzpicture}

%% file: figs/app/elnet_amazon7_nnz_vs_test_acc.tex
\begin{tikzpicture}[scale=1.0]

\definecolor{darkgray153}{RGB}{153,153,153}
\definecolor{darkorange25512714}{RGB}{255,127,14}
\definecolor{forestgreen4416044}{RGB}{44,160,44}
\definecolor{gainsboro229}{RGB}{229,229,229}
\definecolor{steelblue31119180}{RGB}{31,119,180}

\begin{axis}[
width=0.95\textwidth,
height=0.35\textwidth,
legend cell align={left},
legend style={
  fill opacity=0.8,
  draw opacity=1,
  text opacity=1,
  at={(0.97,0.03)},
  anchor=south east,
  draw=none
},
log basis x={10},
tick align=outside,
tick pos=left,
x grid style={gainsboro229},
xlabel={\small Number of non-zeros},
xmajorgrids,
xmin=10, xmax=10000,
xminorgrids,
xmode=log,
xtick style={color=black},
y grid style={gainsboro229},
ylabel={\small Accuracy},
ymajorgrids,
ymin=0.8, ymax=1,
yminorgrids,
ytick style={color=black}
]
\path [fill=steelblue31119180, fill opacity=0.2]
(axis cs:5.6,0.716710701319331)
--(axis cs:5.6,0.714853698680669)
--(axis cs:6,0.714853708258749)
--(axis cs:6,0.714853698680669)
--(axis cs:6,0.714853698680669)
--(axis cs:6,0.714853698680669)
--(axis cs:6,0.714853698680669)
--(axis cs:9,0.726958703837111)
--(axis cs:16,0.828790837081867)
--(axis cs:19.4,0.883630333029495)
--(axis cs:22.6,0.893672880390767)
--(axis cs:29.6,0.9010543539562)
--(axis cs:38.8,0.906570679145987)
--(axis cs:48.8,0.912704842233775)
--(axis cs:64,0.918206020130396)
--(axis cs:86.8,0.923246010241784)
--(axis cs:131,0.927961971298867)
--(axis cs:188.4,0.932045746027476)
--(axis cs:276.2,0.936021178738972)
--(axis cs:442.6,0.939971809257924)
--(axis cs:694.6,0.944099436591238)
--(axis cs:1051.8,0.948037174098592)
--(axis cs:1570.2,0.951605751203968)
--(axis cs:2264.4,0.954849118137307)
--(axis cs:3194.2,0.957767743257961)
--(axis cs:4414.6,0.960143710825591)
--(axis cs:5921,0.962240188822741)
--(axis cs:7646.2,0.963908453389001)
--(axis cs:9698.2,0.965340507203892)
--(axis cs:12666.6,0.966464889462049)
--(axis cs:15598,0.96743964402839)
--(axis cs:19587.8,0.968153833242376)
--(axis cs:24033.4,0.968823541615924)
--(axis cs:28315.2,0.969413473186556)
--(axis cs:32538,0.969913704439621)
--(axis cs:36716,0.970341998793218)
--(axis cs:40887.4,0.970514706384703)
--(axis cs:45550.8,0.970418142310503)
--(axis cs:50372.4,0.970107545852646)
--(axis cs:55256.4,0.96987452491219)
--(axis cs:59696.6,0.969629014686982)
--(axis cs:62300.2,0.969762203461042)
--(axis cs:64186.2,0.969833977009186)
--(axis cs:69161,0.969560280952153)
--(axis cs:75890.4,0.968802155635064)
--(axis cs:81899.8,0.968103276072559)
--(axis cs:91222.8,0.966916465712458)
--(axis cs:111862,0.964513707874428)
--(axis cs:131773.2,0.961005773055181)
--(axis cs:141100.8,0.959790031108814)
--(axis cs:141690.6,0.960425035515037)
--(axis cs:141690.6,0.961479764484963)
--(axis cs:141690.6,0.961479764484963)
--(axis cs:141100.8,0.960916368891186)
--(axis cs:131773.2,0.962033826944819)
--(axis cs:111862,0.965236292125571)
--(axis cs:91222.8,0.967680334287542)
--(axis cs:81899.8,0.968546723927441)
--(axis cs:75890.4,0.969246244364936)
--(axis cs:69161,0.970058519047847)
--(axis cs:64186.2,0.970501222990815)
--(axis cs:62300.2,0.970357396538958)
--(axis cs:59696.6,0.970177785313018)
--(axis cs:55256.4,0.97037987508781)
--(axis cs:50372.4,0.970777254147354)
--(axis cs:45550.8,0.971055057689497)
--(axis cs:40887.4,0.971155293615297)
--(axis cs:36716,0.970955201206782)
--(axis cs:32538,0.970600695560379)
--(axis cs:28315.2,0.970121726813444)
--(axis cs:24033.4,0.969610458384076)
--(axis cs:19587.8,0.968874566757623)
--(axis cs:15598,0.96815595597161)
--(axis cs:12666.6,0.967133910537951)
--(axis cs:9698.2,0.965949892796108)
--(axis cs:7646.2,0.964558346610999)
--(axis cs:5921,0.962977411177259)
--(axis cs:4414.6,0.961011089174409)
--(axis cs:3194.2,0.958574256742039)
--(axis cs:2264.4,0.955515681862693)
--(axis cs:1570.2,0.952488248796032)
--(axis cs:1051.8,0.948668825901408)
--(axis cs:694.6,0.944576563408762)
--(axis cs:442.6,0.940462190742076)
--(axis cs:276.2,0.936569621261028)
--(axis cs:188.4,0.932850653972524)
--(axis cs:131,0.928858828701133)
--(axis cs:86.8,0.924255589758216)
--(axis cs:64,0.918975979869604)
--(axis cs:48.8,0.913513157766225)
--(axis cs:38.8,0.907476520854013)
--(axis cs:29.6,0.9022480460438)
--(axis cs:22.6,0.894948319609233)
--(axis cs:19.4,0.884230066970505)
--(axis cs:16,0.831944362918133)
--(axis cs:9,0.730359696162889)
--(axis cs:6,0.716710701319331)
--(axis cs:6,0.716710701319331)
--(axis cs:6,0.716710701319331)
--(axis cs:6,0.716710701319331)
--(axis cs:6,0.716707491741251)
--(axis cs:5.6,0.716710701319331)
--cycle;

\path [fill=darkorange25512714, fill opacity=0.2]
(axis cs:5.6,0.716710701319331)
--(axis cs:5.6,0.714853698680669)
--(axis cs:6,0.714853698680669)
--(axis cs:6,0.714853698680669)
--(axis cs:6,0.714853698680669)
--(axis cs:6,0.714853698680669)
--(axis cs:6,0.714853698680669)
--(axis cs:9,0.726954662342542)
--(axis cs:16,0.828754710068227)
--(axis cs:19.4,0.883627399896836)
--(axis cs:22.6,0.893675536162086)
--(axis cs:29.6,0.901041398717529)
--(axis cs:38.8,0.907834002081452)
--(axis cs:48.8,0.914258572006401)
--(axis cs:64,0.932953640518109)
--(axis cs:86.8,0.92699016157667)
--(axis cs:131,0.938339738223956)
--(axis cs:188.4,0.946568553281138)
--(axis cs:276.2,0.950919579639998)
--(axis cs:442.6,0.952458589652965)
--(axis cs:694.6,0.95782686474659)
--(axis cs:1051.8,0.960087538182202)
--(axis cs:1570.2,0.959474394144389)
--(axis cs:2264.4,0.961854447235694)
--(axis cs:3194.2,0.963174504450986)
--(axis cs:4414.6,0.963806494625754)
--(axis cs:5921,0.965033074918933)
--(axis cs:7646.2,0.966639523505557)
--(axis cs:9698.2,0.966439619903117)
--(axis cs:12666.6,0.966672610207731)
--(axis cs:15598,0.967542451211528)
--(axis cs:19587.8,0.968447155500336)
--(axis cs:24033.4,0.967629121628285)
--(axis cs:28315.2,0.969484589153543)
--(axis cs:32538,0.970194685712361)
--(axis cs:36716,0.970472638782768)
--(axis cs:40887.4,0.970535603991029)
--(axis cs:45550.8,0.970400739218291)
--(axis cs:50372.4,0.970119546572743)
--(axis cs:55256.4,0.969889445451567)
--(axis cs:59696.6,0.969631268984751)
--(axis cs:62300.2,0.969685480701915)
--(axis cs:64186.2,0.96988224583646)
--(axis cs:69161,0.969630497095197)
--(axis cs:75890.4,0.968812970860471)
--(axis cs:81899.8,0.968118360952383)
--(axis cs:91222.8,0.966996654046862)
--(axis cs:111862,0.964867915288232)
--(axis cs:131773.2,0.961133285823394)
--(axis cs:141100.8,0.959854954807801)
--(axis cs:141690.6,0.960479345028419)
--(axis cs:141690.6,0.961500654971581)
--(axis cs:141690.6,0.961500654971581)
--(axis cs:141100.8,0.9609578451922)
--(axis cs:131773.2,0.962169514176606)
--(axis cs:111862,0.965607684711768)
--(axis cs:91222.8,0.967792945953138)
--(axis cs:81899.8,0.968598839047617)
--(axis cs:75890.4,0.969274229139529)
--(axis cs:69161,0.970115902904803)
--(axis cs:64186.2,0.97055415416354)
--(axis cs:62300.2,0.970914119298085)
--(axis cs:59696.6,0.970237131015249)
--(axis cs:55256.4,0.970416554548434)
--(axis cs:50372.4,0.970813253427257)
--(axis cs:45550.8,0.971066460781709)
--(axis cs:40887.4,0.971247196008971)
--(axis cs:36716,0.970960961217232)
--(axis cs:32538,0.970830914287639)
--(axis cs:28315.2,0.970618610846457)
--(axis cs:24033.4,0.969848878371715)
--(axis cs:19587.8,0.969578444499664)
--(axis cs:15598,0.969046748788472)
--(axis cs:12666.6,0.968714189792269)
--(axis cs:9698.2,0.968375980096883)
--(axis cs:7646.2,0.966936876494444)
--(axis cs:5921,0.965950525081067)
--(axis cs:4414.6,0.965203905374246)
--(axis cs:3194.2,0.964502295549014)
--(axis cs:2264.4,0.963257152764306)
--(axis cs:1570.2,0.962555205855611)
--(axis cs:1051.8,0.960569661817798)
--(axis cs:694.6,0.95879673525341)
--(axis cs:442.6,0.955847010347035)
--(axis cs:276.2,0.951674020360002)
--(axis cs:188.4,0.947727446718862)
--(axis cs:131,0.944436661776044)
--(axis cs:86.8,0.93854463842333)
--(axis cs:64,0.936037159481891)
--(axis cs:48.8,0.932253427993599)
--(axis cs:38.8,0.926878397918548)
--(axis cs:29.6,0.902257801282472)
--(axis cs:22.6,0.894954463837914)
--(axis cs:19.4,0.884209000103164)
--(axis cs:16,0.831922089931773)
--(axis cs:9,0.730359337657458)
--(axis cs:6,0.716710701319331)
--(axis cs:6,0.716710701319331)
--(axis cs:6,0.716710701319331)
--(axis cs:6,0.716710701319331)
--(axis cs:6,0.716710701319331)
--(axis cs:5.6,0.716710701319331)
--cycle;

\path [fill=forestgreen4416044, fill opacity=0.2]
(axis cs:11,0.890523352084578)
--(axis cs:11,0.889689847915422)
--(axis cs:11.6,0.890735520736085)
--(axis cs:14.4,0.89979685587627)
--(axis cs:15.8,0.899903810108026)
--(axis cs:16,0.90790865043415)
--(axis cs:19,0.909385693840223)
--(axis cs:21.6,0.913553707162602)
--(axis cs:24.8,0.917024668269292)
--(axis cs:27,0.917197840732351)
--(axis cs:31,0.923668544042528)
--(axis cs:37,0.926326854668033)
--(axis cs:44.4,0.932713574653097)
--(axis cs:51.6,0.937116801546864)
--(axis cs:62,0.938489014147338)
--(axis cs:77.2,0.940985930428544)
--(axis cs:109,0.944710135712317)
--(axis cs:141.2,0.948655845044727)
--(axis cs:191.4,0.952959041847661)
--(axis cs:275.8,0.955491877111362)
--(axis cs:394.6,0.959687060381937)
--(axis cs:560.4,0.962531069508256)
--(axis cs:789.8,0.964553675978168)
--(axis cs:1059.6,0.965807462688824)
--(axis cs:1360.4,0.966438728124103)
--(axis cs:1683.4,0.967393639500172)
--(axis cs:1988.2,0.968280693341788)
--(axis cs:2313.8,0.968948443001302)
--(axis cs:2646.4,0.96963514908336)
--(axis cs:3083,0.970012203552456)
--(axis cs:3774.2,0.970837793789729)
--(axis cs:4459,0.97077243370924)
--(axis cs:4897.8,0.970883227388281)
--(axis cs:5955.4,0.971063178460247)
--(axis cs:7269,0.970781365326229)
--(axis cs:8635.4,0.970167894405458)
--(axis cs:9865.8,0.969328048021153)
--(axis cs:10167.6,0.969401734036345)
--(axis cs:11250,0.968786180906398)
--(axis cs:13254.6,0.968175615536369)
--(axis cs:16073.8,0.967131125010114)
--(axis cs:18917,0.966124127962779)
--(axis cs:22529.6,0.965495891707004)
--(axis cs:30416,0.965119119562494)
--(axis cs:37524.4,0.963366999777117)
--(axis cs:44257.4,0.962550143406853)
--(axis cs:51788.6,0.962267516417772)
--(axis cs:58597,0.961666737711442)
--(axis cs:66447.4,0.961036167953866)
--(axis cs:75450.2,0.960750618747816)
--(axis cs:85671,0.960180929938225)
--(axis cs:85671,0.960813870061775)
--(axis cs:85671,0.960813870061775)
--(axis cs:75450.2,0.961734181252184)
--(axis cs:66447.4,0.961933032046134)
--(axis cs:58597,0.962512862288558)
--(axis cs:51788.6,0.963552083582228)
--(axis cs:44257.4,0.963881456593147)
--(axis cs:37524.4,0.964443800222883)
--(axis cs:30416,0.965936480437506)
--(axis cs:22529.6,0.966365308292997)
--(axis cs:18917,0.96690907203722)
--(axis cs:16073.8,0.967687274989886)
--(axis cs:13254.6,0.969089584463631)
--(axis cs:11250,0.969731019093602)
--(axis cs:10167.6,0.970071865963655)
--(axis cs:9865.8,0.970155551978847)
--(axis cs:8635.4,0.970452105594541)
--(axis cs:7269,0.971287834673771)
--(axis cs:5955.4,0.971418821539753)
--(axis cs:4897.8,0.971440372611719)
--(axis cs:4459,0.97127196629076)
--(axis cs:3774.2,0.971153406210271)
--(axis cs:3083,0.970839796447544)
--(axis cs:2646.4,0.97059245091664)
--(axis cs:2313.8,0.969753956998698)
--(axis cs:1988.2,0.969153306658212)
--(axis cs:1683.4,0.968326760499828)
--(axis cs:1360.4,0.967647271875897)
--(axis cs:1059.6,0.966887337311176)
--(axis cs:789.8,0.965131924021831)
--(axis cs:560.4,0.963285730491744)
--(axis cs:394.6,0.960565739618063)
--(axis cs:275.8,0.957933722888638)
--(axis cs:191.4,0.954202558152339)
--(axis cs:141.2,0.949640154955273)
--(axis cs:109,0.945513464287683)
--(axis cs:77.2,0.941604069571456)
--(axis cs:62,0.939375385852662)
--(axis cs:51.6,0.937995998453137)
--(axis cs:44.4,0.935162025346903)
--(axis cs:37,0.928709545331967)
--(axis cs:31,0.924363055957472)
--(axis cs:27,0.920531759267649)
--(axis cs:24.8,0.917763731730708)
--(axis cs:21.6,0.914713892837397)
--(axis cs:19,0.910702306159777)
--(axis cs:16,0.90883574956585)
--(axis cs:15.8,0.906051389891974)
--(axis cs:14.4,0.90044394412373)
--(axis cs:11.6,0.891390879263915)
--(axis cs:11,0.890523352084578)
--cycle;

\addplot [semithick, black, dashed]
table {%
10.6 0.8840844
13 0.8943418
13.4 0.9014318
18 0.9067506
20.2 0.9129974
27.6 0.9185096
29.4 0.9237242
33.8 0.927889
44.6 0.9314856
50 0.9351138
66 0.9386504
80.6 0.9424686
100.4 0.9468102
128.4 0.9506368
159.4 0.9539266
191.6 0.9570204
236.8 0.959891
287.2 0.9625926
359.4 0.9648796
445 0.9672348
521.2 0.9689636
624 0.97034
766.6 0.9716036
960 0.9726336
1215 0.973729
1521.6 0.9746914
1940 0.975513
2452.4 0.9763748
3109.2 0.9770494
3907.2 0.9778184
4900.8 0.9785914
6164 0.9793174
7739.8 0.980094
9494.2 0.9808444
11914.2 0.9815376
14547.6 0.9820858
17602 0.9826642
21167.4 0.9832232
24565.4 0.983745
29966.8 0.984232
34065.8 0.9845726
38596.4 0.984838
43367.8 0.9852002
48794.2 0.9853588
53063.4 0.9855064
55929.8 0.9855048
62398.2 0.9854396
70172 0.9853096
78113.4 0.98491
78325.8 0.9851428
};
\addlegendentry{\scriptsize LIBLINEAR}
\addplot [steelblue31119180, opacity=1.0, mark=*, mark size=0, mark options={solid}]
table {%
5.6 0.7157822
6 0.7157806
6 0.7157822
6 0.7157822
6 0.7157822
6 0.7157822
9 0.7286592
16 0.8303676
19.4 0.8839302
22.6 0.8943106
29.6 0.9016512
38.8 0.9070236
48.8 0.913109
64 0.918591
86.8 0.9237508
131 0.9284104
188.4 0.9324482
276.2 0.9362954
442.6 0.940217
694.6 0.944338
1051.8 0.948353
1570.2 0.952047
2264.4 0.9551824
3194.2 0.958171
4414.6 0.9605774
5921 0.9626088
7646.2 0.9642334
9698.2 0.9656452
12666.6 0.9667994
15598 0.9677978
19587.8 0.9685142
24033.4 0.969217
28315.2 0.9697676
32538 0.9702572
36716 0.9706486
40887.4 0.970835
45550.8 0.9707366
50372.4 0.9704424
55256.4 0.9701272
59696.6 0.9699034
62300.2 0.9700598
64186.2 0.9701676
69161 0.9698094
75890.4 0.9690242
81899.8 0.968325
91222.8 0.9672984
111862 0.964875
131773.2 0.9615198
141100.8 0.9603532
141690.6 0.9609524
};
\addlegendentry{\scriptsize Naive Avg.}
\addplot [darkorange25512714, opacity=1.0, mark=*, mark size=0, mark options={solid}]
table {%
5.6 0.7157822
6 0.7157822
6 0.7157822
6 0.7157822
6 0.7157822
6 0.7157822
9 0.728657
16 0.8303384
19.4 0.8839182
22.6 0.894315
29.6 0.9016496
38.8 0.9173562
48.8 0.923256
64 0.9344954
86.8 0.9327674
131 0.9413882
188.4 0.947148
276.2 0.9512968
442.6 0.9541528
694.6 0.9583118
1051.8 0.9603286
1570.2 0.9610148
2264.4 0.9625558
3194.2 0.9638384
4414.6 0.9645052
5921 0.9654918
7646.2 0.9667882
9698.2 0.9674078
12666.6 0.9676934
15598 0.9682946
19587.8 0.9690128
24033.4 0.968739
28315.2 0.9700516
32538 0.9705128
36716 0.9707168
40887.4 0.9708914
45550.8 0.9707336
50372.4 0.9704664
55256.4 0.970153
59696.6 0.9699342
62300.2 0.9702998
64186.2 0.9702182
69161 0.9698732
75890.4 0.9690436
81899.8 0.9683586
91222.8 0.9673948
111862 0.9652378
131773.2 0.9616514
141100.8 0.9604064
141690.6 0.96099
};
\addlegendentry{\scriptsize OWA}
\addplot [forestgreen4416044, opacity=1.0, mark=*, mark size=0, mark options={solid}]
table {%
11 0.8901066
11.6 0.8910632
14.4 0.9001204
15.8 0.9029776
16 0.9083722
19 0.910044
21.6 0.9141338
24.8 0.9173942
27 0.9188648
31 0.9240158
37 0.9275182
44.4 0.9339378
51.6 0.9375564
62 0.9389322
77.2 0.941295
109 0.9451118
141.2 0.949148
191.4 0.9535808
275.8 0.9567128
394.6 0.9601264
560.4 0.9629084
789.8 0.9648428
1059.6 0.9663474
1360.4 0.967043
1683.4 0.9678602
1988.2 0.968717
2313.8 0.9693512
2646.4 0.9701138
3083 0.970426
3774.2 0.9709956
4459 0.9710222
4897.8 0.9711618
5955.4 0.971241
7269 0.9710346
8635.4 0.97031
9865.8 0.9697418
10167.6 0.9697368
11250 0.9692586
13254.6 0.9686326
16073.8 0.9674092
18917 0.9665166
22529.6 0.9659306
30416 0.9655278
37524.4 0.9639054
44257.4 0.9632158
51788.6 0.9629098
58597 0.9620898
66447.4 0.9614846
75450.2 0.9612424
85671 0.9604974
};
\addlegendentry{\scriptsize ACOWA}
\end{axis}

\end{tikzpicture}

%% file: figs/app/elnet_ember100k_nnz_vs_test_acc.tex
\begin{tikzpicture}[scale=1.0]

\definecolor{darkgray153}{RGB}{153,153,153}
\definecolor{darkorange25512714}{RGB}{255,127,14}
\definecolor{forestgreen4416044}{RGB}{44,160,44}
\definecolor{gainsboro229}{RGB}{229,229,229}
\definecolor{steelblue31119180}{RGB}{31,119,180}

\begin{axis}[
width=0.95\textwidth,
height=0.35\textwidth,
legend cell align={left},
legend style={
  fill opacity=0.8,
  draw opacity=1,
  text opacity=1,
  at={(0.97,0.03)},
  anchor=south east,
  draw=none
},
log basis x={10},
tick align=outside,
tick pos=left,
x grid style={gainsboro229},
xlabel={\small Number of non-zeros},
xmajorgrids,
xmin=10, xmax=10000,
xminorgrids,
xmode=log,
xtick style={color=black},
y grid style={gainsboro229},
ylabel={\small Accuracy},
ymajorgrids,
ymin=0.45, ymax=1.0,
yminorgrids,
ytick style={color=black}
]
\path [fill=steelblue31119180, fill opacity=0.2]
(axis cs:1,0.502175706918333)
--(axis cs:1,0.500673893081667)
--(axis cs:1,0.500673893081667)
--(axis cs:1,0.500673893081667)
--(axis cs:1,0.500673893081667)
--(axis cs:1,0.500673893081667)
--(axis cs:1,0.500673893081667)
--(axis cs:1,0.500673893081667)
--(axis cs:1,0.500673893081667)
--(axis cs:1,0.500673893081667)
--(axis cs:1,0.500673893081667)
--(axis cs:1,0.500673893081667)
--(axis cs:1,0.500673893081667)
--(axis cs:1,0.500673893081667)
--(axis cs:1,0.500673893081667)
--(axis cs:1,0.500673893081667)
--(axis cs:1,0.500673893081667)
--(axis cs:1.8,0.500673893081667)
--(axis cs:20.8,0.683231419945159)
--(axis cs:57.8,0.703877195623714)
--(axis cs:117.8,0.722095549698259)
--(axis cs:227.4,0.731767194427443)
--(axis cs:336.8,0.748352597864512)
--(axis cs:538.4,0.757205951478199)
--(axis cs:876.8,0.781239841625932)
--(axis cs:1585.2,0.79292870840758)
--(axis cs:2726.6,0.815137671081727)
--(axis cs:4258.6,0.827519107887064)
--(axis cs:6279.6,0.841289059722483)
--(axis cs:8839.8,0.852540933049139)
--(axis cs:12159,0.867554863192803)
--(axis cs:16271.6,0.878058092864207)
--(axis cs:20887.2,0.887042581791966)
--(axis cs:25504.4,0.894448154372003)
--(axis cs:28684.4,0.899770082267326)
--(axis cs:31722.2,0.904295520445268)
--(axis cs:36203.6,0.908264573383507)
--(axis cs:41034.2,0.911837158294441)
--(axis cs:45821.4,0.914205093307462)
--(axis cs:50108.4,0.916189563139618)
--(axis cs:50902.4,0.918668625501702)
--(axis cs:52386,0.917611917993352)
--(axis cs:53764.6,0.920379633681547)
--(axis cs:58252.2,0.921699252856012)
--(axis cs:63143.6,0.922514622379945)
--(axis cs:68616.8,0.923123049166767)
--(axis cs:72450.2,0.924105136126135)
--(axis cs:74558.6,0.923487032482822)
--(axis cs:76661,0.923992105080018)
--(axis cs:77851.8,0.924201189209124)
--(axis cs:78899,0.923647928737671)
--(axis cs:78899,0.924898871262329)
--(axis cs:78899,0.924898871262329)
--(axis cs:77851.8,0.925318810790877)
--(axis cs:76661,0.924954694919981)
--(axis cs:74558.6,0.924473367517179)
--(axis cs:72450.2,0.924968463873865)
--(axis cs:68616.8,0.923960150833233)
--(axis cs:63143.6,0.923151777620055)
--(axis cs:58252.2,0.922260347143988)
--(axis cs:53764.6,0.920823166318453)
--(axis cs:52386,0.918164882006648)
--(axis cs:50902.4,0.919168174498298)
--(axis cs:50108.4,0.916864036860382)
--(axis cs:45821.4,0.914744506692538)
--(axis cs:41034.2,0.912049641705558)
--(axis cs:36203.6,0.909235426616493)
--(axis cs:31722.2,0.905181279554732)
--(axis cs:28684.4,0.900570317732674)
--(axis cs:25504.4,0.895505445627997)
--(axis cs:20887.2,0.888763818208034)
--(axis cs:16271.6,0.879278707135793)
--(axis cs:12159,0.868845136807197)
--(axis cs:8839.8,0.853885866950861)
--(axis cs:6279.6,0.842184140277517)
--(axis cs:4258.6,0.828471292112936)
--(axis cs:2726.6,0.816432728918273)
--(axis cs:1585.2,0.79483769159242)
--(axis cs:876.8,0.782670158374068)
--(axis cs:538.4,0.771427648521801)
--(axis cs:336.8,0.749167402135488)
--(axis cs:227.4,0.732549605572557)
--(axis cs:117.8,0.724340850301741)
--(axis cs:57.8,0.706779204376286)
--(axis cs:20.8,0.687892180054841)
--(axis cs:1.8,0.502175706918333)
--(axis cs:1,0.502175706918333)
--(axis cs:1,0.502175706918333)
--(axis cs:1,0.502175706918333)
--(axis cs:1,0.502175706918333)
--(axis cs:1,0.502175706918333)
--(axis cs:1,0.502175706918333)
--(axis cs:1,0.502175706918333)
--(axis cs:1,0.502175706918333)
--(axis cs:1,0.502175706918333)
--(axis cs:1,0.502175706918333)
--(axis cs:1,0.502175706918333)
--(axis cs:1,0.502175706918333)
--(axis cs:1,0.502175706918333)
--(axis cs:1,0.502175706918333)
--(axis cs:1,0.502175706918333)
--(axis cs:1,0.502175706918333)
--cycle;

\path [fill=darkorange25512714, fill opacity=0.2]
(axis cs:1,0.502175706918333)
--(axis cs:1,0.500673893081667)
--(axis cs:1,0.500673893081667)
--(axis cs:1,0.500673893081667)
--(axis cs:1,0.500673893081667)
--(axis cs:1,0.500673893081667)
--(axis cs:1,0.500673893081667)
--(axis cs:1,0.500673893081667)
--(axis cs:1,0.500673893081667)
--(axis cs:1,0.500673893081667)
--(axis cs:1,0.500673893081667)
--(axis cs:1,0.500673893081667)
--(axis cs:1,0.500673893081667)
--(axis cs:1,0.500673893081667)
--(axis cs:1,0.500673893081667)
--(axis cs:1,0.500673893081667)
--(axis cs:1,0.500673893081667)
--(axis cs:1.8,0.500673893081667)
--(axis cs:20.8,0.685964290382632)
--(axis cs:57.8,0.70280912345082)
--(axis cs:117.8,0.721145207039672)
--(axis cs:227.4,0.734847929718617)
--(axis cs:336.8,0.755270383448345)
--(axis cs:538.4,0.772104883169577)
--(axis cs:876.8,0.793811289119731)
--(axis cs:1585.2,0.786142727468637)
--(axis cs:2726.6,0.813571102844058)
--(axis cs:4258.6,0.828771450597299)
--(axis cs:6279.6,0.842515666219536)
--(axis cs:8839.8,0.85685525498468)
--(axis cs:12159,0.86789222974778)
--(axis cs:16271.6,0.882314190289003)
--(axis cs:20887.2,0.889892171554239)
--(axis cs:25504.4,0.892274820686755)
--(axis cs:28684.4,0.899073357654417)
--(axis cs:31722.2,0.907576323763087)
--(axis cs:36203.6,0.919929205214865)
--(axis cs:41034.2,0.912432726425105)
--(axis cs:45821.4,0.919103480854085)
--(axis cs:50108.4,0.918152397620806)
--(axis cs:50902.4,0.922130228568921)
--(axis cs:52386,0.919906133457194)
--(axis cs:53764.6,0.920790380932897)
--(axis cs:58252.2,0.921908957818668)
--(axis cs:63143.6,0.923194974722311)
--(axis cs:68616.8,0.924363706169183)
--(axis cs:72450.2,0.923537151986496)
--(axis cs:74558.6,0.922950163176193)
--(axis cs:76661,0.924433761285433)
--(axis cs:77851.8,0.886816395804698)
--(axis cs:78899,0.925996524558166)
--(axis cs:78899,0.930849875441834)
--(axis cs:78899,0.930849875441834)
--(axis cs:77851.8,0.943913604195302)
--(axis cs:76661,0.932703038714567)
--(axis cs:74558.6,0.930790236823807)
--(axis cs:72450.2,0.932549248013504)
--(axis cs:68616.8,0.930936693830817)
--(axis cs:63143.6,0.930395025277689)
--(axis cs:58252.2,0.930811042181332)
--(axis cs:53764.6,0.928886019067103)
--(axis cs:52386,0.930007066542806)
--(axis cs:50902.4,0.929779771431079)
--(axis cs:50108.4,0.926281202379194)
--(axis cs:45821.4,0.926196919145915)
--(axis cs:41034.2,0.924470873574895)
--(axis cs:36203.6,0.922740394785135)
--(axis cs:31722.2,0.922730476236914)
--(axis cs:28684.4,0.916809842345583)
--(axis cs:25504.4,0.912101979313245)
--(axis cs:20887.2,0.916681428445761)
--(axis cs:16271.6,0.912265809710997)
--(axis cs:12159,0.896450970252221)
--(axis cs:8839.8,0.89778114501532)
--(axis cs:6279.6,0.890724333780464)
--(axis cs:4258.6,0.834635349402701)
--(axis cs:2726.6,0.874382097155942)
--(axis cs:1585.2,0.845854072531363)
--(axis cs:876.8,0.837588710880269)
--(axis cs:538.4,0.830238316830423)
--(axis cs:336.8,0.791269216551655)
--(axis cs:227.4,0.775458470281383)
--(axis cs:117.8,0.729091592960328)
--(axis cs:57.8,0.73487407654918)
--(axis cs:20.8,0.689602509617368)
--(axis cs:1.8,0.502175706918333)
--(axis cs:1,0.502175706918333)
--(axis cs:1,0.502175706918333)
--(axis cs:1,0.502175706918333)
--(axis cs:1,0.502175706918333)
--(axis cs:1,0.502175706918333)
--(axis cs:1,0.502175706918333)
--(axis cs:1,0.502175706918333)
--(axis cs:1,0.502175706918333)
--(axis cs:1,0.502175706918333)
--(axis cs:1,0.502175706918333)
--(axis cs:1,0.502175706918333)
--(axis cs:1,0.502175706918333)
--(axis cs:1,0.502175706918333)
--(axis cs:1,0.502175706918333)
--(axis cs:1,0.502175706918333)
--(axis cs:1,0.502175706918333)
--cycle;

\path [fill=forestgreen4416044, fill opacity=0.2]
(axis cs:1,0.502175706918333)
--(axis cs:1,0.500673893081667)
--(axis cs:1,0.500673893081667)
--(axis cs:1,0.500673893081667)
--(axis cs:1,0.500673893081667)
--(axis cs:1,0.500673893081667)
--(axis cs:1,0.500673893081667)
--(axis cs:1,0.500673893081667)
--(axis cs:23,0.686717400584087)
--(axis cs:32,0.687323809625377)
--(axis cs:44.2,0.704871008904478)
--(axis cs:59.6,0.707038795002801)
--(axis cs:71.6,0.723358000848681)
--(axis cs:78,0.742025154131308)
--(axis cs:91,0.743170885981569)
--(axis cs:107.2,0.765244320330685)
--(axis cs:126.2,0.773525762173738)
--(axis cs:147,0.779896244026388)
--(axis cs:170,0.781368297058292)
--(axis cs:203.8,0.803405502259958)
--(axis cs:246,0.812416800112527)
--(axis cs:312.2,0.821632739679243)
--(axis cs:378.8,0.837175164526419)
--(axis cs:463,0.838192887379457)
--(axis cs:566.8,0.864203320274961)
--(axis cs:730.8,0.859265694035048)
--(axis cs:985.4,0.881410252307721)
--(axis cs:1336.6,0.884247373339265)
--(axis cs:1818.2,0.90023387200702)
--(axis cs:2420.2,0.899745703119784)
--(axis cs:3070,0.910419552682092)
--(axis cs:3466.2,0.915582925842427)
--(axis cs:4298,0.913866616594332)
--(axis cs:5228.6,0.922507838938652)
--(axis cs:6312,0.918785281273801)
--(axis cs:7443.8,0.923356850324164)
--(axis cs:8725.2,0.923820184402957)
--(axis cs:9991.6,0.927047524479193)
--(axis cs:11154,0.927346265640811)
--(axis cs:13293.2,0.927654960353447)
--(axis cs:15660.6,0.925920156510036)
--(axis cs:17981.8,0.926605736442347)
--(axis cs:20470,0.927249154687886)
--(axis cs:23180.6,0.925075306393654)
--(axis cs:26217.4,0.924996562913715)
--(axis cs:28986.8,0.925374553231383)
--(axis cs:31743,0.927377858830892)
--(axis cs:35431,0.929544930130257)
--(axis cs:39542.8,0.926170323367381)
--(axis cs:43774.6,0.926191307497482)
--(axis cs:48151.2,0.925351709795707)
--(axis cs:48151.2,0.933574690204293)
--(axis cs:48151.2,0.933574690204293)
--(axis cs:43774.6,0.933515092502518)
--(axis cs:39542.8,0.931942876632619)
--(axis cs:35431,0.933298269869743)
--(axis cs:31743,0.934398941169108)
--(axis cs:28986.8,0.932245446768617)
--(axis cs:26217.4,0.932323837086285)
--(axis cs:23180.6,0.931157893606346)
--(axis cs:20470,0.932561245312113)
--(axis cs:17981.8,0.933494663557653)
--(axis cs:15660.6,0.933263043489964)
--(axis cs:13293.2,0.932718239646553)
--(axis cs:11154,0.931650534359189)
--(axis cs:9991.6,0.930012875520807)
--(axis cs:8725.2,0.931206215597043)
--(axis cs:7443.8,0.928053549675836)
--(axis cs:6312,0.924734718726199)
--(axis cs:5228.6,0.925038961061348)
--(axis cs:4298,0.921143783405668)
--(axis cs:3466.2,0.920183474157573)
--(axis cs:3070,0.919070447317908)
--(axis cs:2420.2,0.911127896880216)
--(axis cs:1818.2,0.907582527992981)
--(axis cs:1336.6,0.900775826660735)
--(axis cs:985.4,0.890996547692279)
--(axis cs:730.8,0.878714305964952)
--(axis cs:566.8,0.867863479725039)
--(axis cs:463,0.858246712620543)
--(axis cs:378.8,0.853098035473581)
--(axis cs:312.2,0.833530860320757)
--(axis cs:246,0.823436399887473)
--(axis cs:203.8,0.815401297740042)
--(axis cs:170,0.803501702941708)
--(axis cs:147,0.789337355973612)
--(axis cs:126.2,0.787237437826262)
--(axis cs:107.2,0.770852079669315)
--(axis cs:91,0.760782714018431)
--(axis cs:78,0.751204445868692)
--(axis cs:71.6,0.753828799151319)
--(axis cs:59.6,0.744888004997199)
--(axis cs:44.2,0.718475391095522)
--(axis cs:32,0.691412990374623)
--(axis cs:23,0.688532199415913)
--(axis cs:1,0.502175706918333)
--(axis cs:1,0.502175706918333)
--(axis cs:1,0.502175706918333)
--(axis cs:1,0.502175706918333)
--(axis cs:1,0.502175706918333)
--(axis cs:1,0.502175706918333)
--(axis cs:1,0.502175706918333)
--cycle;

\addplot [semithick, black, dashed]
table {%
1 0.5014248
1 0.5014248
1 0.5014248
1 0.5014248
1 0.5014248
1 0.5014248
2.2 0.5014248
20 0.6905032
32 0.7124718
54.6 0.7340316
90.8 0.7499966
111.8 0.7775584
168.4 0.7898468
223.8 0.815595
275 0.8297118
382.6 0.8483566
476.4 0.8627766
594.6 0.8789484
754.8 0.89338
892.2 0.90511
1086.2 0.9149668
1321.6 0.922025
1598.2 0.92847
1932 0.9343282
2428.2 0.9391632
2984.2 0.9444882
3570.4 0.9484666
4419.6 0.9519532
5182.6 0.9547266
6197.6 0.9570266
7427 0.9591934
8880.2 0.961325
10616 0.9629884
12678 0.9644966
14523.8 0.9658866
17545.6 0.9668834
20352 0.96789
23566.4 0.9687234
27086.6 0.969315
30981.2 0.9697234
35104.4 0.9699398
39647.6 0.969975
43806.2 0.9700518
47450 0.9700018
51908.8 0.9699982
57765.6 0.9697884
64195.2 0.9695048
69477.8 0.9692932
75273.6 0.9691348
81270.2 0.9690418
};
\addlegendentry{\scriptsize LIBLINEAR}
\addplot [steelblue31119180, opacity=1.0, mark=*, mark size=0, mark options={solid}]
table {%
1 0.5014248
1 0.5014248
1 0.5014248
1 0.5014248
1 0.5014248
1 0.5014248
1 0.5014248
1 0.5014248
1 0.5014248
1 0.5014248
1 0.5014248
1 0.5014248
1 0.5014248
1 0.5014248
1 0.5014248
1 0.5014248
1.8 0.5014248
20.8 0.6855618
57.8 0.7053282
117.8 0.7232182
227.4 0.7321584
336.8 0.74876
538.4 0.7643168
876.8 0.781955
1585.2 0.7938832
2726.6 0.8157852
4258.6 0.8279952
6279.6 0.8417366
8839.8 0.8532134
12159 0.8682
16271.6 0.8786684
20887.2 0.8879032
25504.4 0.8949768
28684.4 0.9001702
31722.2 0.9047384
36203.6 0.90875
41034.2 0.9119434
45821.4 0.9144748
50108.4 0.9165268
50902.4 0.9189184
52386 0.9178884
53764.6 0.9206014
58252.2 0.9219798
63143.6 0.9228332
68616.8 0.9235416
72450.2 0.9245368
74558.6 0.9239802
76661 0.9244734
77851.8 0.92476
78899 0.9242734
};
\addlegendentry{\scriptsize Naive Avg.}
\addplot [darkorange25512714, opacity=1.0, mark=*, mark size=0, mark options={solid}]
table {%
1 0.5014248
1 0.5014248
1 0.5014248
1 0.5014248
1 0.5014248
1 0.5014248
1 0.5014248
1 0.5014248
1 0.5014248
1 0.5014248
1 0.5014248
1 0.5014248
1 0.5014248
1 0.5014248
1 0.5014248
1 0.5014248
1.8 0.5014248
20.8 0.6877834
57.8 0.7188416
117.8 0.7251184
227.4 0.7551532
336.8 0.7732698
538.4 0.8011716
876.8 0.8157
1585.2 0.8159984
2726.6 0.8439766
4258.6 0.8317034
6279.6 0.86662
8839.8 0.8773182
12159 0.8821716
16271.6 0.89729
20887.2 0.9032868
25504.4 0.9021884
28684.4 0.9079416
31722.2 0.9151534
36203.6 0.9213348
41034.2 0.9184518
45821.4 0.9226502
50108.4 0.9222168
50902.4 0.925955
52386 0.9249566
53764.6 0.9248382
58252.2 0.92636
63143.6 0.926795
68616.8 0.9276502
72450.2 0.9280432
74558.6 0.9268702
76661 0.9285684
77851.8 0.915365
78899 0.9284232
};
\addlegendentry{\scriptsize OWA}
\addplot [forestgreen4416044, opacity=1.0, mark=*, mark size=0, mark options={solid}]
table {%
1 0.5014248
1 0.5014248
1 0.5014248
1 0.5014248
1 0.5014248
1 0.5014248
1 0.5014248
23 0.6876248
32 0.6893684
44.2 0.7116732
59.6 0.7259634
71.6 0.7385934
78 0.7466148
91 0.7519768
107.2 0.7680482
126.2 0.7803816
147 0.7846168
170 0.792435
203.8 0.8094034
246 0.8179266
312.2 0.8275818
378.8 0.8451366
463 0.8482198
566.8 0.8660334
730.8 0.86899
985.4 0.8862034
1336.6 0.8925116
1818.2 0.9039082
2420.2 0.9054368
3070 0.914745
3466.2 0.9178832
4298 0.9175052
5228.6 0.9237734
6312 0.92176
7443.8 0.9257052
8725.2 0.9275132
9991.6 0.9285302
11154 0.9294984
13293.2 0.9301866
15660.6 0.9295916
17981.8 0.9300502
20470 0.9299052
23180.6 0.9281166
26217.4 0.9286602
28986.8 0.92881
31743 0.9308884
35431 0.9314216
39542.8 0.9290566
43774.6 0.9298532
48151.2 0.9294632
};
\addlegendentry{\scriptsize ACOWA}
\end{axis}

\end{tikzpicture}